\documentclass{article}

\usepackage{fullpage}
\usepackage[round,authoryear]{natbib}

\usepackage{graphicx}
\usepackage{caption}
\usepackage{subcaption}
\usepackage{booktabs} 
\usepackage{amsfonts,mathtools}
\usepackage{algorithmic}
\usepackage[]{algorithm}
\usepackage{array}
\usepackage{url}
\usepackage{bm}
\usepackage{mathrsfs}
\usepackage{amsmath,bm}
\usepackage{xcolor}
\usepackage{float}
\usepackage[font={small}]{caption}
\usepackage{epstopdf}
\usepackage{amssymb}
\usepackage{enumitem}
\usepackage[colorlinks=true,citecolor=blue,urlcolor=black,linkcolor=black,bookmarks=false,hypertexnames=false]{hyperref}

\usepackage{amsthm}

\theoremstyle{plain}
\newtheorem{theorems}{Theorem}[section]
\newtheorem{lemmas}[theorems]{Lemma}
\newtheorem{corollaries}[theorems]{Corollary}

\theoremstyle{definition}
\newtheorem{definitions}[theorems]{Definition}
\newtheorem{assumptions}[theorems]{Assumption}
\newtheorem{remarks}[theorems]{Remark}

\newcommand{\norm}[1]{\ensuremath{\left\|#1\right\|}}	
\providecommand{\tr}[1]{\text{tr}\left(#1\right)}
\providecommand{\norm}[1]{\left \| #1 \right \|}

\newcommand{\normI}[1]{{\left\vert\kern-0.25ex\left\vert\kern-0.25ex\left\vert #1 
    \right\vert\kern-0.25ex\right\vert\kern-0.25ex\right\vert}}

\def \G {{\mathcal{G}}}

\def \S {{\mathcal{S}}}
\def \Sm {{\bm {S}}}

\def \E {{\mathcal{E}}}

\def \L {{\mathcal{L}}}
\def \R {{\mathbb{R}}}

\begin{document}

	\title{Does the $\ell_1$-norm Learn a Sparse Graph under Laplacian Constrained Graphical Models?}

	\author{ Jiaxi Ying 
        \thanks{\noindent Department of Mathematics, The Hong Kong University of Science and Technology, Clear Water Bay, Hong Kong; Email: \href{mailto:jx.ying@connect.ust.hk}{\texttt{jx.ying@connect.ust.hk}}.}
	\and
	Jos\'{e} Vin\'{i}cius de M. Cardoso
	\thanks{\noindent Department of Electronic and Computer Engineering, The Hong Kong University of Science and Technology, Clear Water Bay, Hong Kong; Email: \href{mailto:jvdmc@connect.ust.hk}{\texttt{jvdmc@connect.ust.hk}}.}
	\and
	Daniel P. Palomar
	\thanks{\noindent Department of Electronic and Computer Engineering, Department of Industrial Engineering and Data Analytics, The Hong Kong University of Science and Technology,
 Clear Water Bay, Hong Kong; Email: \href{mailto:palomar@ust.hk}{\texttt{palomar@ust.hk}}.}
	}

\maketitle

\begin{abstract}

We consider the problem of learning a sparse graph under the Laplacian constrained Gaussian graphical models. This problem can be formulated as a penalized maximum likelihood estimation of the Laplacian constrained precision matrix. Like in the classical graphical lasso problem, recent works made use of the $\ell_1$-norm regularization with the goal of promoting sparsity in Laplacian constrained precision matrix estimation. However, we find that the widely used $\ell_1$-norm is not effective in imposing a sparse solution in this problem. Through empirical evidence, we observe that the number of nonzero graph weights grows with the increase of the regularization parameter. From a theoretical perspective, we prove that a large regularization parameter will surprisingly lead to a complete graph, i.e., every pair of vertices is connected by an edge. To address this issue, we introduce the nonconvex sparsity penalty, and propose a new estimator by solving a sequence of weighted $\ell_1$-norm penalized sub-problems. We establish the non-asymptotic optimization performance guarantees on both optimization error and statistical error, and prove that the proposed estimator can recover the edges correctly with a high probability. To solve each sub-problem, we develop a projected gradient descent algorithm which enjoys a linear convergence rate. Finally, an extension to learn disconnected graphs is proposed by imposing additional rank constraint. We propose a numerical algorithm based on based on the alternating direction method of multipliers, and establish its theoretical sequence convergence. Numerical experiments involving synthetic and real-world data sets demonstrate the effectiveness of the proposed method.

\vspace{.2cm}

\textbf{Keywords:} Sparse graph learning, Laplacian constrained Gaussian graphical model, Graph Laplacian, Nonconvex optimization, Graph signal processing.
\end{abstract}

\vspace{.2cm}

\section{Introduction} \label{Introduction}
Gaussian graphical models (GGM) have been widely used in a number of fields such as finance, bioinformatics, and image analysis \citep{banerjee2008model, hartemink2000using, li2006gradient, park2017learning}. Graph learning under GGM can be formulated to estimate the precision matrix that captures the conditional dependency relations between random variables \citep{dempster1972covariance, lauritzen1996graphical}. In this paper, the goal is to learn a sparse graph under the Laplacian constrained GGM, where the precision matrix obeys the Laplacian constraints.

The general GGM has received broad interest in statistical machine learning, where the problem can be formulated as a sparse precision matrix estimation. The authors \citet{banerjee2008model, d2008first, yuan2007model} proposed the $\ell_1$-norm penalized maximum likelihood estimation method, also known as graphical lasso, to encourage sparsity in its entries. Numerous methods have been developed for solving this optimization problem. To solve the primal problem, first-order methods including Nesterov’s smooth gradient method \citep{d2008first}, augmented Lagrangian method \citep{scheinberg2010sparse}, and second-order methods like inexact interior point method \citep{li2010inexact}, and Newton's method \citep{dinh2013proximal, hsieh2014quic, oztoprak2012newton, schmidt2009optimizing,wang2010solving} have been proposed for sparse precision matrix estimation. To solve the dual problem, block coordinate ascent method \citep{banerjee2008model,friedman2008sparse}, projected subgradient method \citep{Duchi2008}, and accelerated gradient descent method \citep{lu2009smooth} have been proposed to estimate a sparse precision matrix. In addition, various extensions of graphical lasso and their theoretical properties have also been studied \citep{honorio2012variable,mazumder2012graphical, ravikumar2011high, shojaie2010penalized, yang2012graphical,yang2014elementary}. However, those methods mentioned above focus on general graphical models and cannot be directly extended to Laplacian constrained GGM because of the multiple constraints on the precision matrix. Moreover, unlike the case of GGM, this paper will show that the $\ell_1$-norm is not effective in promoting sparsity in the penalized maximum likelihood estimation under Laplacian constrained GGM.   

In recent years, Laplacian constrained GGM has received increasing attention in signal processing and machine learning over graphs \citep{dong2019learning,ortega2018graph,shuman2013emerging,pmlr-v130-ying21a}. Under Laplacian constrained GGM, graph learning can be formulated as Laplacian constrained precision matrix estimation. Unlike general GGM with a general positive definite precision matrix, the precision matrix in Laplacian constrained GGM enjoys the spectral property that its eigenvalues and eigenvectors can be interpreted as spectral frequencies and Fourier basis \citep{shuman2013emerging}, which is very useful in computing graph Fourier transform in graph signal processing \citep{ortega2018graph,shuman2013emerging}, and graph convolutional networks \citep{locallyBruna,niepert2016learning,ruiz2019invariance}. \citet{dong2016learning,egilmez2017graph,gadde2015probabilistic,zhang2012analyzing} formulated the graph signals as random variables under the Laplacian constrained GGM. The learned graph under Laplacian constrained GGM favours smooth graph signal representations \citep{dong2016learning}, since the graph Laplacian quadratic term quantifies the smoothness of graph signals \citep{kalofolias2016learn,kumar2019unified}. \citet{kumar2019structured} proposed to learn structured graphs under the Laplacian constrained GGM where a number of graph structures can be learned by imposing different Laplacian spectral constraints. A regularized Laplacian constrained GGM was proposed for graph structure discovery \citep{lake2010discovering} and dimensionality reduction \citep{lawrence2012unifying}, where the precision matrix is a Laplacian matrix plus a very small diagonal matrix. The underlying assumption in \citep{lake2010discovering,lawrence2012unifying} is that each data feature is independent and identically distributed sampled from a regularized Laplacian constrained GGM. \citet{NEURIPS2021_a64a034c,NEURIPS2022_5adff4d5} extended the Laplacian constrained GGM to handle heavy-tailed settings for the scenario of learning graphs in financial markets. However, sparse graph learning under Laplacian constrained GGM remains to be further explored. For example, how to effectively and efficiently learn a sparse graph and how to characterize the statistical error of the estimation under Laplacian constrained GGM are to be investigated.


This paper focuses on learning a sparse graph under the Laplacian constrained GGM, which can be formulated as the penalized maximum likelihood estimation of the Laplacian constrained precision matrix. The main contributions of this paper are fourfold:
\begin{itemize}
\item We find an unexpected behavior of the $\ell_1$-norm in learning a sparse connected graph under the Laplacian constrained GGM. More specifically, through empirical evidence, we observe that the number of nonzero graph weights grows as the regularization parameter increases. From a theoretical perspective, we prove that a large regularization parameter of the $\ell_1$-norm will surprisingly lead to a solution representing a complete graph, i.e., every pair of vertices is connected by an edge, instead of a sparse graph.

\item To overcome the issue of the $\ell_1$ norm, we introduce the nonconvex sparsity penalty, and propose a new estimator by solving a sequence of weighted $\ell_1$-norm penalized sub-problems. We establish the non-asymptotic optimization performance guarantees on both optimization error and statistical error. We prove that the proposed estimator can recover the graph edges correctly with a high probability.

\item To obtain the new estimator, we develop a projected gradient descent algorithm to solve each sub-problem, and prove that the algorithm enjoys a linear convergence rate. Then we extend our method to learn disconnected graphs by imposing the rank constraint. We propose an algorithm based on the Alternating Direction Method of Multipliers (ADMM), and establish the theoretical sequence convergence.

\item Numerical experiments on both synthetic and real-world data sets demonstrate the effectiveness of the proposed method in learning sparse and interpretable graphs. 
\end{itemize}

A short version of this paper \citep{ying2020nonconvex} has been published in NeurIPS 2020. On what concerns the new contributions of this paper in comparison to the short version, we would like to highlight the following novelties: first, it was unknown in the short version whether the learned graph edges are correct or not, while this paper establishes the edge recovery consistency; second, this paper establishes the linear convergence rate of the proposed algorithm; third, the method proposed in the short version is limited to learning connected graphs, while this paper proposes a new algorithm to learn disconnected graphs, and establishes its theoretical convergence; finally, extensive numerical experiments including new graph structures and new data sets are considered in this paper.

The paper proceeds as follows: Section \ref{background} formulates the problem and reviews related work. Section \ref{proposed} discusses the $\ell_1$-norm issue in learning sparse graphs and proposes nonconvex methods. We present the theoretical results in Section \ref{theoretical results}. An extension to learn disconnected graphs and the experimental results are presented in Sections \ref{Sec k-component} and \ref{experimental section}, respectively. Section \ref{conclusion} concludes and discusses the findings. Proofs and technical lemmas are in Appendixes \ref{Proof} and \ref{Lemmas}. The corresponding open source $\mathsf{R}$ package is available on GitHub at \url{https://github.com/mirca/sparseGraph}.

\vspace{-0.2cm}

\paragraph{Notation}
Lower case bold letters denote vectors and upper case bold letters denote matrices. Both $X_{ij}$ and $[\bm X]_{ij}$ denote the $(i, j)$-th entry of $\bm X$. $\bm X^\top$ denotes transpose of $\bm X$. $[p]$ denotes the set $\{1, \ldots, p\}$. Let $\mathrm{supp}^+(\bm x)=\{ i \in [p] | x_i>0\}$ for $\bm x \in \mathbb{R}^p$. We use $\bm x \geq \bm 0$ to denote that each element of $\bm x$ is non-negative. $\lceil x \rceil$ denotes the least integer greater than or equal to $x$. The all-zero and all-one vectors or matrices are denoted by $\bm 0$ and $\bm 1$. $\norm{\bm x}$, $\norm{\bm X}_{\mathrm{F}}$ and $\norm{\bm X}_2$ denote Euclidean norm, Frobenius norm and operator norm, respectively. $\normI{\bm X}_\infty$ denotes the $\ell_\infty / \ell_\infty$-operator norm given by $\normI{\bm X}_\infty := \max_{i=1, \ldots, p} \sum_{j=1}^p | X_{ij}|$, and $\normI{\bm X}_1$ denotes the $\ell_1 / \ell_1$-operator norm given by $\normI{\bm X}_1:= \max_{j=1, \ldots, p} \sum_{i=1}^p | X_{ij}|$. Let $\lambda_{\max}(\bm X)$ denote the maximum eigenvalue of $\bm X$. The inner product of two vectors is defined as
$\langle \bm x, \bm y\rangle= \sum_{i} x_i y_i$. Let $\norm{\bm x}_{\max} = \max_{i} |x_i|$ and $\norm{\bm x}_{\min} = \min_{i} |x_i|$. For functions $f(n)$ and $g(n)$, we use $f(n) \lesssim g(n)$ if $f(n) \leq C g(n)$ for some constant $C \in (0, +\infty)$. $\S_+^p$ and $\S_{++}^p$ denote the sets of positive semi-definite and positive definite matrices with size $p \times p$, respectively. $\mathbb{R}_+$ denotes the set of positive real numbers. 


\vspace{-0.2cm}

\section{Problem Formulation and Related Work} \label{background}

\vspace{-0.1cm}

We first present the definition of Laplacian constrained Gaussian Markov random fields and formulate the problem of learning a graph under the Laplacian constrained GGM. After that, we discuss related work.

\vspace{-0.2cm}

\subsection{Laplacian constrained Gaussian Graphical Model}
We define a weighted, undirected graph $\G= \left( \mathcal{V},\mathcal{E},\bm W \right)$ containing no graph loops or multiple edges, where $\mathcal{V}$ denotes the set of nodes, and the pair $(i, j) \in \mathcal{E}$ if and only if there is an edge between node $i$ and node $j$. $\bm W \in \mathbb{R}_+^{p \times p}$ is the weighted adjacency matrix with $W_{ij} \geq 0$ denoting the graph weight between node $i$ and node $j$. Note that the graph is undirected, and thus $\bm W$ is symmetric. The graph Laplacian $\bm L \in \mathbb{R}^{p \times p}$, also known as combinatorial graph Laplacian, is defined as
\begin{equation}
\bm L=\bm D-\bm W, \label{Lap}
\end{equation}
where $\bm D$ is a diagonal matrix with $D_{ii}=\sum_{j=1}^p W_{ij}$. A graph is called connected if there is a path between every pair of vertices. The sparse connected graphs include tree graph, modular graphs, grid graph, line graph, star graphs and so on. A graph is called complete if every pair of vertices is connected by an edge.

In this paper, we will focus on the problem of learning sparse connected graphs first, and then we will extend our method to learn multiple component graphs. From spectral graph theory \citep{chung1997spectral}, the rank of the Laplacian matrix for any connected graph with $p$ nodes is $p-1$. Therefore, the set of Laplacian matrices for connected graphs can be formulated as
\begin{equation}\label{Lap-set}
\S_L  = \lbrace \bm\Theta \in \S^{p}_+| \, \Theta_{ij} =\Theta_{ji} \leq 0, \, \forall \ i\neq j,\, \bm\Theta \cdot \bm 1 =\bm 0,\ \mathrm{rank} (\bm \Theta) = p-1 \rbrace,
\end{equation}
where $\bm 0$ and $\bm 1$ denote the constant zero and one vectors, respectively.  Next, we define the Laplacian constrained Gaussian Markov random fields, and without loss of generality we assume that the random vector $\bm x$ has zero mean.

\begin{definitions}\label{def-LGMRF}
A zero-mean random vector $\bm x =[x_1, \ldots, x_p]^{\top} \in V^{p-1}$ is called a Laplacian constrained Gaussian Markov Random Fields (LGMRF) with parameters $(\bm 0, \bm{\Theta})$ with $\bm \Theta \in \S_L$, if and only if its density function $q_L: V^{p-1} \to \mathbb{R}$ follows
\begin{equation}\label{dens-LGMRF}
q_L({\bm x})=(2\pi)^{-\frac{p-1}{2}}{\det}^\star({\bm\Theta})^{\frac{1}{2}}\exp\left(-\frac{1}{2} \bm x ^{\top} \bm \Theta \bm x \right),
\end{equation}
where ${\det}^\star$ denotes the pseudo determinant defined by the product of nonzero eigenvalues \citep{holbrook2018differentiating}, and $V^{p-1}$ is a $(p-1)$-dimensional subspace of the coordinate space $\mathbb{R}^{p}$ defined by $V^{p-1}:= \lbrace \bm x \in \mathbb{R}^p | \, \bm 1^{\top} \bm x  = 0 \rbrace$. 
\end{definitions}
Note that we restrict $\bm x$ into a subspace because the LGMRF does not have a density with respect to the $p$-dimensional Lebesgue measure. According to the disintegration theorem, we can construct a conditional probability measure defined on $V^{p-1}$ and then the density of LGMRF with this measure satisfies \eqref{dens-LGMRF}. In this sense, the LGMRF can be interpreted as a GMRF conditioned on the linear constraint $\bm 1^{\top} \bm x =0$ and thus each observation $\bm x^{(k)}$ of an LGMRF also satisfies $\bm 1^{\top} \bm x^{(k)} =0$. For convenience, we still denote $\bm \Theta$ in \eqref{dens-LGMRF} as the precision matrix, though it formally does not exist \citep{rue2005gaussian}.

Unlike GGM, the elements of the precision matrix in the Laplacian constrained GGM can quantify the similarity between two nodes. More specifically, because of the Laplacian constraints, its probability density function can be written as 
\begin{equation}\label{pdf}
q_L({\bm x}) = (2\pi)^{-\frac{p-1}{2}}{\det}^\star \left({\bm\Theta} \right)^{\frac{1}{2}} \exp \bigg(- \frac{1}{2}\sum_{i \neq j} W_{i j} \left( x_i - x_j \right)^2 \bigg),
\end{equation}
where $W_{ij} = - \Theta_{ij} \geq 0$. If the element $W_{ij}$ is large, then there will be a relatively high probability that $\left( x_i - x_j \right)^2$ is small. This property is desired in modelling smooth graphs where a large graph weight between two nodes indicates a significant similarity between their signal values. The Laplacian constrained GGM has been widely explored in graph signal processing \citep{dong2016learning} and semi-supervised learning \citep{zhu2003semi}, where the underlying graphs are usually assumed smooth.


Compared with GGM, there is an additional sign assumption on the Laplacian constrained GGM. More specifically, define $\bm x := \tilde{\bm x} - \frac{1}{p} \bm 1 \bm 1^\top \tilde{\bm x}$, the projection of $\tilde{\bm x}$ into the subspace $V^{p-1}$, where $\tilde{\bm x}$ follows a general multivariate Gaussian distribution. Then the precision matrix of $\bm x$ will satisfy $\bm \Theta \cdot \bm 1 = \bm 0$. If we add sign assumption that $\Theta_{ij} \leq 0$ $\forall i \neq j$, then $\bm x$ forms a Laplacian constrained GGM. The sign constraints on the entries of precision matrices are also explored in Gaussian distributions or general exponential families of distributions with multivariate total positivity \citep{fallat2017total,lauritzen2019maximum, lauritzen2019total, slawski2015estimation,wang2020learning,cai2021fast,pmlr-v202-ying23a}. Exploring the advantages of the sign constraint in precision matrices is currently an active research topic. \citet{lauritzen2019maximum,slawski2015estimation} proved that the maximum likelihood estimator (MLE) for Gaussian distributions with multivariate total positivity exists if the sample size $n \geq 2$, irrespective of the underlying dimension. This yields a drastic reduction from $n \geq p$ in general Gaussian graphical models. More recently, \citet{lauritzen2019total} showed that the MLE for the totally positive binary distributions may exist with $n = p$ observations, which is striking given that $2^p$ is required for the MLE to exist in the unconstrained binary exponential families.

Sparse graph learning under the Laplacian constrained Gaussian graphical model can be formulated as the penalized maximum likelihood of the Laplacian constrained precision matrix,
\begin{equation}\label{cost-theta}
\min_{\bm{\Theta} \in \S_L} - \log {\det} ( \bm\Theta + \bm J) +\tr{\bm\Theta \bm S}+ \sum_{i > j} h_{\lambda} \left(\Theta_{ij} \right),
\end{equation}
where $\Sm$ is the sample covariance matrix, $h_{\lambda}$ is a \textit{regularizer}, depending on a regularization parameter $\lambda \geq 0$, which serves to enforce sparsity, e.g.,  $h_{\lambda} (\Theta_{ij}) = \lambda |\Theta_{ij}|$. $\bm J=\frac{1}{p} \bm 1_{p \times p}$ is a constant matrix with each element equal to $\frac{1}{p}$. Note that we replace $ {\det}^{\star} ( \bm\Theta )$ with $ {\det} ( \bm\Theta + \bm J)$ in \eqref{cost-theta} as done in \citep{egilmez2017graph}, which follows from the fact that the matrix $\bm J$ is rank one, and the nonzero eigenvalue of $\bm J$ is $1$ whose eigenvector is orthogonal to the row and column spaces of $\bm \Theta \in \S_L$. It is worth mentioning that the estimation method in \eqref{cost-theta} only uses the component in $V^{p-1}$ of the data samples for estimation, and the mean of each data sample is filtered out. Therefore, the estimator obtained by \eqref{cost-theta} is invariant to the mean of the data samples.

We finish the subsection with discussions on the motivations of imposing Laplacian constraints in \eqref{cost-theta}. We consider data that fits the Laplacian constrained GGM well, for example the stock data which has been justified in Section \ref{Sec-Stock}. Imposing Laplacian constraints usually achieves a smaller estimation error than the GGM method without constraints, because the former introduces more prior knowledge and has a smaller search space. Furthermore, the learned graph weights under Laplacian constraints can quantify the similarity between two nodes, which is useful in community detection. Experimental results in Section \ref{Sec-Stock} show that imposing Laplacian constraints can produce a more representative graph on the stock data than the GGM method without constraints.

\subsection{Related Work}
The $\ell_1$-norm regularized maximum likelihood estimation under GGM has been extensively studied \citep{friedman2008sparse, hsieh2014quic,mazumder2012graphical,rothman2008sparse,shojaie2010penalized, yuan2007model}. To reduce estimation bias, nonconvex regularizers have been introduced in estimating a sparse precision matrix \citep{breheny2011coordinate,chen2018covariate,lam2009sparsistency,loh2015regularized,shen2012likelihood,zhang2010nearly}. Some popular nonconvex penalties include the smooth clipped absolute deviation (SCAD) \citep{fan2001variable}, minimax concave penalty (MCP) \citep{zhang2010nearly}, and capped $\ell_1$-penalty \citep{zhang2010analysis}. Several types of algorithms including local quadratic appoximation \citep{fan2001variable}, minimization-maximization \citep{hunter2005variable,SunBabPal2017-MM}, and local linear approximation \citep{zou2008one} were proposed to solve the nonconvex optimization. Recently, \citet{fan2018lamm,loh2015regularized, sun2018graphical} presented theoretical analysis to characterize nonconvex estimators with desired statistical guarantees. However, all those methods cannot be directly extended to Laplacian constrained GGM because we aim to learn a precision matrix that must satisfy the constraints in \eqref{Lap-set}, while the learned precision matrix under GGM is a general positive definite matrix.

Recent works \citep{egilmez2017graph,liu2019block,zhao2019optimization} proposed the $\ell_1$-norm penalized maximum likelihood estimation under Laplacian constrained GGM with the goal of learning a sparse graph. The authors \citet{egilmez2017graph,zhao2019optimization} designed a primal-dual algorithm that introduces additional variables to handle Laplacian constraints. The authors \citet{liu2019block} proposed a block coordinate descent method to solve the optimization problem. The authors \citet{kumar2019structured} proposed a $\ell_1$-norm regularized maximum likelihood estimation method with Laplacian spectral constrains to learn structured graphs such as $k$-component graphs. More recently, the authors \citet{kumar2019unified} proposed a framework with re-weighted $\ell_1$-norm to learn structured graphs by imposing spectral constraints on graph matrices. But, the proposed algorithms in \citep{kumar2019unified,kumar2019structured} have to compute the eigenvalue decomposition in each iteration which is computationally expensive in the high-dimensional regime. Note that all the methods mentioned above lack theoretical analysis on estimation error and algorithm convergence rate. In this paper, through theoretical analysis and empirical experiments, we show that the $\ell_1$-norm is not effective in promoting sparsity in the Laplacian constrained GGM, and further propose a nonconvex estimator with theoretical guarantees on the estimation consistency of the precision matrix and the selection consistency of the graph edges. The approach of theoretical analysis in this paper may be extended to characterize statistical properties of other estimators under the Laplacian constrained GGM.

\section{Proposed Method} \label{proposed}
In this section, we first focus on sparse connected graph learning. In Section \ref{Sec L1}, we present an unexpected behavior of the $\ell_1$-norm in learning a sparse connected graph under the Laplacian constrained GGM. Then, in Section \ref{Sec proposed algorithm}, we introduce the nonconvex sparsity penalty, and propose a new estimator by solving a sequence of weighted $\ell_1$-norm penalized sub-problems. We further develop a projected gradient descent algorithm with backtracking line search to solve each sub-problem.

\subsection{$\ell_1$-norm Regularizer} \label{Sec L1}
Sparsity is often explored in high-dimensional Gaussian graphical models in order to reduce the number of samples required. The effectiveness of the $\ell_1$-norm regularized maximum likelihood estimation, also known as graphical lasso, has been widely demonstrated in a number of fields. One common rule of thumb for graphical lasso is that the estimated graph will get sparser if a larger regularization parameter is used. However, we find an unexpected behavior of the $\ell_1$-norm in sparse graph learning under the Laplacian constrained GGM.

The $\ell_1$-norm regularized maximum likelihood estimation under the Laplacian constrained Gaussian graphical model \citep{egilmez2017graph,zhao2019optimization} can be formulated as 
\begin{equation}\label{Lap-est-l1}
\min_{\bm{\Theta} \in \S_L}- \log {\det} ( \bm\Theta + \bm J) +\tr{\bm\Theta \Sm}+ \lambda \sum_{i>j} |\Theta_{ij}|,
\end{equation}
where $\bm S$ is the sample covariance matrix with each sample independently sampled from LGMRF, and $\lambda$ is the regularization parameter. An intuition of using the $\ell_1$-norm in \eqref{Lap-est-l1} could be that increasing the parameter $\lambda$ will make the graph sparser, and finally leads to the sparsest connected graph, i.e., a tree graph. However, through theoretical derivations and empirical experiments, we show that a large regularization parameter will lead to a complete graph instead, i.e., every pair of vertices is connected by an edge.


\begin{theorems} \label{Theorem 1}
Let $\hat{\bm{\Theta}} \in \mathbb{R}^{p \times p}$ be the global minimum of \eqref{Lap-est-l1} with $p>3$. Define $s_1 = \max_k S_{kk}$ and $s_2 = \min_{ij}S_{ij}$. If the regularization parameter $\lambda$ in \eqref{Lap-est-l1} satisfies $\lambda \in [ (2+2\sqrt{2})(p+1)(s_1 - s_2), + \infty)$, then the estimated graph weight $\hat{W}_{ij} = - \hat{\Theta}_{ij}$ obeys
\begin{equation}
\hat{W}_{ij} \geq \frac{1}{(s_1 - (p+1) s_2 + \lambda ) p } >0, \quad \forall \ i \neq j.  \nonumber
\end{equation}
\end{theorems}

Theorem \ref{Theorem 1}, proved in Appendix \ref{sec-prof-L1-norm}, states that a large regularization parameter of the $\ell_1$-norm will force every graph weight to be strictly positive, thus the resultant graph will be a complete graph. 


\begin{figure}[!htb]
    \captionsetup[subfigure]{justification=centering}
    \centering
    \begin{subfigure}[t]{0.24\textwidth}
        \centering
        \includegraphics[scale=.45]{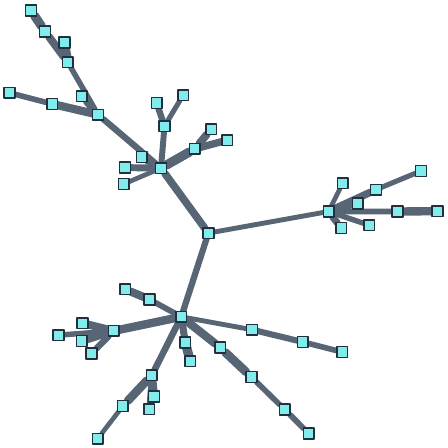}
        \caption{Ground-truth}
    \end{subfigure}%
    ~
    \begin{subfigure}[t]{0.24\textwidth}
        \centering
        \includegraphics[scale=.45]{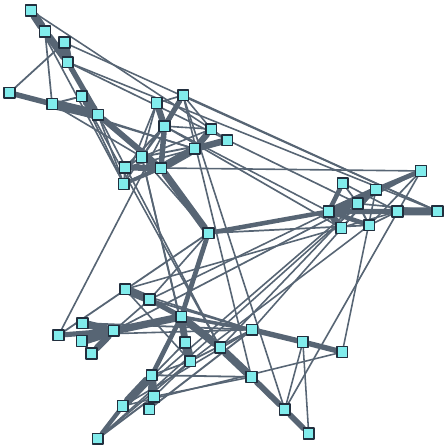}
        \caption{$\lambda=0$}
    \end{subfigure}%
    ~
    \begin{subfigure}[t]{0.24\textwidth}
        \centering
        \includegraphics[scale=.45]{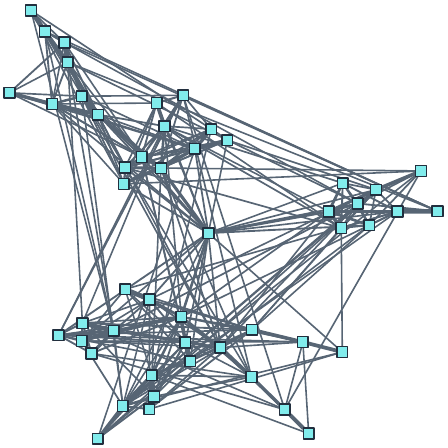}
        \caption{$\lambda = 0.1$}
    \end{subfigure}%
    ~
    \begin{subfigure}[t]{0.24\textwidth}
        \centering
        \includegraphics[scale=.45]{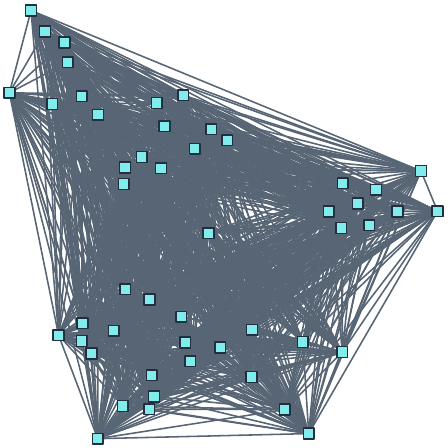}
        \caption{$\lambda = 10$}
    \end{subfigure}
    \caption{ Graph learning using $\ell_1$-norm with different regularization parameters. The number of positive edges in (a), (b), (c) and (d) are 49, 135, 286 and 1225, respectively, and (d) is a complete graph. The relative errors of the graphs in (b), (c) and (d) are 0.14, 0.64 and 0.99, respectively.}
    \label{fig:L1}
\end{figure}

\begin{figure}[H]
    \captionsetup[subfigure]{justification=centering}
    \centering
        \begin{subfigure}[t]{0.24\textwidth}
        \centering
        \includegraphics[scale=.27]{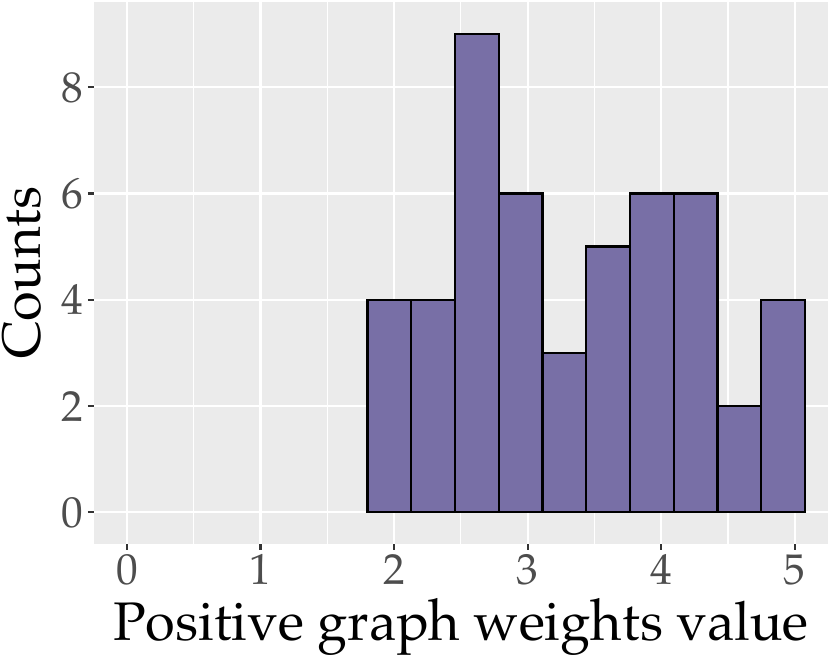}
        \caption{Ground-truth}
    \end{subfigure}%
    ~
    \begin{subfigure}[t]{0.24\textwidth}
        \centering
        \includegraphics[scale=.27]{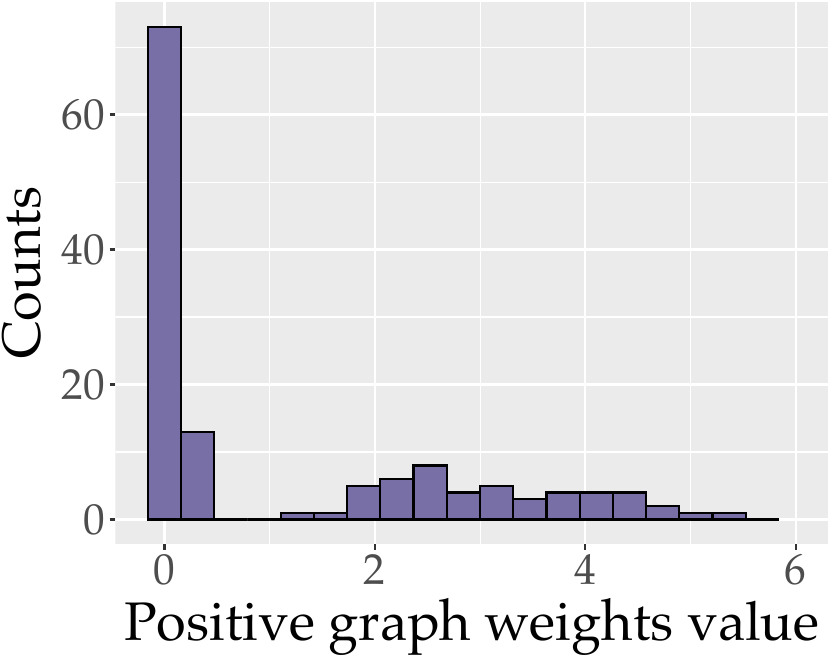}
        \caption{$\lambda=0$}
    \end{subfigure}%
    ~
    \begin{subfigure}[t]{0.24\textwidth}
        \centering
        \includegraphics[scale=.27]{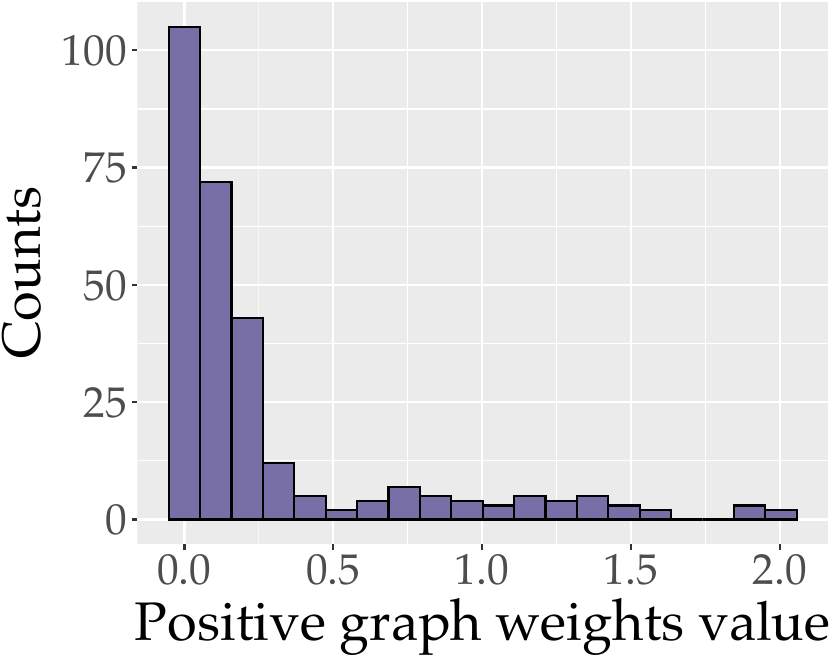}
        \caption{$\lambda=0.1$}
    \end{subfigure}%
    ~
    \begin{subfigure}[t]{0.24\textwidth}
        \centering
        \includegraphics[scale=.27]{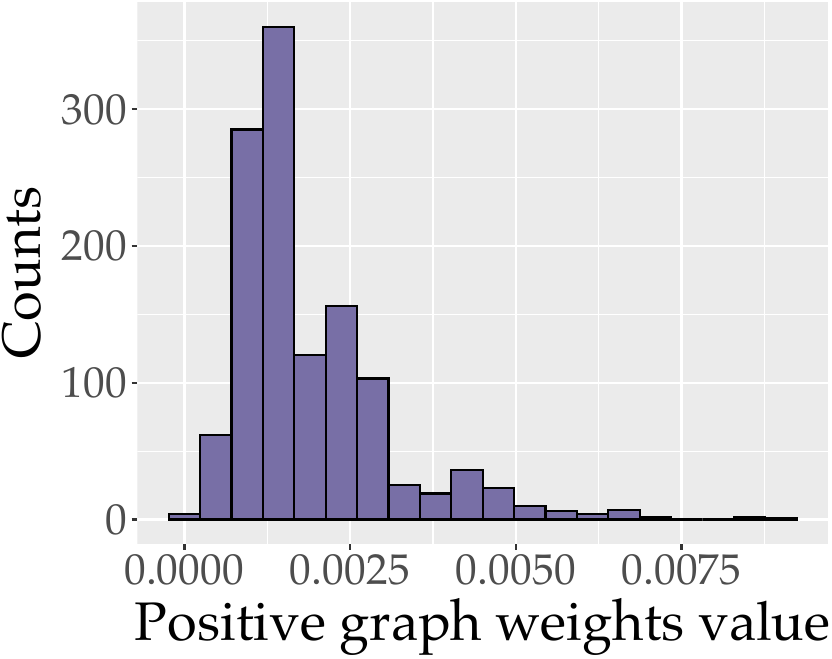}
        \caption{$\lambda = 10$}
    \end{subfigure}
    \caption{ Histograms of nonzero graph weights learned by the $\ell_1$-norm regularization method with different regularization parameters, corresponding to the graphs learned in Figure \ref{fig:L1}. The histograms count the number of nonzero graph weights falling into each interval.}
    \label{fig:L2}
\end{figure}


The theoretical result in Theorem \ref{Theorem 1} is consistent with empirical observations depicted in Figure~\ref{fig:L1}, which shows that the number of positive edges learned by \eqref{Lap-est-l1} grows along with the increase of $\lambda$, and finally the estimated graph in Figure~\ref{fig:L1} (d) is a complete graph. Figure \ref{fig:L2} illustrates the histograms of the nonzero graph weights (associated with the positive edges) learned in Figure~\ref{fig:L1}. Figure \ref{fig:L2} (d) depicts the histogram for the graph in Figure~\ref{fig:L1} (d), where every weight is strictly positive. It is further observed in Figure \ref{fig:L2} (d) that all the graph weights are very small. Therefore, a large regularization parameter will lead to a graph with every weight strictly positive and small. We can see that the histogram in Figure~\ref{fig:L2} (d) is significantly different from the true histogram in Figure \ref{fig:L2} (a), implying that the estimated model fails to identify the true relationships among the data variables.

It is well-known that a larger regularization parameter of graphical lasso will lead to a larger threshold, and the elements in the solution with their absolute values less than the threshold will be shrunk to zero. Therefore, the resultant solution of the graphical lasso will get sparser. The unexpected behavior of the $\ell_1$-norm characterized in Theorem \ref{Theorem 1} is due to the Laplacian constraints in the optimization \eqref{Lap-est-l1}. More specifically, because of the constrains $\bm\Theta \cdot \bm 1 = \bm 0$ and $\Theta_{ij} =\Theta_{ji} \leq 0$ for any $i \neq j$, then we have
\begin{equation}\label{Lap-L1}
\tr{\bm\Theta \Sm}+ \lambda \sum_{i>j} |\Theta_{ij}| = \sum_{i>j} (\lambda + S_{ii} + S_{jj} - S_{ij} - S_{ji} ) |\Theta_{ij}|.
\end{equation}
To intuitively understand the behavior of the $\ell_1$-norm with a large regularization parameter imposed in \eqref{Lap-est-l1}, suppose that $\lambda$ is sufficiently large such that $\tr{\bm\Theta \Sm}+ \lambda \sum_{i>j} |\Theta_{ij}| $ can be approximated by $ \sum_{i>j} \lambda  |\Theta_{ij}|$ well. Then \eqref{Lap-est-l1} will be reduced to the optimization problem as below
\begin{equation}\label{Lap-est-app}
\min_{\bm{\Theta} \in \S_L}- \log {\det} ( \bm\Theta + \bm J) + \lambda \sum_{i>j} |\Theta_{ij}|.
\end{equation}
The global minimizer of the optimization \eqref{Lap-est-app} is shown in Corollary \ref{corollary-L1}.
\begin{corollaries} \label{corollary-L1}
Let $\tilde{\bm{\Theta}}$ be the optimal solution of \eqref{Lap-est-app}. Then $\tilde{W}_{ij} = -\tilde{\Theta}_{ij} $ obeys
\begin{equation}
\tilde{W}_{ij} = \frac{2}{p \lambda}, \quad  \mathrm{for \  any} \ i \neq j.
\end{equation}
\end{corollaries}
Corollary \ref{corollary-L1}, proved in Appendix \ref{sec-prof-L1-norm}, shows that every estimated graph weight $\tilde{W}_{ij}$ is strictly positive, and equal with each other. Moreover, the affect of the $\bm S$ is negligible if a large $\lambda$ is chosen. Therefore, no matter what kind of graph structure or connectivity the underlying graph has, the $\ell_1$ norm does not work as intended, i.e., as a sparsity promoting regularization. It is worth mentioning that it is the two constraints $\bm\Theta \cdot \bm 1 = \bm 0$ and $\Theta_{ij} =\Theta_{ji} \leq 0$ together that make \eqref{Lap-L1} hold, implying that the $\ell_1$ norm can still work well under the GGM with the generalized Laplacian constraints \citep{egilmez2017graph, pavez2018learning, pavez2016generalized,ying2021fast}, where $\bm\Theta \cdot \bm 1 = \bm 0$ does not hold.


To solve the issue of the $\ell_1$-norm in Laplacian constrained GGM, we introduce the nonconvex regularizer, and the effectiveness of the proposed method with the nonconvex regularizer has been demonstrated by numerical experiments in Section \ref{experimental section}.

\subsection{Proposed Algorithm} \label{Sec proposed algorithm}
Problem \eqref{cost-theta} is a constrained optimization problem with $\bm{\Theta} \in \S_L$ including multiple constraints. We first simplify the Laplacian constraints in \eqref{Lap-set} as follows.  

The constraints $\Theta_{ij} = \Theta_{ji}$ and $\bm \Theta \cdot \bm 1 =\bm 0$ in \eqref{Lap-set} are linear and there are only $\frac{p(p-1)}{2}$ free variables in $\bm \Theta$. Therefore, we use a linear operator defined in \citep{kumar2019unified} that maps a vector $\bm x \in \R ^ {p(p-1)/2}$ to a matrix $\mathcal{L} \bm x \in \R^{p \times p}$ as below.
\begin{definitions}
The linear operator $\mathcal{L}: \mathbb{R}^ {p(p-1)/2}\rightarrow \mathbb{R}^{p \times p}$, $\bm x \mapsto \mathcal{L} \bm x$, is defined by
\begin{equation}\label{operator}
[\mathcal{L} \bm x]_{ij} =
\begin{cases}
-x_{k}\; & \; \;i >j,\\
\hspace{.2cm}[\mathcal{L}\bm x]_{ji} \; & \; \;i<j,\\
-\sum_{j \neq i}[\mathcal{L} \bm x]_{ij}\; &\;\; i=j,
\end{cases}
\end{equation}
where $k=i-j+\frac{j-1}{2}(2p-j)$.
\end{definitions}

A simple example is given below which illustrates the definition of the operator $\mathcal{L}$. Let $\bm x \in \mathbb{R}^3$. Then we have  
\begin{equation*} 
\mathcal{L} \bm x= \left[
\begin{array}{ccc}
\sum_{i=1,2} x_i & -x_1 &  -x_2\\
-x_1 & \sum_{i=1,3} x_i &  -x_3\\
-x_2 & -x_3 & \sum_{i=2,3}x_i\\
\end{array}
\right].
\end{equation*}

The adjoint operator $\mathcal{L}^*$ of $\mathcal{L}$ is defined so as to satisfy $\langle \mathcal{L}\bm x,\bm Y \rangle=\langle \bm x,\mathcal{L}^* \bm Y \rangle$, $\forall \bm x \in \mathbb{R}^{p(p-1)/2}$ and $\bm Y \in \mathbb{R}^{p \times p}$.
\begin{definitions}
The adjoint operator $\mathcal{L}^*: \mathbb{R}^{p\times p} \rightarrow \mathbb{R}^{p(p-1)/2}$, $\bm Y \mapsto \mathcal{L}^* \bm Y$, is defined by
	\begin{equation}\label{adj-operator}
	[\mathcal{L}^* \bm Y]_k=Y_{i,i}-Y_{i,j}-Y_{j,i}+Y_{j,j},
	\end{equation}
	where $i,j \in [p]$ obeying $ k=i-j+\frac{j-1}{2}(2p-j)$ and $i>j$.
\end{definitions}

By introducing the linear operator $\mathcal{L}$, we simplify the definition of $\S_L$ in \eqref{Lap-set} in the following theorem, which is proved in Appendix \ref{sec-prof-operator}.
\begin{theorems} \label{Theorem 2}
The Laplacian set $\S_{L}$ defined in \eqref{Lap-set} can be written as 
\begin{equation}\label{Lap-set-new}
\S_L  =\left\lbrace \mathcal{L} \bm x | \ \bm x \geq \bm 0, \ ( \mathcal{L} \bm x + \bm J) \in \S^{p}_{++} \right\rbrace.
\end{equation}
where $\bm J=\frac{1}{p} \bm 1_{p \times p}$ and $\bm x \geq \bm 0$ means every entry of $\bm x$ is non-negative.
\end{theorems}

As a result of Theorem \ref{Theorem 2}, we introduce the linear operator $\L$ defined in \eqref{operator} and reformulate the optimization \eqref{cost-theta} as
\begin{equation}\label{new_cost}
\min_{\bm w \geq \bm 0} - \log {\det}(\mathcal{L} \bm w + \bm J)+\tr{\Sm \mathcal{L} \bm w } + \sum_{i} h_{\lambda} (w_i).
\end{equation}
We can see that there is only non-negativity constraint in \eqref{new_cost}, which can be handled by simple projection. The variable $\bm w$ contains all the graph weights, thus we transfer the problem of graph learning into the optimization problem in \eqref{new_cost}.

Notice that we remove the constraint $( \mathcal{L} \bm x + \bm J) \in \S^{p}_{++}$ in \eqref{new_cost} compared with the constraint set in \eqref{Lap-set-new}, because any $\bm w$ in the feasible set of \eqref{new_cost} must obey $( \mathcal{L} \bm x + \bm J) \in \S^{p}_{++}$ as shown as follows. Let $F(\bm w) =- \log {\det}(\mathcal{L} \bm w + \bm J)+\tr{\Sm \mathcal{L} \bm w } + \sum_{i} h_{\lambda} (w_i)$. The feasible set of \eqref{new_cost} is $\mathcal{S}_{\bm w} = \{ \bm w \, | \bm w \geq \bm 0, \bm w \in \mathrm{dom}(F) \}$, where $\mathrm{dom}(F)$ denotes the domain of the function $F$. One can verify that 
\begin{equation}
\mathrm{dom}(F)=\left\lbrace \bm w \in \mathbb{R}^{p(p-1)/2} \, | \, \det(\L \bm w + \bm J) \in \mathbb{R}_+ \right\rbrace. \label{F_dom}
\end{equation}
The set $\mathcal{S}_{\bm w}$ can be equivalently written as 
\begin{equation}
\mathcal{S}_{\bm w} = \left\lbrace \bm w \in \mathbb{R}_+^{p(p-1)/2} \, |  (\L \bm w + \bm J) \in \mathcal{S}^p_{++} \right\rbrace, \label{Sw-pd}
\end{equation}
which is due to the reason that $\L \bm w + \bm J$ must be positive semi-definite for any $\bm w \geq \bm 0$. The positive semi-definiteness of $\L \bm w + \bm J$ follows from the facts that $\L \bm w$ is positive semi-definite for any $\bm w \geq \bm 0$ because $\L \bm w$ is a diagonally dominant matrix, and the matrix $\bm J$ is rank one with the nonzero eigenvalue being 1 whose eigenvector is orthogonal to the row and column spaces of $\L \bm w$. Thus, on one hand, the condition $\det(\L \bm w + \bm J) \in \mathbb{R}_+$ in \eqref{F_dom} guarantees that $\L \bm w + \bm J$ is non-singular. The non-singularity and positive semi-definiteness of $\L \bm w + \bm J$ together lead to $(\L \bm w + \bm J) \in \mathcal{S}^p_{++}$ in \eqref{Sw-pd}; On the other hand, the positive definiteness of $(\L \bm w + \bm J)$ can guarantee that $\det(\L \bm w + \bm J) \in \mathbb{R}_+$ in \eqref{F_dom}. Therefore, the feasible set of \eqref{new_cost} is equivalent to \eqref{Sw-pd}.

To solve the problem \eqref{new_cost}, we follow the majorization-minimization framework \citep{SunBabPal2017-MM}, which consists of two steps. In the majorization step, we design a majorized function $f \left(\bm w | \hat{\bm w}^{(k-1)} \right)$ that locally approximates the objective function $F(\bm w)$ at $\hat{\bm w}^{(k-1)}$ satisfying 
\begin{equation}\label{mm-1}
f  \left(\bm w | \hat{\bm w}^{(k-1)} \right) \geq F(\bm w)  \quad \mathrm{and} \quad f \left(\hat{\bm w}^{(k-1)} | \hat{\bm w}^{(k-1)} \right) = F \left(\hat{\bm w}^{(k-1)} \right). 
\end{equation}
Then in the minimization step, we minimize the majorized function $f \left(\bm w | \hat{\bm w}^{(k-1)} \right)$. In this paper, we assume $h_{\lambda}$ is concave (refer to Assumption \ref{assumption 1} for more details about the choices of $h_{\lambda}$). Here we find the majorized function $f \left(\bm w | \hat{\bm w}^{(k-1)} \right)$ by linearizing $\sum_{i} h_{\lambda}(w_{i})$. Set $f_{k} (\bm w) = f \left(\bm w | \hat{\bm w}^{(k-1)} \right)$ to simplify the notation and obtain
\begin{equation}\label{mm-3}
f_{k}(\bm w) = - \log {\det} (\mathcal{L}\bm w + \bm J) +\tr{ \Sm\mathcal{L}\bm w } + \sum_i h'_\lambda \left( \hat{w}_i^{(k-1)} \right) w_i.
\end{equation}
By minimizing $f_{k}(\bm w)$, we establish a sequence $\left\lbrace \hat{\bm{w}}^{(k)} \right\rbrace_{k \geq 1}$ by
\begin{equation}\label{cost}
\hat{\bm{w}}^{(k)} = \arg \min_{\bm w \geq \bm 0} - \log {\det}(\mathcal{L} \bm w + \bm J)+\tr{\Sm \mathcal{L} \bm w } + \sum_{i} z_i^{(k-1)}w_i, 
\end{equation}
where $z_i^{(k-1)} = h'_{\lambda} \left(\hat{w}_i^{(k-1)} \right)$, $i \in [p(p-1)/2]$. We can see $\sum_{i} z_{i}^{(k-1)}w_{i}$ is equivalent to $\sum_{i} z_{i}^{(k-1)} |w_{i}|$ because $\bm w \geq \bm 0$ and $z_{i}^{(k-1)} \geq 0$ by Assumption \ref{assumption 1}. Thus the problem \eqref{cost} can be viewed as a weighted $\ell_1$-norm penalized maximum likelihood estimation under Laplacian constrained Gaussian graphical model. The final estimator $\hat{\bm w}$ is obtained by solving a sequence of optimization problems \eqref{cost}. The iteration procedure is summarized in Algorithm \ref{algo-1}.

To solve the optimization problem in \eqref{cost}, we develop an algorithm based on projected gradient descent with backtracking line search. To obtain $\hat{\bm{w}}^{(k)}$, the algorithm starts with ${\bm w}^{(k)}_{0}$ and then establishes the sequence $\left\lbrace \bm w^{(k)}_{t} \right\rbrace_{t \geq 0}$ by the projected gradient descent as below. In the $t$-th iteration, we update $\bm w^{(k)}_{t}$ by
\begin{equation}\label{pgd}
\bm w^{(k)}_{t}  = \mathcal{P}_+ \left( \bm w^{(k)}_{t-1} - \eta \nabla f_{k} \left(\bm w^{(k)}_{t-1} \right) \right),
\end{equation}
where $\mathcal{P}_+ (a) =\max(a, 0)$ and $\nabla f_{k} \left(\bm w^{(k)}_{t-1} \right) = - \mathcal{L}^{\ast} \left(\mathcal{L} \bm w^{(k)}_{t-1} + \bm J \right)^{-1} + \mathcal{L}^{\ast} \bm S +\bm z^{(k-1)}$. The sequence $\left\lbrace \bm w^{(k)}_{t}\right\rbrace_{t \geq 0}$ will converge to $\hat{\bm{w}}^{(k)}$. Here we set ${\bm w}^{(k)}_{0} = \hat{\bm w}^{(k-1)}$ which is the limit point of the sequence $\left\lbrace \bm w^{(k-1)}_{t}\right\rbrace_{t \geq 0}$. The algorithm is summarized in Algorithm \ref{algo-2}. Note that the users do not need to tune the step size manually because of the backtracking line search.

To establish the theoretical results in Section \ref{theoretical results}, the initial point $\hat{\bm w}^{(0)}$ of Algorithm \ref{algo-1} is chosen such that $\left|\mathrm{supp}^+ \left(\hat{\bm w}^{(0)} \right) \right| \leq s$, where $s$ is the number of nonzero graph weights in the true graph. In other words, $\hat{\bm w}^{(0)}$ has no more than $s$ positive elements. Through the analysis of Algorithm \ref{algo-2} in Section \ref{theoretical results}, we will see that the sequence $\left\lbrace \bm w^{(k)}_{t} \right\rbrace_{t \geq 0}$ can be guaranteed in the feasible set of \eqref{new_cost}, implying that the estimated graph must be connected (refer to the proof of Theorem \ref{theorem 4} for more details). 

Though the proposed algorithm has two loops, we prove in Section \ref{theoretical results} that for the outer loop, only $\lceil 4 \log(4 \alpha c_0^{-1}) \rceil$ iterations in Algorithm \ref{algo-1} are needed to achieve the desired order of estimation error, where $\alpha$ and $c_0$ are two constants, independent of the dimension size $p$ and sample size $n$. For the inner loop, the projected gradient descent in Algorithm \ref{algo-2} enjoys a linear convergence rate. In addition, the proposed method can be extended to estimate other structured matrices such as Hankel matrices with the usage of Hankel linear operator \citep{cai2016robust,7903737}.
 
\begin{algorithm}
	\caption{\textsf{Nonconvex Graph Learning (NGL) }}\label{algo-1} 
	\begin{algorithmic}[1]
		\REQUIRE Sample covariance $\bm S$, $\lambda$, $\hat{\bm w}^{(0)}$; \\
		$k\leftarrow 1$;
		\WHILE {Stopping criteria not met}
		\STATE Update $z^{(k-1)}_i = h'_{\lambda} \left(\hat{w}^{(k-1)}_i \right)$, for $i=1, \ldots, p(p-1)/2$;
		\STATE Update $\hat{\bm w}^{(k)}  = \arg \min_{\bm w \geq \bm 0} - \log \det(\mathcal{L} \bm w + \bm J) + \tr{\bm S \mathcal{L} \bm w } + \sum_{i} z_i^{(k-1)}w_i $;
		
		\STATE $k\leftarrow k+1$;
		\ENDWHILE
		\ENSURE $\hat{\bm w}$.
	\end{algorithmic}
\end{algorithm}

\begin{algorithm}

	\caption{\textsf{ Update $\hat{\bm w}^{(k)} $} }\label{algo-2} 
		\begin{algorithmic}[1]
		\REQUIRE Sample covariance $\bm S$, $\lambda$, $\hat{\bm w}^{(k-1)}$, $\beta$; \\
		${\bm w}^{(k)}_{0} = \hat{\bm w}^{(k-1)}$;\\
		$t\leftarrow 1$;

		\WHILE {Stopping criteria not met}
		
		\STATE Update ${\bm w}^{(k)}_{t}  = \mathcal{P}_+ \left( {\bm w}^{(k)}_{t-1} - \eta \nabla f_{k} \left({\bm w}^{(k)}_{t-1} \right) \right)$;
		\IF { $f_{k} \left({\bm w}^{(k)}_{t}  \right) > f_{k} \left({\bm w}^{(k)}_{t-1} \right) + \left\langle \nabla f_{k} \left( {\bm w}^{(k)}_{t-1} \right), {\bm w}^{(k)}_{t} - {\bm w}^{(k)}_{t-1} \right\rangle + \frac{1}{2 \eta} \norm{{\bm w}^{(k)}_{t} - {\bm w}^{(k)}_{t-1}}^2$}
		\STATE $\eta \leftarrow \beta \eta$;
		\STATE Back to Step 2;
		\ENDIF
		
		\STATE $t\leftarrow t+1$;
		\ENDWHILE
		\ENSURE $\hat{\bm w}^{(k)}$.
	\end{algorithmic}
\end{algorithm}

\section{Theoretical Results} \label{theoretical results}
In this section, we first present the non-asymptotic optimization performance guarantees of Algorithm \ref{algo-1} on both optimization error and statistical error in Section \ref{Sec estimation-error}, and its edge recovery consistency in Section \ref{Sec edge recovery}. The theoretical convergence of Algorithms \ref{algo-2} and \ref{algo-3} is established in Section \ref{Sec convergence}.

We first list the assumptions needed for establishing our theoretical results. We denote the underlying true graph weights by $\bm w^{\star} \in \mathbb{R}_+^{p(p-1)/2}$, which are non-negative. Let $\mathcal{S}^\star= \{ i\in [p(p-1)/2] \, | w^{\star}_i > 0 \}$ be the support set of $\bm{w}^{\star}$ and $s$ be the number of the nonzero weights, i.e., $|\mathcal{S}^\star| = s$. This paper focuses on learning connected graphs, and thus $\bm w^{\star}$ corresponds to the weights of a connected graph, i.e., every pair of vertices in the graph is connected. We impose some mild conditions on the sparsity-promoting function $h_{\lambda}$ in Assumption \ref{assumption 1} and the underlying graph weights $\bm w^\star$ in Assumption \ref{assumption 2}.
\begin{assumptions} \label{assumption 1}
The function $h_{\lambda} : \mathbb{R} \to \mathbb{R}$ satisfies the following conditions:
\begin{enumerate}
\item $h_{\lambda} (0)=0$, and $h'_{\lambda} (x)$ is monotone and Lipschitz continuous for $x \in [0, +\infty)$;
\item There exists a $\gamma > 0$ such that $h'_{\lambda} (x)= 0$ for $x \geq \gamma \lambda$;
\item $h'_{\lambda} (x)= \lambda$ for $x \leq 0$ and $h'_{\lambda} (c \lambda) \geq \lambda/2$, where $c= \left(2+\sqrt{2} \right) \lambda^2_{\max} \left(\L \bm w^{\star} \right)$ is a constant.
\end{enumerate}
\end{assumptions}

\begin{assumptions} \label{assumption 2}
The graph weights $\bm w^{\star}$ represent a connected graph, i.e., $\mathcal{L} \bm w^{\star} \in \mathcal{S}_L$ defined in \eqref{Lap-set}. The minimal nonzero graph weight satisfies $\min_{i\in \mathcal{S}^\star} w_i^{\star} \geq  (c +\gamma) \lambda \gtrsim \lambda$, where $c$ and $\gamma$ are defined in Assumption \ref{assumption 1}. There exists $\tau \geq 1$ such that 
\begin{equation}
1/\tau \leq \lambda_{2} \left(\mathcal{L} \bm w^{\star} \right) \leq \lambda_{\max} \left(\mathcal{L} \bm w^{\star} \right) \leq \tau, \label{tau-def}
\end{equation}
where $\lambda_{2} (\mathcal{L} \bm w^{\star} )$ and $\lambda_{\max} \left(\mathcal{L} \bm w^{\star} \right)$ are the second smallest eigenvalue and maximum eigenvalue of $\mathcal{L} \bm w^\star $, respectively. Note that the smallest eigenvalue of $\mathcal{L} \bm w^\star$ is $0$.
\end{assumptions}

\begin{remarks}
In Assumption \ref{assumption 1}, the conditions on $h_{\lambda}(x)$ are mainly made over $x \in [0, +\infty)$ because of the nonnegativity constraint in the optimization \eqref{cost}. In Assumption \ref{assumption 1}, the first two conditions are made to promote sparsity and unbiasedness \citep{loh2017support}, and hold for a variety of nonconvex sparsity-promoting functions including the popular choices SCAD \citep{fan2001variable} and MCP \citep{zhang2010nearly}. The two conditions also ensure that the derivative of $h_{\lambda}(x)$ is not constant to avoid the issue in Theorem \ref{Theorem 1}. In the third condition, we specify $h'_{\lambda} (x)$ for $x \leq 0$ only for theoretical analysis. The condition $h'_{\lambda}(c \lambda) \geq \lambda/2$ can always hold by tuning parameters due to the conditions $h'_{\lambda}(0) = \lambda$ and $h'_{\lambda}(\gamma \lambda) = 0$. Therefore, it is flexible to design a sparsity-promoting function satisfying the conditions stated in \ref{assumption 1}. In Assumption \ref{assumption 2}, the conditions on the true graph weights $\bm w^\star$ are mild. In our theorems, the regularization parameter $\lambda$ is taken with the order $\sqrt{ \log p /n}$ that could be very small when the sample size $n$ increases. The assumptions on the minimal magnitude of signals are often employed in the analysis of nonconvex optimization \citep{fan2018lamm,wang2016precision}. The assumptions on the minimum nonzero eigenvalue are frequently made \citep{chen2018covariate,lyu2019tensor,wang2016precision} to establish a local contraction region.
\end{remarks}

\subsection{Analysis of Estimation Error} \label{Sec estimation-error}
We will characterize the estimation error of our proposed nonconvex optimization method. In the following theorem, proved in Appendix \ref{sec-estimation-error}, the choice of the regularization parameter $\lambda$ is set according to a user-defined parameter $\alpha>2$. A larger $\alpha$ yields a larger probability with which the claims hold, but also leads to a more stringent requirement on the number of samples.
\begin{theorems} \label{theorem 3}
Under Assumptions \ref{assumption 1} and \ref{assumption 2}, take the regularization parameter $\lambda = \sqrt{4 \alpha c_0^{-1} \log p /n} $ for some $\alpha >2$. If the sample size $n$ is lower bounded by 
\begin{equation}
n\geq \max \left(94 \alpha c_0^{-1} \lambda_{\max}^2 \left(\L \bm w^{\star} \right) s \log p, 8 \alpha \log p \right), \nonumber
\end{equation} 
then with probability at least $1-1/p^{\alpha-2}$, the sequence $\hat{\bm w}^{(k)}$ returned by Algorithm \ref{algo-1} satisfies
\begin{equation}
\norm{ \hat{\bm w}^{(k)} - \bm w^{\star} }   \leq \underbrace{ 2(3\sqrt{2}+4)  \lambda_{\max}^2 \left(\L \bm w^{\star} \right) \sqrt{ \alpha c_0^{-1} s\log p /n} }_{\mathrm{Statistical \; error}}
+ \underbrace{ \left( \frac{3}{2+\sqrt{2}} \right )^{k} \norm{ \hat{\bm w}^{(0)} - \bm w^{\star}  } }_{\mathrm{Optimization \;  error}}, \nonumber
\end{equation}
where $c_0 = 1/\left( 8 \norm{\mathcal{L}^{\ast}(\mathcal{L} \bm w^{\star} + \bm J)^{-1}}_{\max}^2\right)$ is a constant.
\end{theorems}

Theorem \ref{theorem 3} holds with overwhelming probability. According to Theorem \ref{theorem 3}, the estimation error between the estimated and underlying true graph weights is bounded by two terms, i.e., the optimization error and statistical error. The optimization error, $\left( \frac{3}{2+\sqrt{2}} \right)^{k} \norm{\hat{\bm w}^{(0)} - \bm w^{\star} }$, decays to zero at a linear rate with respect to the iteration number $k$. The statistical error depends on the sample size $n$, and a large sample size will lead to a small statistical error. We can see the statistical error is independent of $k$, and thus will not decrease during iterations in the algorithm. Therefore, the estimation error of the output of the algorithm is mainly from the statistical error. 


\begin{corollaries} \label{corollary}
Under the same assumptions and conditions as stated in Theorem \ref{theorem 3}, the sequence $\hat{\bm w}^{(k)}$ returned by Algorithm \ref{algo-1} satisfies
\begin{equation}
\norm{ \L \hat{\bm w}^{(k)} - \L \bm w^{\star} }_{\mathrm{F}}  \leq \underbrace{ 4(2\sqrt{2} + 3)  \lambda_{\max}^2 \left(\L \bm w^{\star} \right) \sqrt{ \alpha c_0^{-1} s\log p /n} }_{\mathrm{Statistical \; error}}
+ \underbrace{ \left( \frac{3}{2+\sqrt{2}} \right )^{k}  \norm{ \L \hat{\bm w}^{(0)} - \L \bm w^{\star} }_{\mathrm{F}}}_{\mathrm{Optimization \; error}}, \nonumber
\end{equation}
with probability at least $1-1/p^{\alpha-2}$. If $k \geq \lceil 4 \log \left(4 \alpha c_0^{-1} \right) \rceil$, then the estimation error is dominated by the statistical error and we further obtain
\begin{equation}
\norm{ \L \hat{\bm w}^{(k)} - \L \bm w^{\star} }_{\mathrm{F}} \lesssim \sqrt{s\log p /n},  \nonumber
\end{equation} 
where $c_0 = 1/\left( 8 \norm{\mathcal{L}^{\ast}(\mathcal{L} \bm w^{\star} + \bm J)^{-1}}_{\max}^2\right)$ is a constant.
\end{corollaries}

Corollary \ref{corollary} presents the estimation error between the estimated and underlying true precision matrices in Laplacian constrained GGM. Similar to Theorem \ref{theorem 3}, the optimization error decays to zero at a linear rate. The order of statistical error is upper bounded by $\sqrt{s\log p /n}$, which is consistent with the one obtained for Gaussian graphical models \citep{cai2016estimating,ravikumar2011high,rothman2008sparse}. It is interesting to explore if $\sqrt{s\log p /n}$ is optimal for estimating sparse precision matrices with Laplacian constraints in future work. Furthermore, the proposed estimation method can achieve this order by solving only $\lceil 4 \log(4 \alpha c_0^{-1})\rceil$ sub-problems. Notice that $4 \log \left(4 \alpha c_0^{-1} \right)$ is independent of the dimension size $p$ and sample size $n$.


\subsection{Analysis of Edge Recovery} \label{Sec edge recovery}
We show in Theorem \ref{theorem 3} that the proposed estimator is consistent in terms of parameter estimation. However, it does not mean that the estimator necessarily consistently select the correct edges. In this subsection, we present Theorem \ref{oracle}, proved in Appendix \ref{sec-prof-oracle}, showing that the proposed estimator can recover the edges correctly with a high probability.

An oracle knows the true support set, and the oracle estimator is defined as
\begin{equation} \label{oracle-estimator}
\bm w^{\mathrm{oracle}}: = \arg \min_{\bm w \geq \bm 0} - \log \det \left(\mathcal{L} \bm w + \bm J \right) + \tr{ \bm S \mathcal{L} \bm w }, \quad \mathrm{subject\ to} \ w_i =0, \ \forall i\in \left\lbrace \mathcal{S}^\star \right\rbrace^c,
\end{equation}
where $\mathcal{S}^\star$ is the true support set. Note that the oracle estimator \eqref{oracle-estimator} exists and is unique with probability one (please refer to the proof of Theorem \ref{oracle} for more details).

We define some constants which are only related to the underlying true graph weights $\bm w^\star$. Let 
\begin{equation}\label{constant_K}
K_1 = \normI{ \left( \mathcal{L} \bm w^\star \right)^\dagger }_\infty,  \ K_2 = \normI{\left( \bm H_{\mathcal{S}^\star \mathcal{S}^\star} \right)^{-1} }_\infty \  \mathrm{and} \quad K_3 = \normI{\bm H_{\left\lbrace \mathcal{S}^\star \right\rbrace^c \mathcal{S}^\star} \left( \bm H_{\mathcal{S}^\star \mathcal{S}^\star} \right)^{-1} }_\infty,
\end{equation}
where $\bm H = \left[ \mathcal{L}^\ast \bm B_1, \mathcal{L}^\ast \bm B_2, \ldots, \mathcal{L}^\ast \bm B_{p(p-1)/2}  \right]$,
and $\bm B_i = \left(  \mathcal{L} \bm w^\star \right)^\dagger \mathcal{L} \bm e_i \left(  \mathcal{L} \bm w^\star\right)^\dagger$ for $i \in [p(p-1)/2]$, with $\bm e_i$ the $i$-th canonical basis of $\mathbb{R}^{p(p-1)/2}$. Let $d$ be the maximum node degree of the graph, i.e., the maximum number of edges for each node.

\begin{theorems}\label{oracle}
Under Assumptions \ref{assumption 1} and \ref{assumption 2}, take the regularization parameter $\lambda = \sqrt{4 \alpha c_0^{-1} \log p /n}$. If the sample size $n$ is lower bounded by
\begin{equation*}
n \geq \max \left( 576 \alpha c_0^{-1} c_H^2 \log p \max \left(16 K_1^6 K_2^4 d^4, K_1^2 K_2^2 d^2 \right), 2 \alpha \log p, 32 \alpha c_H^2 \log p \right),
\end{equation*}
and the initial point $\hat{\bm w}^{(0)}$ satisfies $\norm{\hat{\bm w}^{(0)} - \bm w^\star}_{\max} \leq c \lambda$, then with probability at least $1- \left(  p(p-1) - 2s \right) p^{- \frac{1}{4}\alpha } - 2s \cdot p^{-4 c_H^2 \alpha }$, the sequence $\hat{\bm w}^{(k)}$ returned by Algorithm \ref{algo-1} satisfies
\begin{equation*}
\hat{\bm w}^{(k)} = \bm w^{\mathrm{oracle}} \quad \mathrm{and} \quad \mathrm{supp} \left( \hat{\bm w}^{(k)} \right) = \mathrm{supp} \left( \bm w^\star \right), \quad \forall k \geq 1,
\end{equation*}
where $c_H =\min \left( \frac{c}{2 K_2}, \ \frac{1}{4(2 K_3 + 1)} \right)$, and $c$ is a constant defined in \ref{assumption 1}.
\end{theorems}

The statement in Theorem \ref{oracle} holds with a high probability for a reasonably large $\alpha$. Theorem \ref{oracle} shows that the oracle estimator $\bm w^{\mathrm{oracle}}$ defined in \eqref{oracle-estimator} is a limit point of the sequence generated by Algorithm \ref{algo-1}, and Algorithm \ref{algo-1} will converge to $\bm w^{\mathrm{oracle}}$. Furthermore, we can see that the proposed estimator can recover the edges correctly with a high probability. Note that both $\hat{\bm w}^{(k)}$ and $\bm w^\star$ are non-negative. Thus the proposed estimator also enjoys the sign consistency. In addition, the condition on the initial point that $\norm{\hat{\bm w}^{(0)} - \bm w^\star}_{\max} \leq c \lambda$ could be always met if $\hat{\bm w}^{(0)}$ is set by some consistent estimator with sufficient samples.

\subsection{Analysis of Algorithm Convergence} \label{Sec convergence}
In this subsection, we establish the theoretical convergence of Algorithms \ref{algo-2} for solving each sub-problem. 

In the theorem statement, the parameter $\delta >1$ is  a user-defined parameter. A larger $\delta$ leads to a faster convergence rate, but a larger number of samples are required.
\begin{theorems}\label{theorem 4}
Under Assumptions \ref{assumption 1} and \ref{assumption 2}, take the regularization parameter $\lambda = \sqrt{4 \alpha c_0^{-1} \log p /n}$ for some $\alpha >2$. If the sample size $n$ satisfies
\begin{equation}
n\geq \max \left(840 \alpha c_0^{-1} \frac{(\delta \tau^2 +1)^4}{\delta^2 \tau^2} s p \log p, 8 \alpha \log p\right), \nonumber
\end{equation}
then with probability at least $1-1/p^{\alpha-2}$, the sequence $\left\lbrace {\bm w}^{(k)}_{t} \right\rbrace_{t \geq 1}$ returned by Algorithm \ref{algo-2} obeys
\begin{equation}
\norm{ {\bm w}^{(k)}_{t} -\hat{\bm w}^{(k)} }^2 \leq \rho^t \norm{ {\bm w}^{(k)}_{0} -\hat{\bm w}^{(k)} }^2, \quad \forall \ k \geq 2, \nonumber
\end{equation}
where $\rho = 1- \frac{\beta (1-\delta^{-1})^2}{p \tau^4 (1+\delta^{-1})^2} < 1$ with $\delta >1$ and $\beta \in (0, 1)$, and $c_0 = 1/\left( 8 \norm{\mathcal{L}^{\ast}(\mathcal{L} \bm w^{\star} + \bm J)^{-1}}_{\max}^2 \right)$ is a constant.
\end{theorems}

Theorem \ref{theorem 4}, proved in Appendix \ref{sec-prof-gradient-converg}, shows that the designed projected gradient descent algorithm enjoys a linear convergence rate in solving sub-problems \eqref{cost}. Note that when $k=1$, Algorithm \ref{algo-2} may not enjoy a linear convergence rate because we only set a very mild condition on the initial point $\hat{\bm w}^{(0)}$ that $ \left|\mathrm{supp}^+ \left(\hat{\bm w}^{(0)} \right) \right| \leq s$. Thus $\hat{\bm w}^{(0)}$ may not be in the local region $\mathcal{B} \left(\bm w^{\star}; \frac{1}{\sqrt{2p}\delta \tau} \right)= \left\lbrace \bm w | \bm w \in \mathbb{B} \left(\bm w^{\star};  \frac{1}{\sqrt{2p}\delta \tau} \right) \cap \S_{\bm w} \right\rbrace$, where $f_k(\bm w)$ is $\mu$-strongly convex and $L$-smooth with $\mu=\frac{2}{ \left(1+\delta^{-1} \right)^2\tau^2}$ and $L=\frac{2p \tau^2}{ \left(1-\delta^{-1} \right)^2}$ according to Lemma \ref{lem12}. One may impose more conditions on the initial point such that $\hat{\bm w}^{(0)} \in \mathcal{B} \left(\bm w^{\star}; \frac{1}{\sqrt{2p}\delta \tau} \right)$, then Algorithm \ref{algo-2} also has a linear convergence rate for $k=1$. Here $\mathbb{B} \left(\bm w^{\star};r \right) = \left\lbrace \bm w \, | \norm{ \bm w - \bm w^{\star}} \leq r \right\rbrace$, and $\tau$ is defined in Assumption \ref{assumption 2}.

It is observed that the required sample size in Theorem \ref{theorem 4} is larger than the one in Theorem \ref{theorem 3}. Using the sample size in Theorem \ref{theorem 3}, one can still establish a local contraction region $\mathcal{B}_{f_k} \left(\bm w^{\star}; \tau, \bm w^{(k)}_0 \right)=\left\lbrace \bm w \, | \bm w \in \mathbb{B}_{\bm M} \left(\bm w^{\star}; \tau \right) \cap \mathcal{C}_{f_k} \left(\bm w^{(k)}_0 \right) \cap \mathcal{S}_{\bm w} \right\rbrace$ by introducing the lower level set $\mathcal{C}_{f_k} \left(\bm w^{(k)}_0 \right) = \left\lbrace \bm w \, | f_k (\bm w) \leq f_k \left(\bm w^{(k)}_0 \right) \right\rbrace$ such that Algorithm \ref{algo-2} also has a linear convergence. However, the region $\mathcal{B}_{f_k} \left(\bm w^{\star}; \tau, \bm w^{(k)}_0 \right)$ is much larger than $\mathcal{B} \left(\bm w^{\star}; \frac{1}{\sqrt{2p}\delta \tau} \right)$, and the smoothness parameter $L$ of $f_k(\bm w)$ may be very large in $\mathcal{B}_{f_k} \left(\bm w^{\star}; \tau, \bm w^{(k)}_0 \right)$. Here $\mathbb{B}_{\bm M} \left(\bm w^{\star};r \right) = \left\lbrace \bm w \, | \norm{ \bm w - \bm w^{\star}}_{\bm M} \leq r \right\rbrace$, where $\bm M$ is defined in Lemma \ref{lem11}.

\vspace{0.25cm}

\section{Extension to Disconnected Graph Learning}\label{Sec k-component}
We focus on the problem of learning sparse connected graphs in Sections \ref{proposed} and \ref{theoretical results}. In this section, we extend our method to learn sparse disconnected graphs.

\subsection{Algorithm}
We consider the disconnected graph consisting of $k$ components. Following from the spectral graph theory, we obtain that the rank of Laplacian matrix is equal to the number of components in the graph. Then we can formulate the sparse $k$-component graph learning problem as follows:
\begin{equation}\label{k-comp}
\min_{\bm \Theta} -\log {\det}^\star (\bm \Theta) + \tr{\bm \Theta \bm S} +  \sum_{i > j} h_{\lambda}(\Theta_{ij}), \quad \mathrm{subject \ to} \ \bm \Theta  \in \mathcal{S}_{L_k},
\end{equation}
where $\mathcal{S}_{L_k}$ is the set of all the Laplacian matrices with rank $p-k$, defined by
\begin{equation}\label{Lap-set-k}
\S_{L_k}  = \lbrace \bm\Theta \in \S^{p}_+| \, \Theta_{ij} =\Theta_{ji} \leq 0, \, \forall \ i\neq j,\, \bm\Theta \cdot \bm 1 =\bm 0,\ \mathrm{rank} (\bm \Theta) = p-k \rbrace.
\end{equation}

Note that we have to impose the rank constraint in \eqref{k-comp}, because the pseudo determinant function ${\det}^\star (\cdot)$ is not continuous between sets of matrices of different ranks. It is worth mentioning that the GGM or the GGM with generalized Laplacian constraints does not need the rank constraint in learning multiple component graphs \citep{hsieh2012divide,pavez2018learning}, because the precision matrices can keep positive definite for disconnected graphs.

The challenge in solving \eqref{k-comp} is that the Laplacian constraints and rank constraint in \eqref{Lap-set-k} are simultaneously imposed on one variable. To decouple those constraints in \eqref{Lap-set-k}, we introduce an auxiliary variable $\bm w$, and reformulate \eqref{k-comp} as
\begin{equation}\label{k-comp-new}
\min_{\bm \Theta \succeq \bm 0, \bm w \geq \bm 0} -\log {\det}^\star (\bm \Theta) + \tr{\mathcal{L} \bm w \bm S} +  \sum_{i} h_{\lambda}(w_i), \ \mathrm{subject \ to}  \ \bm \Theta = \mathcal{L} \bm w, \ \mathrm{rank} (\bm \Theta) = p-k.
\end{equation}
Now we can see that the Laplacian constraints are imposed on $\mathcal{L} \bm w$, and the rank constraint is imposed on $\bm \Theta$. To solve the optimization \eqref{k-comp-new}, we propose an algorithm based on ADMM. 
The augmented Laplagian function of \eqref{k-comp-new} is
\begin{equation}\label{Laglagian-k-comp}
L_{\rho} (\bm \Theta, \bm w, \bm G) = -\log {\det}^\star (\bm \Theta) + \tr{\mathcal{L} \bm w \bm S} +  \sum_{i} h_{\lambda}(w_i) + \left\langle \bm G, \bm \Theta - \mathcal{L} \bm w \right\rangle + \frac{\rho}{2} \norm{\bm \Theta - \mathcal{L} \bm w }_{\mathrm{F}}^2,
\end{equation}
where $\bm G$ is the dual variable. We successively update each variable as follows.

To update $\bm \Theta$, we have the following sub-problem,
\begin{equation}\label{sub-theta}
\hat{\bm \Theta}  = \arg \min_{\bm \Theta \succeq \bm 0} -\log {\det}^\star (\bm \Theta) + \left\langle \bm G, \bm \Theta  \right\rangle + \frac{\rho}{2} \norm{\bm \Theta - \mathcal{L} \bm w }_{\mathrm{F}}^2, \quad \mathrm{subject \ to} \quad \mathrm{rank} (\bm \Theta) = p-k.
\end{equation}
The closed-form solution of \eqref{sub-theta} is 
\begin{equation} \label{Theta-update}
\hat{\bm \Theta} = \frac{1}{2} \bm U \left( \bm \Lambda + \sqrt{\bm {\Lambda}^2 + \frac{4}{\rho} \bm{I}}\right) \bm U^\top,
\end{equation}
where $\bm U \bm {\Lambda}\bm{U}^\top$ is the eigenvalue decomposition of $\mathcal{L}\bm{w} - \frac{1}{\rho}\bm{G}$,
and $\bm \Lambda$ is a diagonal matrix containing the largest $p - k$ eigenvalues on the diagonal and
$\bm{U} \in \mathbb{R}^{p\times (p-k)}$ contains the corresponding eigenvectors.

To update $\bm w$, we have the following sub-problem,
\begin{equation}\label{subpro-w-kcomp}
\min_{\bm w \geq \bm 0}  \tr{\mathcal{L} \bm w \bm S} +  \sum_{i} h_{\lambda}(w_i) + \left\langle \bm G, \bm \Theta - \mathcal{L} \bm w \right\rangle + \frac{\rho}{2} \norm{ \mathcal{L} \bm w - \bm \Theta }_{\mathrm{F}}^2.
\end{equation}
To solve the problem \eqref{subpro-w-kcomp}, we follow the majorization-minimization framework. In the $(t+1)$-th iteration, we minimize the majorized function at $\bm w_{t}$ and get
\begin{equation}\label{sub_w_majorized}
\bm w_{t+1} = \arg \min_{\bm w \geq \bm 0}  \left\langle \bm a_t, \bm w \right\rangle + \frac{L}{2} \norm{\bm w - \bm w_t}_2^2,
\end{equation}
where $\bm a_t = \bm z_t + \mathcal{L}^\ast \left( \bm S + \rho \mathcal{L} \bm w_t - \rho \bm \Theta - \bm G\right)$ with $[\bm z_t]_i = h'_{\lambda} ([ \bm w_{t}]_i)$, $i \in [p(p-1)/2]$, and $L = 2p$ following from Lemma 5.3. The closed-form solution of \eqref{sub_w_majorized} is
\begin{equation}\label{updating_sub_w}
\bm w_{t+1} = \mathcal{P}_+ \Big( \bm w_t - \frac{1}{L} \bm a_t \Big).
\end{equation} 
We can establish the sequence $\left\lbrace \bm w_t \right\rbrace_{t \geq 1}$, and set its limit point as $\hat{\bm w}$. 

The dual variable is updated by
\begin{equation}\label{G-update}
\hat{\bm G }= \bm G  + \rho \left( \bm \Theta - \mathcal{L} \bm w  \right).
\end{equation}
The algorithm is summarized in Algorithm \ref{algo-3}. 


\vspace{-0.15cm}

\begin{algorithm}
	\caption{\textsf{Disconnected Graph Learning}}\label{algo-3}
	\begin{algorithmic}[1]
		\REQUIRE Sample covariance $\bm S$, $\lambda$, $\hat{\bm w}^{(0)}$; \\
		$l \leftarrow 1$;
		\WHILE {Stopping criteria not met}
		\STATE Update $\bm \Theta^{(l)}$ by \eqref{Theta-update};
		\STATE Update $\bm w^{(l)} $ by iterating \eqref{updating_sub_w};		
		\STATE Update $\bm G^{(l)} $ by \eqref{G-update};
		\STATE $l \leftarrow l+1$;
		\ENDWHILE
		\ENSURE $\hat{\bm \Theta}$, $\hat{\bm w}$.
	\end{algorithmic}
\end{algorithm}

	\vspace{-0.55cm}

\subsection{Convergence}

\vspace{-0.1cm}

We present the theoretical convergence of Algorithm \ref{algo-3} in Theorem \ref{k-component-convergence}, which is proved in Appendix \ref{sec-prof-k-comp}. 
\begin{theorems}\label{k-component-convergence}
Suppose $h_\lambda$ satisfies Assumption \ref{assumption 1} and has $L_h$-Lipschitz continuous gradient. If the parameter $\rho$ obeys
\begin{equation*}
\rho \geq \max_l \left( L_h + \frac{2\norm{\bm G^{(l+1)} - \bm G^{(l)}}_{\mathrm{F}}}{\norm{ \mathcal{L} \bm w^{(l+1)} - \mathcal{L} \bm w^{(l)}}_{\mathrm{F}}} \right),
\end{equation*}
then Algorithm~\ref{algo-3} will subsequently converges, that is, the sequence $\left\lbrace \left( \bm{\Theta}^{(l)}, \bm w^{(l)}, \bm G^{(l)} \right)\right\rbrace$ generated by Algorithm~\ref{algo-3} has at least one limit point, and each limit point is a stationary point of \eqref{Laglagian-k-comp}.
\end{theorems} 
The optimization problem \eqref{k-comp} is nonconvex because of the nonconvex sparsity penalty and the rank constraint. Theorem \ref{k-component-convergence} shows that Algorithm \ref{algo-3} will subsequently converge to a stationary point of \eqref{Laglagian-k-comp} with a sufficiently large $\rho$. In practice, we could iteratively increase the $\rho$ to a decently large value and treat this procedure as an initialization as adopted in \citep{ying2018vandermonde}.

\vspace{-0.4cm}

\section{Experimental Results} \label{experimental section}

\vspace{-0.25cm}

In this section, we conduct numerical experiments on both synthetic data and real-world data sets to verify the performance of the proposed method. The compared methods include the state-of-the-art GLE-ADMM algorithm \citep{zhao2019optimization} and the baseline projected gradient descent with $\ell_1$-norm. 


We employ the performance measures: Relative Error (\textsf{RE}) and F-score (\textsf{FS}), which are defined as follows:
\vspace{-0.25cm}
\begin{equation}
\textsf{RE}=\frac{ \big\| \hat{\bm \Theta}-\bm \Theta^{\star} \big\|_{\mathrm{F}}}{\norm{\bm \Theta^{\star}}_{\mathrm{F}}},\quad \mathrm{and} \quad  \textsf{FS}=\frac{2\textsf{tp}}{2\textsf{tp}+\textsf{fp}+\textsf{fn}}, \label{measure}
\end{equation}
where $\hat{\bm \Theta}=\mathcal{L}\hat{\bm w}$ and $\bm \Theta^{\star} = \mathcal{L} \bm w^\star$ denote the estimated and true precision matrices, respectively.

The true positive number is denoted as \textsf{tp}, i.e., the case that there is an actual edge and the algorithm detects it, the false positive is denoted as \textsf{fp}, i.e., the case that there is no actual edge but algorithm detects one, and the false negative is denoted as \textsf{fn}, i.e., the case that the algorithm failed to detect an actual edge. The F-score takes values in $[0, 1]$, where $1$ indicates perfect structure recovery. For our method, we test two nonconvex penalties, MCP and SCAD, defined respectively by
\begin{equation}
h'_{\textsf{MCP},\lambda}(x)=\left\{
\begin{array}{lcl}
\lambda - \frac{x}{\gamma}  &  & x \in [0, \gamma \lambda],\\
 0  & & x \in [ \gamma \lambda, +\infty),
\end{array}
\right.
\text{and} \quad h'_{\textsf{SCAD},\lambda}(x)=\left\{
\begin{array}{lcl}
\ \lambda & & x \in [0, \lambda], \\
\dfrac{(\gamma \lambda -x )}{\gamma -1}  &  & x \in [ \lambda, \gamma \lambda],\\
 0  & & x \in [ \gamma \lambda, +\infty),
\end{array}
\right. \nonumber
\end{equation}
where we define $h'_{\textsf{MCP},\lambda}(x)$ and $h'_{\textsf{SCAD},\lambda}(x)$ only for $x \geq 0$ because of the nonnegativity constraint in \eqref{cost}. We set $\gamma$ equal to $1.01$ and $2.01$ in $h'_{\textsf{MCP},\lambda}(x)$ and $h'_{\textsf{SCAD},\lambda}(x)$ in all the experiments, respectively. Note that the proposed method is not limited to using MCP and SCAD regularizers, and one can explore other sparsity-inducing functions \citep{bach2012optimization} for interest.

 \begin{figure}[htb]
    \captionsetup[subfigure]{justification=centering}
    \centering
    \begin{subfigure}[t]{0.32\textwidth}
        \centering
        \includegraphics[scale=.25]{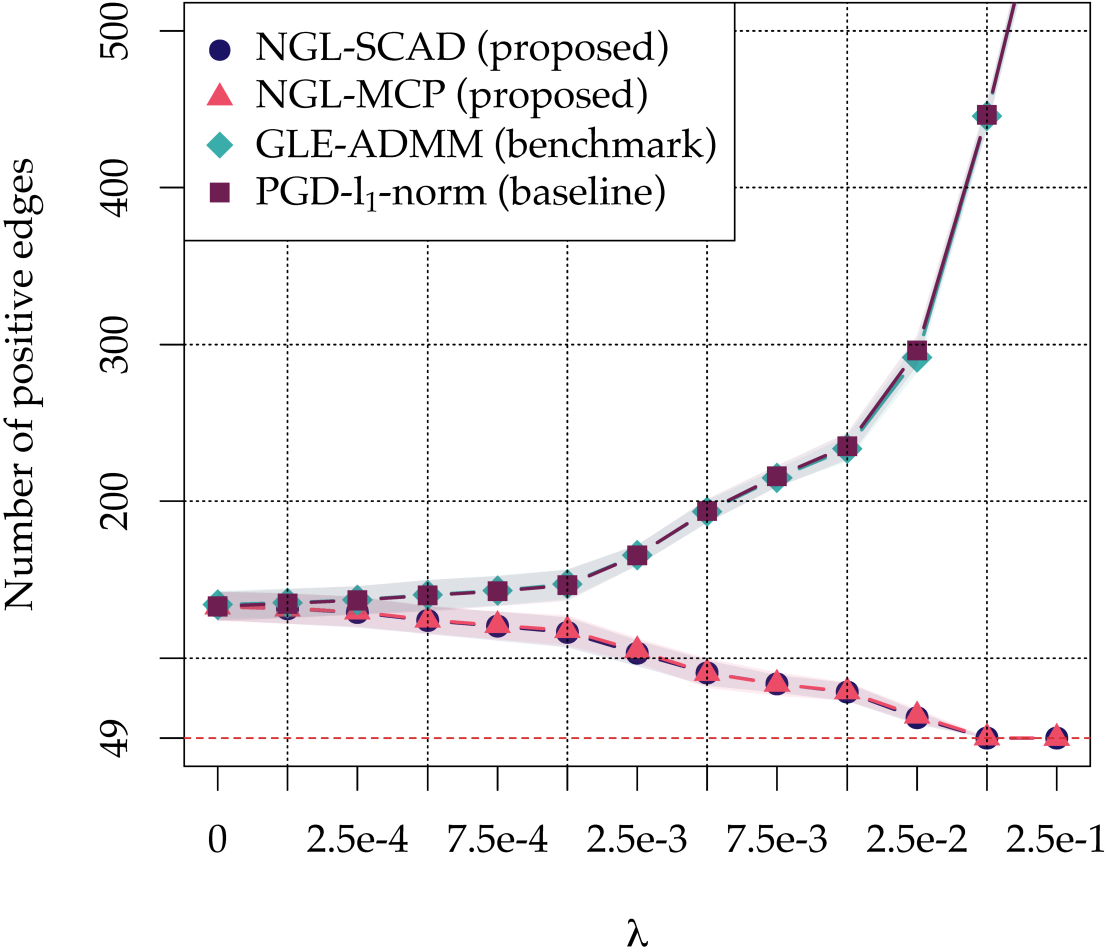}
        \caption{}
    \end{subfigure}%
   ~
    \begin{subfigure}[t]{0.32\textwidth}
        \centering
        \includegraphics[scale=.25]{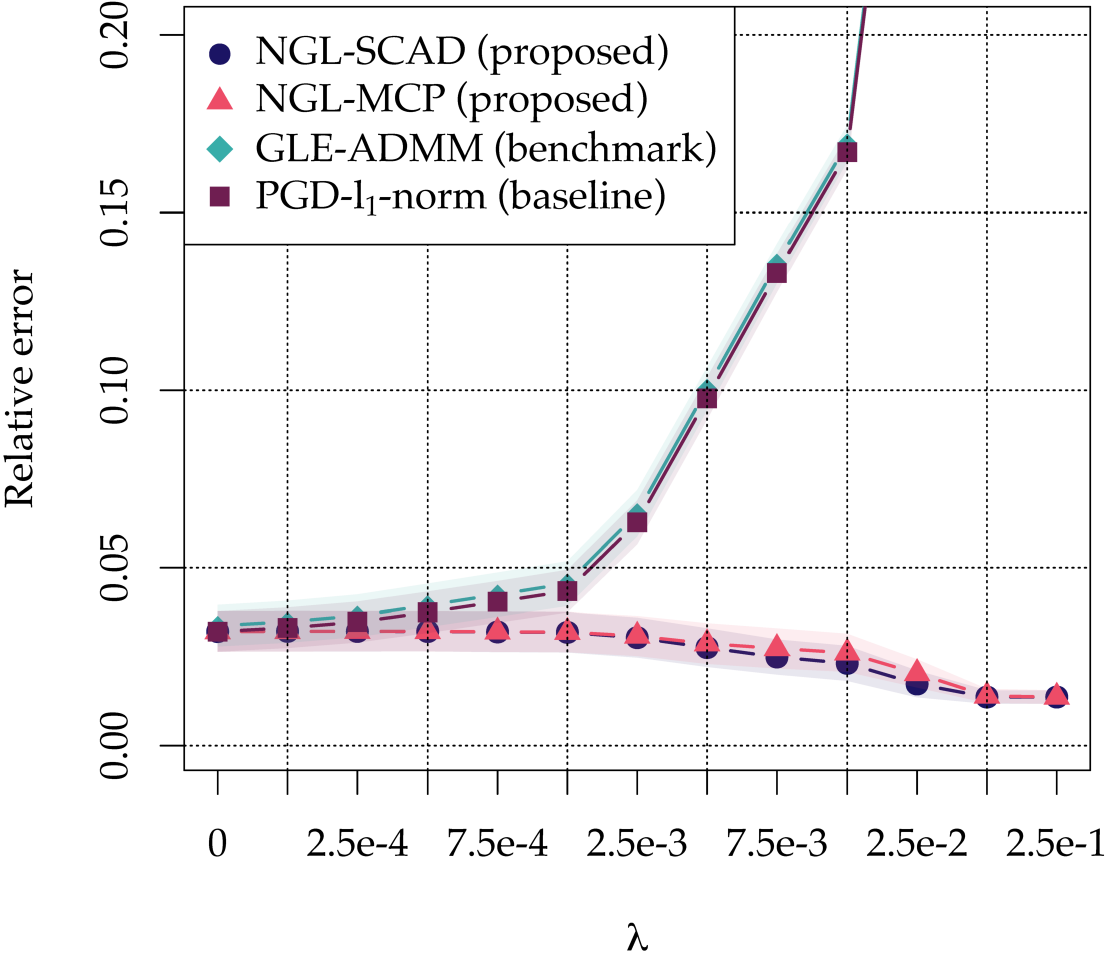}
        \caption{}
    \end{subfigure}%
    ~
    \begin{subfigure}[t]{0.32\textwidth}
        \centering
        \includegraphics[scale=.25]{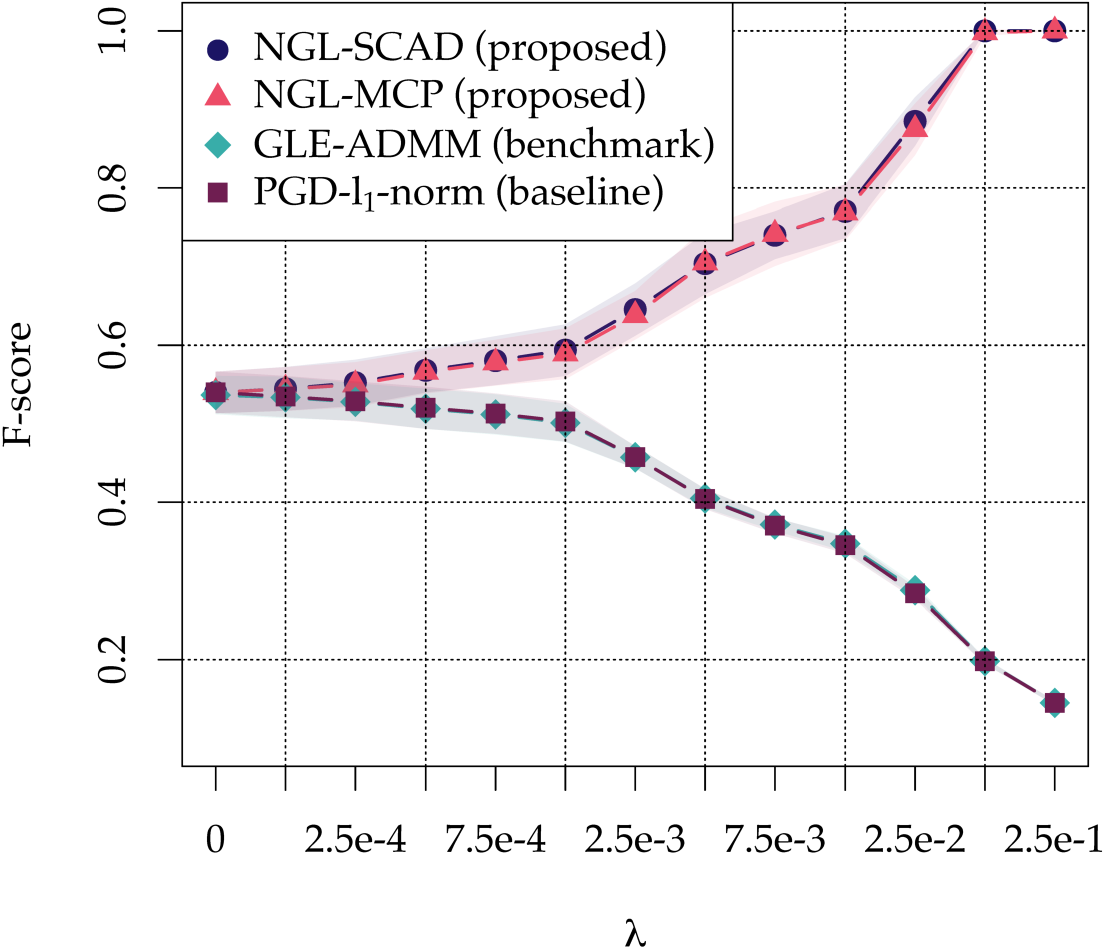}
        \caption{}
    \end{subfigure}%
     \vspace{-0.25cm}
    \caption{Performance measures (a) Number of positive edges, (b) Relative error and (c) F-score as a function of regularization parameter $\lambda$ in learning the Barabasi-Albert graph. The true number of positive edges in (a) is $49$ and the sample size ratio is $n / p = 100$.}
    \label{fig:sparsity-lambda}
    \vspace{-0.2cm}
\end{figure}

\vspace{-0.3cm}

\begin{figure}[!htb]
    \captionsetup[subfigure]{justification=centering}
    \centering
        \begin{subfigure}[t]{0.32\textwidth}
        \centering
        \includegraphics[scale=.29]{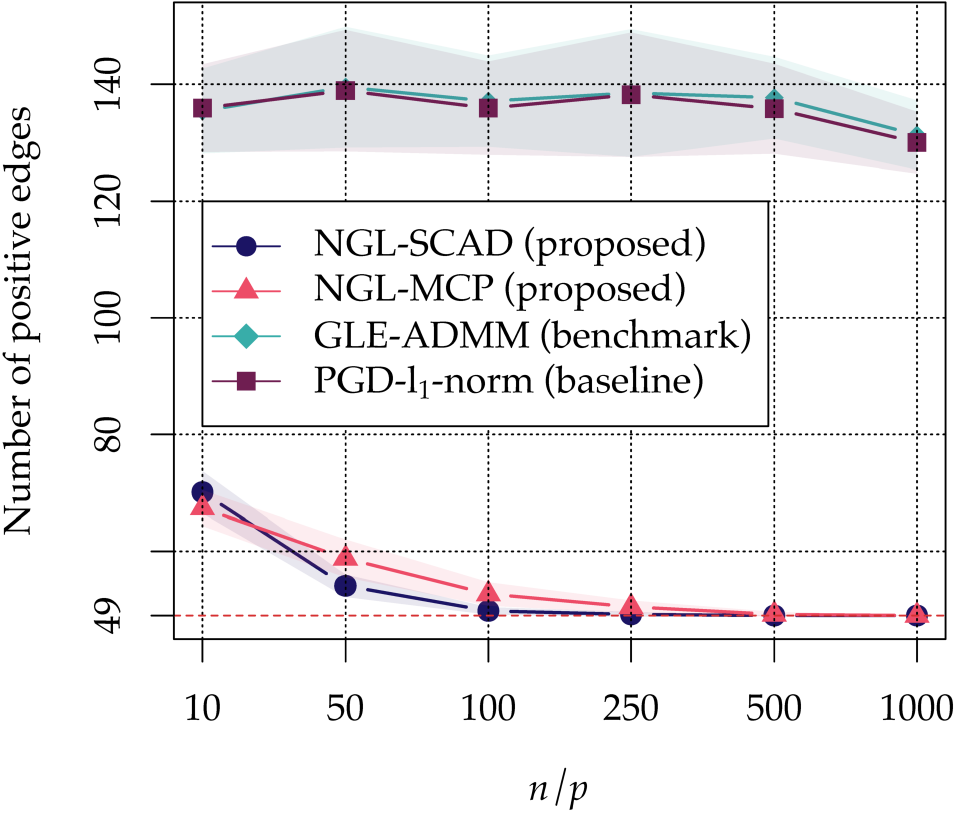}
        \caption{}
    \end{subfigure}%
    ~
    \begin{subfigure}[t]{0.32\textwidth}
        \centering
        \includegraphics[scale=.29]{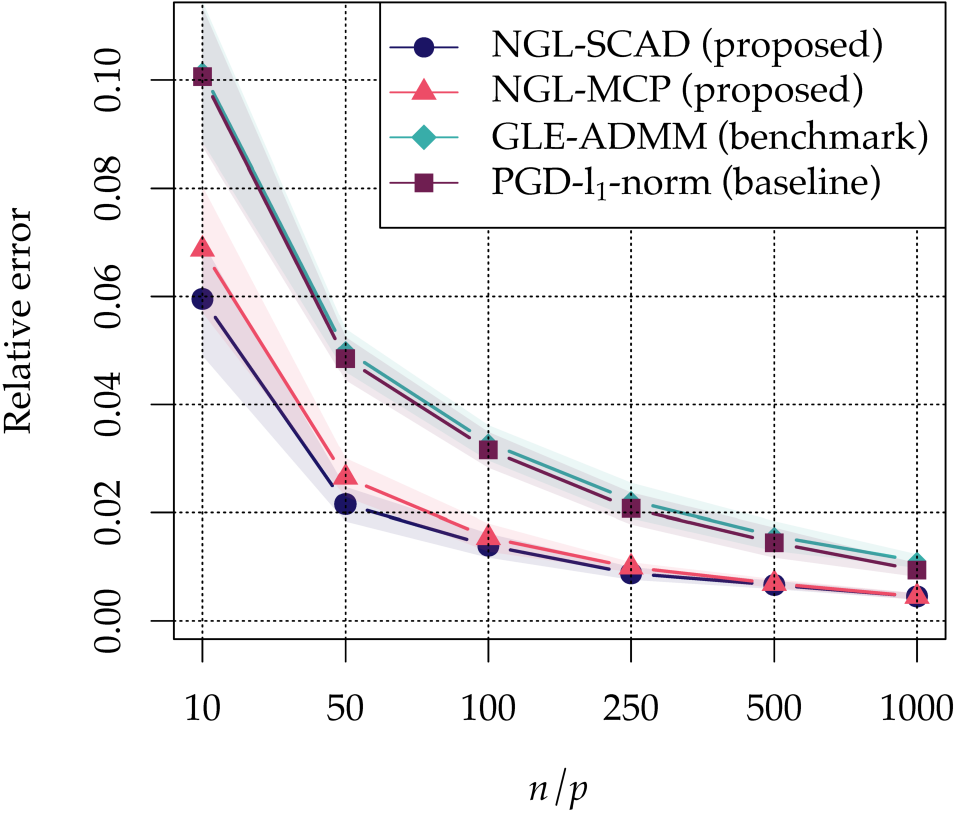}
        \caption{}
    \end{subfigure}%
   ~
    \begin{subfigure}[t]{0.32\textwidth}
        \centering
        \includegraphics[scale=.29]{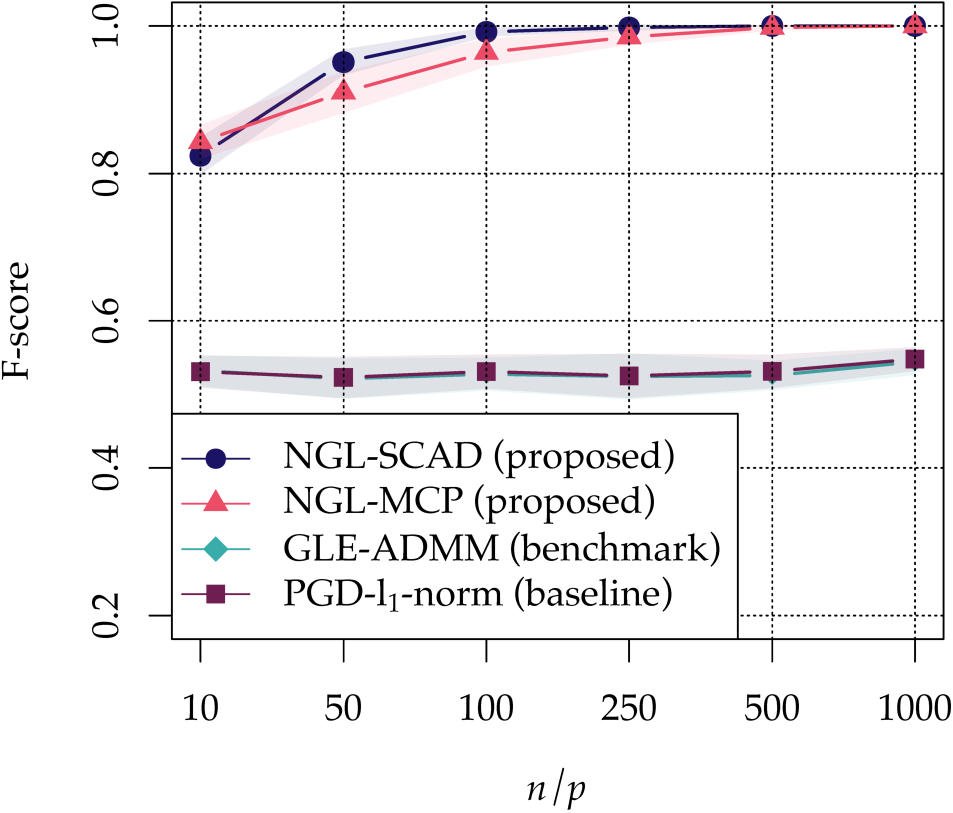}
        \caption{}
    \end{subfigure}%
    \vspace{-0.25cm}
    \caption{Performance measures (a) Number of positive edges, (b) Relative error and (c) F-score as a function of the sample size ratio of $n / p$ in learning the Barabasi-Albert graph. The true number of positive edges in (a) is $49$. The regularization parameter $\lambda$ for each algorithm is fine-tuned.}
    \label{fig:sparsity-alpha}
    \vspace{-0.3cm}
\end{figure}

\vspace{-0.4cm}

\subsection{Synthetic Data}

\vspace{-0.15cm}

For experiments on synthetic data, we consider two types of graphs: Barabasi-Albert graphs of degree one \citep{albert2002statistical}, and modular graphs.
Estimating tree graphs is a crucial task that has an impact in the performance of many applications such as image segmentation \citep{wu1993optimal}. Modular graph consists of a connected graph which contains clusters of nodes that share similar properties. The sparsity level in modular graphs are twofold: (1) intra-modular and (2) inter-modular. The first one refers to the number of connections among nodes that belong to the same module, whereas the second one refers to the number of connections among nodes in different modules. Modular graphs have been applied to a myriad of social network tasks such as community detection \citep{fortunato2010community}.

\begin{figure}[htb]
    \captionsetup[subfigure]{justification=centering}
    \centering
    \begin{subfigure}[t]{0.32\textwidth}
        \centering
        \includegraphics[scale=.25]{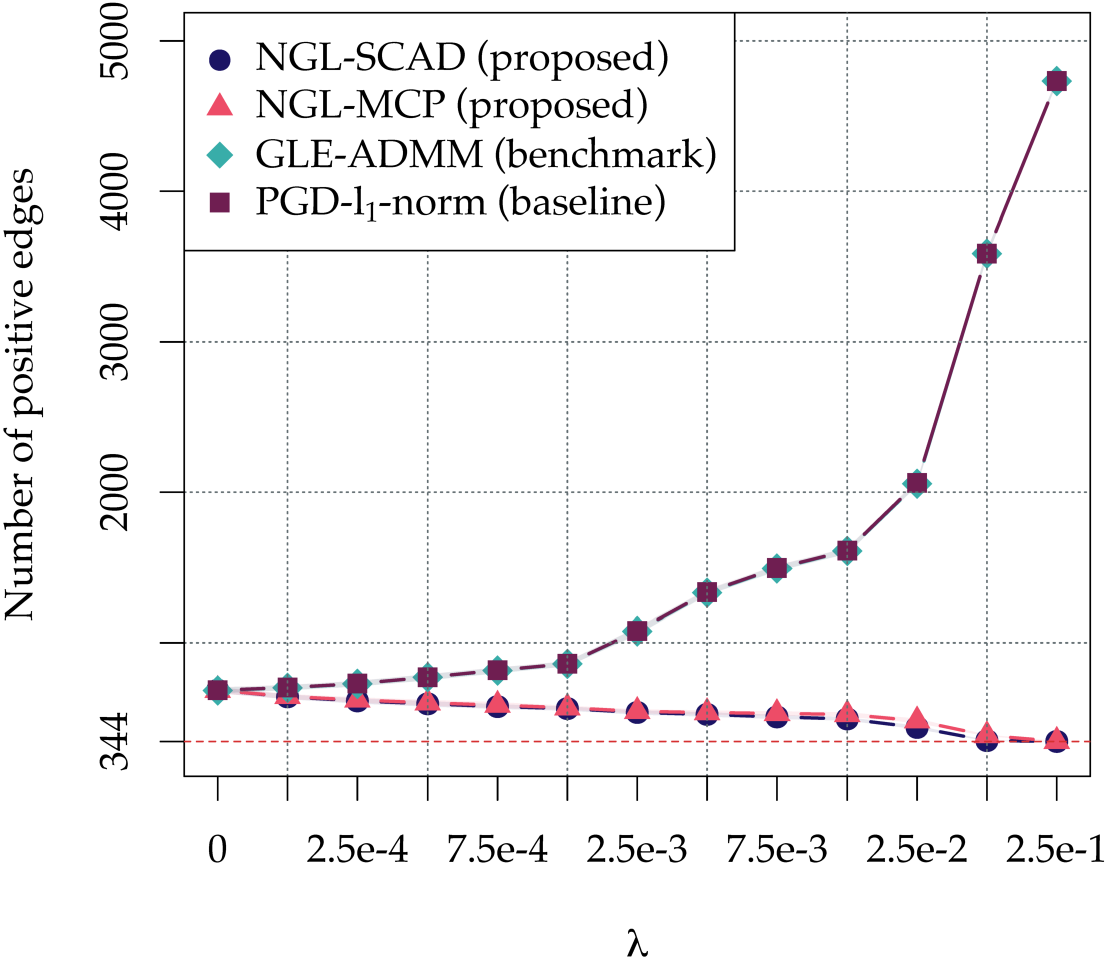}
        \caption{}
    \end{subfigure}%
   ~
    \begin{subfigure}[t]{0.32\textwidth}
        \centering
        \includegraphics[scale=.25]{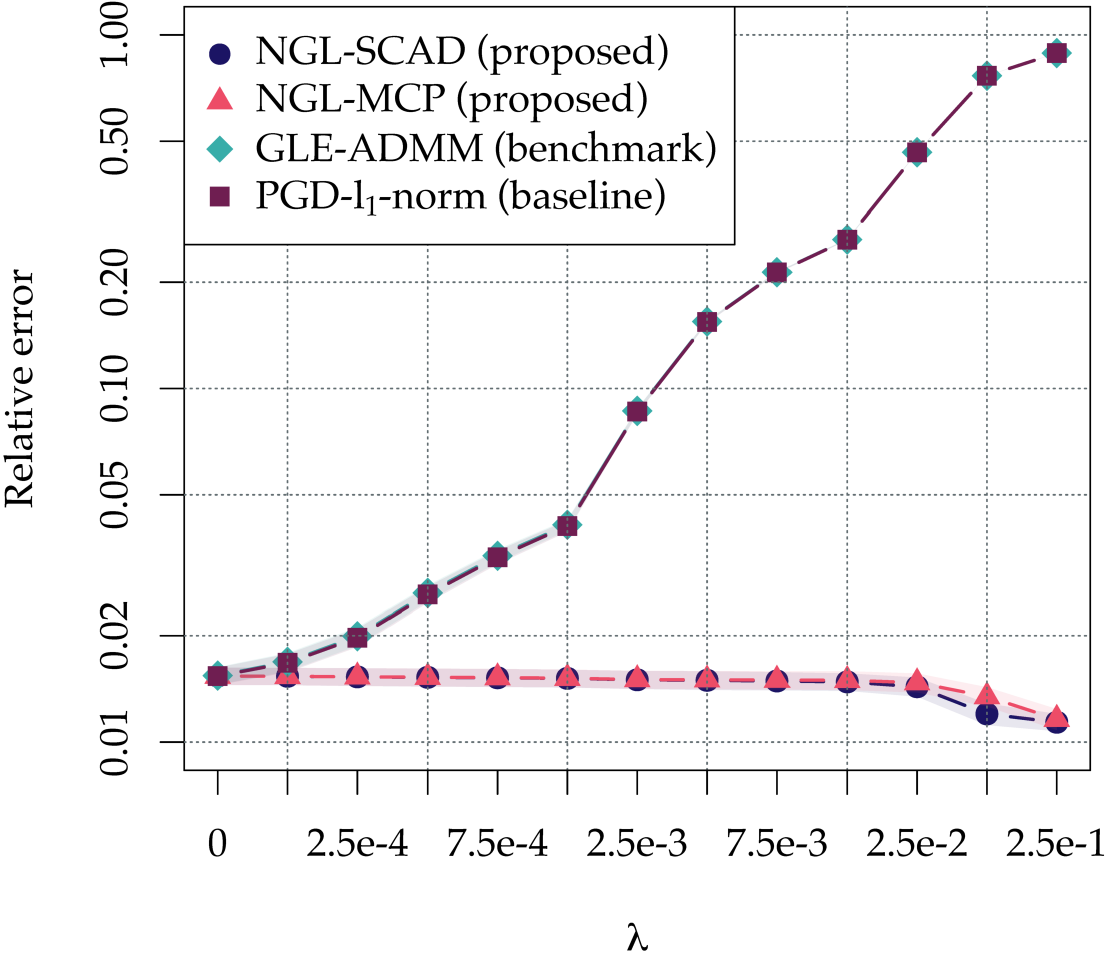}
        \caption{}
    \end{subfigure}%
    ~
    \begin{subfigure}[t]{0.32\textwidth}
        \centering
        \includegraphics[scale=.25]{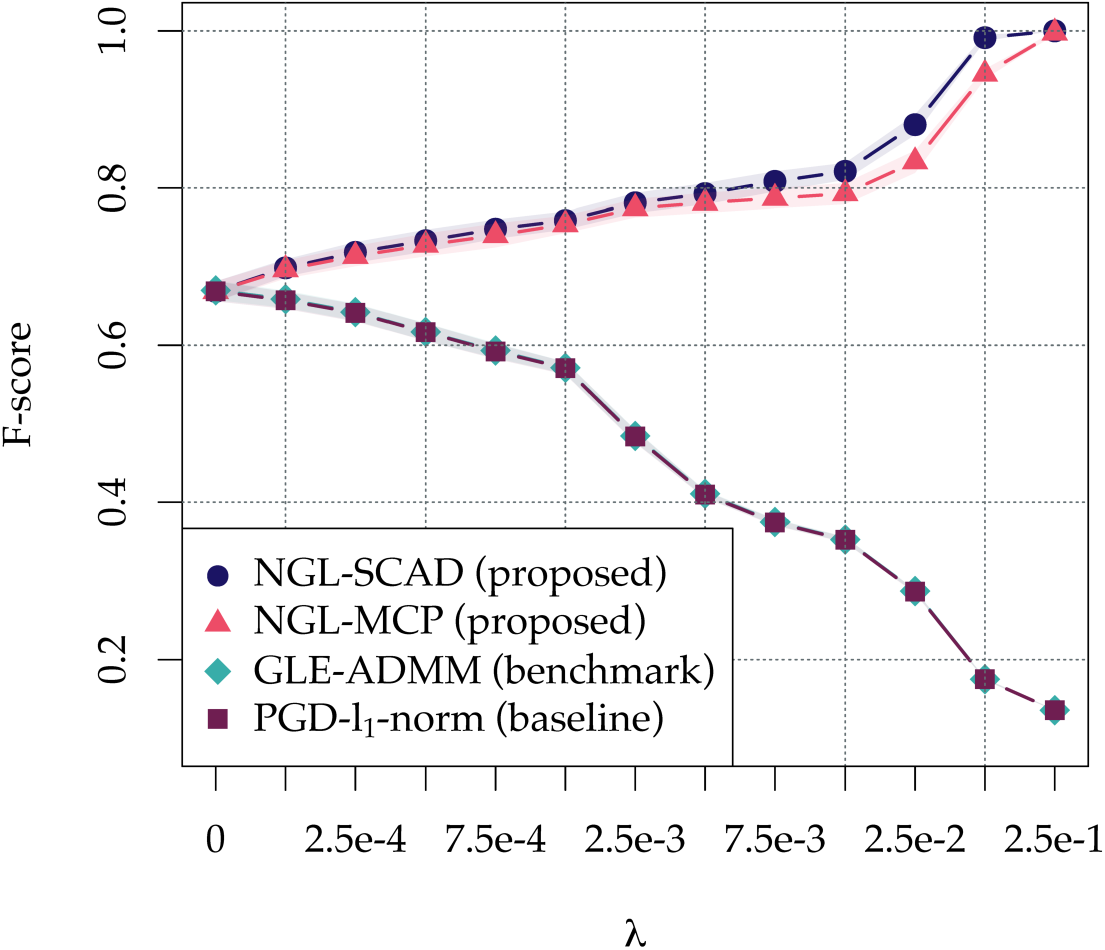}
        \caption{}
    \end{subfigure}%
    \vspace{-0.25cm}
    \caption{Performance measures (a) Number of positive edges, (b) Relative error and (c) F-score as a function of regularization parameter $\lambda$ in learning modular graphs. The true number of positive edges in (a) is $344$ and the sample size ratio is $n / p = 500$.}
    \label{fig:sparsity-lambda-modular}
     \vspace{-0.2cm}
\end{figure}

\begin{figure*}[!htb]
    \captionsetup[subfigure]{justification=centering}
    \centering
        \begin{subfigure}[t]{0.32\textwidth}
        \centering
        \includegraphics[scale=.29]{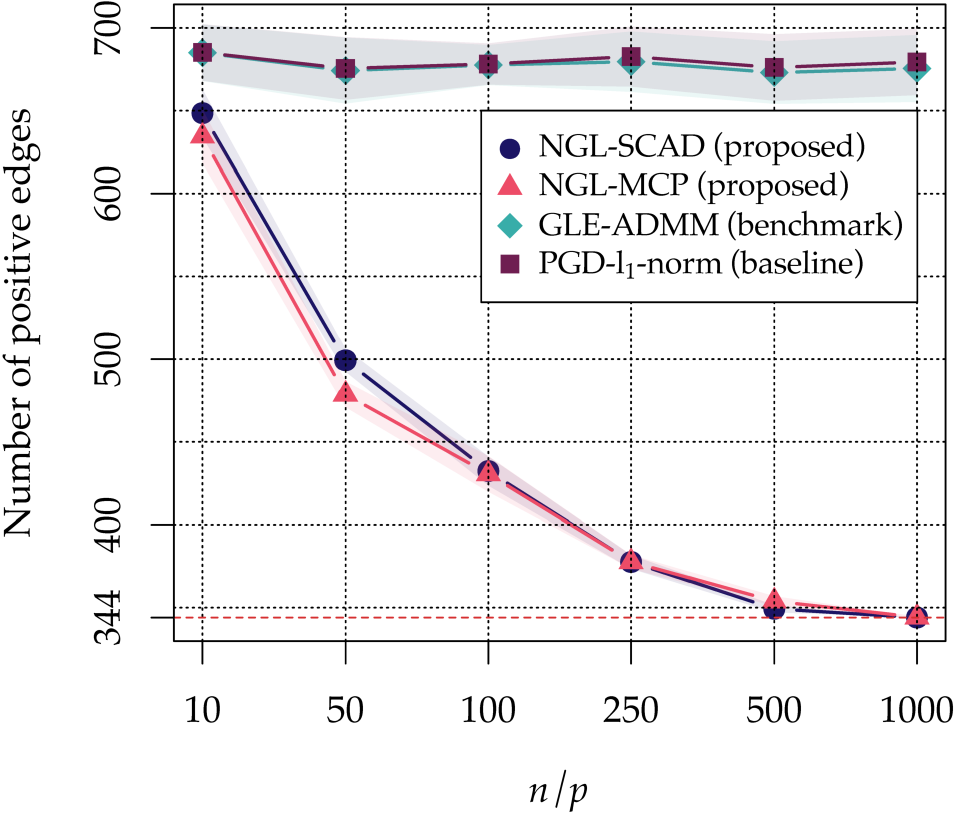}
        \caption{}
    \end{subfigure}%
    ~
    \begin{subfigure}[t]{0.32\textwidth}
        \centering
        \includegraphics[scale=.29]{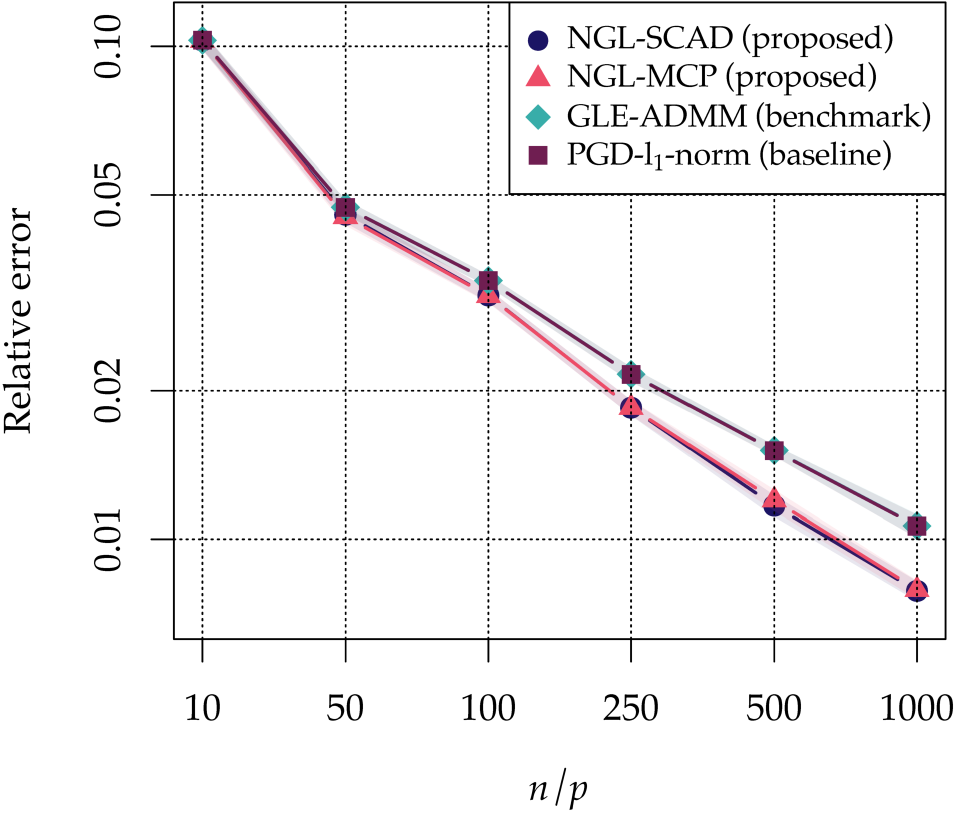}
        \caption{}
    \end{subfigure}%
   ~
    \begin{subfigure}[t]{0.32\textwidth}
        \centering
        \includegraphics[scale=.29]{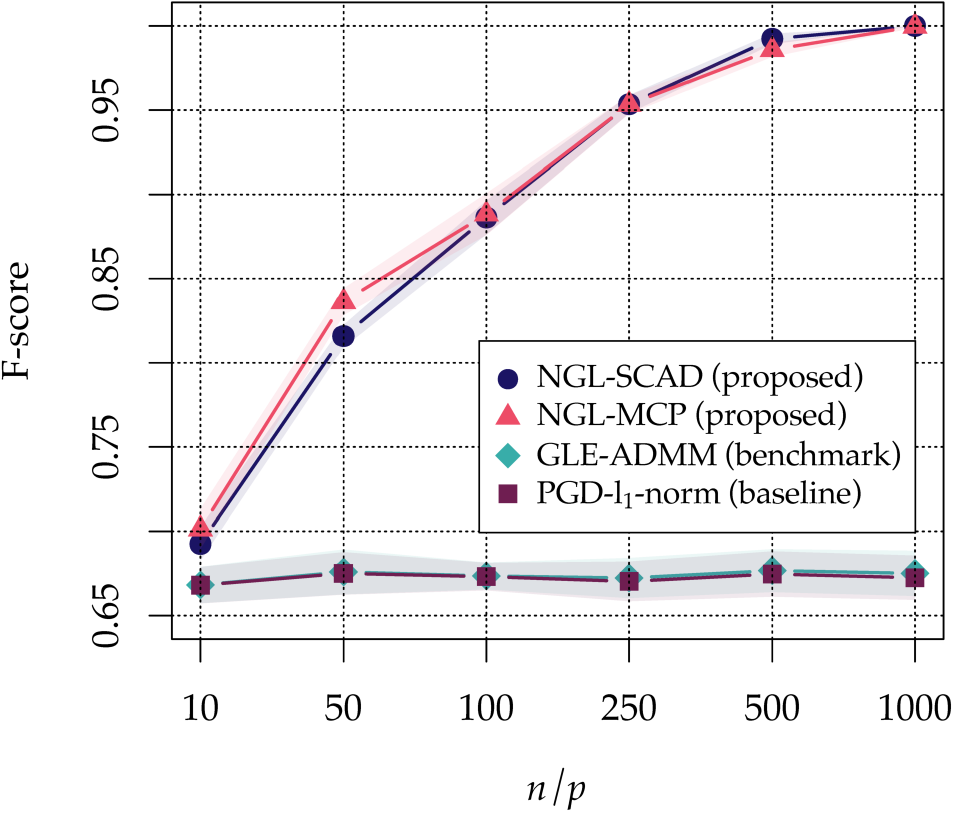}
        \caption{}
    \end{subfigure}%
     \vspace{-0.25cm}
    \caption{Performance measures (a) Number of positive edges, (b) Relative error and (c) F-score as a function of the sample size ratio of $n / p$ in learning modular graphs. The true number of positive edges in (a) is $344$. The regularization parameter $\lambda$ for each algorithm is fine-tuned.}
    \label{fig:sparsity-alpha-modular} 
     \vspace{-0.2cm}
\end{figure*}

We generate the data matrix $\bm{X} \in \mathbb{R}^{p \times n}$ with each column independently sampled from LGMRF with $\bm \Theta = \mathcal{L} \bm w^\star$, where $\bm w^\star$ contains the true weights from Barabasi-Albert graphs of degree one and modular graphs. The number of nodes in the Barabasi-Albert graph and modular graph is $p=50$ and $100$, respectively. The weights associated with positive edges in both Barabasi-Albert graph and modular graph are uniformly sampled from $U(2, 5)$. The modular graph is generated with probabilities of intra-module and inter-module connections $0.25$ and $0.005$, respectively. The sample covariance matrix is constructed by $\bm S = \frac{1}{n} \bm X \bm X^{\top}$, where $n$ is the number of samples. The curves in Figures~\ref{fig:sparsity-lambda}, \ref{fig:sparsity-alpha}, \ref{fig:sparsity-lambda-modular} and \ref{fig:sparsity-alpha-modular} are the results of an average of $100$ Monte Carlo realizations and the shaded areas represent the one-standard deviation confidence interval.

Figure~\ref{fig:sparsity-lambda} presents the results of learning Barabasi-Albert graphs by GLE-ADMM, projected gradient descent with $\ell_1$-norm, and the proposed method with the two regularizers. It is observed that the numbers of positive edges learned by GLE-ADMM and projected gradient descent with $\ell_1$-norm rise along with the increase of $\lambda$, implying that the graphs will get denser as $\lambda$ increases. Figure~\ref{fig:sparsity-lambda} shows that both GLE-ADMM and projected gradient descent with $\ell_1$-norm achieve the best performance in terms of sparsity, relative error, and F-score when $\lambda=0$, which defies the purpose of introducing the $\ell_1$-norm regularizer. In contrast, the proposed NGL-SCAD and NGL-MCP enhance sparsity when $\lambda$ increases. It is observed that, with $\lambda$ equal to $0.1$ or $0.25$, NGL-SCAD and NGL-MCP achieve the true number of positive edges, and an F-score of $1$, implying that all connections and disconnections between nodes in the true graph are correctly identified.

Figure~\ref{fig:sparsity-alpha} shows that the proposed NGL-SCAD and NGL-MCP always outperform both GLE-ADMM and the baseline projected gradient descent in terms of sparsity, relative error, and F-score under different samples size ratios.

 \vspace{-0.4cm}

Figure~\ref{fig:sparsity-lambda-modular} presents the results of learning modular graphs by GLE-ADMM, projected gradient descent with $\ell_1$-norm, and the proposed NGL-SCAD and NGL-MCP. It is observed that a larger regularization parameter $\lambda$ for GLE-ADMM and projected gradient descent with $\ell_1$-norm will lead to a worse performance in terms of sparsity, relative error, and F-score. On the contrary, NGL-SCAD and NGL-MCP enhance the sparsity and improve the relative error and F-score as $\lambda$ increases.

Figure~\ref{fig:sparsity-alpha-modular} shows that the proposed method always leads to a better performance in learning modular graphs in terms of sparsity, relative error, and F-score, than the compared methods under different sample size ratios.

\begin{figure}[htb]
    \captionsetup[subfigure]{justification=centering}
    \centering
    \begin{subfigure}[t]{0.25\textwidth}
        \centering
        \includegraphics[scale=.37]{images-final/tree-original-new-eps-converted-to.pdf}
        \caption{Ground-truth graph, $\textsf{NE}=49$}
    \end{subfigure}%
    ~
    \begin{subfigure}[t]{0.25\textwidth}
        \centering
        \includegraphics[scale=.37]{images-final/tree-admm0-new-eps-converted-to.pdf}
        \caption{$\textsf{NE}=135$, $\textsf{RE} = 0.14$, $\textsf{FS} = 0.53$}
    \end{subfigure}%
    ~
    \begin{subfigure}[t]{0.25\textwidth}
        \centering
        \includegraphics[scale=.37]{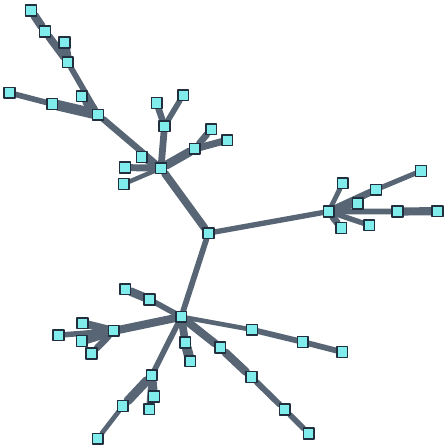}
        \caption{$\textsf{NE}=49$, $\textsf{RE} = 0.08$, $\textsf{FS} = 1$}
    \end{subfigure}%
    ~
    \begin{subfigure}[t]{0.25\textwidth}
        \centering
        \includegraphics[scale=.37]{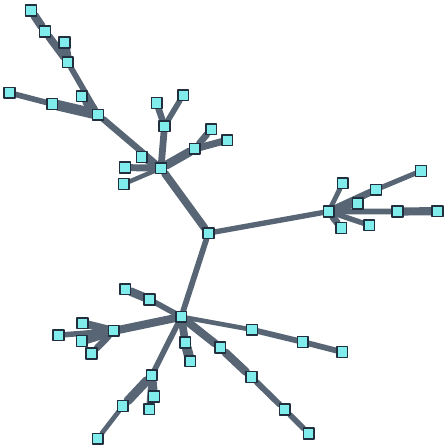}
        \caption{$\textsf{NE}=49$, $\textsf{RE} = 0.08$, $\textsf{FS} = 1$}
    \end{subfigure}
    \caption{A sample result in learning a Barabasi-Albert graph of degree one by
             (b) GLE-ADMM \citep{zhao2019optimization}, (c) NGL-SCAD (proposed) and (d) NGL-MCP (proposed).  The sample size ratio is $n / p = 6$. $\textsf{NE}$ denotes the number of positive edges in the graph. The regularization parameters for each method are set as $\lambda_{\textsf{ADMM}} = 0$, $\lambda_{\textsf{SCAD}} = \lambda_{\textsf{MCP}} = 0.5$.}
    \label{fig:sparsity}
\end{figure}

\vspace{-0.2cm}

\begin{figure}[!htb]
    \captionsetup[subfigure]{justification=centering}
    \centering
    \begin{subfigure}[t]{0.24\textwidth}
        \centering
        \includegraphics[scale=.29]{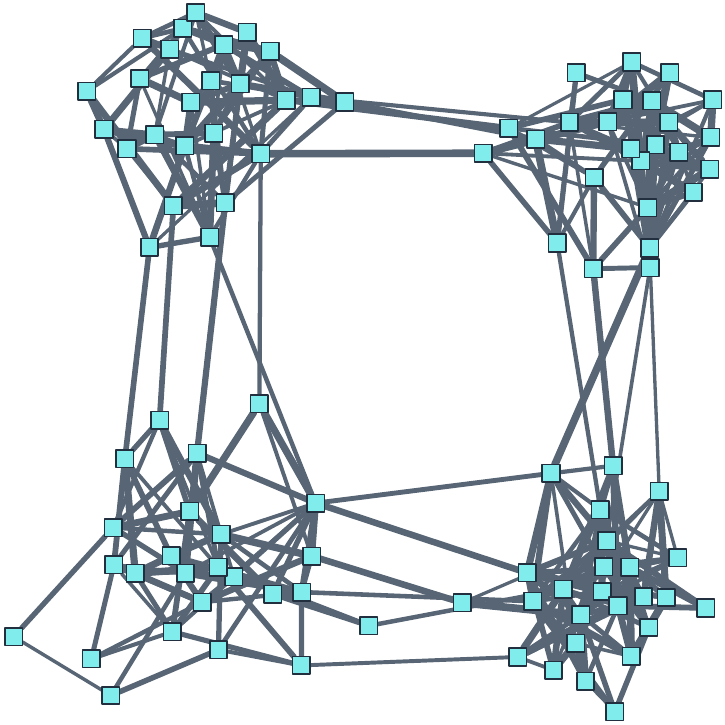}
        \caption{Ground-truth graph, $\textsf{NE}= 344$}
    \end{subfigure}%
    ~
    \begin{subfigure}[t]{0.24\textwidth}
        \centering
        \includegraphics[scale=.46]{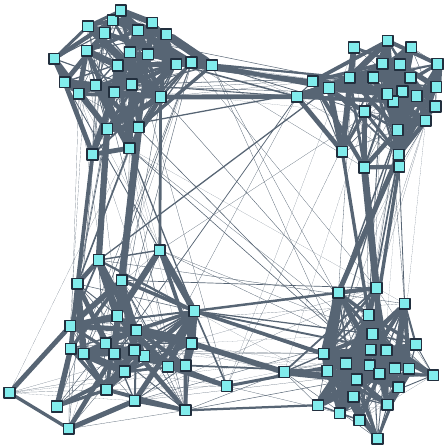}
        \caption{$\textsf{NE}= 681$, $\textsf{RE} = 0.14$, $\textsf{FS} = 0.66$}
    \end{subfigure}%
    ~
    \begin{subfigure}[t]{0.24\textwidth}
        \centering
        \includegraphics[scale=.46]{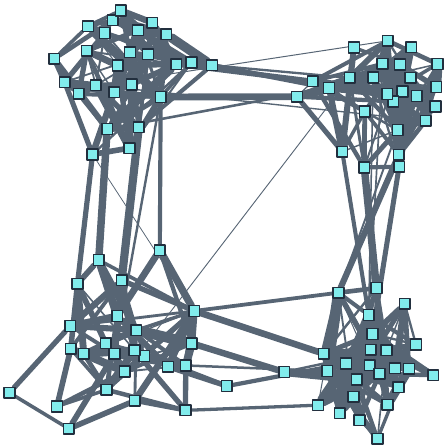}
        \caption{$\textsf{NE}= 398$, $\textsf{RE} = 0.15$, $\textsf{FS} = 0.85$.}
    \end{subfigure}%
    ~
    \begin{subfigure}[t]{0.24\textwidth}
        \centering
        \includegraphics[scale=.46]{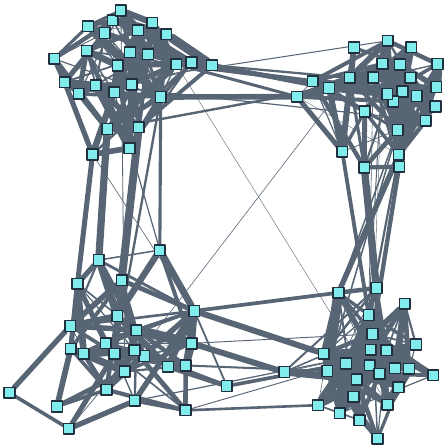}
        \caption{$\textsf{NE}= 453$, $\textsf{RE} = 0.14$, $\textsf{FS} = 0.81$.}
    \end{subfigure}
    \caption{ A sample result of learning a modular graph by
             (b) GLE-ADMM \citep{zhao2019optimization}, (c) NGL-SCAD (proposed) and (d) NGL-MCP (proposed). The sample size ratio is $n / p = 6$. $\textsf{NE}$ denotes the number of positive edges in the graph. The regularization parameters for each method are set as $\lambda_{\textsf{ADMM}} = 0$, $\lambda_{\textsf{SCAD}} = \lambda_{\textsf{MCP}} = 0.1$.}
    \label{fig:sample-modular}
\end{figure}

Figure~\ref{fig:sparsity} shows a sample result of learning a Barabasi-Albert graph via GLE-ADMM, NGL-SCAD and NGL-MCP. It is observed that the learned graphs via NGL-SCAD and NGL-MCP present the connection between any two nodes correctly, while there are many incorrect connections in the graph learned via GLE-ADMM. In addition, performance measures including sparsity, relative error, and F-score also indicate a better performance of the proposed method.

Figure~\ref{fig:sample-modular} shows a sample result of learning a modular graph via GLE-ADMM, NGL-SCAD and NGL-MCP. It is observed that the learned graphs via NGL-SCAD and NGL-MCP are much more representative than the one learned via GLE-ADMM.

\subsection{Real-world Data Sets}
In this subsection, we conduct numerical experiments on real-world data sets including stock data sets and \textsf{COVID-19} data set for connected graph learning, and \textsf{animals} data set for $k$-component graph learning.

\subsubsection{Stock data sets}\label{Sec-Stock}

\begin{figure}[htb]
    \captionsetup[subfigure]{justification=centering}
    \centering
        \begin{subfigure}[t]{0.46\textwidth}
        \centering
        \includegraphics[scale=.45]{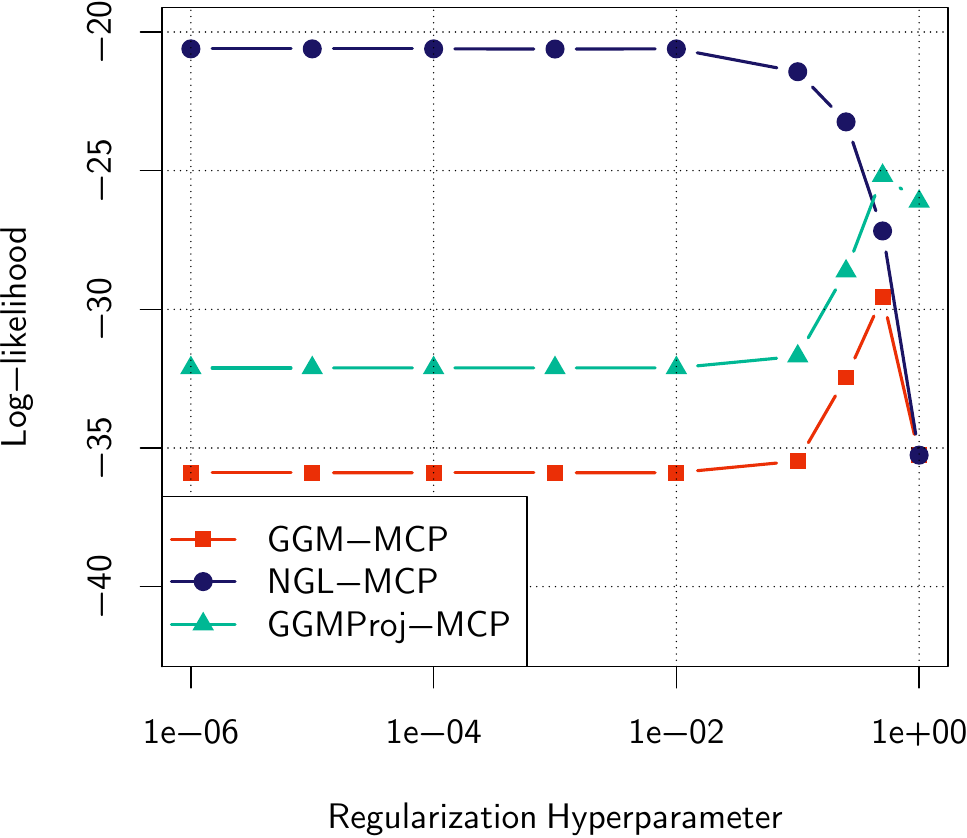}
        \caption{}
    \end{subfigure}%
    ~~
    \begin{subfigure}[t]{0.49\textwidth}
        \centering
        \includegraphics[scale=.52]{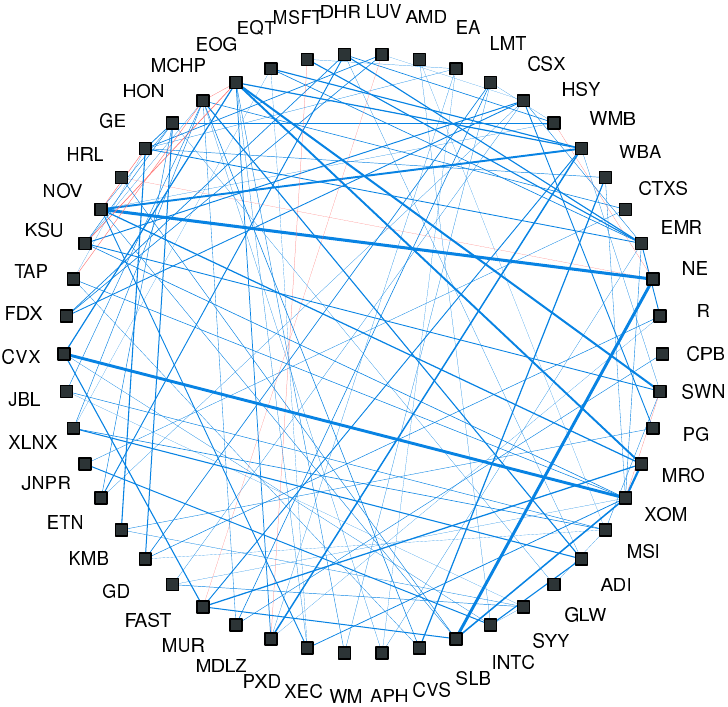}
        \caption{}
    \end{subfigure}%
    \caption{ (a) Average log-likelihood of the estimated precision matrices as a function of the regularization parameter; (b) The learned graph corresponding to the precision matrix obtained by GGMProj-MCP with the highest likelihood. The widths of the edges in (b) are proportional to the absolute value of the graph weights. Blue edges represent positive weights, while red edges represent negative ones. The minor weights are shrunk to zero. The data set in Figure \ref{fig:Likelihood} consists of log-returns time series from $50$ stocks randomly chosen stocks from the S\&P500 index during the period between Jan. 4th 2005 to Feb. 7th 2020 with 3800 observations. We then divide this data set into $19$ sequential sub-datasets with each of which containing $n = 200$ observations. For the $i$-th sub-dataset, we estimate the precision matrices under different values of the regularization parameter $\lambda$, and compute their log-likelihood using the $(i + 1)$-th dataset. Finally, we average the log-likelihood values over the sub-datasets.}
    \label{fig:Likelihood}
\end{figure}

We first present an experiment to verify that the Laplacian constrained GGM could describe the stock data set better than the general GGM. We compare the log-likelihood of the general precision matrices and the Laplacian constrained precision matrices, estimated by the GGM method with the MCP penalty (GGM-MCP) and the proposed NGL-MCP, respectively, on a test set of stock data. Note that the log-likelihood of the general precision matrices is computed with the function $\log \det (\cdot)$, while that of the Laplacian constrained precision matrices is with $\log \det^\star (\cdot)$ because of the singularity of the Laplacian matrix. To make the comparison fairer, we add a comparison wit GGMProj-MCP which is obtained by projecting the precision matrices estimated by GGM-MCP such that they satisfy $\bm \Theta \cdot \bm 1 = \bm 0$, then compute the log-likelihood using $\log \det^\star (\cdot)$. 

Figure \ref{fig:Likelihood} (a) shows that the NGL-MCP achieves the highest log-likelihood, indicating that the Laplacian constrained precision matrices describe the stock data better than the general precision matrices. Figure \ref{fig:Likelihood} (b) presents the graph weights $W_{ij} = -\Theta_{ij}$ with $i \neq j$, where the precision matrix $\bm \Theta$ is obtained by GGMProj-MCP with the highest log-likelihood in Figure \ref{fig:Likelihood} (b). It is observed that most of the graph weights are non-negative, implying that the sign assumption is satisfied well on the stock data set. 

\begin{figure}[!htb]
    \captionsetup[subfigure]{justification=centering}
    \centering
      \begin{subfigure}[t]{0.33\textwidth}
        \centering
        \includegraphics[scale=.34]{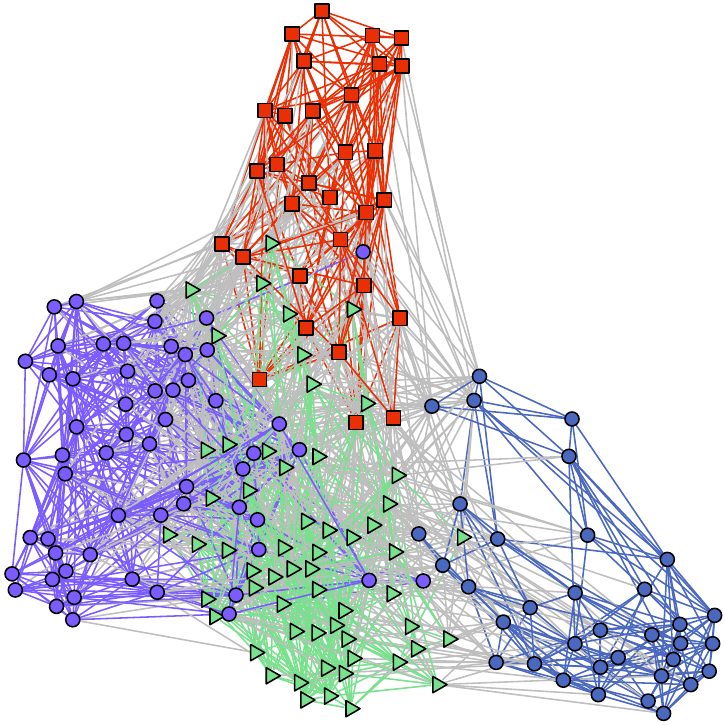}
        \caption{}
    \end{subfigure}%
    ~ 
        \begin{subfigure}[t]{0.33\textwidth}
        \centering
        \includegraphics[scale=.34]{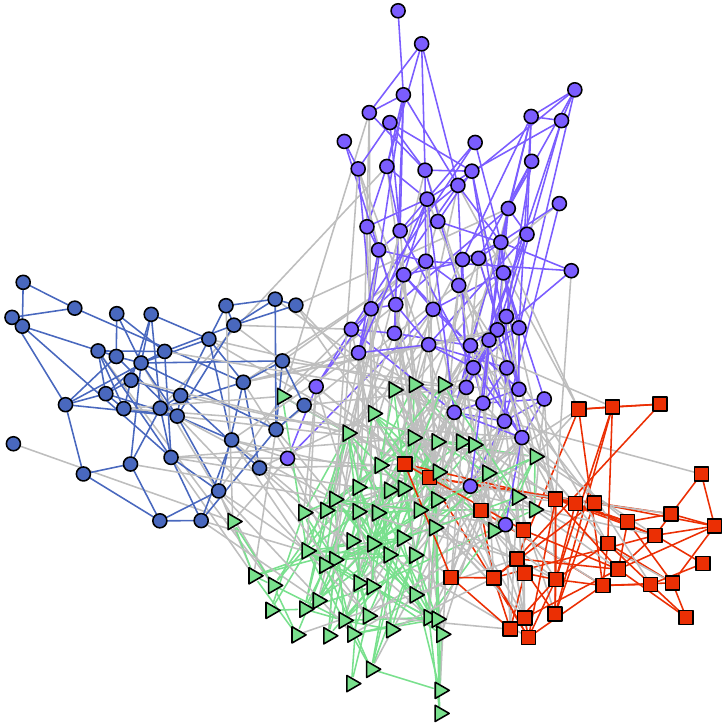}
        \caption{}
    \end{subfigure}%
    ~ 
        \begin{subfigure}[t]{0.33\textwidth}
        \centering
        \includegraphics[scale=.34]{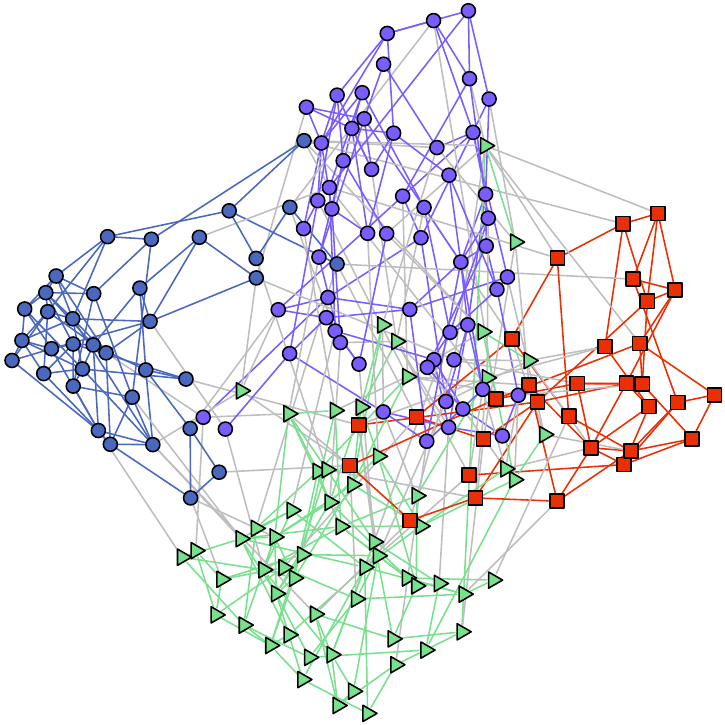}
        \caption{}
    \end{subfigure}%
    \caption{ Stock graphs learned via (a) GLE-ADMM, (b) GGM-MCP,  and (c) NGL-MCP. The modularity values for (a), (b) and (c) are 0.36, 0.37 and 0.51, respectively. The data set is from stocks composing the S\&P 500 index. The regularization parameters for GLE-ADMM, GGM-MCP and NGL-MCP are 0, 0.4 and 0.25, respectively. The regularization parameter is tuned such that the estimated graph has the highest modularity value. The weights below $10^{-5}$ are shrinkaged for the graph obtained by GLE-ADMM to produce a sparser graph.}
    \label{fig:ggm-mcp}
\end{figure}

Next, we compare the proposed NGL-MCP with GLE-ADMM and GGM-MCP in learning graphs on a stock data set. It is observed in Figure~\ref{fig:ggm-mcp} that the performance of NGL-MCP is the most significant, because most connections are between nodes within the same community, and only a few connections (gray edges) are between nodes from distinct communities. The modularity values for GLE-ADMM, GGM-MCP and NGL-MCP are 0.36, 0.37 and 0.51, respectively. A higher modularity value indicates a better representation of the actual network of stocks. The data set is from stocks composing the S\&P 500 index. We select log-returns from 181 stocks from 4 sectors, namely: "Industrials", "Consumer Staples", "Energy", "Information Technology", during a period of 4 years from January 1st 2016  to May 20th 2020, with a total of 1101 observations. Then the data matrix $\bm{X}$ has a size of $181 \times 1101$.

\subsubsection{COVID-19 data sets}
We conduct numerical experiments on the \textsf{COVID-19} data set\footnote{2019-nCoV data is available in a queryable format via the \textsf{R} package \textsf{nCov2019} which lives on GitHub: \url{https://github.com/GuangchuangYu/nCov2019}.} from 98 anonymous Chinese patients affected by the outbreak of \textsf{COVID-19} on early February, 2020. The features include age (integer), gender (categorical), and location (categorical). The label is a binary variable representing the life status of patients, alive (green) or no longer alive (red). Our goal is to construct a graph from the data features. To this end, we first pre-process the feature matrix so as to transform the categorical features into numerical ones via one-hot-encoding. The pre-processed feature matrix $\bm{X}$ has the dimension $98 \times 32$, i.e., $p=98$ and $n=32$. We then compute the sample covariance matrix and learn the graphs. 

\begin{figure}[htb]
    \captionsetup[subfigure]{justification=centering}
    \centering
    \begin{subfigure}[t]{0.32\textwidth}
        \centering
        \includegraphics[scale=.25]{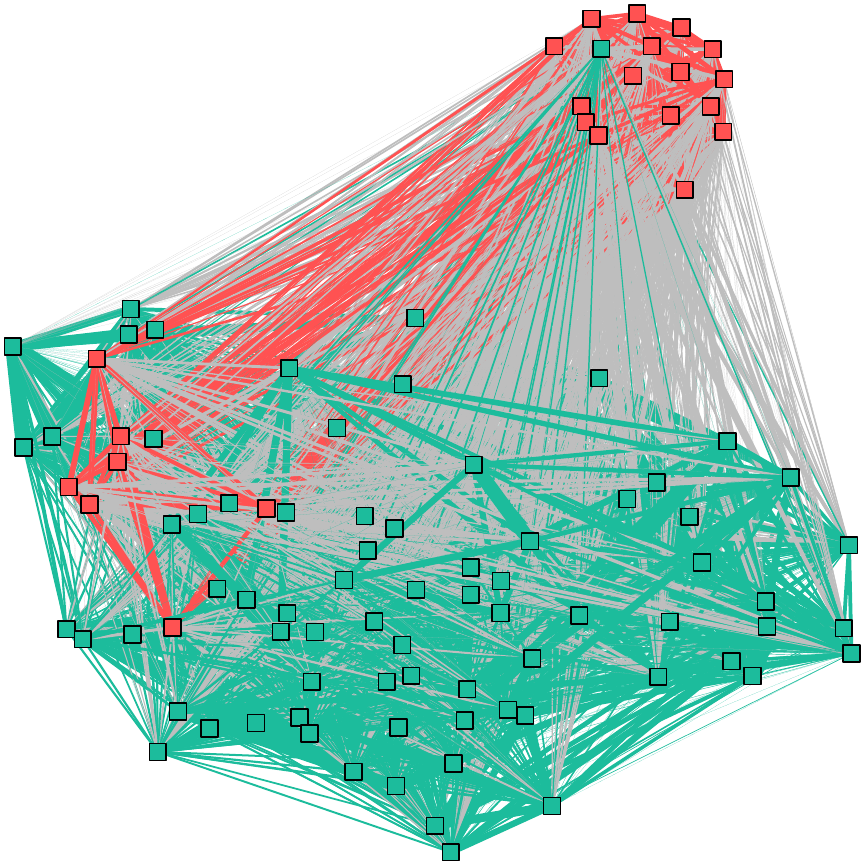}
        \caption{}
    \end{subfigure}%
    ~
    \begin{subfigure}[t]{0.32\textwidth}
        \centering
        \includegraphics[scale=.25]{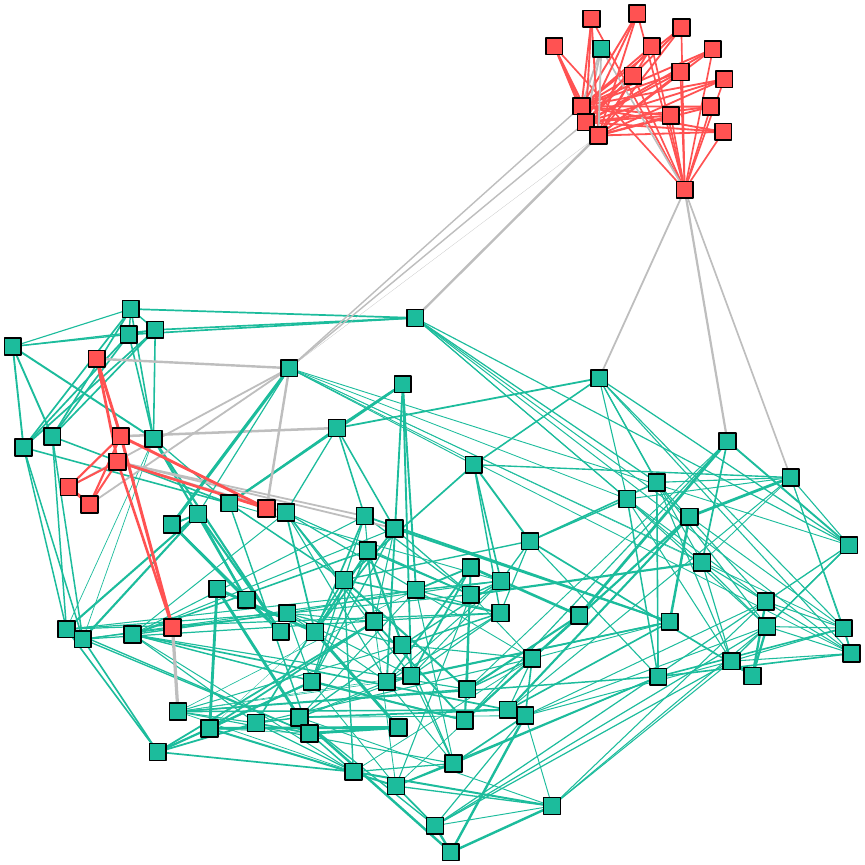}
        \caption{}
    \end{subfigure}%
        ~
        \begin{subfigure}[t]{0.32\textwidth}
        \centering
        \includegraphics[scale=.25]{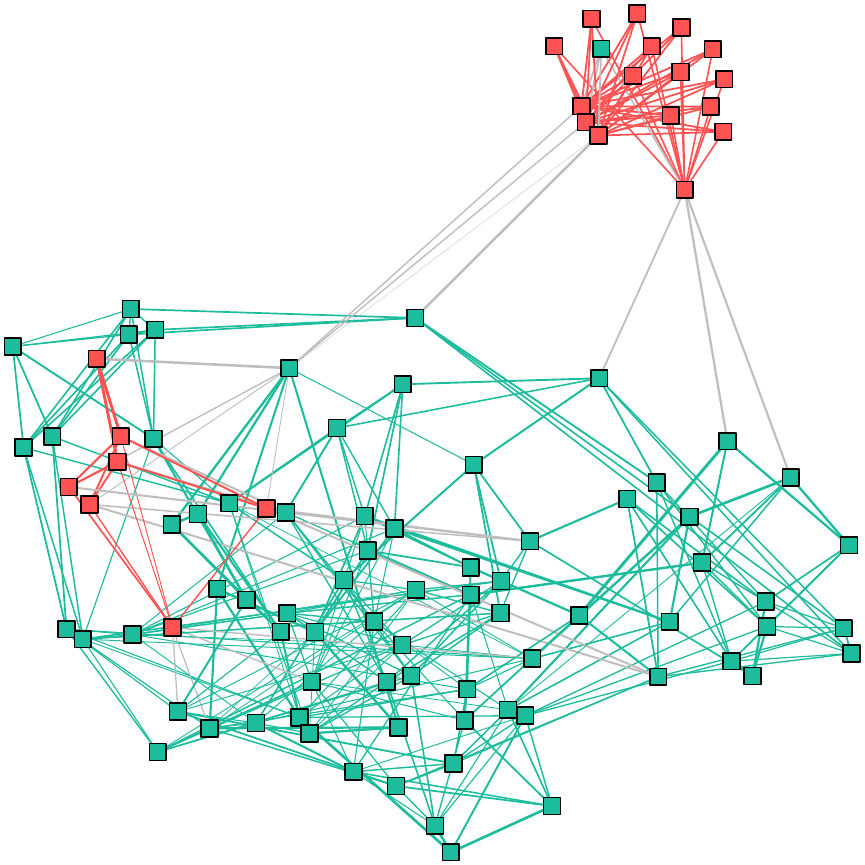}
        \caption{}
    \end{subfigure}%
    
    \caption{The learned graphs using the \textsf{COVID-19} data set from Chinese patients via (a) GLE-ADMM, (b) NGL-SCAD (proposed), and (c) NGL-MCP (proposed). The computational time for GLE-ADMM, NGL-SCAD and NGL-MCP are 2.9, 0.7 and 0.8 seconds, respectively. The regularization parameters are set as $\lambda_{\textsf{ADMM}} = 0$, $\lambda_{\textsf{SCAD}} = 0.6$, and $\lambda_{\textsf{MCP}} = 1.2$. }
    \label{fig:corona}
\end{figure}

Figure~\ref{fig:corona} shows that the benchmark GLE-ADMM is unable to impose sparsity, diminishing interpretation capabilities of the graph severely. On the other hand, the proposed NGL-SCAD and NGL-MCP obtain sparse graphs with clearer connections. The learned graphs possibly provide guidance on priority setting in health care because green nodes (patients alive) that have stronger connections with red nodes (patients that passed away) may suffer a higher health risk.

\begin{figure}[!htb]
    \captionsetup[subfigure]{justification=centering}
    \centering
    \begin{subfigure}[t]{0.33\textwidth}
        \centering
        \includegraphics[scale=.23]{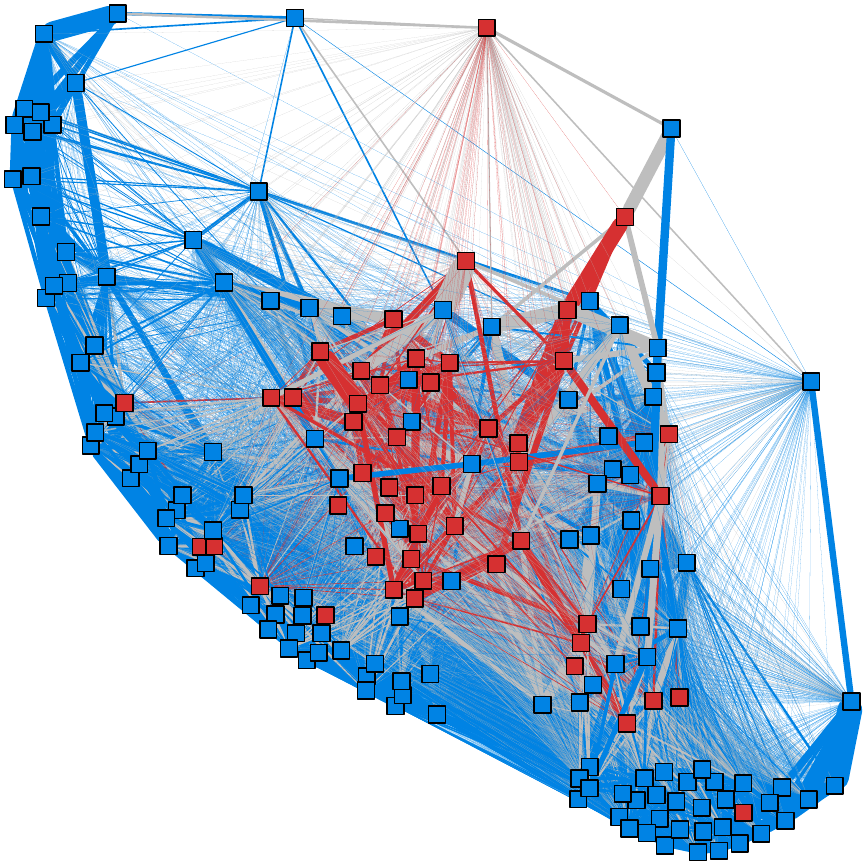}
        \caption{}
    \end{subfigure}%
    ~
    \begin{subfigure}[t]{0.33\textwidth}
        \centering
        \includegraphics[scale=.23]{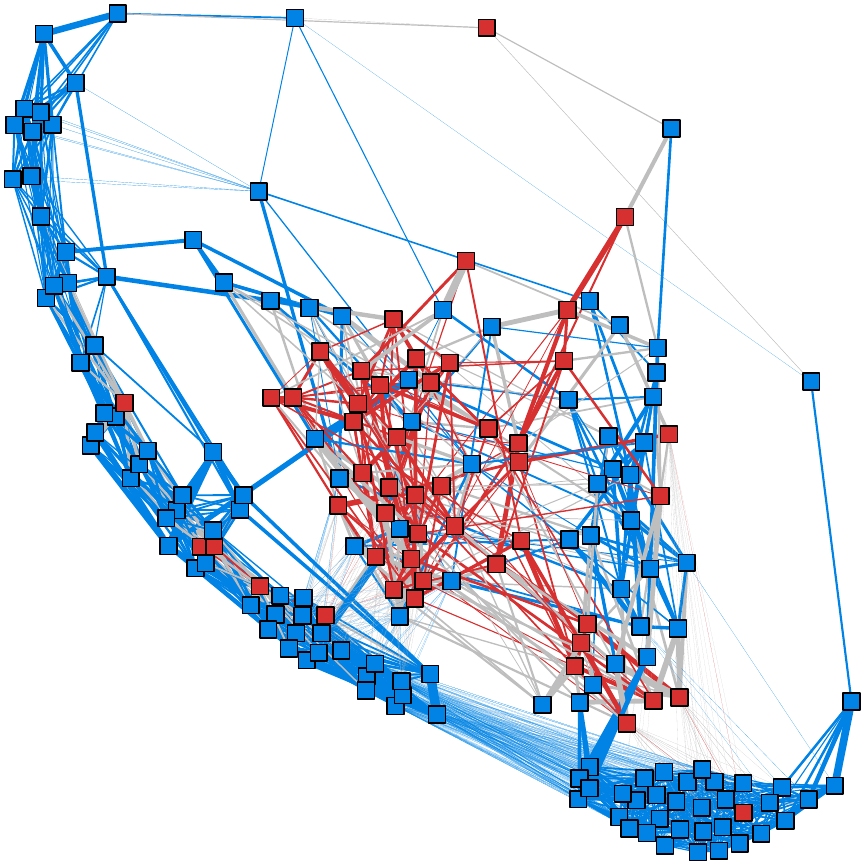}
        \caption{}
    \end{subfigure}%
        ~
    \begin{subfigure}[t]{0.33\textwidth}
        \centering
        \includegraphics[scale=.23]{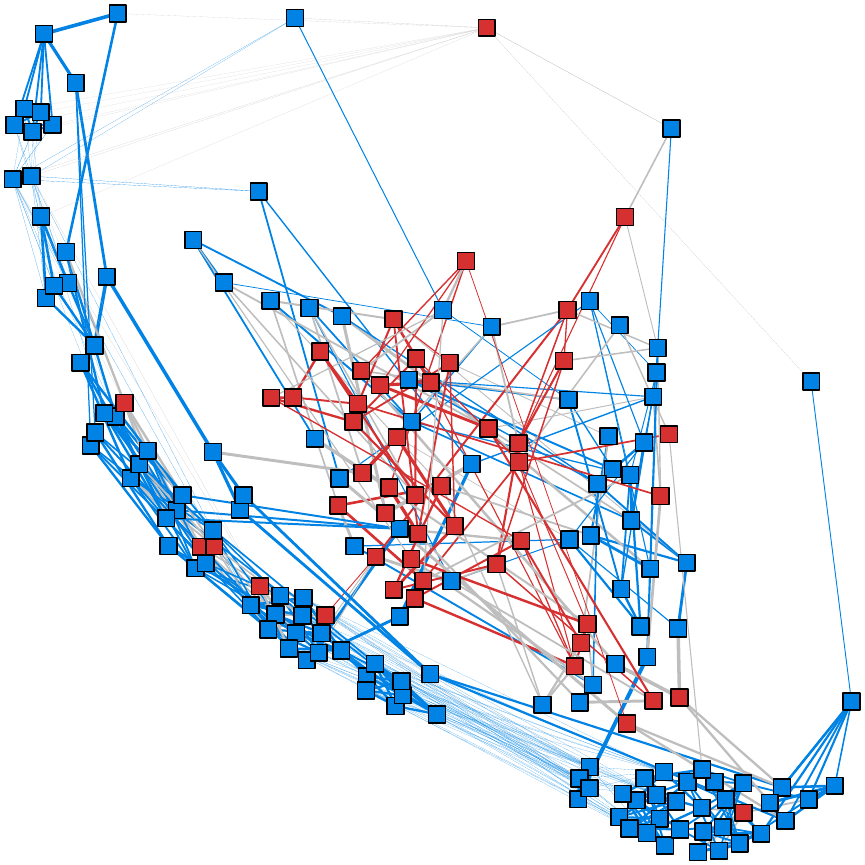}
        \caption{}
    \end{subfigure}%
    \caption{The learned graphs using the \textsf{COVID-19} data set provided by the Israelite Hospital Albert Einstein via (a) GLE-ADMM, (b) NGL-SCAD (proposed), and (c) NGL-MCP (proposed). The computational time for GLE-ADMM, NGL-SCAD and NGL-MCP are 9.9, 55.8 and 57.9 seconds, respectively. The regularization parameters are set as $\lambda_{\textsf{ADMM}} = 0$, $\lambda_{\textsf{SCAD}} = 0.1$ and $\lambda_{\textsf{MCP}} = 0.5$.}
    \label{fig:corona1}
    \vspace{-0.4cm}
\end{figure}

We next perform experiments on the \textsf{COVID-19} data set\footnote{The data set is freely available at: \url{https://www.kaggle.com/einsteindata4u/covid19}.} provided by the Israelite Hospital Albert Einstein in Brazil. The data set contains anonymized data from patients who had samples collected to perform the test for \textsf{SARS-CoV-2}. The features in the data set are mainly clinical coming from blood, urine, and saliva exams, e.g., hemoglobin level, platelets, red blood cells, etc. The original data set contains 108 features from 558 patients. Due to the high number of missing values, we do not consider features that were measured for at most 10 patients. In addition, a large number of patients had no record of any features. Finally, we end up with a data matrix of 182 patients with 57 features, i.e., $p=182$ and $n=57$. The remaining missing values were filled in with zeros. We then compare the proposed method with the GLE-ADMM method on this data set. It is observed from Figure~\ref{fig:corona1} that the proposed NGL-SCAD and NGL-MCP output a more interpretable representation of the network, where blue and red nodes denote patients who tested negative and positive for \textsf{SARS-CoV-2}, respectively.

It is worth mentioning that the real-world data sets may not exactly follow the Laplacian constrained Gaussian graphical models. In this case, our formulation in \eqref{cost-theta} can be related to the regularized log-determinant Bregman divergence optimization, and the learned graph weights can quantify the similarity between nodes. This is because the trace term can be written as Laplacian quadratic \citep{dong2016learning,kalofolias2016learn}, which tends to assign a large weight between nodes if their signal values are similar.

\vspace{-0.2cm}

\subsubsection{Disconnected Graph Learning}

\vspace{-0.1cm}

\begin{figure}[!htb]
    \captionsetup[subfigure]{justification=centering}
    \centering
    \begin{subfigure}[t]{0.33\textwidth}
        \centering
        \includegraphics[scale=.42]{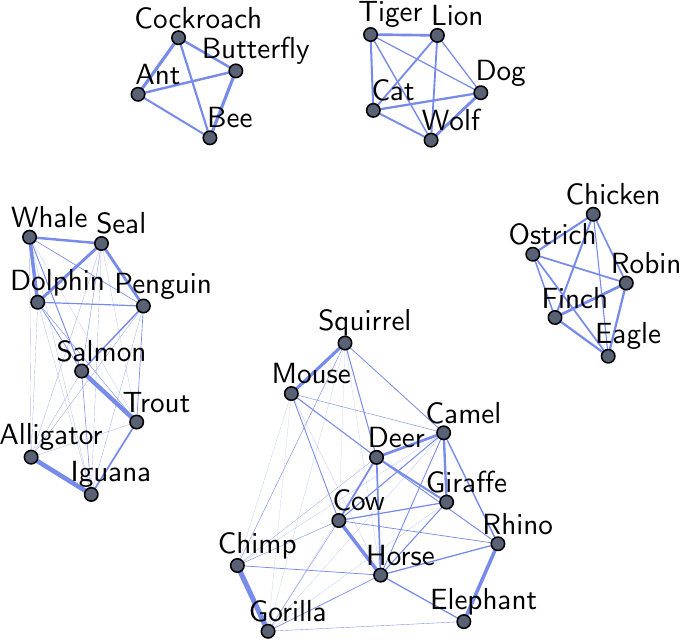}
        \caption{CLR with $k=5$}
    \end{subfigure}%
    ~
        \begin{subfigure}[t]{0.33\textwidth}
        \centering
        \includegraphics[scale=.42]{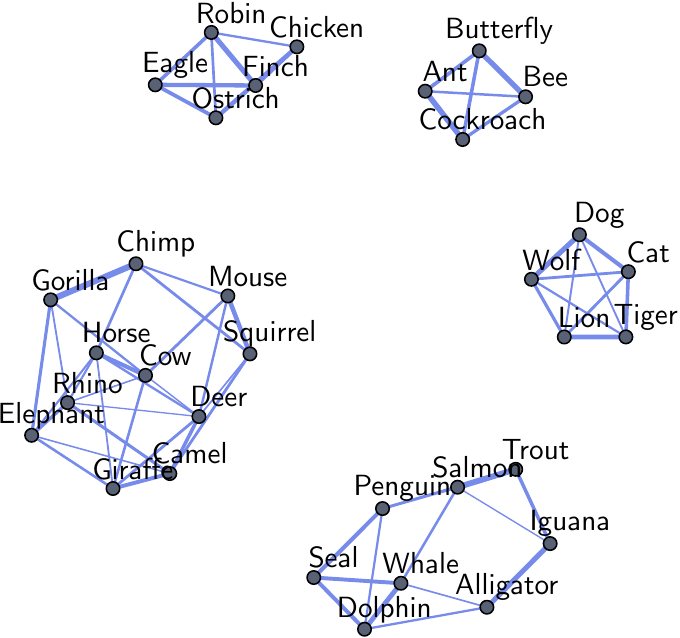}
        \caption{SGL with $k=5$}
    \end{subfigure}%
    ~
        \begin{subfigure}[t]{0.33\textwidth}
        \centering
        \includegraphics[scale=.42]{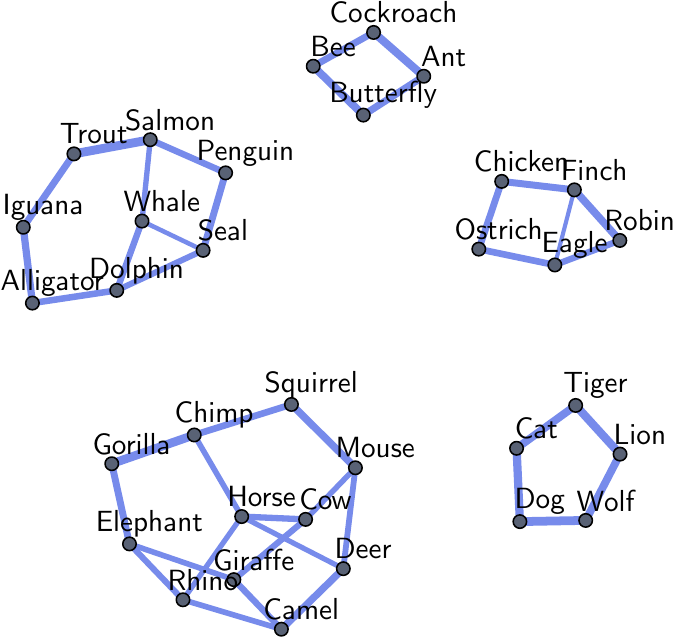}
        \caption{NGL-MCP with $k=5$}
    \end{subfigure}%
    \\
         \vspace{0.4cm}
        \begin{subfigure}[t]{0.33\textwidth}
        \centering
        \includegraphics[scale=.42]{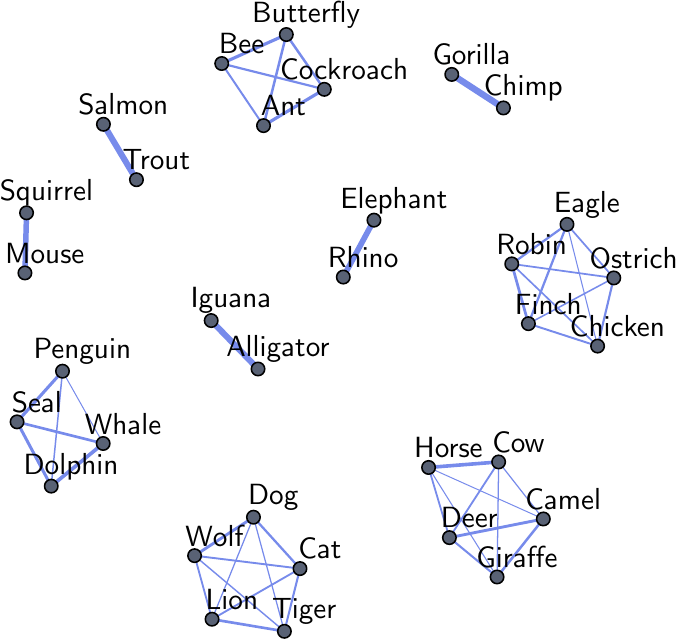}
        \caption{CLR with $k=10$}
    \end{subfigure}%
    ~
    \begin{subfigure}[t]{0.33\textwidth}
        \centering
        \includegraphics[scale=.42]{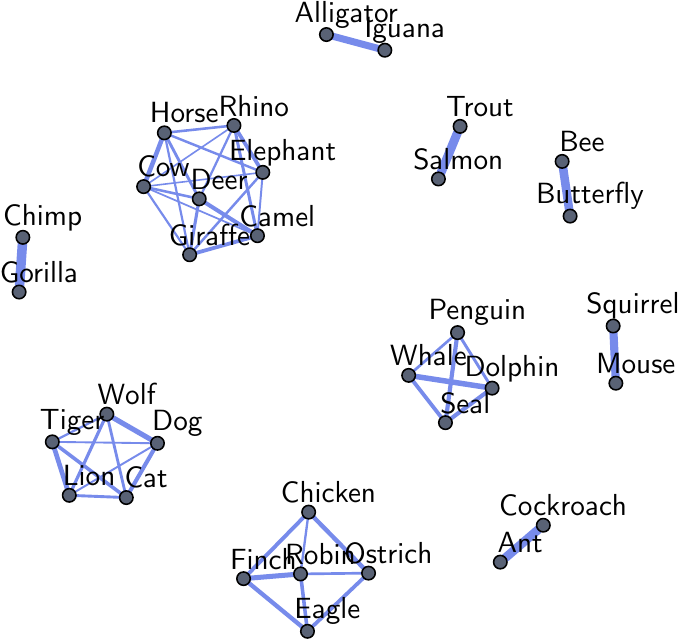}
        \caption{SGL with $k=10$}
     \end{subfigure}%
     ~
      \begin{subfigure}[t]{0.33\textwidth}
        \centering
        \includegraphics[scale=.42]{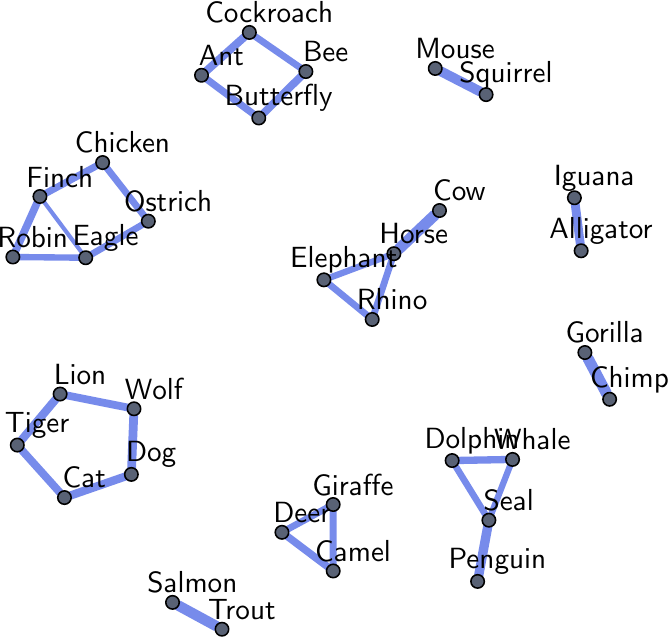}
        \caption{NGL-MCP with $k=10$}
    \end{subfigure}%
    \caption{ Learning the connectivity of the \textsf{animals} data set via CLR \citep{nie2016constrained}, SGL \citep{kumar2019unified}, and the proposed NGL-MCP with the number of components $k=5, 10$. The widths of the edges are proportional to the value of the graph weights. }
    \label{fig:animals}
\end{figure}

In this subsection, we conduct an experiment on the \textsf{animals} data set \citep{lake2010discovering} for $k$-component graph learning, and compare our proposed method with the state-of-the-art CLR \citep{nie2016constrained} and SGL \citep{kumar2019unified}. In this data set, every node denotes one animal and edge weights represent similarities among them. The \textsf{animals} data set consists of 33 animals and 102 features. The regularization parameter of the proposed method is set as 0.5 in Figure~\ref{fig:animals}.

It is observed in Figure~\ref{fig:animals} that the animal groupings among the graphs learned by CLR, SGL and the proposed NGL-MCP vary slightly. The learned graphs in Figure~\ref{fig:animals} are meaningful, because similar animals such as (chimp, gorilla) and (salmon, trout) are grouped together in Figure~\ref{fig:animals} (d), (e) and (f) when setting $k=10$. Note that the CLR method aims to learn a similarity graph, where a large graph weight between two nodes indicates a large similarity between them, and there is no underlying statistical assumption, while both SGL and NGL-MCP have the assumption of the Laplacian constrained Gaussian graphical models. Therefore, this experimental result justifies our statement in Section \ref{background} that the graph weights in Laplacian constrained Gaussian graphical models can quantify the similarity between two nodes. 

Figure~\ref{fig:animals} shows that our method can learn a more interpretable graph than CLR and SGL, because the connectivity within groups in the graphs learned by CLR and SGL is dense, which reduces the interpretability of graphs. For example, in Figure ~\ref{fig:animals} (d) and (e), the connectivity in the group (tiger, lion, wolf, dog and cat) is full, i.e., every two nodes are connected by an edge. This is because CLR does not impose the sparsity explicitly, and SGL suffers the interplay between the spectral regularization term and the sparsity term.

\section{Conclusions and Discussions}\label{conclusion}
In this paper, we have considered learning a sparse graph under the Laplacian constrained Gaussian graphical models. We have proved that a large regularization parameter of the $\ell_1$-norm leads to a complete graph. To overcome the issue of the $\ell_1$-norm, we have proposed a new estimator by introducing the nonconvex sparsity penalty. We have established the non-asymptotic optimization performance guarantees on both optimization error and statistical error, and proved that the proposed estimator can recover the graph edges correctly with a high probability. A projected gradient descent algorithm has designed to solve each sub-problem which enjoys a linear convergence rate. We have further extended our method to learn disconnected graphs. Numerical results have demonstrated the effectiveness of the proposed method.

Finally, we propose several viable directions for future research. First, it would be intriguing to establish non-asymptotic optimization performance guarantees on both optimization and statistical errors in the context of learning disconnected graphs. Notably, the rank constraint may introduce numerous local minima, and the techniques required would significantly differ from those used for learning connected graphs. Second, this paper focuses on Laplacian constraints within Gaussian graphical models. Future work could expand this to explore Laplacian constraints within other Markov random fields, such as Ising models \citep{ravikumar2010high}.

\appendix

\section{Proofs of Theorems} \label{Proof}
This section includes the proofs of Theorems \ref{Theorem 1}, \ref{Theorem 2}, \ref{theorem 3}, \ref{oracle}, \ref{theorem 4} and \ref{k-component-convergence}, and Corollaries \ref{corollary-L1} and \ref{corollary}. Before proving the theorems, we first present some technical lemmas.

\subsection{Technical Lemmas}

\begin{lemmas}\label{lem6} 
Let $f(\bm w) = - \log \det(\L \bm w + \bm J)$, with $\bm w \in \mathbb{R}^{p(p-1)/2}$. Then for any $\bm x \in  \mathbb{R}^{p(p-1)/2}$, we have
\begin{equation}
\bm x^\top \nabla^2 f(\bm w) \bm x = \mathrm{vec} (\L \bm x)^\top \left( \left( \L \bm w + \bm J \right)^{-1} \otimes \left( \L \bm w + \bm J \right)^{-1} \right)  \mathrm{vec} (\L \bm x). \nonumber
\end{equation} 
\end{lemmas}

\vspace*{1pt} 

\begin{lemmas}\label{lem18}
For any given $\bm w \in \mathbb{R}^{p(p-1)/2}$ satisfying $(\L \bm w + \bm J) \in \mathcal{S}^p_{++}$, there must exist an unique $\bm x \in \mathbb{R}^{p(p-1)/2}$ such that
\begin{equation}\label{inverse}
\L \bm x + \frac{1}{b}\bm J = \left( \L \bm w + b \bm J \right)^{-1}
\end{equation} 
holds for any $b \neq 0$, where $\bm J = \frac{1}{p} \bm 1_{p \times p}$, in which $\bm 1_{p \times p} \in \mathbb{R}^{p \times p}$ with each element equal to 1.
\end{lemmas}



\begin{lemmas}\label{lem-L}
Denote $\normI{\mathcal{L}^\ast}_\infty$ by
\begin{equation*}
\normI{\mathcal{L}^\ast}_\infty : = \underset{\substack{ \norm{\bm X}_{\max} = 1 \\ \bm X \in \mathbb{R}^{p \times p}}}{\sup} \norm{ \mathcal{L}^\ast \bm X}_{\max}.
\end{equation*}
 Then we have $\normI{\mathcal{L}^\ast}_\infty =4$.
\end{lemmas}

\vspace*{1pt} 

\begin{lemmas}\label{lem11}
Let $\mathcal{G}=\mathcal{L}^\ast \mathcal{L}: \mathbb{R}^{p(p-1)/2}\rightarrow \mathbb{R}^{p(p-1)/2}$, $\bm x \mapsto \mathcal{L}^\ast \mathcal{L} \bm x$. For any $\bm x \in \mathbb{R}^{p(p-1)/2}$, $\mathcal{G} \bm x =\bm M \bm x$ with $\bm M \in \mathbb{R}^{\frac{p(p-1)}{2} \times \frac{p(p-1)}{2}}$ satisfying
\begin{equation*}
M_{kl} =
\begin{cases}
4 \; &\;\; l=k, \\
1 \; & \; \; l \in \left(\Omega_i \cup \Omega_j\right) \backslash k,\\
0 \; & \; \; \text{Otherwise},\\
\end{cases}
\end{equation*}
where $i, j \in [p]$ satisfying $ k=i-j+\frac{j-1}{2}(2p-j)$ and $i>j$, and $\Omega_t$ is an index set defined by
\begin{equation}
  \Omega_t := \left \{ l \in [p(p-1)/2] \, | [\L \bm x]_{tt} = \sum_{l} x_l \right \}, \quad t \in [p].  \nonumber
\end{equation}
Furthermore, we have $\lambda_{\min} (\bm M) = 2$ and $\lambda_{\max} (\bm M) = 2p$.
\end{lemmas}

\vspace*{1pt} 

\begin{lemmas}\label{lem3}
Take $ \lambda = \sqrt{4 \alpha c_0^{-1} \log p /n}$ and suppose $n \geq 94 \alpha c_0^{-1} \lambda_{\max}^2 \left(\L \bm w^{\star} \right) s \log p $ for some $\alpha>2$, where $c_0$ is a constant defined in Lemma \ref{lem8}. Let 
\begin{align}
\hat{\bm w} = \arg \min_{\bm w \geq \bm 0} - \log \det(\L \bm w + \bm J) + \tr{\L \bm w \bm S} + \bm z^{\top} \bm w, \nonumber
\end{align}
where $\bm z $ obeys $0 \leq z_i \leq \lambda$ for $i \in [p(p-1)/2]$. If $\norm{  \L^{\ast} \left( \left( \L \bm w^{\star} + \bm J  \right)^{-1} - \bm S  \right) }_{\max} \leq  \lambda/2  \leq \norm{\bm z_{\E^c}}_{\min}$ holds with the set $\E$ satisfying $\mathcal{S}^\star \subseteq \E$ and $|\E| \leq 2s$, then $\hat{\bm w}$ obeys
\begin{align}
\norm{\L \hat{\bm w} - \L \bm w^{\star}}_{\mathrm{F}} & \leq 2\sqrt{2} \lambda_{\max}^2(\L \bm w^{\star} ) \left( \norm{\bm z_{\mathcal{S}^\star}} + \norm{ \left(  \L^{\ast} \left( \left( \L \bm w^{\star} + \bm J  \right)^{-1} - \bm S  \right) \right)_{\E} } \right) \nonumber \\
& \leq 2 (1+\sqrt{2}) \lambda_{\max}^2 \left(\L \bm w^{\star} \right) \sqrt{s} \lambda, \nonumber
\end{align}
where $\mathcal{S}^\star$ is the support of $\bm w^{\star}$. 
\end{lemmas}

\vspace*{1pt} 

\begin{lemmas}\label{lem4}
Take $ \lambda = \sqrt{4 \alpha c_0^{-1} \log p /n}$ and suppose $n \geq 94 \alpha c_0^{-1} \lambda_{\max}^2 \left(\L \bm w^{\star} \right) s \log p $ for some $\alpha >2$, where $c_0$ is a constant defined in Lemma \ref{lem8}. Define the set $\E^{(k)}$ by
\begin{equation}\label{e-set}
\E^{(k)} = \left\lbrace \mathcal{S}^\star \cup \S^{(k)} \right\rbrace,   \quad \mathrm{with} \quad \S^{(k)} = \left\lbrace i \in [p(p-1)/2] \, | \hat{w}_i^{(k-1)} \geq  b \right\rbrace,
\end{equation} 
where $\hat{\bm w}^{(k)}$ for $k \geq 1$ is defined in \eqref{cost},  $\mathcal{S}^\star$ is the support of $\bm w^{\star}$ with $ \left|\mathcal{S}^\star \right| \leq s$ and $b = (2+\sqrt{2}) \lambda^2_{\max} \left(\L \bm w^{\star} \right) \lambda$ is a constant. Under Assumption \ref{assumption 1}, if $ \norm{ \L^{\ast} \left( \left( \L \bm w^{\star} + \bm J  \right)^{-1} - \bm S  \right) }_{\max} \leq \lambda/2$ holds and $\hat{\bm w}^{(0)}$ satisfies $ \left| \mathrm{supp}^+ \left(\hat{\bm w}^{(0)} \right) \right| \leq s$, then $\E^{(k)}$ obeys
$\left| \E^{(k)} \right| \leq 2s $, for any $k \geq 1$.
\end{lemmas}


\begin{lemmas}\label{lem5}
Take $ \lambda = \sqrt{4 \alpha c_0^{-1} \log p /n}$ and suppose $n \geq 94 \alpha c_0^{-1} \lambda_{\max}^2 \left(\L \bm w^{\star} \right) s \log p $ for some $\alpha >2$, where $c_0$ is a constant defined in Lemma \ref{lem8}. Under Assumptions \ref{assumption 1} and \ref{assumption 2}, if $ \norm{ \L^{\ast} \left( \left( \L \bm w^{\star} + \bm J  \right)^{-1} - \bm S  \right) }_{\max} \leq \lambda/2 $ holds and $\hat{\bm w}^{(0)}$ satisfies $ \left| \mathrm{supp}^+ \left(\hat{\bm w}^{(0)} \right) \right| \leq s$, then for any $k \geq 1$, $\hat{\bm w}^{(k)}$ defined in \eqref{cost} obeys
\begin{equation}
 \norm{ \hat{\bm{w}}^{(k)} - \bm w^{\star}} \leq 2 \lambda_{\max}^2 \left(\L \bm w^{\star} \right) \norm{ \left(  \L^{\ast} \left( \left( \L \bm w^{\star} + \bm J  \right)^{-1} - \bm S  \right) \right)_{\mathcal{S}^\star} } + \frac{3}{2+\sqrt{2}} \norm{ \hat{\bm w}^{(k-1)} - \bm w^{\star} },  \nonumber
\end{equation}
and
\begin{equation}
\norm{ \L \hat{\bm w}^{(k)} - \L \bm w^{\star} }_{\mathrm{F}} \leq 2\sqrt{2} \lambda_{\max}^2 \left(\L \bm w^{\star} \right) \norm{ \left(  \L^{\ast} \left( \left( \L \bm w^{\star} + \bm J  \right)^{-1} - \bm S  \right) \right)_{\mathcal{S}^\star} } + \frac{3}{2+\sqrt{2}} \norm{ \L \hat{\bm w}^{(k-1)} - \L \bm w^{\star} }_{\mathrm{F}}.  \nonumber
\end{equation} 
\end{lemmas}

\vspace*{1pt} 

\begin{lemmas}\label{lem9} 
Take $ \lambda = \sqrt{4\alpha c_0^{-1} \log p /n} $ and suppose $n \geq 8 \alpha \log p$ for some $\alpha>2$, where $c_0$ is a constant defined in Lemma \ref{lem8}. Then one has
\begin{equation}
\mathbb{P} \left[ \norm{ \L^{\ast} \left( \left( \L \bm w^{\star} + \bm J  \right)^{-1} - \bm S  \right) }_{\max} \leq \lambda/2 \right ] \geq 1-1/p^{\alpha-2}. \nonumber
\end{equation} 
\end{lemmas}

\vspace*{1pt}

\begin{lemmas}\label{lem8}
Consider a zero-mean random vector $\bm x = [x_1, \ldots, x_p]^{\top} \in \mathbb{R}^p$ is a LGMRF with precision matrix $\L \bm w^{\star} \in \S_L$. Given $n$ i.i.d samples $\bm x^{(1)}, \ldots, \bm x^{(n)}$, the associated sample covariance matrix $\bm S = \frac{1}{n} \sum_{k=1}^n \bm x^{(k)} \left( \bm x^{(k)} \right)^\top$ satisfies, for $t \in [0, t_0]$, 
\begin{equation}
 \mathbb{P}\left [ \, \left | [\L^{\ast} \bm S]_i -  \left( \L^{\ast} \left(\L \bm w^{\star} + \bm J \right)^{-1} \right)_i \right | \geq t \right ] 
 \leq 2\exp  \left( - c_0 nt^2 \right),  \quad  \mathrm{for} \ i \in [p(p-1)/2], \nonumber
\end{equation}  
where $t_0 = \norm{\L^{\ast} \left(\L \bm w^{\star} + \bm J \right)^{-1}}_{\max}$ and $c_0 = 1 /\left( 8 \norm{\L^{\ast} \left(\L \bm w^{\star} + \bm J \right)^{-1}}_{\max}^2 \right)$ are two constants.
\end{lemmas}

\vspace*{1pt} 

\begin{lemmas}\label{lem12}
Let $f(\bm w) = -\log \det (\L \bm w + \bm J)$. Define a local region of $\bm w^{\star}$ by
\begin{equation}
\mathcal{B} \left(\bm w^{\star};r \right)= \left\lbrace \bm w | \bm w \in \mathbb{B} \left(\bm w^{\star}; r \right) \cap \mathcal{S}_{\bm w} \right\rbrace, \nonumber
\end{equation}
where $\mathbb{B} \left(\bm w^{\star};r \right) = \left\lbrace \bm w \in \mathbb{R}^{p(p-1)/2} \, | \norm{ \bm w - \bm w^{\star}} \leq r \right\rbrace$, and $\mathcal{S}_{\bm w} = \left\lbrace \bm w \, | \bm w \geq \bm 0, (\L\bm w + \bm J) \in \S_{++}^p \right\rbrace$. Then, under Assumption \ref{assumption 2}, $g(\bm w)$ is $\frac{2}{ \left(1+\delta^{-1} \right)^2\tau^2}$-strongly convex and 
$\frac{2p \tau^2}{ \left(1-\delta^{-1} \right)^2}$-smooth in the region $\mathcal{B} \left(\bm w^{\star}; \frac{1}{\sqrt{2p}\delta \tau} \right)$ where $\tau$ is defined in \eqref{tau-def} and $\delta > 1$. In other words, for any $\bm w_1, \bm w_2 \in \mathcal{B} \left(\bm w^{\star}; \frac{1}{\sqrt{2p}\delta \tau} \right)$, we have
\begin{equation}
\frac{1}{ \left(1+\delta^{-1} \right)^2 \tau^2} \norm{\bm w_2 - \bm w_1}^2 \leq f(\bm w_2) - f(\bm w_1) - \langle \nabla f(\bm w_1), \bm w_2 - \bm w_1 \rangle \leq \frac{p \tau^2}{ \left(1-\delta^{-1} \right)^2} \norm{\bm w_2 - \bm w_1}^2.   \nonumber
\end{equation}
\end{lemmas}

\vspace*{1pt} 

\begin{lemmas}\label{lem2}
Let $f(\bm w) = -\log \det (\L \bm w + \bm J)$. Define a local region of $\bm w^{\star}$ by
\begin{equation}
\mathcal{B}_{\bm M} \left(\bm w^{\star};r \right)= \left\lbrace \bm w | \bm w \in \mathbb{B}_{\bm M} \left(\bm w^{\star}; r \right) \cap \mathcal{S}_{\bm w} \right\rbrace,   \nonumber
\end{equation}
where $\mathbb{B}_{\bm M} \left(\bm w^{\star};r \right) = \left\lbrace \bm w \in \mathbb{R}^{p(p-1)/2} \, | \norm{ \bm w - \bm w^{\star}}_{\bm M} \leq r \right\rbrace$, in which $\norm{\bm x}_{\bm M} = \langle \bm x, \bm M \bm x \rangle^{\frac{1}{2}}$ for any $\bm x \in \mathbb{R}^{p(p-1)/2}$ with $\bm M$ defined in Lemma \ref{lem11}, and $\mathcal{S}_{\bm w} = \left\lbrace \bm w \, | \bm w \geq \bm 0, (\L\bm w + \bm J) \in \S_{++}^p \right\rbrace$. Then for any $\bm w_1, \bm w_2 \in \mathcal{B}_{\bm M}  \left(\bm w^{\star}; r \right)$, we have
\begin{equation}
\left\langle \nabla f(\bm w_1) - \nabla f(\bm w_2), \bm w_1 - \bm w_2 \right \rangle \geq  \left( \norm{\L \bm w^{\star} }_2 + r \right)^{-2} \norm{\mathcal{L} \bm w_1 - \mathcal{L}\bm w_2}_{\mathrm{F}}^2.  \nonumber
\end{equation}
\end{lemmas}

\vspace*{1pt}

\begin{lemmas}\citep{willoughby1977inverse} \label{lem15}
Suppose a positive matrix $\bm A \in \mathbb{R}^{p \times p}$ is diagonally scaled such that $A_{ii} =1$, $i=1, \ldots, p$, and $0 < A_{ij} <1$, $i \neq j$. Let $y$ and $x$ be the lower and upper bounds satisfying
\begin{equation}
0 < y \leq A_{ij} \leq x < 1, \quad \forall \ i \neq j,   \nonumber
\end{equation}
and define $s$ by
\begin{equation}
x^2 = s y + (1-s)y^2.    \nonumber
\end{equation}
Then the inverse matrix of $\bm A$ exists and $\bm A$ is an inverse $M$-matrix if $s^{-1} \geq p-2$ with $p>3$.
\end{lemmas}

\vspace*{1pt} 

\begin{lemmas}\citep{wainwright_2019} \label{lem16}
(Sub-exponential tail bound) Suppose $X$ is sub-exponential with parameters $(\upsilon, \alpha)$. Then 
\begin{equation*}
\mathbb{P}[X-\mu \geq t] \leq \left\{
\begin{array}{lcl}
e^{-\frac{t^2}{2\upsilon^2}} & & t \in [0, \frac{\upsilon^2}{\alpha}], \\
e^{-\frac{t}{2\alpha}}   &  & t \in \left(\frac{\upsilon^2}{\alpha}, + \infty \right).
\end{array}
\right.
\end{equation*}
\end{lemmas}

\subsection{Proofs of Theorem \ref{Theorem 1} and Corollary \ref{corollary-L1}}\label{sec-prof-L1-norm}
\begin{proof}
As a result of Theorem \ref{Theorem 2}, the optimization \eqref{Lap-est-l1} can be equivalently written as  
\begin{equation}
\min_{\bm w \geq \bm 0} -\log \det(\L \bm w + \bm J) + \tr{\L \bm w \bm S} + \lambda \norm{\bm w}_1.  \label{qqL1}
\end{equation}
Due to the non-negativity constraint $\bm w \geq \bm 0$, \eqref{qqL1} can be further rewritten as
\begin{equation} 
\min_{\bm w \geq \bm 0} -\log \det(\L \bm w + \bm J) + \langle \L^{\ast} \bm S + \lambda \bm 1, \bm w \rangle,    \label{L1}
\end{equation}
where $\bm 1 = [1, \ldots, 1]^{\top}$. 

We first prove that the optimization \eqref{L1} has one global minimizer if $\lambda>0$. Let $f(\bm w)= -\log \det(\L \bm w + \bm J) + \langle \L^{\ast} \bm S + \lambda \bm 1, \bm w \rangle$. The feasible set of \eqref{L1} is $\mathcal{S}_{\bm w} = \{ \bm w \in \mathbb{R}^{p(p-1)/2} \, | \bm w \geq \bm 0, (\L \bm w + \bm J) \in \mathcal{S}^p_{++} \}$, which is the same with the feasible set of \eqref{new_cost}. For any $\bm w \in \mathcal{S}_{\bm w}$, the minimum eigenvalue of $\nabla^2 f(\bm w)$ can be lower bounded by
\begin{equation*}
\begin{split}
\lambda_{\min}\left( \nabla^2 f(\bm w)\right) & = \inf_{\norm{\bm x}=1} \bm x^\top \nabla^2 f(\bm w) \bm x \\
& = \inf_{\norm{\bm x}=1} \left(\mathrm{vec} (\L \bm x) \right)^{\top} \left( \left( \L \bm w + \bm J \right)^{-1} \otimes \left( \L \bm w + \bm J \right)^{-1} \right)  \mathrm{vec} (\L \bm x) \\
& \geq \inf_{\norm{\bm x} =1} \frac{ \left(\mathrm{vec} (\L \bm x) \right)^{\top}  \left( (\L \bm w + \bm J)^{-1} \otimes (\L \bm w + \bm J)^{-1} \right) \mathrm{vec} (\L \bm x)}{\left( \mathrm{vec} (\L \bm x) \right)^{\top} \mathrm{vec} (\L \bm x)} \cdot \inf_{\norm{\bm x} =1} \norm{\L \bm x}_{\mathrm{F}}^2 \\
&\geq \lambda_{\min} \left( (\L \bm w + \bm J)^{-1} \otimes (\L \bm w + \bm J)^{-1}\right) \cdot \inf_{\norm{\bm x} =1} \norm{\L \bm x}_{\mathrm{F}}^2\\
&= \lambda_{\min} \left( (\L \bm w + \bm J)^{-1} \otimes (\L \bm w + \bm J)^{-1}\right) \cdot \inf_{\norm{\bm x} =1} \bm x^\top \bm M \bm x\\
& = 2 \lambda_{\min}  \left( (\L \bm w + \bm J)^{-1}\right)^2 \\
& >0,
\end{split}
\end{equation*}
where the second equality is due to Lemma \ref{lem6}; the third equality follows from Lemma \ref{lem11};  the last equality follows from the property of Kronecker product that the eigenvalues of $\bm A \otimes \bm B$ are $\lambda_i \mu_j$ for $i, j \in [p]$, where $\lambda_i$ and $\mu_j$ are the eigenvalues of $\bm A \in \mathbb{R}^{p \times p}$ and $\bm B \in \mathbb{R}^{p \times p}$, respectively, and $\lambda_{\min} (\bm M) = 2$ following from Lemma \ref{lem11}; the last inequality follows from the fact that $\bm w \in \mathcal{S}_{\bm w}$. Therefore, the optimization \eqref{L1} is strictly convex, and thus \eqref{L1} has at most one global minimizer.

The existence of minimizers of \eqref{L1} can be guaranteed by the coercivity of $f(\bm w)$. The function $f(\bm w)$ can be lower bounded by 
\begin{equation}\label{f_lower}
\begin{split}
f(\bm w) &= - \log \left(\prod_{i=1}^p \lambda_i(\L \bm w + \bm J) \right) + \left\langle \L^{\ast} \bm S + \lambda \bm 1, \bm w \right\rangle \\
& = - \log \left( \prod_{i=2}^p\lambda_i(\L \bm w) \right) + \langle \L^{\ast} \bm S + \lambda \bm 1, \bm w \rangle  \\
& \geq -(p-1)\log\left(\sum_{i=1}^p\lambda_i(\L \bm w)\right) + \langle \L^{\ast} \bm S + \lambda \bm 1, \bm w \rangle + (p-1) \log(p-1)  \\
& = -(p-1)\log\left(\sum_{i=1}^p [\L \bm w]_{ii} \right) + \langle \L^{\ast} \bm S + \lambda \bm 1, \bm w \rangle + (p-1) \log(p-1) \\
& = -(p-1)\log\left(2\sum_{t=1}^{p(p-1)/2} w_t \right) + \langle \L^{\ast} \bm S + \lambda \bm 1, \bm w \rangle + (p-1) \log(p-1)   \\
& \geq -(p-1)\log\left(\sum_{t=1}^{p(p-1)/2} w_t \right) + \lambda \sum_{t=1}^{p(p-1)/2} w_t + (p-1) \log\frac{p-1}{2}, 
\end{split}
\end{equation}
where the second equality follows from \eqref{Lw_J} with $b=1$; the forth equality follows from $\L \bm w \cdot \bm 1=\bm 0$; the last inequality holds because $\bm w\geq \bm0$, and $\L^{\ast} \bm S \geq \bm 0$, which follows from \eqref{q19}; the first inequality holds because the smallest eigenvalue $\lambda_1(\L \bm w) =0$ and
\begin{equation}
\frac{a_1+ a_2+\ldots + a_n}{n} \geq \sqrt[n]{a_1 \cdot a_2 \cdots a_n}   \nonumber
\end{equation}
holds for any non-negative real numbers of $a_1, \ldots, a_n$. A function $g: \Omega \to \mathbb{R} \cup \{ + \infty \}$ is called coercive over $\Omega$, if every sequence $\bm x_k \in \Omega$ with $\norm{\bm x_k} \to +\infty$ obeys $\lim_{k\to \infty} g(\bm x_k) = +\infty$, where $\Omega \subset \mathbb{R}^n$. Let
\begin{equation}
h(z) = -(p-1)\log z + \lambda z + (p-1) \log\frac{p-1}{2}.   \nonumber
\end{equation}
A simple calculation yields $\lim_{z \to +\infty} h(z) = +\infty$ if $\lambda>0$. For any sequence $\bm w_k \in \text{cl}(\mathcal{S}_{\bm w})$ with $\norm{\bm w_k} \to +\infty$, where $\text{cl}(\mathcal{S}_{\bm w})$ is the closure of $\mathcal{S}_{\bm w}$, one has $\sum_{t=1}^{p(p-1)/2} [\bm w_k]_t \to +\infty$, because $\sum_{t=1}^{p(p-1)/2} [\bm w_k]_t \geq \norm{\bm w_k}$. Then one obtains
\begin{equation} 
\lim_{k \to \infty} f(\bm w_k) \geq \lim_{k \to \infty} h \left(\sum_{t=1}^{p(p-1)/2} [\bm w_k]_t \right) = \lim_{z \to +\infty} h(z) = +\infty,  \nonumber
\end{equation}
where the first inequality follows from \eqref{f_lower}. Hence, $f(\bm w)$ is coercive over $\mathrm{cl}(\mathcal{S}_{\bm w})$. Following from the Extreme Value Theorem \citep{drabek2007methods}, if $\Omega \subset \mathbb{R}^n$ is non-empty and closed, and $g: \Omega \to \mathbb{R} \cup \{ + \infty \}$ is lower semi-continuous and coercive, then the optimization $\min_{\bm x \in \Omega} g(\bm x)$ has at least one global minimizer. Therefore, by the coercivity of $f(\bm w)$, \eqref{L1} has at least one global minimizer in $\mathrm{cl}(\mathcal{S}_{\bm w})$.

Let $\Omega_A = \{ \bm w \in \mathbb{R}^{p(p-1)/2} \, | \bm w \geq \bm 0 \}$ and $\Omega_B= \left\lbrace \bm w \in \mathbb{R}^{p(p-1)/2} \, | (\L \bm w + \bm J) \in \mathcal{S}^p_{++} \right\rbrace$. $\Omega_A$ is a closed set and $\Omega_B$ is an open set. Then $\mathcal{S}_{\bm w}$ can be rewritten as $\mathcal{S}_{\bm w} = \Omega_A \cap \Omega_B$. Consider the set $V := \mathrm{cl}(\mathcal{S}_{\bm w}) \setminus \mathcal{S}_{\bm w}$, we have
\begin{equation}
V \subseteq \left\lbrace \mathrm{cl}(\Omega_A) \cap \mathrm{cl}(\Omega_B) \right\rbrace \setminus \left\lbrace \Omega_A \cap \Omega_B \right\rbrace = \Omega_A \cap \partial \Omega_B, 
\end{equation}
where $\partial \Omega_B$ is the boundary of $\Omega_B$. Notice that every matrix on the boundary of the set of positive definite matrices is positive semi-definite and has zero determinant. Hence, one has $\partial \Omega_B = \left\lbrace \bm w \in \mathbb{R}^{p(p-1)/2} \, | (\L \bm w + \bm J) \in \mathcal{S}^p_{+}, \, \det (\L \bm w + \bm J)=0 \right\rbrace$. As a result, for any $\bm w_b \in \mathrm{cl}(\mathcal{S}_{\bm w}) \setminus \mathcal{S}_{\bm w}$, $f(\bm w_b) = +\infty$. Therefore, \eqref{L1} has at least one global minimizer and the minimizer must belong to the set $\mathcal{S}_{\bm w}$. On the other hand, by the strict convexity of $f(\bm w)$, \eqref{L1} has at most one global minimizer in $\mathcal{S}_{\bm w}$. Totally, we conclude that \eqref{L1} has an unique global minimizer in $\mathcal{S}_{\bm w}$ if $\lambda>0$.
 
We prove the theorem through the KKT conditions. The Lagrangian of the optimization \eqref{L1} is
\begin{equation}
L(\bm w, \bm \upsilon) = - \log \det(\L \bm w + \bm J) + \langle \L^{\ast} \bm S + \lambda \bm 1, \bm w \rangle - \bm \upsilon^{\top} \bm w,   \nonumber
\end{equation}
where $\bm \upsilon$ is a KKT multiplier. Let $ \left(\hat{\bm w}, \hat{\bm \upsilon} \right)$ be any pair of points that satisfies the KKT conditions. Then we have
\begin{align}
-\L^{\ast} \left(  \left(\L \hat{\bm w} + \bm J \right)^{-1} \right) + \L^{\ast} \bm S + \lambda \bm 1 - \hat{\bm \upsilon} = \bm 0&; \label{kkt1-t1}\\
\hat{w}_i \hat{\upsilon}_i =0, \forall i \in [p(p-1)/2] &; \label{kkt2-t1}\\
\hat{\bm w} \geq \bm 0, \ \hat{\bm \upsilon} \geq \bm 0&. \label{kkt3-t1}
\end{align}
As we know, for any convex optimization with differentiable objective and constraint functions, any point that satisfies the KKT conditions (under Slater's constraint qualification) must be primal and dual optimal. Therefore, $\hat{\bm w}$ must obey $\hat{\bm w} = \arg \min_{\bm w \geq \bm 0} f(\bm w)$. Note that the pair of points $ \left(\hat{\bm w}, \hat{\bm \upsilon} \right)$ that satisfies the KKT conditions is unique. To prove the optimal solution $\hat{\bm w}>\bm 0$ holds for \eqref{L1}, we can equivalently prove that the KKT conditions \eqref{kkt1-t1}-\eqref{kkt3-t1} hold for $ \left(\hat{\bm w}, \hat{\bm \upsilon} \right)$ with $\hat{\bm w}>\bm 0$ and $\hat{\bm \upsilon}=\bm 0$. It is further equivalent to prove that
\begin{equation}\label{q40}
\L^{\ast} \left( (\L \hat{\bm w} + \bm J)^{-1} \right)   = \L^{\ast} \bm S + \lambda \bm 1
\end{equation}
holds for $\hat{\bm w}>\bm 0$. Following from Lemma \ref{lem18} with the fact that $\hat{\bm w} \in \mathcal{S}_{\bm w}$, there must exist an unique $\bm x$ such that
\begin{equation}\label{qq40}
\L \bm x + \frac{1}{b} \bm J = \left(\L \hat{\bm w} + b \bm J \right)^{-1}
\end{equation}
holds for any $b \neq 0$. Thus one has
\begin{equation}\label{q41}
\L^{\ast} \left(  \left(\L \hat{\bm w} + \bm J \right)^{-1} \right) = \L^{\ast} \left( \L \bm x + \bm J \right) = \L^{\ast} \L \bm x, 
\end{equation}
where the first equality follows from \eqref{qq40} with $b=1$; the second equality holds because $\bm J \in \mathcal{N} (\L^{\ast})$ where $\mathcal{N} \left(\L^{\ast} \right)$ is the null space  of $\L^{\ast}$ defined by $\mathcal{N} (\L^{\ast}) := \left\lbrace \bm X \in \mathbb{R}^{p \times p} \; | \; \L^{\ast} \bm X = \bm 0 \right\rbrace$. Combining \eqref{q40} and \eqref{q41} yields
\begin{equation}\label{qq41}
\bm x = \left(\L^{\ast} \L \right)^{-1} \left(\L^{\ast} \bm S + \lambda\bm 1 \right), 
\end{equation}
where $\L^\ast \L$ is invertible according to Lemma \ref{lem11}. Recall that $\bm S$ is the sample covariance matrix defined by
\begin{equation}
\bm S = \sum_{k=1}^n \bm x^{(k)}\left( \bm x^{(k)} \right)^{\top},   \nonumber
\end{equation}
where $\bm x^{(1)}, \ldots, \bm x^{(n)}$ are the samples independently drawn from LGMRF in Definition \ref{def-LGMRF}. According to the density function of LGMRF defined in \eqref{dens-LGMRF}, we get $\bm 1^\top \bm x^{(k)}  = 0$ for $k = 1, \ldots, n$. Therefore, $\bm S$ is symmetric and obeys $\bm S \cdot \bm 1 = \bm 0$. It is easy to verify that $\bm S \in \mathcal{R}(\L)$, where $\mathcal{R}(\L)$ is the range space of $\L$ defined by $\mathcal{R}(\L) := \left\lbrace \L \bm y \; | \; \bm y \in \mathbb{R}^{p(p - 1)/2} \right\rbrace$. Hence, there must exist a $\bm y \in \mathbb{R}^{p(p - 1)/2} $ such that $\bm S = \L \bm y$. Thus $\L^{\ast} \bm S = \L^{\ast} \L \bm y$. One further obtains 
\begin{equation}
\bm y = \left(\L^{\ast} \L \right)^{-1} \L^{\ast} \bm S.   \nonumber
\end{equation} 
Then a simple calculation yields
\begin{equation}\label{L-range}
\L  \left(\L^{\ast} \L \right)^{-1} \L^{\ast} \bm S = \L \bm y = \bm S.
\end{equation}

Next, we construct a matrix $\bm X = \L \bm x + a\bm J$ with $a>0$ and have
\begin{equation}\label{q42}
\begin{split}
\L \bm x + a\bm J & = \L \left(\L^{\ast} \L \right)^{-1} (\L^{\ast} \bm S + \lambda \bm 1) + a\bm J   \\  
& = \bm S + \lambda \L \left(\L^{\ast} \L \right)^{-1} \bm 1 + a\bm J     \\    
& = \bm S + \frac{\lambda}{2p} \L \bm 1 + a \bm J, 
\end{split}
\end{equation}
where the first equality follows from \eqref{qq41}; the second equality follows from \eqref{L-range}; the third equality holds because $\L^{\ast} \L \bm 1 = 2p \bm 1$ and then one has $ \left(\L^{\ast} \L \right)^{-1} \bm 1 = \frac{1}{2p} \bm 1$.

Let $\hat{\bm X} = \bm D^{-\frac{1}{2}} \bm X \bm D^{-\frac{1}{2}}$ be the normalized matrix of $\bm X$, where $\bm D$ is a diagonal matrix containing the diagonals of $\bm X$. Notice that each diagonal element of $\hat{\bm X}$ is $1$. Next, we will prove that, under some conditions, $\hat{\bm X}$ is an inverse $M$-matrix, that is, $ \big(\hat{\bm X} \big)^{-1}$ is a $M$-matrix. We say $\bm A \in \mathbb{R}^{p \times p}$ is an $M$-matrix if
\begin{equation}
\bm A = s \bm I - \bm B, \label{def-Matrix}
\end{equation}
where $\bm B \in \mathbb{R}^{p \times p}$ is an element-wise non-negative matrix and $s > \rho (\bm B)$, the spectral radius of $\bm B$. According to \eqref{q42}, one has
\begin{equation}
X_{ij} =S_{ij} - \frac{\lambda}{2p} + \frac{a}{p},  \quad \mathrm{for} \ i \neq j,    \nonumber
\end{equation}
and 
\begin{equation}
X_{ii} = S_{ii} + \frac{p-1}{2p}\lambda + \frac{a}{p}, \quad i = 1, \ldots, p.      \nonumber
\end{equation}
Define $\widetilde{S}_{ij} = \max_{i \neq j} S_{ij}$, $\bar{S}_{ij} = \min_{i \neq j} S_{ij}$, $\widetilde{S}_{kk} = \max_{k} S_{kk}$ and $\bar{S}_{kk} = \min_{k} S_{kk}$. By the definition of $\hat{\bm X}$, the lower bound $y$ and upper bound $x$ of the elements off the diagonal of $\hat{\bm X}$ can be obtained as below,
\begin{equation}\label{q43}
\begin{split}
\hat{X}_{ij} & = \frac{X_{ij}}{\sqrt{X_{ii} X_{jj}}} = \frac{S_{ij} - \frac{\lambda}{2p} + \frac{a}{p}}{\sqrt{S_{ii} + \frac{p-1}{2p}\lambda + \frac{a}{p}} \cdot \sqrt{S_{jj} + \frac{p-1}{2p}\lambda + \frac{a}{p}}}         \\
& \geq \frac{\bar{S}_{ij} - \frac{\lambda}{2p} + \frac{a}{p}}{\widetilde{S}_{kk} + \frac{p-1}{2p}\lambda + \frac{a}{p}} =: y, \quad \forall \ i \neq j,
\end{split}
\end{equation}
and
\begin{equation}
\hat{X}_{ij} \leq \frac{\widetilde{S}_{ij} - \frac{\lambda}{2p} + \frac{a}{p}}{\bar{S}_{kk} + \frac{p-1}{2p}\lambda + \frac{a}{p}} =: x, \quad \forall \ i \neq j.  \label{q47}
\end{equation}
Define $s$ by $x^2 = s y + (1-s)y^2$. According to Lemma \ref{lem15}, if $0 < y \leq x <1$ and $s^{-1} \geq p-2$ with $p>3$, then $\hat{\bm X}$ is an inverse $M$-matrix. We can see, provided that $y=\frac{1}{p+1}$ and the inequalities
\begin{equation}
0 < y \leq x \leq \sqrt{2}y <1 \label{q46}
\end{equation}
hold, then one has
\begin{equation}
s^{-1} = \frac{y - y^2}{x^2 - y^2} \geq \frac{1 -y}{y} > p-2,    \nonumber
\end{equation}
where the first inequality follows from $x \leq \sqrt{2}y$. Therefore, if $y=\frac{1}{p+1}$ and \eqref{q46} holds, then $\hat{\bm X}$ is an inverse $M$-matrix.

Next, we will prove that if $a=\widetilde{S}_{kk} - (p+1) \bar{S}_{ij} +\lambda$ with $\lambda\geq 2(\sqrt{2}+1)(p+1)\left(\widetilde{S}_{kk} -\bar{S}_{ij} \right)$, then $y=\frac{1}{p+1}$ and \eqref{q46} holds. Substituting $a=\widetilde{S}_{kk} - (p+1) \bar{S}_{ij} +\lambda$ into \eqref{q43} yields $y=\frac{1}{p+1}$. Then it is clear that $y>0$ and $\sqrt{2}y <1$. By comparing $x$ and $y$ defined in \eqref{q43} and \eqref{q47}, respectively, one has $y \leq x$. A simple algebra yields
\begin{equation*}
\begin{split}
x & = \frac{\widetilde{S}_{ij} - \frac{\lambda}{2p} + \frac{a}{p}}{\bar{S}_{kk} + \frac{p-1}{2p}\lambda + \frac{a}{p}}  \leq  \frac{p\widetilde{S}_{kk} - \frac{\lambda}{2} + a}{p\bar{S}_{ij} + \frac{p-1}{2}\lambda + a}  \\
& = \frac{(p+1) \left(\widetilde{S}_{kk} - \bar{S}_{ij} \right) + \frac{\lambda}{2}}{\widetilde{S}_{kk} - \bar{S}_{ij} + \frac{(p+1)\lambda}{2}} \leq  \frac{\sqrt{2}}{p+1} =\sqrt{2}y,   
\end{split}
\end{equation*}
where the first inequality follows from $\widetilde{S}_{ij} \leq \widetilde{S}_{kk}$ and $\bar{S}_{kk} \geq \bar{S}_{ij}$ because $\bm S \in \mathcal{S}_+^p$ and $\bm S \cdot \bm 1 = \bm 0$. It is easy to verify that $\lambda\geq 2(\sqrt{2}+1)(p+1)\left(\widetilde{S}_{kk} -\bar{S}_{ij} \right)$ is large enough to establish the second inequality. Therefore, all the inequalities in \eqref{q46} hold.

Consequently, by Lemma \ref{lem15}, we conclude that $\hat{\bm X}$ is an inverse $M$-matrix when $a=\widetilde{S}_{kk} - (p+1) \bar{S}_{ij} +\lambda$ with $\lambda\geq 2(\sqrt{2}+1)(p+1) \left(\widetilde{S}_{kk} -\bar{S}_{ij} \right)$. Therefore, $\hat{\bm X}^{-1} = \bm D^{\frac{1}{2}} \bm X^{-1} \bm D^{\frac{1}{2}}$ is an $M$-matrix. Notice that the elements off the diagonal of an $M$-matrix are non-positive according to \eqref{def-Matrix}. As a result, the elements off the diagonal of $\bm D^{\frac{1}{2}} \bm X^{-1} \bm D^{\frac{1}{2}}$ are non-positive, implying that the elements off the diagonal of $\bm X^{-1}$ are also non-positive, because $\bm X = \L \bm x + a\bm J$ is positive definite and thus the diagonal elements of $\bm D$ are positive. The application of \eqref{qq40} with $b = \frac{1}{a}$ yields
\begin{equation}  
\bm X^{-1} = \left(\L \bm x +a \bm J \right)^{-1}=\L \hat{\bm w} + \frac{1}{a}\bm J.   \nonumber
 \end{equation}
One further obtains
\begin{equation}  
\left[\L \hat{\bm w}  + \frac{1}{a}\bm J \right]_{ij} = - \hat{w}_k + \frac{1}{ap} \leq 0,   \quad  \forall \ i \neq j,     \nonumber     
\end{equation}
where $k = i -j + \frac{j-1}{2}(2p-j)$. Therefore, we establish 
\begin{equation}  
\hat{w}_k \geq \frac{1}{ap} = \frac{1}{ \left(\widetilde{S}_{kk} - (p+1) \bar{S}_{ij} +\lambda \right) p} >0, \quad \forall \ k,         \nonumber
\end{equation}
concluding that \eqref{q40} holds for $\hat{\bm w}>\bm 0$. Note that $a=\widetilde{S}_{kk} - (p+1) \bar{S}_{ij} +\lambda > 0$ because $\widetilde{S}_{kk}>0, \lambda>0$, and $\bar{S}_{ij} \leq 0$, where $\bar{S}_{ij} \leq 0$ following from $\bm S \cdot \bm 1 =\bm 0$ and each diagonal element $S_{ii}\geq 0$ since $\bm S$ is positive semi-definite.

Finally, we prove the Corollary \ref{corollary-L1}. Recall that $\tilde{\bm{\Theta}}$ is the optimal solution of \eqref{Lap-est-app}. Let $\mathcal{L} \tilde{\bm w} = \tilde{\bm{\Theta}}$. Then we have
\begin{equation}\label{Lap-est-app-sol}
\tilde{\bm w}= \arg \min_{ \bm w \geq \bm 0} - \log {\det} ( \mathcal{L} \bm w + \bm J) + \lambda \sum_k w_k.
\end{equation}
Similar to \eqref{L1}, the optimization \eqref{Lap-est-app-sol} has a unique optimal solution. We can check that the pair $(\tilde{\bm w}, \tilde{\bm \upsilon} )$ with $\tilde{\bm w} = \frac{2}{p \lambda} \bm 1$ and $\tilde{\bm \upsilon} = \bm 0$ satisfies the KKT conditions of \eqref{Lap-est-app-sol} as below:
\begin{align}
-\L^{\ast} \left( (\L \tilde{\bm w} + \bm J)^{-1} \right) +  \lambda \bm 1 - \tilde{\bm \upsilon} = \bm 0&; \nonumber \\
\tilde{w}_i \tilde{\upsilon}_i =0, \forall i \in [p(p-1)/2] &; \nonumber \\
\tilde{\bm w} \geq \bm 0, \ \tilde{\bm \upsilon} \geq \bm 0&. \nonumber
\end{align}
Therefore, $\tilde{W}_{ij} = \tilde{w}_k = \frac{2}{p \lambda} $ with $i, j \in [p]$, and $ k=i-j+\frac{j-1}{2}(2p-j)$ for $i>j$.
\end{proof}

\subsection{Proof of Theorem \ref{Theorem 2} } \label{sec-prof-operator}
\begin{proof}
Let $\bm x \in \mathbb{R}^{p(p-1)/2}$. According to the definition of $\mathcal{L}$ in \eqref{operator}, $\mathcal{L} \bm x$ must obey $[\mathcal{L} \bm x]_{ij} =[\mathcal{L} \bm x]_{ji}$, for any $i\neq j$ and $(\mathcal{L} \bm x) \cdot \bm 1 =\bm 0$. 

Next, we will show that $\mathcal{L} \bm x$ is positive semi-definite for any $\bm x \geq \bm 0$ by the Gershgorin circle theorem \citep{varga2004gershgorin}. Given a matrix $\bm X \in \R^{p \times p}$ with entries $X_{ij}$. Let $R_i(\bm X) = \sum_{j \neq i} |X_{ij}|$ be the sum of the absolute values of the non-diagonal entries in the $i$-th row. Then a Gershgorin disc is the disc $D(X_{ii},R_i(\bm X))$ centered at $X_{ii}$ on the complex plane with radius $R_i(\bm X)$. Gershgorin circle theorem \citep{gershgorin1931uber} shows that each eigenvalue of $\bm X$ lies within at least one of the Gershgorin discs. For any $\bm x \geq \bm 0$, $R_i(\mathcal{L} \bm x) = [\mathcal{L} \bm x]_{ii}$ holds for each $i$ because $(\mathcal{L} \bm x) \cdot \bm 1 =\bm 0$ and $[\mathcal{L} \bm x]_{ij} \leq 0$ for any $i \neq j$. For any given eigenvalue $\lambda$ of $\mathcal{L} \bm x$, by Gershgorin circle theorem, there must exist one Gershgorin disc $D \left([\mathcal{L} \bm x]_{ii},R_i(\mathcal{L} \bm x) \right)$ such that 
\begin{equation} \label{Gershgorin}
|\lambda-[\mathcal{L} \bm x]_{ii}| \leq R_i(\bm X)=[\mathcal{L} \bm x]_{ii},
\end{equation}
indicating that $\lambda \geq 0$. Note that the eigenvalues of $\mathcal{L} \bm x$ are real since $\mathcal{L} \bm x$ is symmetric. Therefore, one has
\begin{equation}
\mathcal{L} \bm x \in \S_{+}^p, \quad \forall \bm x \geq  \bm 0.  \label{Lx-psd}
\end{equation}
Finally, we will prove that $\mathrm{rank}(\mathcal{L} \bm x) = p-1 \Leftrightarrow ( \mathcal{L} \bm x + \bm J) \in \S^{p}_{++}$, for any $\bm x \geq \bm 0$. On one hand, if $\mathrm{rank}(\mathcal{L} \bm x) = p-1$, then $ \mathcal{L} \bm x + \bm J$ admits the eigenvalue decomposition $\bm U \bm \Lambda \bm U^{\top}$, where $\bm U = \left[\bm U_s \ \frac{1}{\sqrt{p}}\bm 1 \right]$ and $\bm \Lambda$ is a diagonal matrix with the diagonal elements $[\lambda_2, \ldots, \lambda_{p}, 1]$. Here $\lambda_{i=2}^{p}$ are the nonzero eigenvalues of $\mathcal{L} \bm x$ and $\bm U_s$ is a $p \times (p-1)$ matrix whose columns are the corresponding eigenvectors of $\mathcal{L} \bm x$. Note that the nonzero eigenvalue $\lambda_{i=2}^{p} >0$ because $\mathcal{L} \bm x \in \S_{+}^p$. Therefore, one has $( \mathcal{L} \bm x + \bm J) \in \S^{p}_{++}$. On the other hand, if $( \mathcal{L} \bm x + \bm J) \in \S^{p}_{++}$, then $\mathrm{rank}(\mathcal{L} \bm x) \geq \mathrm{rank}(\mathcal{L} \bm x + \bm J) - \mathrm{rank}(\bm J) =p-1$ because $\mathcal{L} \bm x + \bm J$ is full rank and $\bm J$ is rank one. Furthermore, $\mathrm{rank}(\mathcal{L} \bm x) \leq p-1$ because $(\mathcal{L} \bm x) \cdot \bm 1 = \bm 0$. Therefore, we conclude that $\mathrm{rank}(\mathcal{L} \bm x) = p-1$, completing the proof.
\end{proof} 

\subsection{Proofs of Theorem \ref{theorem 3} and Corollary \ref{corollary}}\label{sec-estimation-error}

\begin{proof}
We first prove Theorem \ref{theorem 3}. Take the regularization parameter $ \lambda = \sqrt{4\alpha c_0^{-1} \log p /n}$ for some $\alpha > 2$, and the sample size 
\begin{equation}
n\geq \max \left(94 \alpha c_0^{-1} \lambda_{\max}^2 \left(\L \bm w^{\star} \right) s \log p, 8 \alpha \log p \right),   \label{sample size}  
\end{equation}
where $c_0$ is a constant defined in Lemma \ref{lem8}. Notice that the sample size $n$ in \eqref{sample size} satisfies the conditions on the number of samples in Lemmas \ref{lem3}, \ref{lem4}, \ref{lem5} and \ref{lem9}.
Recall that the initial point $\hat{\bm w}^{(0)}$ of Algorithm \ref{algo-1} satisfies $\left| \mathrm{supp}^+ \left(\hat{\bm w}^{(0)} \right) \right| \leq s$. 

Define an event $\mathcal{J} = \left\lbrace \norm{ \L^{\ast} \left( \left( \L \bm w^{\star} + \bm J  \right)^{-1} - \bm S  \right) }_{\max} \leq \lambda/2 \right\rbrace$. According to Lemma \ref{lem9}, $ \norm{ \L^{\ast} \left( \left( \L \bm w^{\star} + \bm J  \right)^{-1} - \bm S  \right) }_{\max} \leq \lambda/2$ holds with probability at least $1-1/p^{\alpha-2}$. Under the event $\mathcal{J}$, one applies Lemma \ref{lem5} and obtains, for any $k \geq 1$,
\begin{equation}
 \norm{ \hat{\bm{w}}^{(k)} - \bm w^{\star} } \leq 2 \lambda_{\max}^2 \left(\L \bm w^{\star} \right) \norm{ \left( \L^{\ast} \left( \left( \L \bm w^{\star} + \bm J  \right)^{-1} - \bm S  \right) \right)_{\mathcal{S}^\star} } + \frac{3}{2+\sqrt{2}} \norm{ \hat{\bm w}^{(k-1)} - \bm w^{\star} }.  \nonumber
\end{equation}
By induction, if $d_k \leq a_0 +\rho d_{k-1}$ for any $k \geq 1$ with $\rho \in [0, 1)$, then
\begin{equation}\label{induction}
d_k \leq \frac{1-\rho^{k}}{1-\rho}a_0 + \rho^{k}d_0.         
\end{equation}
Taking $a_0 = 2\lambda_{\max}^2 \left(\L \bm w^{\star} \right) \norm{ \left( \L^{\ast} \left( \left( \L \bm w^{\star} + \bm J  \right)^{-1} - \bm S  \right) \right)_{\mathcal{S}^{\star}} }  $, $\rho = \frac{3}{2+\sqrt{2}}$ and $d_k = \norm{ \hat{\bm w}^{(k)} - \bm w^{\star} }$, one obtains
\begin{equation*}
\begin{split}
\norm{ \hat{\bm w}^{(k)} - \bm w^{\star} } \leq & ~2(3\sqrt{2}+4) \lambda_{\max}^2 \left(\L \bm w^{\star} \right) \norm{ \left(\L^{\ast} \left( \left( \L \bm w^{\star} + \bm J  \right)^{-1} - \bm S  \right) \right)_{\mathcal{S}^{\star}} }     \\
  & \qquad + \left( \frac{3}{2+\sqrt{2}} \right )^{k} \norm{ \hat{\bm w}^{(0)} - \bm w^{\star} }.      
  \end{split} 
\end{equation*}
Under the event $\mathcal{J}$, $\norm{ \left( \L^{\ast} \left( \left( \L \bm w^{\star} + \bm J  \right)^{-1} - \bm S  \right) \right)_{\mathcal{S}^{\star}} }$ can be bounded by
\begin{equation}
\norm{ \left( \L^{\ast} \left( \left( \L \bm w^{\star} + \bm J  \right)^{-1} - \bm S  \right) \right)_{\mathcal{S}^{\star}} } \leq \sqrt{s}\lambda/2 \leq \sqrt{\alpha c_0^{-1} s\log p /n}. \label{bnd}
\end{equation}
Therefore, under the event $\mathcal{J}$, which holds with probability at least $1-1/p^{\alpha-2}$, one has
\begin{equation}
\norm{ \hat{\bm w}^{(k)} - \bm w^{\star} } \leq 2(3\sqrt{2}+4)  \lambda_{\max}^2 \left(\L \bm w^{\star} \right) \sqrt{ \alpha c_0^{-1} s\log p /n}+ \left( \frac{3}{2+\sqrt{2}} \right)^{k} \norm{ \hat{\bm w}^{(0)} - \bm w^{\star}}.   \nonumber
\end{equation}

Next, we prove the Corollary 3.9. Under the event $\mathcal{J}$, one applies Lemma \ref{lem5} and obtains
\begin{equation}\label{t3_1}    
\begin{split}
 \norm{ \L \hat{\bm w}^{(k)} - \L \bm w^{\star} }_{\mathrm{F}} \leq  & ~ 2\sqrt{2} \lambda_{\max}^2 \left(\L \bm w^{\star} \right) \norm{ \left( \L^{\ast} \left( \left( \L \bm w^{\star} + \bm J  \right)^{-1} - \bm S  \right) \right)_{\mathcal{S}^\star} } \\
&   \qquad +\frac{3}{2+\sqrt{2}} \norm{ \L \hat{\bm w}^{(k-1)} - \L \bm w^{\star} }_{\mathrm{F}}   
\end{split}
\end{equation}
holds for any $k \geq 1$. Taking $a_0 = 2\sqrt{2} \lambda_{\max}^2 \left(\L \bm w^{\star} \right) \norm{ \left( \L^{\ast} \left( \left( \L \bm w^{\star} + \bm J  \right)^{-1} - \bm S  \right) \right)_{\mathcal{S}^{\star}} }$, $\rho = \frac{3}{2+\sqrt{2}}$ and $d_k = \norm{ \L \hat{\bm w}^{(k)} - \L \bm w^{\star} }_{\mathrm{F}}$, by \eqref{induction} one has
\begin{equation*}
\begin{split}
\norm{ \L \hat{\bm w}^{(k)} - \L \bm w^{\star} }_{\mathrm{F}} \leq & ~ 4(\sqrt{2}+1)^2 \lambda_{\max}^2 \left(\L \bm w^{\star} \right) \norm{ \left( \L^{\ast} \left( \left( \L \bm w^{\star} + \bm J  \right)^{-1} - \bm S  \right) \right)_{\mathcal{S}^{\star}} } \\
&  \qquad + \left( \frac{3}{2+\sqrt{2}} \right )^{k} \norm{ \L \hat{\bm w}^{(0)} - \L \bm w^{\star} }_{\mathrm{F}}.    
\end{split}      
\end{equation*}
Similarly, according to \eqref{bnd}, one obtains,
\begin{equation}
\norm{ \L \hat{\bm w}^{(k)} - \L \bm w^{\star} }_{\mathrm{F}} \leq 4(3+2\sqrt{2})  \lambda_{\max}^2 \left(\L \bm w^{\star} \right) \sqrt{ \alpha c_0^{-1} s\log p /n}+ \left( \frac{3}{2+\sqrt{2}} \right )^{k} \norm{ \L \hat{\bm w}^{(0)} - \L \bm w^{\star} }_{\mathrm{F}}   \nonumber
\end{equation}
holds at least $1-1/p^{\alpha-2}$.

Alternative to \eqref{induction}, one obtains
\begin{equation}
d_k \leq \frac{1-\rho^{k-1}}{1-\rho}a_0 + \rho^{k-1}d_1,           \nonumber 
\end{equation}
and correspondingly establishes
\begin{equation}\label{t3_2}   
\norm{ \L \hat{\bm w}^{(k)} - \L \bm w^{\star} }_{\mathrm{F}} \leq 4(3+2\sqrt{2})  \lambda_{\max}^2 \left(\L \bm w^{\star} \right) \sqrt{ \alpha c_0^{-1} s\log s /n}  + \left( \frac{3}{2+\sqrt{2}} \right )^{k-1} \norm{ \L \hat{\bm w}^{(1)} -\L \bm w^{\star} }_{\mathrm{F}}.        
\end{equation}

To apply Lemma \ref{lem3}, we first check the necessary conditions of the lemma. Let $\bm z^{(0)}$ satisfy $z_i^{(0)} = h'_{\lambda} \left( \hat{w_i}^{(0)} \right)$, $i \in [p(p-1)/2]$. Notice that $z_i^{(0)}  \in [0, \lambda]$ for $i \in [p(p-1)/2]$ by Assumption \ref{assumption 1}. According to \eqref{e-set}, $\E^{(1)} = \left\lbrace \mathcal{S}^\star \cup \S^{(1)} \right\rbrace$, where $\S^{(1)} = \left\lbrace i \in [p(p-1)/2] \ | \hat{w}_i^{(0)} \geq  b \right\rbrace$. For any $i \in \left\lbrace \S^{(1)}\right\rbrace^c$, one has
\begin{equation}
z_i^{(0)} = h'_{\lambda} \left(\hat{w}_i^{(0)} \right) \geq h'_{\lambda} (b) \geq \frac{\lambda}{2},   \nonumber
\end{equation} 
where the first inequality holds because $\hat{w}_i^{(0)} < b$ for any $i \in \left\{ \S^{(1)}\right\}^c$, and $h'_{\lambda}$ is non-increasing according to Assumption \ref{assumption 1}; the second inequality directly follows from Assumption \ref{assumption 1}. Hence one has 
\begin{equation}
\norm{ \bm z_{\left\{ \E^{(1)}\right\}^c}^{(0)} }_{\min} \geq \norm{ \bm z_{\left\{ \S^{(1)}\right\}^c}^{(0)} }_{\min} \geq \lambda/2.   \nonumber
\end{equation} 
One also obtains $\left| \E^{(1)} \right|<2s$ by Lemma \ref{lem4}, and $\mathcal{S}^\star \subseteq \E^{(1)}$ by the definition of $\E^{(1)}$. Therefore, one can apply Lemma \ref{lem3} with $\E = \E^{(1)}$ and $\bm z =\bm z^{(0)}$ and obtains
\begin{equation}
 \norm{ \L \hat{\bm w}^{(1)} - \L \bm w^{\star} }_{\mathrm{F}}  \leq 2(1+\sqrt{2}) \lambda_{\max}^2 \left(\L \bm w^{\star} \right) \sqrt{s} \lambda.          \nonumber
\end{equation}
If $t \geq \log_{\frac{2+\sqrt{2}}{3}} \left(\lambda  \sqrt{n/ \log p} \right) = \log \left( \sqrt{4\alpha c_0^{-1} } \right)/\log \frac{2+\sqrt{2}}{3}$, a simple algebra yields
\begin{equation}
\left( \frac{3}{2+\sqrt{2}} \right)^{t-1}  \norm{ \L \hat{\bm w}^{(1)} - \L \bm w^{\star} }_{\mathrm{F}} \leq \frac{6\sqrt{2}+8}{3} \left( \lambda \sqrt{n / \log p } \right)^{-1} \lambda_{\max}^2 \left(\L \bm w^{\star} \right) \sqrt{s} \lambda   \lesssim \sqrt{s \log p /n}.     \label{t3_3}     
\end{equation}
Taking $k \geq \lceil 4 \log \left(4 \alpha c_0^{-1} \right) \rceil \geq \log \left( \sqrt{4\alpha c_0^{-1} } \right)/\log \frac{2+\sqrt{2}}{3}$, and combining \eqref{t3_2} and \eqref{t3_3} together, we can conclude that
\begin{equation}
\norm{ \L \hat{\bm w}^{(k)} - \L \bm w^{\star} }_{\mathrm{F}} \lesssim \sqrt{s\log p /n},   \nonumber
\end{equation}
completing the proof.
\end{proof}


\subsection{Proof of Theorem \ref{oracle}} \label{sec-prof-oracle}
\begin{proof}
The proof consists of two prats: In Part I, we prove that the sequence $\left\lbrace \hat{\bm w}^{(k)} \right\rbrace_{k \geq 1}$ will converge to the oracle estimator $\bm w^{\mathrm{oracle}}$ and $\mathrm{supp} \left( \hat{\bm w}^{(k)} \right) = \mathrm{supp} \left( \bm w^\star \right)$ under some conditions; In part II, we bound the probabilities that the conditions introduced in Part I hold.

\paragraph{Part I} 
The sequence $\left\lbrace \hat{\bm w}^{(k)} \right\rbrace_{k \geq 1}$ is established by solving a sequence of sub-problems
\begin{equation} \label{w_l+1}
\hat{\bm w}^{(k)}: = \arg \min_{\bm w \geq \bm 0} - \log \det \left(\mathcal{L} \bm w + \bm J \right) + \tr{ \bm S \mathcal{L} \bm w } + \sum_i h'_\lambda \left( \hat{w}^{(k-1)}_i \right) w_i.
\end{equation}
We first show that the global minimizers of the optimization problems \eqref{oracle-estimator} and \eqref{w_l+1}, i.e., $\hat{\bm w}^{(k)}$ and $\bm w^{\mathrm{oracle}}$, exist and are unique with probability one. The uniqueness can be established by proving that the optimization problems \eqref{oracle-estimator} and \eqref{w_l+1} are strictly convex, and the existence can be guaranteed by proving that the \eqref{oracle-estimator} and \eqref{w_l+1} are coercive. The proof procedure is similar to the proof of the uniqueness and existence of the minimizer of the optimization \eqref{L1} in Theorem \ref{Theorem 1}, except one main modification in \eqref{f_lower} for proving the coercivity. Let 
\begin{equation*}
f(\bm w) :=  - \log \det \left(\mathcal{L} \bm w + \bm J \right) + \tr{ \bm S \mathcal{L} \bm w }.
\end{equation*}
Then $f(\bm w)$ can be be lower bounded by
\begin{equation}\label{f_lower_new}
\begin{split}
f(\bm w) & = - \log \left( \prod_{i=2}^p\lambda_i(\mathcal{L} \bm w) \right) + \langle \L^{\ast} \bm S , \bm w \rangle  \\
& \geq -(p-1)\log\left(\sum_{i=1}^p\lambda_i(\mathcal{L} \bm w)\right) + \langle \L^{\ast} \bm S, \bm w \rangle + (p-1) \log(p-1) \\
& \geq -(p-1)\log\left(\sum_{t=1}^{p(p-1)/2} w_t \right) + \alpha \sum_{t=1}^{p(p-1)/2} w_t + (p-1) \log\frac{p-1}{2}, 
\end{split}
\end{equation}
where $\alpha= \min_k \left[ \mathcal{L}^{\ast} \bm S \right]_k$. Note that the sample covariance matrix is computed by $\bm S= \frac{1}{n} \sum_{t=1}^n \bm x_t \bm x_t^\top$, where $\bm x_1, \ldots, \bm x_n$ are $i.i.d.$ samples drawn from $p$-dimensional LGMRF with the parameters $\left(\bm 0, \bm \Omega \right)$. By calculation, one obtains, for any $k \in [p(p-1)/2]$,
\begin{equation}\label{LS}
\left[\mathcal{L}^\ast \bm S \right]_k =  \frac{1}{n} \sum_{t=1}^n \left[ \mathcal{L}^\ast \left( \bm x_t \bm x_t^\top \right) \right]_k =\frac{1}{n}\sum_{t=1}^n \left( [\bm x_t]_i - [\bm x_t]_j \right)^2,
\end{equation}
where $i, j \in [p]$ satisfy $k=i-j+\frac{j-1}{2}(2p-j)$ and $i>j$. Define a set $\mathcal{A}_{ij}:= \left\lbrace \bm x \in V^{p-1} | x_i = x_j \right\rbrace$, where $V^{p-1}= \lbrace \bm x \in \mathbb{R}^p | \, \bm 1^{\top} \bm x  = 0 \rbrace$. Obviously, $\mathcal{A}_{ij}$ has measure zero in $V^{p-1}$ for any $i \neq j$. Therefore, for $n \geq 1$, one has
\begin{equation} \label{bound_s}
\mathbb{P} \left[ \min_{k \in [p(p-1)/2]} \left[\mathcal{L}^\ast \bm S \right]_k > 0 \right] =1.
\end{equation}
In other words, $\alpha>0$ in \eqref{f_lower_new} with probability one. The other proof procedure is similar to that for the case of the optimization \eqref{L1}, and thus is omitted.

Define two events on $\bm w^{\mathrm{oracle}}$ that
\begin{equation}\label{event1}
\mathcal{J}_1:= \left\lbrace \norm{\nabla_{\left\lbrace \mathcal{S}^\star \right\rbrace^c} f \left( \bm w^{\mathrm{oracle}} \right)}_{\max} \leq \frac{1}{2} \lambda \right\rbrace,
\end{equation}
and
\begin{equation} \label{event2}
\mathcal{J}_2:= \left\lbrace \norm{\bm w_{\mathcal{S}^\star}^{\mathrm{oracle}}}_{\min} \geq  \gamma \lambda \right\rbrace.
\end{equation}

Next, we will prove by induction that $\hat{\bm w}^{(k)} = \bm w^{\mathrm{oracle}}$ for any $k \geq 1$ on the conditions that the events $\mathcal{J}_1$ and $\mathcal{J}_2$ hold. 

We first prove that $\hat{\bm w}^{(1)} = \bm w^{\mathrm{oracle}}$. Under the assumption that $\norm{\hat{\bm w}^{(0)} - \bm w^\star}_{\max} \leq c \lambda$, for any $i \in \left\lbrace \mathcal{S}^\star \right\rbrace^c$, we have
\begin{equation*}
 \hat{w}_i^{(0)} \leq \norm{\hat{\bm w}^{(0)} - \bm w^\star}_{\max} \leq c \lambda.
\end{equation*}
Thus one has
\begin{equation}\label{eq_lambda}
h'_\lambda \left( \hat{w}_i^{(0)} \right) \geq \frac{1}{2} \lambda, \quad \forall i \in \left\lbrace \mathcal{S}^\star \right\rbrace^c.
\end{equation}
On the other hand, for any $i \in \mathcal{S}^\star$, 
\begin{equation*}
 \hat{w}_i^{(0)}  \geq  \norm{\bm w_{\mathcal{S}^\star}^\star}_{\min} - \norm{\hat{\bm w}^{(0)} - \bm w^\star}_{\max}  \geq \gamma \lambda,
\end{equation*}
which yields $h'_\lambda \left( \hat{w}_i^{(0)} \right) =0$ by Assumption \ref{assumption 1}. Therefore, in the first iteration, we can rewrite \eqref{w_l+1} as
\begin{equation} \label{nw_l+1}
\hat{\bm w}^{(1)} = \arg \min_{\bm w \geq \bm 0} - \log \det \left(\mathcal{L} \bm w + \bm J \right) + \tr{ \bm S \mathcal{L} \bm w } + \sum_{i \in \left\lbrace\mathcal{S}^\star \right\rbrace^c } h'_\lambda \left( \hat{w}^{(0)}_i \right) w_i.
\end{equation}
For any $\bm w \in\mathcal{S}_w$, one has
\begin{equation} \label{f_ineq}
\begin{split}
f(\bm w) &\geq f \left(\bm w^{\mathrm{oracle}} \right) + \sum_i \nabla_i f \left( \bm w^{\mathrm{oracle}} \right) \left( w_i -w_i^{\mathrm{oracle}} \right) \\
& \geq f \left(\bm w^{\mathrm{oracle}} \right) + \sum_{i \in \left\lbrace \mathcal{S}^\star \right\rbrace^c } \nabla_i f \left( \bm w^{\mathrm{oracle}} \right) \left( w_i  - w_i^{\mathrm{oracle}}\right), 
\end{split}
\end{equation}
where the second inequality follows from the facts that $\sum_{i \in \mathcal{S}^\star } \nabla_i f \left( \bm w^{\mathrm{oracle}} \right) \left( w_i  - w_i^{\mathrm{oracle}}\right) \geq 0$ which can be established by the KKT conditions as follows. Let $(\bm w^{\mathrm{oracle}}, \hat{\bm \upsilon}^{\mathrm{o}})$ be any pair of points that satisfies the KKT conditions of \eqref{oracle-estimator}. Then one has
\begin{equation}\label{oracle_KKT}
\begin{split}
 \nabla f \left(\bm w^{ \mathrm{oracle} } \right) - \hat{\bm \upsilon}^{\mathrm{o}} &= \bm 0; \\
 w_i^{\mathrm{oracle}}  \hat{\upsilon}^{\mathrm{o}}_i =0, \ w_i^{\mathrm{oracle}}  \geq 0, \  \hat{\upsilon}_i^{\mathrm{o}} & \geq 0, \ \mathrm{for} \ i \in \mathcal{S}^\star, \\
 w_i^{\mathrm{oracle}}  & = 0, \ \mathrm{for} \ i \in \left\lbrace \mathcal{S}^\star \right\rbrace^c,
\end{split}
\end{equation}
It is easy to verify that 
\begin{equation*}
\nabla_i f \left(\bm w^{ \mathrm{oracle} } \right) \left( w_i - w_i^{\mathrm{oracle} } \right) = \hat{\upsilon}_i^{\mathrm{o}} \left( w_i - w_i^{\mathrm{oracle}} \right) = \hat{\upsilon}_i^{\mathrm{o}} w_i \geq 0, \quad \forall i \in \mathcal{S}^\star.
\end{equation*}
For any $\bm w \in\mathcal{S}_w$, we obtain
\begin{equation}\label{p_ineq}
\begin{split}
q :& = \left( f(\bm w) + \sum_{i \in \left\lbrace \mathcal{S}^\star \right\rbrace^c} h'_\lambda \left( w_i^{(0)} \right)w_i \right) - \left( f \left(\bm w^{\mathrm{oracle}} \right) + \sum_{i \in \left\lbrace \mathcal{S}^\star \right\rbrace^c } h'_\lambda \left( w_i^{(0)} \right) w_i^{\mathrm{oracle}} \right) \\
& \geq \sum_{i \in \left\lbrace \mathcal{S}^\star \right\rbrace^c } \nabla_i f\left( \bm w^{\mathrm{oracle}} \right) \left( w_i - w_i^{\mathrm{oracle}} \right) + \sum_{i \in \left\lbrace \mathcal{S}^\star \right\rbrace^c } h'_\lambda \left( w_i^{(0)} \right) \left( w_i - w_i^{\mathrm{oracle}} \right) \\
& = \sum_{i \in \left\lbrace \mathcal{S}^\star \right\rbrace^c } \left( \nabla_i f \left( \bm w^{\mathrm{oracle}} \right)  +  h'_\lambda \left( w_i^{(0)} \right) \right) w_i \geq 0,
\end{split}
\end{equation}
where the first inequality follows from \eqref{f_ineq}, the second equality follows from \eqref{oracle_KKT}, and the last inequality is established by
\begin{equation*} 
\nabla_i f \left( \bm w^{\mathrm{oracle}} \right)  +  h'_\lambda \left( w_i^{(0)} \right) \geq 0, \quad \forall i \in \left\lbrace \mathcal{S}^\star \right\rbrace^c,
\end{equation*}
which follows from \eqref{eq_lambda} together with the event $\mathcal{J}_1$ and the fact that $\bm w \geq \bm 0$. Therefore, we conclude that the objective function in \eqref{nw_l+1} achieves the optimal value when $\bm w = \bm w^{\mathrm{oracle}}$. By the uniqueness of $\hat{\bm w}^{(1)}$, we obtain that 
\begin{equation}\label{w_l+1_orc}
\hat{\bm w}^{(1)} = \bm w^{\mathrm{oracle}}.
\end{equation}

Assume $\hat{\bm w}^{(k_o)} = \bm w^{\mathrm{oracle}}$ holds for $k_o \in \mathbb{N}^+$. We will show that $\hat{\bm w}^{(k_o+1)} = \bm w^{\mathrm{oracle}}$ also holds. 
Under the event $\mathcal{J}_2$ defined in \eqref{event2}, one has
\begin{equation}\label{eq_lambda_orc}
h'_\lambda \left( w_i^{\mathrm{oracle}} \right) =0, \quad \forall i \in \mathcal{S}^\star.
\end{equation}
Therefore, we can rewrite \eqref{w_l+1} as
\begin{equation} \label{nw_l+2}
\hat{\bm w}^{(k_o+1)} = \arg \min_{\bm w \geq \bm 0} - \log \det \left(\mathcal{L} \bm w + \bm J \right) + \tr{ \bm S \mathcal{L} \bm w } + \sum_{i \in \left\lbrace\mathcal{S}^\star \right\rbrace^c } h'_\lambda \left( w^{\mathrm{oracle}}_i \right) w_i.
\end{equation} 
Note that for any $i \in \left\lbrace \mathcal{S}^\star \right\rbrace^c$, $ w_i^{\mathrm{oracle}} =0$ and thus 
\begin{equation*}
h'_\lambda \left( w_i^{\mathrm{oracle}} \right)= \lambda.
\end{equation*}
Under the event $\mathcal{J}_1$, similar to \eqref{p_ineq},
\begin{equation*}
\begin{split}
q' :& = \left( f(\bm w) + \sum_{i \in \left\lbrace \mathcal{S}^\star \right\rbrace^c} h'_\lambda \left( w_i^{\mathrm{oracle}} \right)w_i \right) - \left( f \left(\bm w^{\mathrm{oracle}} \right) + \sum_{i \in \left\lbrace \mathcal{S}^\star \right\rbrace^c } h'_\lambda \left( w_i^{\mathrm{oracle}} \right) w_i^{\mathrm{oracle}} \right) \\
& \geq \sum_{i \in \left\lbrace \mathcal{S}^\star \right\rbrace^c } \left( \nabla_i f \left( \bm w^{\mathrm{oracle}} \right)  +  h'_\lambda \left( w_i^{\mathrm{oracle}} \right) \right) w_i \geq 0
\end{split}
\end{equation*}
holds for any $\bm w \in \mathcal{S}_w$. With the same argument for \eqref{w_l+1_orc}, we have
\begin{equation*}
\hat{\bm w}^{(k_o+1)} = \bm w^{\mathrm{oracle}}.
\end{equation*}
To sum up, under the events $\mathcal{J}_1$ and $\mathcal{J}_2$, we obtain that
 \begin{equation}\label{eq_seq-oracle}
\hat{\bm w}^{(k)} = \bm w^{\mathrm{oracle}}, \quad \forall k \geq 1.
\end{equation}
By the uniqueness of $\bm w^{\mathrm{oracle}}$, we can conclude that the sequence $\left\lbrace \hat{\bm w}^{(k)} \right\rbrace_{k \geq 1}$ converges to $\bm w^{\mathrm{oracle}}$. Following from \eqref{eq_seq-oracle}, we obtain
\begin{equation}
\mathrm{supp} \left( \hat{\bm w}^{(k)} \right) \subseteq \mathrm{supp} \left( \bm w^\star \right) , \quad \forall k \geq 1.
\end{equation}
Under the event $\mathcal{J}_2$, there is no zero element for $\bm w^{\mathrm{oracle}}$ in the set $\mathcal{S}^\star$. Therefore, together with \eqref{eq_seq-oracle}, we further obtain
\begin{equation}
\mathrm{supp} \left( \hat{\bm w}^{(k)} \right) = \mathrm{supp} \left( \bm w^\star \right) , \quad \forall k \geq 1.
\end{equation}

\paragraph{Part II}  In this part, we bound the probability that the events $\mathcal{J}_1$ and $\mathcal{J}_2$ hold.

We first prove that $\norm{\bm w_{\mathcal{S}^\star}^{\mathrm{oracle}}}_{\min} \geq  \gamma \lambda$ holds under the condition that 
\begin{equation}\label{cond_sam}
\norm{ \left[ \mathcal{L}^\ast \left( \mathcal{L} \bm w^\star + \bm J \right)^{-1} \right]_{\mathcal{S}^\star} - \left[ \mathcal{L}^\ast \bm S \right]_{\mathcal{S}^\star} }_{\max} \leq \min \left( \frac{1}{12 K_1 K_2 d},  \frac{1}{48 K_1^3 K_2^2 d^2}, \frac{c \lambda}{2 K_2} \right),
\end{equation}
where $K_1$ and $K_2$ are constants defined in \eqref{constant_K}.

By the inequality
\begin{equation}\label{w_oracle_bound}
\norm{\bm w_{\mathcal{S}^\star}^{\mathrm{oracle}}}_{\min}  \geq \norm{\bm w_{\mathcal{S}^\star}^\star}_{\min}  - \norm{\bm w_{\mathcal{S}^\star}^{\mathrm{oracle}} - \bm w_{\mathcal{S}^\star}^\star}_{\max},
\end{equation}
we can bound $\norm{\bm w_{\mathcal{S}^\star}^{\mathrm{oracle}} - \bm w_{\mathcal{S}^\star}^\star}_{\max}$ instead. 

Define a continuous map $F: \bm \Delta_{\mathcal{S}^\star} \mapsto F \left(\bm \Delta_{\mathcal{S}^\star} \right)$,
\begin{equation*}
F \left(\bm \Delta_{\mathcal{S}^\star} \right) = \left( \bm H_{\mathcal{S}^\star \mathcal{S}^\star} \right)^{-1} \left( \left[  \mathcal{L}^\ast \left(  \mathcal{L} \left( \bm w^\star + \bm \Delta  + \bm J \right)^{-1} \right) \right]_{\mathcal{S}^\star} - \left[ \mathcal{L}^\ast \bm S \right]_{\mathcal{S}^\star} \right) + \bm \Delta_{\mathcal{S}^\star},
\end{equation*}
where the entries of $\bm \Delta $ in $\mathcal{S}^\star$ equal to $\bm \Delta_{\mathcal{S}^\star}$ and the entries in $ \left\lbrace \mathcal{S}^\star \right\rbrace^c$ equal to zero.
Define
\begin{equation} \label{def_Br}
\mathbb{B}(r) = \left\lbrace \bm \Delta_{\mathcal{S}^\star} \in \mathbb{R}^{ \left| \mathcal{S}^\star \right|} | \norm{\bm \Delta_{\mathcal{S}^\star}}_{\max} \leq r\right\rbrace,
\end{equation}
where $r$ is defined by
\begin{equation}\label{def_r}
r= 2 K_2 \norm{ \left[ \mathcal{L}^\ast \left( \mathcal{L} \bm w^\star + \bm J \right)^{-1} \right]_{\mathcal{S}^\star}  - \left[  \mathcal{L}^\ast \bm S \right]_{\mathcal{S}^\star} }_{\max}.
\end{equation}
Note that $\mathbb{B}(r)$ is a convex compact set. We say $F$ maps the set $\mathbb{B}(r)$ to itself, if 
\begin{equation} \label{fixed point}
F(\mathbb{B}(r) )  \subset \mathbb{B}(r)
\end{equation}
holds. In the following, we will prove that $\norm{\bm w_{\mathcal{S}^\star}^{\mathrm{oracle}}}_{\min} \geq  \gamma \lambda$ with the assumption that \eqref{fixed point} holds. After that, we will turn to prove \eqref{fixed point}. 

By Brouwer's Fixed Point Theorem, there exists a $\widehat{\bm \Delta}_{\mathcal{S}^\star} \in \mathbb{B}(r)$ such that $F \left( \widehat{\bm \Delta}_{\mathcal{S}^\star}    \right) = \widehat{\bm \Delta}_{\mathcal{S}^\star} $, i.e., 
\begin{equation*}
\left( \bm H_{\mathcal{S}^\star \mathcal{S}^\star} \right)^{-1} \left( \left[  \mathcal{L}^\ast \left(  \mathcal{L} \left( \bm w^\star + \widehat{\bm \Delta} \right) + \bm J \right)^{-1}  \right]_{\mathcal{S}^\star} - \left[ \mathcal{L}^\ast \bm S \right]_{\mathcal{S}^\star} \right) = \bm 0,
\end{equation*}
where $\widehat{\bm \Delta}  \in \mathbb{R}^{p(p-1)/2}$ with entries in $\mathcal{S}^\star$ equal to $\widehat{\bm \Delta} _{\mathcal{S}^\star}$ and entries in $ \left\lbrace \mathcal{S}^\star \right\rbrace^c$ equal to zero. Then one further obtains 
\begin{equation*}
\left[  \mathcal{L}^\ast \left(  \mathcal{L} \left( \bm w^\star + \widehat{\bm \Delta} \right)  + \bm J \right)^{-1}  \right]_{\mathcal{S}^\star} - \left[ \mathcal{L}^\ast \bm S \right]_{\mathcal{S}^\star}  = \bm 0.
\end{equation*}
We construct an $\hat{\bm \upsilon}$ such that 
\begin{equation*}
\hat{\bm \upsilon}_{\left\{ \mathcal{S}^\star \right\}^c } = - \left[  \mathcal{L}^\ast \left(  \mathcal{L} \left( \bm w^\star + \widehat{\bm \Delta} \right)  + \bm J \right)^{-1}  \right]_{\left\{ \mathcal{S}^\star \right\}^c } + \left[ \mathcal{L}^\ast \bm S \right]_{\left\{ \mathcal{S}^\star \right\}^c } \quad \mathrm{and} \quad \hat{\bm \upsilon}_{\mathcal{S}^\star } = \bm 0.
\end{equation*}
Take $\hat{\bm w} = \bm w^\star + \widehat{\bm \Delta}$. Note that $\hat{\bm w}_{ \left\{\mathcal{S}^\star \right\}^c }= \bm 0$ and $\hat{\bm w}_{ \mathcal{S}^\star  } \geq \bm 0$ because
\begin{equation*}
\min_{i \in \mathcal{S}^\star } \hat{w}_i \geq \min_{i \in \mathcal{S}^\star }  w^\star_i - \norm{ \widehat{\bm \Delta}}_{\max} \geq \gamma \lambda \geq 0.
\end{equation*}
We can verify that the pair $\left( \hat{\bm w}, \hat{\bm \upsilon} \right)$ satisfies all the KKT conditions in \eqref{oracle_KKT}. By the uniqueness of $\bm w^{\mathrm{oracle}}$, one has $\hat{\bm w} = \bm w^{\mathrm{oracle}}$, thus $\widehat{\bm \Delta} = \bm w^{\mathrm{oracle}} - \bm w^\star$ and $\norm{\bm w^{\mathrm{oracle}} - \bm w^\star}_{\max} = \norm{\widehat{\bm \Delta}}_{\max} \leq r$. Therefore, under \eqref{cond_sam},  we can establish that
\begin{equation*}
\norm{\bm w_{\mathcal{S}^\star}^{\mathrm{oracle}}}_{\min} \geq \norm{\bm w_{\mathcal{S}^\star}^\star}_{\min} - \norm{\widehat{\bm \Delta}}_{\max}  \geq (c+\gamma) \lambda - 2K_2 \norm{\left[ \mathcal{L}^\ast \left( \mathcal{L} \bm w^\star + \bm J \right)^{-1} \right]_{\mathcal{S}^\star} - \left[ \mathcal{L}^\ast \bm S \right]_{\mathcal{S}^\star} }_{\max} \geq \gamma \lambda,
\end{equation*}
completing the proof.

Next, we will prove \eqref{fixed point}. By the sub-multiplicativity of the norm $\normI{\cdot}_\infty$, we have $\normI{\bm A \bm B}_\infty \leq \normI{\bm A}_\infty \normI{\bm B}_\infty$, for any $\bm A, \bm B \in \mathbb{R}^{p \times p}$. Thus, for any $\bm \Delta $ with $ \bm \Delta_{\mathcal{S}^\star} \in \mathbb{B}(r)$ and $\bm \Delta_{ \left\lbrace \mathcal{S}^\star \right\rbrace^c } = \bm 0$, one has
\begin{equation}\label{expan}
\begin{split}
\normI{ \left( \mathcal{L} \bm w^\star \right)^\dagger   \mathcal{L} \bm \Delta }_\infty & \leq \normI{ \left( \mathcal{L} \bm w^\star \right)^\dagger }_\infty \normI{\mathcal{L} \bm \Delta}_\infty  \leq 2K_1 d \norm{\bm \Delta}_{\max}  \\
& \leq 2K_1d r =   4 K_1 K_2 d \norm{ \left[ \mathcal{L}^\ast \left( \mathcal{L} \bm w^\star + \bm J \right)^{-1} \right]_{\mathcal{S}^\star} - \left[ \mathcal{L}^\ast \bm S \right]_{\mathcal{S}^\star} }_{\max} \leq \frac{1}{3},    
\end{split}
\end{equation}
where we use the definition of $K_1$ in \eqref{constant_K}, and the fact that the graph associated with $ \bm \Delta$ has at most $d$ edges for each node. The last inequality in \eqref{expan} follows from \eqref{cond_sam}.  As a result, we have the convergent matrix expansion,
\begin{equation}\label{expansion}
\begin{split}
\left(  \mathcal{L} \bm w^\star + \mathcal{L} \bm \Delta + \bm J  \right)^{-1} & = \left(  \left(  \mathcal{L} \bm w^\star + \bm J \right) \left( \bm I + \left(  \mathcal{L} \bm w^\star + \bm J \right)^{-1}  \mathcal{L} \bm \Delta \right)  \right)^{-1}  \\
& = \left(  \bm I + \left( \mathcal{L} \bm w^\star \right)^\dagger \mathcal{L} \bm \Delta   \right)^{-1}  \left( \mathcal{L} \bm w^\star + \bm J \right)^{-1} \\
& = \sum_{k=0}^\infty \left( -1 \right)^k \left(  \left( \mathcal{L} \bm w^\star \right)^\dagger \mathcal{L} \bm \Delta  \right)^k  \left( \mathcal{L} \bm w^\star + \bm J \right)^{-1}  \\
& = \left( \mathcal{L} \bm w^\star + \bm J \right)^{-1}- \left( \mathcal{L} \bm w^\star \right)^\dagger \mathcal{L} \bm \Delta  \left( \mathcal{L} \bm w^\star \right)^\dagger + \sum_{k=2}^\infty \left( -1 \right)^k \left( \left( \mathcal{L} \bm w^\star \right)^\dagger \mathcal{L} \bm \Delta \right)^k  \left( \mathcal{L} \bm w^\star \right)^\dagger \\
& = \left( \mathcal{L} \bm w^\star + \bm J \right)^{-1}- \left( \mathcal{L} \bm w^\star \right)^\dagger \mathcal{L} \bm \Delta  \left( \mathcal{L} \bm w^\star \right)^\dagger +  \left( \mathcal{L} \bm w^\star \right)^\dagger \mathcal{L} \bm \Delta   \left( \mathcal{L} \bm w^\star \right)^\dagger \mathcal{L} \bm \Delta  \bm N \left( \mathcal{L} \bm w^\star \right)^\dagger,
\end{split}
\end{equation} 
where $\bm N = \sum_{k=0}^\infty \left( -1 \right)^k \left( \left( \mathcal{L} \bm w^\star \right)^\dagger \mathcal{L} \bm \Delta \right)^k$. We can write $\bm \Delta$ as 
\begin{equation*}
\bm \Delta = \delta_1 \bm e_1 + \delta_2 \bm e_2 + \cdots + \delta_{p(p-1)/2} \bm e_{p(p-1)/2},
\end{equation*}
where $\bm e_i$ is the unit vector with $1$ in position $i$ and zeros elsewhere. Then one obtains
\begin{equation*}
\begin{split}
\mathcal{L}^\ast \left(  \left( \mathcal{L} \bm w^\star \right)^\dagger \mathcal{L} \bm \Delta \left(  \mathcal{L} \bm w^\star\right)^\dagger  \right) &= \sum_{i=1}^{p(p-1)/2} \delta_i \mathcal{L}^\ast \left(  \left(  \mathcal{L} \bm w^\star \right)^\dagger \mathcal{L} \bm e_i \left(  \mathcal{L} \bm w^\star\right)^\dagger  \right) = \sum_{i=1}^{p(p-1)/2}  \delta_i \mathcal{L}^\ast \bm B_i \\
& = \left[ \mathcal{L}^\ast \bm B_1, \mathcal{L}^\ast \bm B_2, \ldots, \mathcal{L}^\ast \bm B_{p(p-1)/2}  \right] \bm \Delta = \bm H \bm \Delta,
\end{split}
\end{equation*}
where $\bm B_i = \left(  \mathcal{L} \bm w^\star \right)^\dagger \mathcal{L} \bm e_i \left(  \mathcal{L} \bm w^\star\right)^\dagger$ for $i=1, \ldots, p(p-1)/2$. Note that $\bm \Delta_{ \left\lbrace \mathcal{S}^\star \right\rbrace^c } = \bm 0$, implying that $\delta_i = 0$ for any $i \in \left\lbrace \mathcal{S}^\star \right\rbrace^c $. Then one has
\begin{equation} \label{Hdelta}
\left[ \mathcal{L}^\ast \left(   \left(  \mathcal{L} \bm w^\star  \right)^\dagger \mathcal{L} \bm \Delta  \left( \mathcal{L} \bm w^\star \right)^\dagger  \right) \right]_{\mathcal{S}^\star} = \bm H_{\mathcal{S}^\star \mathcal{S}^\star} \bm \Delta_{\mathcal{S}^\star}.
\end{equation}
Define 
\begin{equation} \label{R_delta}
\bm R^\Delta := \left( \mathcal{L} \bm w^\star \right)^\dagger \mathcal{L} \bm \Delta   \left( \mathcal{L} \bm w^\star \right)^\dagger \mathcal{L} \bm \Delta  \bm N \left( \mathcal{L} \bm w^\star \right)^\dagger.
\end{equation}
Together with \eqref{expansion}, \eqref{Hdelta} and \eqref{R_delta}, one has
\begin{equation} \label{sep_inv}
\left[ \mathcal{L}^\ast \left(  \left(  \mathcal{L} \bm w^\star + \mathcal{L} \bm \Delta + \bm J  \right)^{-1} \right)  \right]_{\mathcal{S}^\star} = \left[ \mathcal{L}^\ast \left(  \left( \mathcal{L} \bm w^\star \right)^\dagger \right) \right]_{\mathcal{S}^\star}- \bm H_{\mathcal{S}^\star \mathcal{S}^\star} \bm \Delta_{\mathcal{S}^\star} + \left[ \mathcal{L}^\ast\bm R^\Delta \right]_{\mathcal{S}^\star}.
\end{equation}
We bound $F(\bm \Delta_{\mathcal{S}^\star})$ as below,
\begin{equation} \label{F_bound}
\begin{split}
\norm{F \left( \bm \Delta_{\mathcal{S}^\star} \right)}_{\max} &= \norm{  \left( \bm H_{\mathcal{S}^\star \mathcal{S}^\star} \right)^{-1} \left( \left[ \mathcal{L}^\ast \left( \mathcal{L} \left( \bm w^\star + \bm \Delta \right) + \bm J\right)^{-1}  \right]_{\mathcal{S}^\star} - \left[ \mathcal{L}^\ast \bm S \right]_{\mathcal{S}^\star} \right) + \bm \Delta_{\mathcal{S}^\star}}_{\max} \\
& = \norm{ \left( \bm H_{\mathcal{S}^\star \mathcal{S}^\star} \right)^{-1} \left( \left[  \mathcal{L}^\ast \left(  \left( \mathcal{L} \bm w^\star \right)^\dagger \right) \right]_{\mathcal{S}^\star} - \bm H_{\mathcal{S}^\star \mathcal{S}^\star} \bm \Delta_{\mathcal{S}^\star} +\left[ \mathcal{L}^\ast \bm R^\Delta \right]_{\mathcal{S}^\star} - \left[ \mathcal{L}^\ast \bm S \right]_{\mathcal{S}^\star} \right)  + \bm \Delta_{\mathcal{S}^\star} }_{\max} \\
& \leq \underbrace{ \normI{ \left( \bm H_{\mathcal{S}^\star \mathcal{S}^\star} \right)^{-1}}_{\infty}}_{I_1}   \left(  \underbrace{ \norm{\left[  \mathcal{L}^\ast \left(  \left( \mathcal{L} \bm w^\star \right)^\dagger \right) \right]_{\mathcal{S}^\star} - \left[ \mathcal{L}^\ast \bm S \right]_{\mathcal{S}^\star}}_{\max} }_{I_2}  +  \underbrace{\norm{\mathcal{L}^\ast \bm R^\Delta}_{\max} }_{I_3} \right). 
\end{split}
\end{equation}
We need to bound the terms $I_1$, $I_2$ and $I_3$. The term $I_1$ is a constant term and equals to $K_2$ defined in \eqref{constant_K}. The term $I_2$ can be written as $\frac{r}{2 K_2}$ according to \eqref{def_Br}. The term $I_3$ can be bounded as below,
\begin{equation}\label{Bound_LR}
\begin{split}
\norm{\mathcal{L}^\ast \bm R^\Delta}_{\max} & = \norm{\mathcal{L}^\ast \left[ \left( \mathcal{L} \bm w^\star \right)^\dagger \mathcal{L} \bm \Delta \left( \mathcal{L} \bm w^\star \right)^\dagger \mathcal{L} \bm \Delta \bm N \left(  \mathcal{L} \bm w^\star \right)^\dagger	 \right]}_{\max}  \\
& \leq \normI{\mathcal{L}^\ast}_{\infty} \cdot \norm{ \left(  \mathcal{L} \bm w^\star \right)^\dagger  \mathcal{L}  \bm \Delta \left( \mathcal{L} \bm w^\star \right)^\dagger \mathcal{L} \bm \Delta \bm N \left( \mathcal{L} \bm w^\star \right)^\dagger  }_{\max}  \\
& = \normI{\mathcal{L}^\ast}_{\infty} \cdot \max_{i,j} \left| \bm e_i^\top \left(  \mathcal{L} \bm w^\star \right)^\dagger \mathcal{L} \bm \Delta \left( \mathcal{L} \bm w^\star \right)^\dagger \mathcal{L} \bm \Delta  \bm N \left(  \mathcal{L} \bm w^\star \right)^\dagger  \bm e_j\right|  \\
& \leq \normI{\mathcal{L}^\ast}_{\infty} \cdot \max_i \norm{\bm e_i^\top \left( \mathcal{L} \bm w^\star \right)^\dagger \mathcal{L} \bm \Delta}_{\max} \cdot \max_j \norm{ \left(   \mathcal{L} \bm w^\star \right)^\dagger \mathcal{L} \bm \Delta \bm N \left( \mathcal{L} \bm w^\star \right)^\dagger \bm e_j  }_1 \\
& \leq \normI{\mathcal{L}^\ast}_{\infty} \cdot \max_i \norm{\bm e_i^\top \left( \mathcal{L} \bm w^\star \right)^\dagger }_1 \cdot \norm{\mathcal{L} \bm \Delta}_{\max} \\
 & \qquad \qquad \qquad \qquad \qquad \times \max_j \norm{\left( \mathcal{L} \bm w^\star \right)^\dagger \mathcal{L} \bm \Delta \bm N \left( \mathcal{L}w^\star \right)^\dagger \bm e_j }_1 \\
& \leq \normI{\mathcal{L}^\ast}_{\infty} \cdot \normI{\left(  \mathcal{L} \bm w^\star \right)^\dagger }_{\infty} \cdot \norm{\mathcal{L} \bm \Delta}_{\max} \cdot \normI{\left(  \mathcal{L} \bm w^\star \right)^\dagger \mathcal{L} \bm \Delta \bm N \left( \mathcal{L} \bm w^\star \right)^\dagger }_1 \\
& \leq \normI{\mathcal{L}^\ast}_{\infty} \cdot \normI{\left( \mathcal{L} \bm w^\star \right)^\dagger }_{\infty}^3 \cdot \normI{\mathcal{L} \bm \Delta}_\infty \cdot \norm{\mathcal{L} \bm \Delta}_{\max} \cdot \normI{\bm N^\top}_{\infty} \\
& \leq 8 d^2 \normI{\left( \mathcal{L} \bm w^\star \right)^\dagger }_{\infty}^3 \cdot  \normI{\bm N^\top}_{\infty} \cdot \norm{\bm \Delta}_{\max}^2,
\end{split}
\end{equation}
where the first inequality follows from the fact that $\left|\bm a^\top \bm b \right| \leq \norm{\bm a}_{\max} \norm{\bm b}_1$ holds for any $\bm a, \bm b \in \mathbb{R}^p$; the second inequality is established by the inequality that $\norm{\bm a^\top \bm B}_{\max} \leq \norm{\bm a}_1 \norm{\bm B}_{\max}$ for any $\bm a \in \mathbb{R}^p$ and $\bm B \in \mathbb{R}^{p \times q}$; the forth inequality follows from $\normI{\bm A}_1 = \normI{\bm A^\top}_\infty$ and the sub-multiplicativity of $\normI{\cdot}_\infty$; the last inequality follows from $\normI{\mathcal{L}^\ast}_{\infty}=4$ by Lemma \ref{lem-L}, $\normI{\mathcal{L}\bm \Delta}_\infty \leq 2 d \norm{\bm \Delta}_{\max}$, and $\norm{\mathcal{L}\bm \Delta}_{\max} \leq d \norm{\bm \Delta}_{\max}$. We further have
\begin{equation} \label{Bound_N}
\normI{\bm N^\top}_\infty \leq \sum_{k=0}^\infty \normI{  \mathcal{L} \bm \Delta \left( \mathcal{L} \bm w^\star \right)^\dagger }_\infty^k \leq \frac{1}{1- \normI{\left( \mathcal{L} \bm w^\star \right)^\dagger}_\infty \normI{\mathcal{L}\bm \Delta}_\infty} \leq \frac{1}{1 -2K_1 r d} \leq \frac{3}{2},
\end{equation}
where the last inequality follows from \eqref{expan}. Together with \eqref{Bound_LR} and \eqref{Bound_N}, we obtain
\begin{equation}\label{L_Rhat}
\norm{\mathcal{L}^\ast \bm R^\Delta}_{\max} \leq 12 K_1^3 r^2 d^2.
\end{equation}
Thus, $F\left(\bm \Delta_{\mathcal{S}^\star} \right)$ in \eqref{F_bound} can  be bounded by
\begin{equation*}
\begin{split}
\norm{F \left( \bm \Delta_{\mathcal{S}^\star} \right)}_{\max}  &\leq K_2 \left( \frac{r}{2K_2} + 12 K_1^3 r^2 d^2 \right) \\
& \leq \frac{r}{2} + 24 K_1^3 K_2^2 r d^2 \norm{ \left[  \mathcal{L}^\ast \left( \mathcal{L} \bm w^\star + \bm J \right)^{-1} \right]_{\mathcal{S}^\star} - \left[ \mathcal{L}^\ast \bm S \right]_{\mathcal{S}^\star} }_{\max}.
\end{split}
\end{equation*}
Following from \eqref{cond_sam}, we obtain
\begin{equation*}
\norm{F \left( \bm \Delta_{\mathcal{S}^\star} \right)}_{\max} \leq r.
\end{equation*}
Therefore, we have established \eqref{fixed point}.


Next, we will show that $ \norm{ \nabla_{\left\lbrace \mathcal{S}^\star \right\rbrace^c}  f \left( \bm w^{\mathrm{oracle}}  \right)  }_{\max}  \leq \frac{1}{2} \lambda$ holds under the conditions that
\begin{equation}\label{cond1}
\norm{ \left[  \mathcal{L}^\ast \left( \mathcal{L} \bm w^\star + \bm J \right)^{-1} \right]_{\mathcal{S}^\star}  - \left[ \mathcal{L}^\ast \bm S \right]_{\mathcal{S}^\star} }_{\max} \leq \min \left( \frac{1}{48 K_1^3 K_2^2 d^2}, \ \frac{1}{12 K_1 K_2 d}, \ \frac{c \lambda}{2 K_2}, \ \frac{\lambda}{4(2 K_3 + 1)} \right),
\end{equation}
and
\begin{equation}\label{cond2}
\norm{ \left[  \mathcal{L}^\ast  \left(  \mathcal{L} \bm w^\star + \bm J \right)^{-1}  \right]_{ \left\lbrace \mathcal{S}^\star \right\rbrace^c }  - \left[  \mathcal{L}^\ast \bm S\right]_{ \left\lbrace  \mathcal{S}^\star \right\rbrace^c} }_{\max} \leq \frac{\lambda}{4}.
\end{equation}
We have
\begin{equation*}
\nabla_{ \left\lbrace \mathcal{S}^\star \right\rbrace^c }  f \left( \bm w^{\mathrm{oracle}} \right)  = \left[  - \mathcal{L}^\ast \left( \mathcal{L} \bm w^{\mathrm{oracle}} + \bm J \right)^{-1} \right]_{ \left\lbrace \mathcal{S}^\star  \right\rbrace^c }  + \left[ \mathcal{L}^\star \bm S \right]_{ \left\lbrace \mathcal{S}^\star \right\rbrace^c }.
\end{equation*}
Then we get
\begin{equation} \label{B_nabla}
\begin{split}
& \norm{ \nabla_{ \left\lbrace \mathcal{S}^\star  \right\rbrace^c } f \left( \bm w^{ \mathrm{oracle} } \right) }_{\max}  \leq \norm{ - \left[ \mathcal{L}^\ast \left( \mathcal{L} \bm w^\star + \bm J \right)^{-1} \right]_{ \left\lbrace \mathcal{S}^\star \right\rbrace^c } + \left[ \mathcal{L}^\ast \bm S \right]_{ \left\lbrace \mathcal{S}^\star \right\rbrace^c } }_{\max} \\
& \qquad \qquad \qquad \qquad + \norm{ \left[ - \mathcal{L}^\ast \left( \mathcal{L} \bm w^{\mathrm{oracle}} + \bm J \right)^{-1} \right]_{ \left\lbrace  \mathcal{S}^\star \right\rbrace^c }   +  \left[ \mathcal{L}^\ast \left( \mathcal{L} \bm w^\star + \bm J \right)^{-1} \right]_{ \left\lbrace \mathcal{S}^\star \right\rbrace^c } }_{\max}.  \\
\end{split}
\end{equation}
Recall that $\bm w^{ \mathrm{oracle}} = \bm w^\star +  \widehat{\bm \Delta}$. By \eqref{sep_inv}, we can obtain
\begin{equation} \label{S-part}
\left[  \mathcal{L}^\ast \left( \mathcal{L} \bm w^{\mathrm{oracle}} + \bm J \right)^{-1} \right]_{\mathcal{S}^\star}  - \left[ \mathcal{L}^\ast \left( \mathcal{L} \bm w^\star \right)^\dagger \right]_{\mathcal{S}^\star} = - \bm H_{\mathcal{S}^\star \mathcal{S}^\star}  \widehat{\bm \Delta}_{\mathcal{S}^\star}  + \left[ \mathcal{L}^\ast \widehat{\bm R}^\Delta \right]_{\mathcal{S}^\star},
\end{equation}
where $\widehat{\bm R}^\Delta = \left( \mathcal{L} \bm w^\star \right)^\dagger \mathcal{L} \widehat{\bm \Delta}   \left( \mathcal{L} \bm w^\star \right)^\dagger \mathcal{L} \widehat{\bm \Delta}  \widehat{\bm N} \left( \mathcal{L} \bm w^\star \right)^\dagger$ with $\widehat{\bm N} = \sum_{k=0}^\infty \left( -1 \right)^k \left( \left( \mathcal{L} \bm w^\star \right)^\dagger \mathcal{L} \widehat{\bm \Delta} \right)^k$. Note that $\widehat{\bm N}$ is also a convergent matrix series under the condition \eqref{cond_sam}. Similarly, we can obtain
\begin{equation}\label{Sc-part}
\left[  \mathcal{L}^\ast \left( \mathcal{L} \bm w^{\mathrm{oracle}} + \bm J \right)^{-1} \right]_{ \left\lbrace \mathcal{S}^\star \right\rbrace^c }  - \left[ \mathcal{L}^\ast \left( \mathcal{L} \bm w^\star \right)^\dagger \right]_{ \left\lbrace  \mathcal{S}^\star \right\rbrace^c} = - \bm H_{ \left\lbrace \mathcal{S}^\star \right\rbrace^c \mathcal{S}^\star}  \widehat{\bm \Delta}_{\mathcal{S}^\star}  + \left[ \mathcal{L}^\ast \widehat{\bm R}^\Delta \right]_{ \left\lbrace \mathcal{S}^\star \right\rbrace^c }.
\end{equation}
Combining \eqref{S-part} and \eqref{Sc-part}, we get
\begin{equation}\label{S-Sc-Com}
\begin{split}
& \left[  \mathcal{L}^\ast \left( \mathcal{L} \bm w^{\mathrm{oracle}} + \bm J \right)^{-1} \right]_{ \left\lbrace \mathcal{S}^\star \right\rbrace^c }  - \left[ \mathcal{L}^\ast \left( \mathcal{L} \bm w^\star \right)^\dagger \right]_{ \left\lbrace  \mathcal{S}^\star \right\rbrace^c}  = \left[  \mathcal{L}^\ast \widehat{\bm R}^\Delta \right]_{  \left\lbrace \mathcal{S}^\star \right\rbrace^c} + \bm H_{ \left\lbrace \mathcal{S}^\star \right\rbrace^c \mathcal{S}^\star }  \left( \bm H_{ \mathcal{S}^\star \mathcal{S}^\star } \right)^{-1} \\
& \qquad \qquad \qquad \qquad \times \left( \left[ \mathcal{L}^\ast \left( \mathcal{L} \bm w^{\mathrm{oracle}} +\bm J \right)^{-1} \right]_{\mathcal{S}^\star} - \left[ \mathcal{L}^\ast \left( \mathcal{L} \bm w^\star \right)^\dagger \right]_{\mathcal{S}^\star} - \left[ \mathcal{L}^\ast \widehat{\bm R}^\Delta \right]_{\mathcal{S}^\star } \right) .
\end{split}
\end{equation}
The KKT conditions in \eqref{oracle_KKT} together with $\norm{\bm w_{\mathcal{S}^\star}^{\mathrm{oracle}}}_{\min} \geq  \gamma \lambda$ yield $\hat{\bm \upsilon}^{\mathrm{o}}_{\mathcal{S}^\star} = \bm 0$, and thus
\begin{equation*}
\left[ \mathcal{L}^\ast \left( \mathcal{L} \bm w^{\mathrm{oracle}} +\bm J \right)^{-1} \right]_{\mathcal{S}^\star} - \left[  \mathcal{L}^\ast \bm S  \right]_{\mathcal{S}^\star}  = \bm 0.
\end{equation*} 
Then, we have
\begin{equation}\label{S-par}
\left[ \mathcal{L}^\ast \left( \mathcal{L} \bm w^{\mathrm{oracle}} +\bm J \right)^{-1} \right]_{\mathcal{S}^\star} - \left[ \mathcal{L}^\ast \left( \mathcal{L} \bm w^\star \right)^\dagger \right]_{\mathcal{S}^\star} =  \left[  \mathcal{L}^\ast \bm S  \right]_{\mathcal{S}^\star}  - \left[   \mathcal{L}^\ast \left(   \left(  \mathcal{L} \bm w^\star + \bm J  \right)^{-1}  \right)  \right]_{\mathcal{S}^\star}.
\end{equation}
Together with \eqref{S-Sc-Com} and \eqref{S-par}, we obtain
\begin{equation}\label{Bound1}
\begin{split}
& \norm{ \left[ \mathcal{L}^\ast \left( \mathcal{L} \bm w^{\mathrm{oracle}} + \bm J \right)^{-1} \right]_{ \left\lbrace \mathcal{S}  \right\rbrace^c } - \left[  \mathcal{L}^\ast \left(   \mathcal{L} \bm w^\star \right)^\dagger \right]_{  \left\lbrace \mathcal{S}^\star  \right\rbrace^c }   }_{\max}  \leq  \norm{ \left[ \mathcal{L}^\ast \widehat{\bm R}^\Delta \right]_{ \left\lbrace  \mathcal{S}^\star \right\rbrace^c } }_{\max}   \\ 
& \quad  +\normI{ \bm H_{ \left\lbrace \mathcal{S}^\star \right\rbrace^c \mathcal{S}^\star }  \left( \bm H_{\mathcal{S}^\star  \mathcal{S}^\star }  \right)^{-1} }_{\infty}  \left(   \norm{ \left[ \mathcal{L}^\ast \left( \mathcal{L} \bm w^\star + \bm J \right)^{-1} \right]_{\mathcal{S}^\star} -\left[ \mathcal{L}^\ast \bm S \right]_{\mathcal{S}^\star} }_{\max}  +  \norm{  \left[ \mathcal{L}^\ast \widehat{\bm R}^\Delta \right]_{\mathcal{S}^\star } }_{\max} \right) .
\end{split}
\end{equation}
Note that \eqref{L_Rhat} also holds for $\widehat{\bm R}^\Delta$. Thus, we have
\begin{equation}\label{Bound2}
\begin{split}
\norm{ \mathcal{L}^\ast \widehat{\bm R}^\Delta }_{\max} & \leq 12  K_1^3 r^2d^2  = 48 K_1^3 K_2^2 d^2 \norm{ \left[ \mathcal{L}^\ast \left( \mathcal{L} \bm w^\star + \bm J \right)^{-1} \right]_{\mathcal{S}^\star} - \left[ \mathcal{L}^\ast \bm S \right]_{\mathcal{S}^\star}  }_{\max}^2 \\
& \leq \norm{ \left[ \mathcal{L}^\ast \left(  \mathcal{L} \bm w^\star + \bm J  \right)^{-1} \right]_{\mathcal{S}^\star} - \left[ \mathcal{L}^\ast \bm S \right]_{\mathcal{S}^\star}  }_{\max},
\end{split}
\end{equation}
where the last inequality follows from \eqref{cond1}. Using the definition of $K_3$ in \eqref{constant_K} and combining \eqref{Bound1} and \eqref{Bound2} yield
\begin{equation*}
\begin{split}
& \norm{ \left[ \mathcal{L}^\ast \left( \mathcal{L} \bm w^{\mathrm{oracle}} + \bm J \right)^{-1} \right]_{ \left\lbrace \mathcal{S}  \right\rbrace^c } - \left[  \mathcal{L}^\ast \left(   \mathcal{L} \bm w^\star \right)^\dagger \right]_{  \left\lbrace \mathcal{S}^\star  \right\rbrace^c }   }_{\max}   \leq \left (  2K_3 + 1 \right) \\
& \qquad \qquad \qquad \qquad \qquad \qquad \qquad \qquad \times \norm{ \left[  \mathcal{L}^\ast  \left(  \mathcal{L} \bm w^\star + \bm J \right)^{-1}  \right]_{\mathcal{S}^\star}  - \left[  \mathcal{L}^\ast \bm S\right]_{\mathcal{S}^\star} }_{\max}.
\end{split}
\end{equation*}
According to \eqref{B_nabla}, we obtain
\begin{equation}
\norm{ \nabla_{ \left\lbrace \mathcal{S}^\star  \right\rbrace^c } f \left( \bm w^{ \mathrm{oracle} } \right) }_{\max} \leq \frac{\lambda}{4} + \left (  2K_3 + 1 \right) \norm{ \left[  \mathcal{L}^\ast  \left(  \mathcal{L} \bm w^\star + \bm J \right)^{-1}  \right]_{\mathcal{S}^\star}  - \left[  \mathcal{L}^\ast \right]_{\mathcal{S}^\star} }_{\max} \leq \frac{\lambda}{2},
\end{equation}
where the first and second inequalities follow from \eqref{cond1} and \eqref{cond2}, respectively.

To sum up, we conclude that $\norm{\bm w_{\mathcal{S}^\star}^{\mathrm{oracle}}}_{\min} \geq  \gamma \lambda $ and $ \norm{ \nabla_{\left\lbrace \mathcal{S}^\star \right\rbrace^c}  f \left( \bm w^{\mathrm{oracle}}  \right)  }_{\max}  \leq \frac{1}{2} \lambda$ hold under the conditions in \eqref{cond1} and \eqref{cond2}. To guarantee the \eqref{cond1} hold with a high probability, we impose the condition that $c_H \lambda \leq \min \left(\frac{1}{48 K_1^3 K_2^2 d^2}, \ \frac{1}{12 K_1 K_2 d} \right)$, where $c_H =\min \left( \frac{c}{2 K_2}, \ \frac{1}{4(2 K_3 + 1)} \right)$ is  a constant depending on the underlying true graph. Then \eqref{cond1} could be written as
\begin{equation}\label{cond3}
\norm{ \left[  \mathcal{L}^\ast \left( \mathcal{L} \bm w^\star + \bm J \right)^{-1} \right]_{\mathcal{S}^\star}  - \left[ \mathcal{L}^\ast \bm S \right]_{\mathcal{S}^\star} }_{\max} \leq c_H \lambda.
\end{equation}
Note that $\lambda = \sqrt{4 \alpha c_0^{-1} \log p /n}$. Thus, to guarantee that $c_H \lambda \leq \min \left(\frac{1}{48 K_1^3 K_2^2 d^2}, \ \frac{1}{12 K_1 K_2 d} \right)$, the sample size should be lower bounded by 
\begin{equation*}
n \geq 576 \alpha c_0^{-1} c_H^2 \log p \max \left(16 K_1^6 K_2^4 d^4, K_1^2 K_2^2 d^2 \right).
\end{equation*}

Finally, we compute the probability that \eqref{cond2} and \eqref{cond3} hold. We apply Lemma 5.8 and obtain
\begin{equation*}
\begin{split}
\mathbb{P} \left[ \norm{ \left[  \mathcal{L}^\ast  \left(  \mathcal{L} \bm w^\star + \bm J \right)^{-1}  \right]_{ \left\lbrace \mathcal{S}^\star \right\rbrace^c }  - \left[  \mathcal{L}^\ast \bm S\right]_{ \left\lbrace  \mathcal{S}^\star \right\rbrace^c} }_{\max} \geq \frac{\lambda}{4}\right] & \leq \left(  p(p-1)- 2s \right) \exp \left( - \frac{1}{16} c_0 n \lambda^2  \right) \\
&=\left(  p(p-1) - 2s \right) p^{- \frac{1}{4} \alpha},
\end{split}
\end{equation*}
with the condition on the sample size that 
\begin{equation*}
n \geq 2 \alpha \log p.
\end{equation*}
Similarly, we can get
\begin{equation*}
\mathbb{P} \left[ \norm{ \left[  \mathcal{L}^\ast  \left(  \mathcal{L} \bm w^\star + \bm J \right)^{-1}  \right]_{ \mathcal{S}^\star }  - \left[  \mathcal{L}^\ast \bm S\right]_{ \mathcal{S}^\star} }_{\max} \geq c_H \lambda \right] \leq 2s\exp \left( - c_0 c_H^2 n \lambda^2  \right) = 2s \cdot p^{-4  c_H^2 \alpha},
\end{equation*}
with the condition on the sample size that 
\begin{equation*}
n \geq 32 \alpha c_H^2 \log p.
\end{equation*}
Therefore, \eqref{cond2} and \eqref{cond3} hold with probability at least $1- \left(  p(p-1) - 2s \right) p^{- \frac{1}{4}\alpha } - 2s \cdot p^{-4 c_H^2 \alpha }$, with the sample size lower bounded by
\begin{equation*}
n \geq \max \left( 576 \alpha c_0^{-1} c_H^2 \log p \max \left(16 K_1^6 K_2^4 d^4, K_1^2 K_2^2 d^2 \right), 2 \alpha \log p, 32 \alpha c_H^2 \log p \right).
\end{equation*}

\end{proof}

\subsection{Proof of Theorem \ref{theorem 4}}\label{sec-prof-gradient-converg}
%

\begin{proof}
Take the regularization parameter $\lambda = \sqrt{4\alpha c_0^{-1} \log p /n}$ for some $\alpha > 2$, and the sample size 
\begin{equation}
n \geq \max \left(840 \alpha c_0^{-1} \frac{ \left(\delta \tau^2 +1 \right)^4}{\delta^2 \tau^2} s p \log p, 8 \alpha \log p \right),  \label{sample}   
\end{equation}
where $c_0$ is a constant defined in Lemma \ref{lem8}. The sample size $n$ in \eqref{sample} satisfies the conditions on the number of samples in Lemmas \ref{lem3}, \ref{lem4}, \ref{lem5} and \ref{lem9}. Define an event 
\begin{equation*}
\mathcal{J} = \left\lbrace \norm{ \L^{\ast} \left( \left( \L \bm w^{\star} + \bm J  \right)^{-1} - \bm S  \right) }_{\max} \leq \lambda/2 \right\rbrace.
\end{equation*}
According to Lemma \ref{lem9}, the event $\mathcal{J} $ holds with probability at least $1-1/p^{\alpha-2}$. Recall that the initial point $\hat{\bm w}^{(0)}$ of Algorithm \ref{algo-1} satisfies $ \left| \mathrm{supp}^+ \left( \hat{\bm w}^{(0)} \right) \right| \leq s$.

Define a local region around $\bm w^{\star}$
\begin{equation}
\mathcal{B} \left (\bm w^{\star}; \frac{1}{\sqrt{2p} \delta \tau} \right)=\left\lbrace \bm w \, \big | \bm w \in \mathbb{B} \left( \bm w^{\star}; \frac{1}{\sqrt{2p} \delta \tau} \right) \cap \mathcal{S}_{\bm w} \right\rbrace,  \nonumber
\end{equation}
where $\tau$ is defined in \eqref{tau-def} and $\delta > 1$, $\mathbb{B} \left( \bm w^{\star}; r \right) =  \left\lbrace \bm w \in \mathbb{R}^{p(p-1)/2} \, | \norm{ \bm w - \bm w^{\star}} \leq r \right\rbrace$ and $\mathcal{S}_{\bm w} = \left\lbrace \bm w  \in \mathbb{R}^{p(p-1)/2} \, | \bm w \geq \bm 0, (\L\bm w + \bm J) \in \S_{++}^p \right\rbrace$.  Notice that $\mathcal{B}\left(\bm w^{\star}; \frac{1}{\sqrt{2p} \delta \tau} \right)$ must be nonempty because $\bm w^\star \in \mathcal{S}_{\bm w}$ according to Theorem \ref{Theorem 2} together with the fact that $\bm w^\star$ contains the weights from a connected graph.  According to Lemma \ref{lem12}, $-\log \det (\L \bm w + \bm J)$ is $\mu$-strongly convex and $L$-smooth in $\mathcal{B} \left(\bm w^{\star}; \frac{1}{\sqrt{2p} \delta \tau} \right)$ with $\mu = \frac{2}{ \left(1+\delta^{-1} \right)^2\tau^2}$ and $L=\frac{2p \tau^2}{ \left(1-\delta^{-1} \right)^2}$. Therefore, $f_k(\bm w) =-\log \det (\L \bm w + \bm J) + \tr{\bm S \L \bm w}+ \langle \bm z^{(k-1)}, \bm w \rangle$ as defined in \eqref{mm-3} is also $\mu$-strongly convex and $L$-smooth in $\mathcal{B} \left(\bm w^{\star}; \frac{1}{\sqrt{2p} \delta \tau} \right)$, for any $k \geq 1$.

We define another local region 
\begin{equation}
\mathcal{B} \left(\hat{\bm w}^{(k)}; \frac{2}{3\sqrt{2p} \delta \tau} \right)=\left\lbrace \bm w \, \big | \bm w \in \mathbb{B} \left(\hat{\bm w}^{(k)};\frac{2}{3\sqrt{2p} \delta \tau} \right) \cap \mathcal{S}_{\bm w} \right\rbrace.  \nonumber
\end{equation}
Next, we will prove that under the event $\mathcal{J}$, 
\begin{equation}
\mathcal{B} \left( \hat{\bm w}^{(k)}; \frac{2}{3\sqrt{2p} \delta \tau} \right) \subseteq \mathcal{B} \left(\bm w^{\star}; \frac{1}{\sqrt{2p} \delta \tau} \right)  \nonumber
\end{equation}
holds for any $k \geq 1$, where $\hat{\bm w}^{(k)}$ is defined in \eqref{cost}. Before applying Lemma \ref{lem3}, we first check the necessary conditions of the lemma. Let $\bm z^{(k-1)}$ satisfy $z_i^{(k-1)} = h'_{\lambda} \left(\hat{w_i}^{(k-1)} \right)$, $i \in [p(p-1)/2]$. One has $z_i^{(k-1)}  \in [0, \lambda]$ for $i \in [p(p-1)/2]$ by Assumption \ref{assumption 1}. According to the definition of $\E^{(k)}$ in \eqref{e-set}, one has $\mathcal{S}^\star \subseteq \E^{(k)}$. For any $i \in \left\{ \S^{(k)}\right\}^c$, one further has
\begin{equation}
z_i^{(k-1)} = h'_{\lambda} \left(\hat{w}_i^{(k-1)} \right) \geq h'_{\lambda} (b) \geq \frac{\lambda}{2},   \nonumber
\end{equation} 
where the first inequality holds because $\hat{w}_i^{(k-1)} < b$ for any $i \in \left\{ \S^{(k)} \right\}^c$, and $h'_{\lambda}$ is non-increasing by Assumption \ref{assumption 1}; the second inequality follows from Assumption \ref{assumption 1}. Hence one obtains
\begin{equation}
\norm{ \bm z_{\left\{ \E^{(k)}\right\}^c}^{(k-1)} }_{\min} \geq \norm{ \bm z_{\left\{ \S^{(k)}\right\}^c}^{(k-1)} }_{\min} \geq \lambda/2.   \nonumber
\end{equation} 
Under the event $\mathcal{J}$, $ \left| \E^{(k)} \right| \leq 2s$ holds for any $k \geq 1$ following from Lemma \ref{lem4}. Therefore, all the conditions in Lemma \ref{lem3} are satisfied with $\E = \E^{(k)}$ and $\bm z =\bm z^{(k-1)}$, and one has
\begin{equation}\label{T3-1}
\norm{ \hat{\bm w}^{(k)} - \bm w^{\star} } \leq \frac{\sqrt{2}}{2} \norm{\L \hat{\bm w}^{(k)} - \L \bm w^{\star}}_{\mathrm{F}} \leq (2+\sqrt{2}) \lambda_{\max}^2(\L \bm w^{\star}) \sqrt{s} \lambda \leq  \frac{1}{3\sqrt{2p} \delta \tau}, 
\end{equation}
where the first inequality follows from \eqref{lw-w} and the third inequality is established by plugging $\lambda = \sqrt{4\alpha c_0^{-1} \log p /n}$ with $n\geq 840 \alpha c_0^{-1} \frac{ \left(\delta \tau^2 +1 \right)^4}{\delta^2 \tau^2} s p \log p$. The \eqref{T3-1} indicates that
\begin{equation}\label{T3-2}
\hat{\bm w}^{(k)} \in \mathbb{B}  \left(\bm w^{\star}; \frac{1}{3\sqrt{2p} \delta \tau} \right) 
\end{equation}
holds for any $k \geq 1$. For any $\bm w \in \mathbb{B} \left(\hat{\bm w}^{(k)}; \frac{2}{3\sqrt{2p} \delta \tau} \right)$,
\begin{equation}
\norm{\bm w - \bm w^{\star}} \leq \norm{\bm w - \hat{\bm w}^{(k)} } + \norm{ \bm w^{\star} - \hat{\bm w}^{(k)}} \leq \frac{1}{\sqrt{2p} \delta \tau},  \nonumber
\end{equation}
indicating that $\mathbb{B} \left(\hat{\bm w}^{(k)}; \frac{2}{3\sqrt{2p} \delta \tau} \right) \subseteq \mathbb{B} \left(\bm w^{\star}; \frac{1}{\sqrt{2p} \delta \tau} \right)$. Consequently, one has
\begin{equation}\label{subset}
\mathcal{B} \left( \hat{\bm w}^{(k)}; \frac{2}{3\sqrt{2p} \delta \tau} \right) \subseteq \mathcal{B} \left(\bm w^{\star}; \frac{1}{\sqrt{2p} \delta \tau} \right), \quad \forall \, k\geq 1. 
\end{equation}
Therefore, $f_k(\bm w)$ is $\mu$-strongly convex and $L$-smooth in $\mathcal{B} \left(\hat{\bm w}^{(k)}; \frac{2}{3\sqrt{2p} \delta \tau} \right)$, for any $k \geq 1$.

Then we will establish that the whole sequence $\big\{ \bm w^{(k)}_{t} \big\}_{t \geq 0}$ returned from Algorithm 2 is within the region $\mathcal{B} \left(\hat{\bm w}^{(k)}; \frac{2}{3\sqrt{2p} \delta \tau} \right)$ for any $k \geq 2$. We first prove that $\bm w^{(k)}_{t} \in \mathcal{S}_{\bm w}$ for any $k \geq 2$ and $t \geq 0$. Any $\bm w^{(k)}_{t}$ returned from Algorithm 2 must obey $\bm w^{(k)}_{t} \geq \bm 0$ because of the projection $\mathcal{P}_+$. Notice that $\L \bm w + \bm J$ must be positive semi-definite for any $\bm w \geq \bm 0$, because $\L \bm w$ is positive semi-definite for any $\bm w \geq \bm 0$ according to \eqref{Lx-psd}, and $\bm J$ has an unique nonzero eigenvalue $1$ with the eigenvector orthogonal to the row and column spaces of $\L \bm w$. Therefore, any $\bm w^{(k)}_{t} \notin \mathcal{S}_{\bm w}$ will lead $\L \bm w + \bm J$ to be singular and positive semi-definite, and thus the objective function $f_{k} \left({\bm w}^{(k)}_{t} \right)$ will go to infinity. On the other hand, one has 
\begin{align}
& f_{k} \left({\bm w}^{(k)}_{t} \right)  \leq f_{k} \left({\bm w}^{(k)}_{t-1} \right)  + \left\langle \nabla  f_{k} \left( \bm w^{(k)}_{t-1} \right), \bm w^{(k)}_{t}-  \bm w^{(k)}_{t-1} \right\rangle +\frac{1}{2 \eta} \norm{{\bm w}^{(k)}_{t} - {\bm w}^{(k)}_{t-1}}^2  \nonumber \\
& \qquad = f_{k} \left({\bm w}^{(k)}_{t-1} \right)  + \frac{1}{\eta} \left\langle \bm w^{(k)}_{t}-  \bm w^{(k)}_{t-1} + \eta \nabla  f_{k} \left( \bm w^{(k)}_{t-1} \right), \bm w^{(k)}_{t}-  \bm w^{(k)}_{t-1} \right\rangle - \frac{1}{2 \eta} \norm{{\bm w}^{(k)}_{t} - {\bm w}^{(k)}_{t-1}}^2 \nonumber \\
& \qquad \leq f_{k} \left({\bm w}^{(k)}_{t-1} \right) - \frac{1}{2 \eta} \norm{{\bm w}^{(k)}_{t} - {\bm w}^{(k)}_{t-1}}^2 \leq f_{k} \left({\bm w}^{(k)}_{t-1} \right), \label{nonincreasing}
\end{align}
where the first inequality is established by the backtracking exit inequality in Algorithm \ref{algo-2}; the second inequality is established by  
\begin{equation}
\frac{1}{\eta} \left\langle \bm w^{(k)}_{t} - \left(\bm w^{(k)}_{t-1} - \eta \nabla f_{k} \left(\bm w^{(k)}_{t-1} \right) \right), \bm w^{(k)}_{t}- \bm w^{(k)}_{t-1} \right\rangle  
 = \frac{1}{\eta} \left\langle \mathcal{P}_{+} \left(\tilde{ \bm w}\right ) - \tilde{\bm w}, \mathcal{P}_{+}\left ( \tilde{\bm w} \right) - \bm w^{(k)}_{t-1} \right\rangle  
\leq 0,      \nonumber
\end{equation}
where $\tilde{\bm w} =  \bm w^{(k)}_{t-1}  - \eta \nabla f_{k} \left(\bm w^{(k)}_{t-1} \right)$. The equality follows from the updating rule $ \bm w^{(k)}_{t}  =  \mathcal{P}_{+}  \left( \bm w^{(k)}_{t-1}  - \eta \nabla f_{k} \left(\bm w^{(k)}_{t-1} \right) \right)$ and the inequality holds because of the projection theorem together with the fact that $\bm w^{(k)}_{t-1} \geq \bm 0$. More specifically, let $S \subseteq \mathbb{R}^n$ be a closed and convex set. Then $\mathcal{P}_S(\bm{x})$ is the projection of $\bm{x}  \in \mathbb{R}^n$ on $S$, if and only if one can establish
\begin{equation}
\left\langle \mathcal{P}_S \left(\bm{x} \right) - \bm{x}, \mathcal{P}_S \left(\bm{x} \right) - \bm{z} \right\rangle \leq 0, \  \  \  \forall \ \bm{z} \in S. \label{proj_thm}
\end{equation}
The \eqref{nonincreasing} indicates that the objective function $f_{k} \left({\bm w}^{(k)}_{t} \right)$ is not increasing with the increase of $t$, and consequently one has $f_{k} \left(\bm w^{(k)}_{t} \right) \leq f_{k} \left(\bm w^{(k)}_{0} \right)$. Hence, if $f_{k} \left(\bm w^{(k)}_{0} \right)$ is upper bounded, then $f_{k} \left(\bm w^{(k)}_{t} \right)$ is also upper bounded and thus $\bm w^{(k)}_{t} \in \mathcal{S}_{\bm w}$. Notice that Algorithm \ref{algo-2} takes the initial point $\bm w^{(k)}_0 = \hat{\bm w}^{(k-1)}$, which is the minimizer of the optimization \eqref{cost}. Thus, $- \log {\det} \left(\mathcal{L} \hat{\bm w}^{(k-1)} + \bm J \right)+\tr{\Sm \mathcal{L} \hat{\bm w}^{(k-1)} }$ is upper bounded. $\sum_{i} h'_{\lambda} \left(\hat{w}_i^{(k-1)} \right) \hat{w}_i^{(k-1)}$ is also upper bounded following from Assumption \ref{assumption 1}. Therefore, $f_{k} \left(\bm w^{(k)}_{0} \right)$ is upper bounded for any $k\geq 2$. Then, we can conclude that 
\begin{equation}\label{Sw}
\bm w^{(k)}_{t} \in \mathcal{S}_{\bm w} 
\end{equation}
holds for any $k \geq 2$ and $t \geq 0$.

We will prove by induction that the sequence $\big\{ \bm w^{(k)}_{t} \big\}_{t \geq 0} $ is in $\mathcal{B} \left(\hat{\bm w}^{(k)}; \frac{2}{3\sqrt{2p} \delta \tau} \right)$ for any $k \geq 2$. For any given $k \geq 2$, when $t=0$, one has 
\begin{equation}
\norm{\bm w^{(k)}_{0} - \hat{\bm w}^{(k)}} = \norm{\hat{\bm w}^{(k-1)} - \hat{\bm w}^{(k)}} 
\leq \norm{\hat{\bm w}^{(k-1)} - \bm w^{\star}} + \norm{ \hat{\bm w}^{(k)} - \bm w^{\star}} \leq  \frac{2}{3\sqrt{2p} \delta \tau},  \nonumber
\end{equation}
where the second inequality follows from \eqref{T3-2}. Together with \eqref{Sw}, we conclude that $\bm w^{(k)}_{0} \in \mathcal{B} \left(\hat{\bm w}^{(k)}; \frac{2}{3\sqrt{2p} \delta \tau} \right)$. Suppose $\bm w^{(k)}_{t-1} \in \mathcal{B} \left(\hat{\bm w}^{(k)}; \frac{2}{3\sqrt{2p} \delta \tau} \right)$ for some $t \geq 1$. Then one obtains
\begin{equation}\label{q37}
\begin{split}
& \norm{ \bm w^{(k)}_{t} - \hat{\bm w}^{(k)} }^2  = \norm{\bm w^{(k)}_{t} - \bm w^{(k)}_{t-1}}^2 + \norm{ \hat{\bm w}^{(k)} - \bm w^{(k)}_{t-1}}^2 -2 \left\langle \bm w^{(k)}_{t} - \bm w^{(k)}_{t-1}, \hat{\bm w}^{(k)} - \bm w^{(k)}_{t-1} \right\rangle. 
\end{split}
\end{equation}
On the other hand, one has 
\begin{equation}\label{q35}
\begin{split}
 &  f_{k} \left(\bm w^{(k)}_{t} \right)  -  f_{k} \left(\hat{\bm w}^{(k)} \right) = f_{k} \left(\bm w^{(k)}_{t} \right) - f_{k} \left(\bm w^{(k)}_{t-1} \right) + f_{k} \left(\bm w^{(k)}_{t-1} \right) - f_{k} \left(\hat{\bm w}^{(k)} \right)  \\
 \qquad & \leq  \left\langle \nabla  f_{k} \left( \bm w^{(k)}_{t-1} \right), \bm w^{(k)}_{t}-  \bm w^{(k)}_{t-1} \right\rangle + \frac{1}{2 \eta} \norm{ \bm w^{(k)}_{t} - \bm w^{(k)}_{t-1}}^2         \\
 & \qquad \qquad \quad \quad + \left\langle \nabla f_{k} \left( \bm w^{(k)}_{t-1} \right),  \bm w^{(k)}_{t-1} -\hat{\bm w}^{(k)} \right\rangle -\frac{\mu}{2} \norm{\bm w^{(k)}_{t-1} - \hat{\bm w}^{(k)}}^2                \\
 \qquad &=  \left\langle \nabla f_{k}  \left( \bm w^{(k)}_{t-1} \right), \bm w^{(k)}_{t} - \hat{\bm w}^{(k)} \right\rangle + \frac{1}{2\eta} \norm{ \bm w^{(k)}_{t} - \bm w^{(k)}_{t-1}}^2 - \frac{\mu}{2} \norm{\bm w^{(k)}_{t-1} - \hat{\bm w}^{(k)}}^2             \\
 \qquad  & \leq \frac{1}{\eta}  \left\langle \bm w^{(k)}_{t-1} - \bm w^{(k)}_{t}, \bm w^{(k)}_{t} - \hat{\bm w}^{(k)} \right\rangle  + \frac{1}{2\eta} \norm{\bm w^{(k)}_{t} - \bm w^{(k)}_{t-1}}^2 - \frac{\mu}{2} \norm{\bm w^{(k)}_{t-1} - \hat{\bm w}^{(k)}}^2               \\
 \qquad  & =  \frac{1}{\eta} \left \langle  \bm w^{(k)}_{t-1} - \bm w^{(k)}_{t}, \bm w^{(k)}_{t-1}-\hat{\bm w}^{(k)} \right\rangle -  \frac{1}{2\eta} \norm{\bm w^{(k)}_{t} - \bm w^{(k)}_{t-1}}^2 -  \frac{\mu}{2} \norm{\bm w^{(k)}_{t-1} - \hat{\bm w}^{(k)}}^2, 
\end{split}
\end{equation}
where the first inequality follows from the backtracking exit inequality, and $f_{k}(\bm w)$ is $\mu$-strongly convex in the region $\mathcal{B} \left(\hat{\bm w}^{(k)}; \frac{2}{3\sqrt{2p} \delta \tau} \right)$ and $\bm w^{(k)}_{t-1}\in \mathcal{B}\left(\hat{\bm w}^{(k)}; \frac{2}{3\sqrt{2p} \delta \tau} \right)$. The last equality is obtained by plugging 
\begin{equation*}
\frac{1}{\eta} \left\langle \bm w^{(k)}_{t-1} - \bm w^{(k)}_{t}, \bm w^{(k)}_{t} - \hat{\bm w}^{(k)} \right\rangle = \frac{1}{\eta} \left\langle \bm w^{(k)}_{t-1} - \bm w^{(k)}_{t}, \bm w^{(k)}_{t - 1} - \hat{\bm w}^{(k)} \right \rangle - \frac{1}{\eta} \norm{\bm w^{(k)}_{t} - \bm w^{(k)}_{t-1} }^2
\end{equation*}
The second inequality in \eqref{q35} follows from
\begin{equation*}
\begin{split}
&\left\langle \nabla f_{k} \left(\bm w^{(k)}_{t-1} \right) - \frac{1}{\eta} \left(\bm w^{(k)}_{t-1} - \bm w^{(k)}_{t} \right), \bm w^{(k)}_{t} - \hat{\bm w}^{(k)} \right\rangle \\
=& \frac{1}{\eta} \left\langle \bm w^{(k)}_{t} - \left(\bm w^{(k)}_{t-1} - \eta \nabla f_{k} \left(\bm w^{(k)}_{t-1} \right) \right), \bm w^{(k)}_{t}- \hat{\bm w}^{(k)} \right\rangle      \\
=& \frac{1}{\eta} \left\langle \mathcal{P}_{+} \left(\tilde{ \bm w} \right) - \tilde{\bm w}, \mathcal{P}_{+} \left( \tilde{\bm w} \right) - \hat{\bm w}^{(k)} \right\rangle     \leq 0, 
\end{split}
\end{equation*}
where $\tilde{\bm w} =  \bm w^{(k)}_{t-1}  - \eta \nabla f_{k} \left(\bm w^{(k)}_{t-1} \right)$ and $ \bm w^{(k)}_{t}  =  \mathcal{P}_{+} \left(\tilde{\bm w} \right)$. The second equality follows from the updating rule $ \bm w^{(k)}_{t}  =  \mathcal{P}_{+} \left( \bm w^{(k)}_{t-1}  - \eta \nabla f_{k} \left(\bm w^{(k)}_{t-1} \right) \right)$ and the inequality is established by the projection theorem as shown in \eqref{proj_thm}, together with $\hat{\bm w}^{(k)} \geq \bm 0$. Substituting $f_{k} \left(\bm w^{(k)}_{t} \right) -f_{k} \left(\hat{\bm w}^{(k)} \right) \geq 0$ into \eqref{q35}, one obtains
\begin{equation}\label{q36}
2 \left\langle  \bm w^{(k)}_{t-1} - \bm w^{(k)}_{t}, \bm w^{(k)}_{t-1}- \hat{\bm w}^{(k)} \right\rangle -  \norm{\bm w^{(k)}_{t} - \bm w^{(k)}_{t-1}}^2 -  \mu \eta \norm{\bm w^{(k)}_{t-1} - \hat{\bm w}^{(k)}}^2 \geq 0.
\end{equation}
Substituting \eqref{q36} into \eqref{q37} yields
\begin{equation*}
\begin{split}
\norm{ \bm w^{(k)}_{t} - \hat{\bm w}^{(k)}}^2 & \leq \norm{\hat{\bm w}^{(k)} - \bm w^{(k)}_{t-1} }^2 - \mu \eta \norm{ \bm w^{(k)}_{t-1} - \hat{\bm w}^{(k)}
}^2        \\
& = (1 - \mu \eta) \norm{ \bm w^{(k)}_{t-1} - \hat{\bm w}^{(k)}}^2  < \norm{ \bm w^{(k)}_{t-1} - \hat{\bm w}^{(k)}},    
\end{split}
\end{equation*}
indicating that $\bm w^{(k)}_{t}$ is within the region $\mathbb{B} \left(\hat{\bm w}^{(k)}; \frac{2}{3\sqrt{2p} \delta \tau} \right)$. Together with \eqref{Sw}, we conclude that $\bm w^{(k)}_{t} \in \mathcal{B} \left(\hat{\bm w}^{(k)}; \frac{2}{3\sqrt{2p} \delta \tau} \right),$ completing the induction. Therefore, the whole sequence $ \big\{\bm w^{(k)}_{t} \big\}_{t \geq 0}$ is in $\mathcal{B} \left(\hat{\bm w}^{(k)}; \frac{2}{3\sqrt{2p} \delta \tau} \right)$, for any $k \geq 2$.

Finally, we will establish the lower bound of the step size $\eta$ for each iteration. Since $f_{k} (\bm w)$ is $L$-smooth in $\mathcal{B} \left(\hat{\bm w}^{(k)}; \frac{2}{3\sqrt{2p} \delta \tau} \right)$, for any $\bm w_1$ and $\bm w_2$ in $\mathcal{B} \left(\hat{\bm w}^{(k)}; \frac{2}{3\sqrt{2p} \delta \tau} \right)$, one has
\begin{equation} 
f_{k} \left(\bm w_2 \right) \leq f_{k} \left(\bm w_1 \right) + \left\langle \nabla f_{k} \left (\bm w_1 \right), \bm w_2 -\bm w_1 \right\rangle + \frac{L}{2} \norm{\bm w_1 -\bm w_2}^2.   \nonumber
\end{equation}
Thus any $\eta \in [0, \frac{1}{L}]$ must satisfy the backtracking exit condition and the backtracking line search must terminate when $\eta \leq \frac{\beta}{L} $ with $\beta \in (0, 1)$, implying that $\eta \geq \frac{\beta}{L}$. Therefore, we establish
\begin{equation}
\norm{ \bm w^{(k)}_{t} -  \hat{\bm w}^{(k)}}^2 \leq \left (1 - \frac{\beta \mu }{L} \right)^t \norm{ \bm w^{(k)}_{0} -  \hat{\bm w}^{(k)}}^2 = \rho^t \norm{ \bm w^{(k)}_{0} -  \hat{\bm w}^{(k)}}^2,   \nonumber
\end{equation}
where $\rho = 1- \frac{\beta \left(1-\delta^{-1} \right)^2}{p \tau^4 \left( 1+\delta^{-1} \right)^2} <1$, completing the proof.

\end{proof}

\subsection{Proof of Theorem \ref{k-component-convergence}}\label{sec-prof-k-comp}
\begin{proof}
Before proving Theorem \ref{k-component-convergence}, we first establish the boundedness of the sequence $ \left\lbrace \left( \bm \Theta^{(l)}, \bm w^{(l)}, \bm G^{(l)} \right) \right\rbrace$ generated by Algorithm \ref{algo-3}, and the monotonicity of $ \left\lbrace  L_\rho \left( \bm \Theta^{(l)}, \bm w^{(l)}, \bm G^{(l)} \right) \right\rbrace$.

Now we begin to prove by induction that the sequence $ \left\lbrace \left( \bm \Theta^{(l)}, \bm w^{(l)}, \bm G^{(l)} \right) \right\rbrace$ generated by Algorithm \ref{algo-3} is bounded. Let $\bm w^{(0)}$ and $\bm G^{(0)}$ be the initialization of the sequences $\left\lbrace \bm w^{(l)} \right\rbrace$ and $\left\lbrace \bm G^{(l)} \right\rbrace$, respectively, and $\norm{\bm w^{(0)}}$ and $\norm{\bm G^{(0)}}_{\mathrm{F}}$ are bounded. Recall that the sequence $ \left\lbrace \bm \Theta^l \right\rbrace$ is established by 
\begin{equation} \label{Theta-update-l}
\bm \Theta^{(l)} = \frac{1}{2} \bm U^{(l-1)} \left( \bm \Lambda^{(l-1)} + \sqrt{ \left( \bm {\Lambda}^{(l-1)} \right)^2 + \frac{4}{\rho} \bm{I}}\right) \left(\bm U^{(l-1)} \right)^\top,
\end{equation}
where $\bm U^{(l-1)} \bm {\Lambda}^{(l-1)} \left(\bm{U}^{(l-1)} \right)^\top$ is the eigenvalue decomposition of $\mathcal{L}\bm{w}^{(l-1)} - \frac{1}{\rho}\bm{G}^{(l-1)}$. When $l=1$, $\norm{\bm \Lambda^{(0)}}_{\mathrm{F}}$ is bounded because of the boundedness of $\norm{\bm w^{(0)}}$ and $\norm{\bm G^{(0)}}_{\mathrm{F}}$, and thus $\norm{\bm \Theta^{(1)}}_{\mathrm{F}}$ is bounded. The sequence $\left\lbrace \bm w^{(l)} \right\rbrace$ is established by solving the sub-problems
\begin{equation}\label{subpro-w-kcomp-l}
\min_{\bm w}  \tr{\mathcal{L} \bm w \bm S}  + \frac{\rho}{2} \norm{ \mathcal{L} \bm w - \bm \Theta^{(l)} -\frac{1}{\rho} \bm G^{(l-1)} }_{\mathrm{F}}^2 +  \sum_{i} h_{\lambda}(w_i), \quad \mathrm{subject \ to} \quad \bm w \geq \bm 0.
\end{equation}
The problem \eqref{subpro-w-kcomp-l} is solved by majorization minimization framework. Thus the objective function value is monotonically decreasing during iterations and $\bm w^{(l)}$ is a stationary point of \eqref{subpro-w-kcomp-l}. Let $f_l (\bm w) = \tr{\mathcal{L} \bm w \bm S}  + \frac{\rho}{2} \norm{ \mathcal{L} \bm w - \bm \Theta^{(l)} -\frac{1}{\rho} \bm G^{(l-1)} }_{\mathrm{F}}^2 +  \sum_{i} h_{\lambda}(w_i)$. We can check that $f_1 (\bm w)$ is coercive, and thus $\norm{\bm w^{(1)}}$ is bounded. Finally, $\bm G^{(l)} $ is updated by
\begin{equation} \label{update_Yl}
\bm G^{(l)} = \bm G^{(l-1)} + \rho \left( \bm \Theta^{(l)} - \mathcal{L} \bm w^{(l)}  \right).
\end{equation}
Then we can see that $\norm{\bm G^{(1)}}$ is also bounded, because both $\norm{\bm w^{(1)}}$ and $\norm{\bm \Theta^1}_{\mathrm{F}}$ are bounded. Therefore, $\left\lbrace \left( \bm \Theta^{(1)}, \bm w^{(1)}, \bm G^{(1)}\right) \right\rbrace$ is bounded.

Assume that $ \left\lbrace \left( \bm \Theta^{(l-1)}, \bm w^{(l-1)}, \bm G^{(l-1)}\right) \right\rbrace$ is bounded for some $l \geq 1$, i.e., each term in $ \left\lbrace \left( \bm \Theta^{(l-1)}, \bm w^{(l-1)}, \bm G^{(l-1)} \right) \right\rbrace$ is bounded under $\ell_2$-norm or Frobenius norm. Following from \eqref{Theta-update-l}, we can obtain that $\norm{\bm \Theta^{(l)}}_{\mathrm{F}}$ is bounded. The coercivity of $f_l (\bm w)$ can lead to the boundedness of  $\norm{\bm w^{(l)}}$. Then $\norm{\bm G^{(l)}}_{\mathrm{F}}$ is also bounded by \eqref{update_Yl}. Thus, $\left\lbrace \left( \bm \Theta^{(l)}, \bm w^{(l)}, \bm G^{(l)} \right) \right\rbrace$ is bounded, completing the induction. Therefore, we conclude that the sequence $\left\lbrace \left( \bm \Theta^{(l)}, \bm w^{(l)}, \bm G^{(l)}\right) \right\rbrace$ is bounded, and thus there must exist at least one limit point of $\left\lbrace \left( \bm \Theta^{(l)}, \bm w^{(l)}, \bm G^{(l)}\right) \right\rbrace$.

Next, we show that the sequence $ \left\lbrace L_\rho \left( \bm \Theta^{(l)}, \bm w^{(l)}, \bm G^{(l)} \right) \right\rbrace$ generated by Algorithm \ref{algo-3} is lower bounded, and decreasing for a sufficiently large $\rho$. 

According to \eqref{Laglagian-k-comp}, we have
\begin{equation}\label{Laglagian-k-comp-l}
\begin{split}
L_{\rho} \left(\bm \Theta^{(l)}, \bm w^{(l)}, \bm G^{(l)} \right) =  & -\log {\det}^\star \left(\bm \Theta^{(l)} \right) + \tr{\mathcal{L} \bm w^{(l)} \bm S} +  \sum_{i} h_{\lambda} \left(w_i^{(l)} \right) \\
& \qquad \qquad + \left\langle \bm G^{(l)}, \bm \Theta^{(l)} - \mathcal{L} \bm w^{(l)} \right\rangle + \frac{\rho}{2} \norm{\bm \Theta^{(l)} - \mathcal{L} \bm w^{(l)} }_{\mathrm{F}}^2,
\end{split}
\end{equation}
By the definition of $h_\lambda (\cdot)$, the term $\sum_{i} h_{\lambda}(w_i^{(l)})$ is lower bounded. The boundedness of $\left\lbrace \left( \bm \Theta^{(l)}, \bm w^{(l)}, \bm G^{(l)}\right) \right\rbrace$ can establish that $-\log {\det}^\star \left(\bm \Theta^{(l)} \right) + \tr{\mathcal{L} \bm w^{(l)} \bm S} $ and $\left\langle \bm G^{(l)}, \bm \Theta^{(l)} - \mathcal{L} \bm w^{(l)} \right\rangle + \frac{\rho}{2} \norm{\bm \Theta^{(l)} - \mathcal{L} \bm w^{(l)} }_{\mathrm{F}}^2$ are lower bounded. Thus $L_{\rho} \left(\bm \Theta^{(l)}, \bm w^{(l)}, \bm G^{(l)} \right) $ is lower bounded.

For any $l \in \mathbb{N}_+$, we have
\begin{equation}\label{Lap_decreas}
\begin{split}
&L_{\rho} \left(\bm \Theta^{(l+1)}, \bm w^{(l)}, \bm G^{(l)}  \right)  - L_{\rho} \left(\bm \Theta^{(l)}, \bm w^{(l)}, \bm G^{(l)} \right) 
= -\log {\det}^\star \left(\bm \Theta^{(l+1)} \right) + \left\langle \bm G^{(l)}, \bm \Theta^{(l+1)} \right\rangle \\ 
&+ \frac{\rho}{2} \norm{\bm \Theta^{(l+1)} - \mathcal{L} \bm w^{(l)} }_{\mathrm{F}}^2 - \left( -\log {\det}^\star (\bm \Theta^{(l)}) + \left\langle \bm G^{(l)}, \bm \Theta^{(l)} \right\rangle + \frac{\rho}{2} \norm{\bm \Theta^{(l)} - \mathcal{L} \bm w^{(l)} }_{\mathrm{F}}^2 \right) \leq 0,
\end{split}
\end{equation}
where the inequality follows from the fact that
\begin{equation*}
\bm \Theta^{(l+1)} = \arg \min_{\bm \Theta \succeq \bm 0} -\log {\det}^\star (\bm \Theta ) + \left\langle \bm G^{(l)}, \bm \Theta \right\rangle + \frac{\rho}{2} \norm{\bm \Theta - \mathcal{L} \bm w^{(l)} }_{\mathrm{F}}^2,  \ \mathrm{subject \ to} \ \mathrm{rank} (\bm \Theta) = p-k.
\end{equation*}
On the other hand, we obtain
\begin{equation}\label{eq:Lag-dif}
\begin{split}
    &L_{\rho} \left(\bm \Theta^{(l+1)}, \bm w^{(l)}, \bm G^{(l)} \right)  - L_{\rho} \left(\bm \Theta^{(l+1)}, \bm w^{(l+1)}, \bm G^{(l+1)} \right)   \\
    = & \left\langle \mathcal{L}^\ast  \bm S, \bm w^{(l)} -\bm w^{(l+1)} \right\rangle   + \left \langle \bm{G}^{(l)}, \bm \Theta^{(l+1)} - \mathcal{L}\bm{w}^{(l)} \right\rangle - \left \langle \bm{G}^{(l+1)}, \bm \Theta^{(l+1)} - \mathcal{L}\bm{w}^{(l+1)} \right\rangle \\
 & \ + \frac{\rho}{2}\norm{\bm \Theta^{(l+1)} - \mathcal{L}\bm w^{(l)}}^2_{\mathrm{F}} -\frac{\rho}{2} \norm{\bm \Theta^{(l+1)} - \mathcal{L}\bm{w}^{(l+1)}}^2_{\mathrm{F}}  +  \sum_{i}  h_{\lambda} \left(w_i^{(l)} \right) - h_{\lambda} \left(w_i^{(l+1)} \right)  \\
 = & \left\langle \mathcal{L}^\ast  \bm S, \bm w^{(l)} -\bm w^{(l+1)} \right\rangle   + \left\langle \bm G^{(l)}, \mathcal{L} \bm w^{(l+1)} - \mathcal{L} \bm w^{(l)} \right\rangle  +  \sum_{i} h_{\lambda} \left(w_i^{(l)} \right) - h_{\lambda} \left(w_i^{(l+1)} \right) \\
 &  \ + \frac{\rho}{2}\norm{\bm \Theta^{(l+1)} - \mathcal{L}\bm w^{(l)}}^2_{\mathrm{F}} +\frac{\rho}{2} \norm{\bm \Theta^{(l+1)} - \mathcal{L}\bm{w}^{(l+1)}}^2_{\mathrm{F}} \\
 \geq &  \sum_{i} h_{\lambda} \left(w_i^{(l)} \right) - h_{\lambda} \left(w_i^{(l+1)} \right) - h'_\lambda \left(  w^{(l+1)}_i \right) \left( w^{(l)}_i - w^{(l+1)}_i \right) \\
 & \ + \frac{\rho}{2} \norm{ \mathcal{L} \bm w^{(l+1)} - \mathcal{L} \bm w^{(l)}}_{\mathrm{F}}^2 - \frac{1}{\rho} \norm{\bm G^{(l+1)} - \bm G^{(l)}}_{\mathrm{F}}^2 \\
 \geq & \frac{\rho - L_h}{2} \norm{ \mathcal{L} \bm w^{(l+1)} - \mathcal{L} \bm w^{(l)}}_{\mathrm{F}}^2 - \frac{1}{\rho} \norm{\bm G^{(l+1)} - \bm G^{(l)}}_{\mathrm{F}}^2,
 \end{split}
\end{equation}
where $L_h$ is Lipschitz continuous gradient constant of $h_\lambda$. The second equality follows from $\bm G^{(l+1)} = \bm G^{(l)} + \rho \left( \bm \Theta^{(l+1)} - \mathcal{L} \bm w^{(l+1)}  \right)$, the first inequality is established by the fact that $\bm w^{(l+1)}$ is a stationary point of the problem
\begin{equation}\label{subpro-w-kcomp-l1}
\min_{\bm w}  \tr{\mathcal{L} \bm w \bm S}  + \frac{\rho}{2} \norm{ \mathcal{L} \bm w - \bm \Theta^{(l+1)} -\frac{1}{\rho} \bm G^{(l)} }_{\mathrm{F}}^2 +  \sum_{i} h_{\lambda}(w_i), \quad \mathrm{subject \ to} \quad \bm w \geq \bm 0.
\end{equation}
Let $g_{l+1} (\bm w) = \tr{\mathcal{L} \bm w \bm S}  + \frac{\rho}{2} \norm{ \mathcal{L} \bm w - \bm \Theta^{(l+1)} -\frac{1}{\rho} \bm G^{(l)} }_{\mathrm{F}}^2 +  \sum_{i} h_{\lambda}(w_i)$. The set of the stationary points of \eqref{subpro-w-kcomp-l1} is defined by
\begin{equation}\label{station}
\mathcal{X} = \left\lbrace  \bm w | \nabla g_{l+1} (\bm w)^\top (\bm x - \bm w) \geq 0, \forall \bm x \geq \bm 0  \right\rbrace.
\end{equation}
By setting $\bm x = \bm w^{(l)}$ and $\bm w = \bm w^{(l+1)}$ in \eqref{station}, we have
\begin{equation}\label{statio_ineq}
\left\langle \mathcal{L}^\ast \bm S, \bm w^{(l)} - \bm w^{(l+1)} \right\rangle \geq -  \left\langle \rho \mathcal{L}^\ast \left( \mathcal{L} \bm w^{(l+1)} - \bm \Theta^{(l+1)} - \frac{1}{\rho} \bm G^{(l)} \right) + h'_\lambda \left(  \bm w^{(l+1)}\right), \bm w^{(l)} - \bm w^{(l+1)} \right\rangle.
\end{equation}
The first inequality in \eqref{eq:Lag-dif} follows from \eqref{statio_ineq}. The last inequality in \eqref{eq:Lag-dif} is due to the fact that $h_\lambda$ is a concave function and has $L_h$-Lipschitz continuous gradient such that
\begin{equation}
h_{\lambda} \left(w_i^{(l)} \right) - h_{\lambda} \left(w_i^{(l+1)} \right) - h'_\lambda \left(  w^{(l+1)}_i \right) \left( w^{(l)}_i - w^{(l+1)}_i \right) \geq - \frac{L_h}{2} \left( w^{(l)}_i - w^{(l+1)}_i \right)^2
\end{equation}
holds for any $ i \in [p(p-1)/2]$. By calculation, we get that if $\rho$ is sufficiently large such that
\begin{equation}\label{rho_low_bound}
\rho \geq \max_l \left( L_h + \frac{2\norm{\bm G^{(l+1)} - \bm G^{(l)}}_{\mathrm{F}}}{\norm{ \mathcal{L} \bm w^{(l+1)} - \mathcal{L} \bm w^{(l)}}_{\mathrm{F}}} \right),
\end{equation}
then, together with \eqref{Lap_decreas}, we obtain that
\begin{equation}\label{Lap_dif_p}
L_{\rho} \left(\bm \Theta^{(l)}, \bm w^{(l)}, \bm G^{(l)} \right)  - L_{\rho} \left(\bm \Theta^{(l+1)}, \bm w^{(l+1)}, \bm G^{(l+1)} \right) \geq \rho \norm{\bm \Theta^{(l+1)} - \mathcal{L} \bm w^{(l+1)}}_{\mathrm{F}}^2 \geq 0,
\end{equation}
holds for any $l \in \mathbb{N}_+$. Therefore, the sequence $\left\lbrace L_{\rho} (\bm \Theta^{(l)}, \bm w^{(l)}, \bm G^{(l)})  \right\rbrace$ is decreasing.

Now we are ready to prove Theorem \ref{k-component-convergence}. We have proved that the sequence $ \left\lbrace \left( \bm \Theta^{(l)}, \bm w^{(l)}, \bm G^{(l)} \right) \right\rbrace$ is bounded. Therefore, there exists at least one convergent subsequence $ \left\lbrace \left( \bm \Theta^{(l_s)}, \bm w^{(l_s)}, \bm G^{(l_s)} \right) \right\rbrace_{s \in \mathbb{N}}$, which converges to a limit point denoted by $\left\lbrace \left( \bm \Theta^{(l_\infty)}, \bm w^{(l_\infty)}, \bm G^{(l_\infty)} \right) \right\rbrace$. We have also proved that the sequence $L_\rho \left( \bm \Theta^{(l)}, \bm w^{(l)}, \bm G^{(l)} \right)$ is monotonically decreasing and lower bounded, and thus is convergent. Note that the function $\log \det^\star (\bm \Theta)$ is continuous over the set $\mathcal{S}=\left\lbrace \bm \Theta  \in \mathcal{S}_+^p |  \mathrm{rank}(\bm \Theta) = p-k \right\rbrace$. We can get 
\begin{equation}
\lim_{l \to + \infty} L_\rho \left( \bm \Theta^{(l)}, \bm w^{(l)}, \bm G^{(l)} \right) = L_\rho \left( \bm \Theta^{(\infty)}, \bm w^{(\infty)}, \bm G^{(\infty)} \right) = L_\rho \left( \bm \Theta^{(l_\infty)}, \bm w^{(l_\infty)}, \bm G^{(l_\infty)} \right).
\end{equation}
By taking $l \to + \infty$ in \eqref{Lap_dif_p}, we have $\lim_{l \to + \infty} \norm{\mathcal{L} \bm w^{(l)} -\bm \Theta^{(l)} }_{\mathrm{F}} =0$. For any subsequence, we also have
\begin{equation}\label{eq:limit_w}
\lim_{s \to + \infty}  \norm{\mathcal{L}\bm w^{(l_s)} -\bm \Theta^{(l_s)} }_{\mathrm{F}} = 0,
\end{equation}
implying that $\bm G^{(l_\infty)}$ satisfies the condition of stationary point of $L_\rho \left(\bm \Theta, \bm w, \bm G \right)$ with respect to $\bm G$. According to \eqref{update_Yl}, we obtain
\begin{equation}\label{eq:limit_Y}
\lim_{s \to + \infty} \norm{ \bm G^{(l_s)} - \bm G^{(l_s -1)}}_{\mathrm{F}} = 0.
\end{equation}
For any limit point $\left\lbrace \left( \bm \Theta^{(l_\infty)}, \bm w^{(l_\infty)}, \bm G^{(l_\infty)} \right) \right\rbrace$ of the sequence $ \left\lbrace \left( \bm \Theta^{(l)}, \bm w^{(l)}, \bm G^{(l)} \right) \right\rbrace$, $\bm \Theta^{(l_\infty)}$ minimizes the following sub-problem
\begin{equation*}
\begin{split}
      \bm \Theta^{(l_\infty)} &= \underset{\substack{\mathrm{rank}(\bm \Theta) = p-k \\ \bm \Theta \succeq \mathbf{0}}} {\arg \min}   - \log {\det}^\star (\bm \Theta) + \left\langle\bm \Theta, \bm G^{(l_\infty-1)} \right\rangle +
      \frac{\rho}{2}\norm{\bm \Theta - \mathcal{L}\bm w^{(l_\infty-1)}}^2_{\mathrm{F}} \\
      &= \underset{\substack{\mathrm{rank}(\bm \Theta) = p-k \\ \bm \Theta \succeq \mathbf{0}}} {\arg \min}   - \log {\det}^\star (\bm \Theta) + \left\langle\bm \Theta, \bm G^{(l_\infty)} \right\rangle +
      \frac{\rho}{2}\norm{\bm \Theta - \mathcal{L}\bm w^{(l_\infty)}}^2_{\mathrm{F}}, 
      \end{split}
\end{equation*}
where the last equality follows from \eqref{eq:limit_w} and \eqref{eq:limit_Y}. Therefore, $\bm \Theta^{(l_\infty)} $ satisfies the condition of the stationary point of $L_\rho \left(\bm \Theta, \bm w, \bm G \right)$ with respect to $\bm \Theta$. Similarly, we can obtain that $ \bm w^{(l_\infty)} $ is a stationary point of the following problem
\begin{equation}\label{subpro-w-kcomp-l1-inf}
\min_{\bm w}  \tr{\mathcal{L} \bm w \bm S}  + \frac{\rho}{2} \norm{ \mathcal{L} \bm w - \bm \Theta^{(l_\infty)} -\frac{1}{\rho} \bm G^{(l_\infty)} }_{\mathrm{F}}^2 +  \sum_{i} h_{\lambda}(w_i), \quad \mathrm{subject \ to} \quad \bm w \geq \bm 0.
\end{equation}
In other words, $ \bm w^{(l_\infty)} $ satisfies the condition of the stationary point of $L_\rho(\bm \Theta, \bm w, \bm G)$ with respect to $\bm w$. To sum up, we conclude that any limit point $\left\lbrace \left( \bm \Theta^{(l_\infty)}, \bm w^{(l_\infty)}, \bm G^{(\infty)}\right) \right\rbrace$ of the sequence generated by Algorithm \ref{algo-3} is a stationary point of $L_\rho \left(\bm \Theta, \bm w, \bm{G} \right)$.

\end{proof}

\section{Proof of Technical Lemmas}\label{Lemmas}
This section contains the proofs of the technical lemmas used in Section \ref{Proof}.

\subsection{Proof of Lemma \ref{lem6}}
\begin{proof}
The gradient of $f(\bm w)$ is $\nabla f(\bm w) = - \L^{\ast} (\L \bm w + \bm J)^{-1}$ and its Hessian matrix is $\nabla^2 f(\bm w)$ with the $k$-th column being
\begin{equation*}
\begin{split}
\left[\nabla^2 f(\bm w)\right]_{:,k}& =\frac{\partial \left(\nabla f(\bm w) \right)}{\partial w_k} = - \L^{\ast} \left( \frac{ \partial(\L \bm w + \bm J)^{-1}}{\partial w_k}\right) \nonumber \\
& = \L^{\ast} \left(  \left(\L \bm w + \bm J \right)^{-1}  \frac{\partial \left(\L \bm w + \bm J \right)}{\partial w_k} \left(\L \bm w + \bm J \right)^{-1} \right)  \\
& = \L^{\ast} \left( \left(\L \bm w + \bm J \right)^{-1}  \bm A_k \left(\L \bm w + \bm J \right)^{-1} \right), 
\end{split}
\end{equation*} 
where $\bm A_k \in \mathbb{R}^{p \times p}$ is a matrix with $[\bm A_k]_{ii}=[\bm A_k]_{jj}=1$, $[\bm A_k]_{ij}=[\bm A_k]_{ji}=-1$ and zeros for the other elements, in which $i,j \in \mathbb{Z}^+$ obeying $k=i-j+\frac{j-1}{2}(2p-j)$ and $i>j$. Therefore, $\nabla^2 f(\bm w)$ can be written as
\begin{equation}
\nabla^2 f(\bm w)=[\L^{\ast} \bm B_1,\L^{\ast} \bm B_2, \ldots, \L^{\ast} \bm B_{p(p-1)/2}],   \nonumber
\end{equation} 
where $\bm B_k = (\L \bm w + \bm J)^{-1}  \bm A_k (\L \bm w + \bm J)^{-1}$, for $k=1, 2, \ldots, p(p-1)/2$. Then one has
\begin{equation*}
\begin{split}
\bm x^\top \nabla^2 f(\bm w) \bm x &= \bm x^\top \left[\L^{\ast} \bm B_1,\L^{\ast} \bm B_2, \ldots, \L^{\ast} \bm B_{p(p-1)/2} \right] \bm x \\
& = \bm x^\top \L^{\ast} \left( \sum_{k=1}^{p(p-1)/2} x_k \bm B_k \right)   \\
& = \bm x^\top \L^{\ast} \left( (\L \bm w + \bm J)^{-1}  \left( \sum_{k=1}^{p(p-1)/2} x_k \bm A_k \right)(\L \bm w + \bm J)^{-1} \right)  \nonumber \\
& = \bm x^\top \L^{\ast} \left( (\L \bm w + \bm J)^{-1}  \L \bm x (\L \bm w + \bm J)^{-1} \right)    \\
& = \left\langle \L\bm x,  (\L \bm w + \bm J)^{-1}  \L \bm x (\L \bm w + \bm J)^{-1} \right \rangle     \\
& = \mathrm{vec}(\L \bm x)^\top \mathrm{vec}\left( (\L \bm w + \bm J)^{-1}  \L \bm x (\L \bm w + \bm J)^{-1} \right)    \\
&=\mathrm{vec} (\L \bm x)^\top \left( \left( \L \bm w + \bm J \right)^{-1} \otimes \left( \L \bm w + \bm J \right)^{-1} \right)  \mathrm{vec} (\L \bm x), 
\end{split}
\end{equation*} 
where the forth equality follows from the definition of $\L$ in \eqref{operator}, and the last equality follows from the property of Kronecker product that $\mathrm{vec}(\bm A \bm B \bm C) = \left( \bm C^\top \otimes \bm A\right) \mathrm{vec}(\bm B)$.
\end{proof}

\subsection{Proof of Lemma \ref{lem18}}
\begin{proof}
Let $\bm X = \L \bm w + b\bm J$ with any $b \neq 0$ and any given $\bm w \in \mathbb{R}^{p(p-1)/2}$ obeying $(\L \bm w + \bm J) \in \mathcal{S}^p_{++}$. It is easy to verify that the column spaces as well as row spaces of $\L \bm w$ and $\bm J$ are orthogonal with each other. Hence $\bm X$ admits the eigenvalue decomposition
\begin{align} \label{Lw_J}
\bm X = \L \bm w + b \bm J=
\left[
\begin{array}{cc}
\bm U & \bm u 
\end{array}
\right]
\left[
\begin{array}{cc}
\bm \Lambda & \bm 0 \\
\bm 0 & b 
\end{array}
\right]
\left[
\begin{array}{cc}
\bm U & \bm u 
\end{array}
\right]^\top, 
\end{align} 
where $\L \bm w  = \bm U \bm \Lambda \bm U^\top$ and $b \bm J = b \bm u \bm u^\top$ with $\bm u =\frac{1}{\sqrt{p}}\bm 1_p$, in which $\bm 1_{p} \in \mathbb{R}^p$ with each element equal to 1. Notice that $\bm \Lambda$ is non-singular. $\bm X^{-1}$ admits the eigenvalue decomposition
\begin{align}
\bm X^{-1}= 
\left[
\begin{array}{cc}
\bm U & \bm u 
\end{array}
\right]
\left[
\begin{array}{cc} 
\bm \Lambda^{-1} & \bm 0 \\
\bm 0 & \frac{1}{b} 
\end{array}
\right]
\left[
\begin{array}{cc}
\bm U & \bm u 
\end{array}
\right]^\top = \bm U \bm \Lambda^{-1} \bm U^\top + \frac{1}{b} \bm J, \nonumber
\end{align} 
It is easy to check that $\bm U \bm \Lambda^{-1} \bm U^\top$ is symmetric and $\bm U \bm \Lambda^{-1} \bm U^\top \cdot \bm 1_{p} = \bm 0$. Therefore, there must exist a $\bm x$ such that $\L \bm x = \bm U \bm \Lambda^{-1} \bm U^\top$. One further obtains $x_k = \left[\bm U \bm \Lambda^{-1} \bm U^\top \right]_{ij}$, for $k \in[p(p-1)/2]$, where $i, j \in \mathbb{Z}^+$ obeying $k=i-j+\frac{j-1}{2}(2p-j)$ and $i>j$. Hence such $\bm x$ is fixed and unique for a given $\bm w$. Note that $\bm x$ is independent of $b$, and thus $\L \bm x  + \frac{1}{b} \bm J = \left( \L \bm w + b \bm J \right)^{-1}$ holds for any $b \neq 0$, completing the proof.
\end{proof}

%

\subsection{Proof of Lemma \ref{lem-L}}
\begin{proof}
For any $\bm X \in \mathbb{R}^{p \times p}$ obeying $\norm{\bm X}_{\max} = 1$, one has
\begin{equation*}
\begin{split}
\norm{\mathcal{L}^\ast \bm X}_{\max} &= \max_k \left|  \left[  \mathcal{L}^\ast  \bm X \right]_k  \right| \\
& = \max_{i>j} \left|  X_{ii}  + X_{jj} - X_{ij} - X_{ji} \right|  \\
& \leq \max_{i >j} \left|  X_{ii}  \right| + \left|  X_{jj}  \right| + \left|  X_{ij}  \right| + \left|  X_{ji}  \right| \\
& \leq 4,
\end{split}
\end{equation*}
with equality when $X_{ii} = X_{jj} =1$, $X_{ij} = X_{ji} = -1$, for some $i , j \in [p]$. Therefore, we conclude that $\normI{\mathcal{L}^\ast}_\infty = 4$.
\end{proof}

\subsection{Proof of Lemma \ref{lem11}}
\begin{proof}
Define an index set $\Omega_t$ by
	\begin{equation}\label{omega1}
	\Omega_t := \left \{ l \in [p(p-1)/2] \, | \left[\L \bm x \right]_{tt} = \sum_{l} x_l \right \}, \quad t \in [p].
	\end{equation} 
According to the definition of $\mathcal{L}$ in \eqref{operator}, for any $\bm x \in \mathbb{R}^{p(p-1)/2}$, one obtains that $\mathcal{L} \bm x \in \mathbb{R}^{p \times p}$ obeys
\begin{equation} \label{operator_2}
\left[\mathcal{L} \bm x \right]_{ij} =
\begin{cases}
-x_{k}\; & \; \;i >j,\\
\hspace{.2cm}[\mathcal{L}\bm x]_{ji} \; & \; \;i<j,\\
\sum_{l \in \Omega_i} x_l \; &\;\; i=j,
\end{cases}
\end{equation}
where $k=i-j+\frac{j-1}{2}(2p-j)$. By the definition of $\mathcal{L}^\ast$ in \eqref{adj-operator}, one further obtains that $\mathcal{L}^\ast\mathcal{L} \bm x \in \mathbb{R}^{p(p-1)/2}$ satisfies
\begin{equation} \label{def-M}
\left[\mathcal{L}^\ast\mathcal{L} \bm x \right]_k= [\L \bm x]_{ii} + [\L \bm x]_{jj} + 2x_k = \sum_{l \in \Omega_i \cup \Omega_j} x_l + 2x_k = \sum_{l \in \Omega_i \cup \Omega_j \setminus k} x_l + 4x_k,
\end{equation}
where $i,j \in [p]$ satisfying $ k=i-j+\frac{j-1}{2}(2p-j)$ and $i>j$. The last equality holds because
\begin{equation}
\sum_{l \in \Omega_i \cup \Omega_j} x_l = [\L \bm x]_{ii} + [\L \bm x]_{jj} = - \sum_{m \neq i} [\L \bm x]_{im} - \sum_{m \neq j} [\L \bm x]_{jm} = 2x_k - \sum_{m \neq i, j} [\L \bm x]_{im} - \sum_{m \neq i, j} [\L \bm x]_{jm}, \nonumber
\end{equation}
where the last equality follows from $[\L \bm x]_{ij}= -x_k$ according to \eqref{operator_2}. 

According to \eqref{def-M}, we conclude that there exists a matrix $\bm M \in \mathbb{R}^{\frac{p(p-1)}{2} \times \frac{p(p-1)}{2}}$ such that $\L^\ast \L \bm x = \bm M \bm x$ for any $\bm x \in \mathbb{R}^{p(p-1)/2}$, and $\bm M$ obeys
\begin{equation} \label{M}
M_{kl} =
\begin{cases}
4 \; &\;\; l=k, \\
1 \; & \; \; l \in \left(\Omega_i \cup \Omega_j\right) \backslash k,\\
0 \; & \; \; \text{Otherwise},\\
\end{cases}
\end{equation}
where $i,j \in [p]$ satisfying $ k=i-j+\frac{j-1}{2}(2p-j)$ and $i>j$. Note that we use the fact that $\{\Omega_i \setminus k \} \cap \{\Omega_j \setminus k\} = \varnothing$ with $k=i-j+\frac{j-1}{2}(2p-j)$.

Finally, we will compute the minimum and maximum eigenvalues of $\bm M$. To compute the minimum eigenvalue of $\bm M$, one has
	\begin{equation}
	\lambda_{\min}(\bm M) = \inf_{\bm x \neq \bm 0} \frac{\bm x^\top \bm M \bm x}{\norm{\bm x}^2} = \inf_{\bm x \neq \bm 0} \frac{ \norm{\L \bm x}_{\mathrm{F}}^2}{\norm{\bm x}^2} =\inf_{\bm x \neq \bm 0} \frac{ 2\sum_{k=1}^{{p(p-1)/2}}x_k^2 + \sum_{i=1}^{p}( \left[\L \bm x \right]_{ii})^2}{\norm{\bm x}^2} \geq 2,   \nonumber
	\end{equation}
with equality when $[\L \bm x]_{11}= \ldots = [\L \bm x]_{pp} = 0$, which can be written as $\bm Q \bm x =\bm 0$ with $\bm Q \in \mathbb{R}^{p \times \frac{p(p-1)}{2}}$. Obviously, there must exist a nonzero solution to $\bm Q \bm x =\bm 0$, and thus $\lambda_{\min}(\bm M)=2$. To compute the maximum eigenvalue of $\bm M$, one has
	\begin{align}
	\lambda_{\max}(\bm M) &= \sup_{\bm x \neq \bm 0} \frac{\bm x^\top \bm M \bm x}{\norm{\bm x}^2} =  \sup_{\bm x \neq \bm 0} \frac{ 2\sum_{k=1}^{p(p-1)/2}x_k^2 + \sum_{t=1}^{p}([\L \bm x]_{tt})^2}{\norm{\bm x}^2}  \nonumber  \\
	& = \sup_{\bm x \neq \bm 0} \frac{4\sum_{k=1}^{p(p-1)/2}x_k^2 + \sum_{t=1}^p \sum_{i, j \in \Omega_t, \ i \neq j} x_i x_j}{\norm{\bm x}^2}   \nonumber \\
	& \leq \sup_{\bm x \neq \bm 0} \frac{4\sum_{k=1}^{p(p-1)/2}x_k^2 + \frac{1}{2}\sum_{t=1}^p \sum_{i, j \in \Omega_t, \ i \neq j} (x_i^2 + x_j^2) }{\norm{\bm x}^2}  \nonumber  \\
	& = \sup_{\bm x \neq \bm 0} \frac{4\sum_{k=1}^{p(p-1)/2}x_k^2 + \sum_{t=1}^p (|\Omega_t| -1)\sum_{i\in \Omega_t} x_i^2}{\norm{\bm x}^2}  \nonumber  \\
	& = \sup_{\bm x \neq \bm 0} \frac{4\sum_{k=1}^{p(p-1)/2}x_k^2 + ( p-2)\sum_{t=1}^p\sum_{i\in \Omega_t} x_i^2}{\norm{\bm x}^2}  \nonumber  \\
	& = 2p,  \nonumber
	\end{align}
with equality when each element of $\bm x$ is equal with each other, and thus $\lambda_{\max}(\bm M)=2p$. The last second equality is obtained by plugging $\left| \Omega_t \right|=p-1$ with $t \in [p]$, which is easy to verify according to the definition of $\Omega_t$ in \eqref{omega1}; the last equality follows from $\sum_{t=1}^p \sum_{i\in \Omega_t} x_i^2 = 2\sum_{k=1}^{p(p-1)/2}x_k^2$, because for any $k \in [p(p-1)/2]$, $k \in \Omega_t$ only holds with $t= \{i, j\}$, where $i, j \in [p]$ obeying $ k=i-j+\frac{j-1}{2}(2p-j)$ and $i>j$.
\end{proof}

\subsection{Proof of Lemma \ref{lem3}}

\begin{proof}
Take $\lambda = \sqrt{4 \alpha c_0^{-1} \log p /n}$ and $n \geq 94 \alpha c_0^{-1} \lambda_{\max}^2 \left(\L \bm w^{\star} \right) s \log p$. Define a local region 
\begin{equation}
\mathcal{B}_{\bm M} \left(\bm w^{\star}; \lambda_{\max} \left(\L \bm w^{\star} \right)  \right)= \left\lbrace \bm w \, | \bm w \in \mathbb{B}_{\bm M} \left(\bm w^{\star}; \lambda_{\max} \left(\L \bm w^{\star} \right) \right) \cap \mathcal{S}_{\bm w} \right\rbrace, \nonumber
\end{equation}
where $\mathbb{B}_{\bm M} \left(\bm w^{\star}; r \right) =  \left\{ \bm w \in \mathbb{R}^{p(p-1)/2} \, | \norm{ \bm w - \bm w^{\star}}_{\bm M} \leq r \right\}$, in which $\norm{\bm x}_{\bm M}^2 = \left\langle \bm x, \bm M \bm x \right\rangle = \norm{\L \bm x}_{\mathrm{F}}^2$ with $\bm M \succ \bm 0$ defined in Lemma \ref{lem11}, and $\mathcal{S}_{\bm w} = \left\{ \bm w \in \mathbb{R}^{p(p-1)/2}\, | \bm w \geq \bm 0, (\L\bm w + \bm J) \in \S_{++}^p \right\}$.  
It is easy to check that $\bm w^\star \in \mathcal{B}_{\bm M}(\bm w^{\star}; \lambda_{\max}(\L \bm w^{\star}))$. 

Recall that $\hat{\bm w}$ minimizes the optimization
\begin{equation}
\min_{\bm w \geq \bm 0} - \log \det(\L \bm w + \bm J) + \tr{\L \bm w \bm S} + \bm z^{\top} \bm w, \label{w_hat}
\end{equation}
where $0 \leq z_i \leq \lambda$ for $i\in [p(p-1)/2]$. We can see the optimization problems \eqref{new_cost} and \eqref{w_hat} have the same feasible set. Therefore, $\mathcal{S}_{\bm w}$ is also the feasible set of \eqref{w_hat} and thus $\hat{\bm w} \in \mathcal{S}_{\bm w}$ must hold. 

Next, we will prove that $\hat{\bm w} \in \mathcal{B}_{\bm M} \left(\bm w^{\star};\lambda_{\max} \left(\L \bm w^{\star} \right) \right)$. We first construct an intermediate estimator,
\begin{equation}
\bm w_t = \bm w^{\star} + t \left(\hat{\bm w} - \bm w^{\star} \right), \label{wt}
\end{equation}
where $t$ is taken such that $\norm{\bm w_t - \bm w^{\star}}_{\bm M} = \lambda_{\max} \left(\L \bm w^{\star} \right)$ if $\norm{\hat{\bm w} - \bm w^{\star}}_{\bm M} > \lambda_{\max} \left(\L \bm w^{\star} \right)$, and $t=1$ otherwise. Hence $\norm{\bm w_t - \bm w^{\star}}_{\bm M} \leq \lambda_{\max} \left(\L \bm w^{\star} \right)$ always holds and $t \in [0, 1]$. One further has $\bm w_t \in \mathcal{S}_{\bm w}$ because both $\bm w^\star, \hat{\bm w} \in \mathcal{S}_{\bm w}$ and $\mathcal{S}_{\bm w}$ is a convex set as shown in \eqref{Lxt}. Therefore, we conclude that $\bm w_t \in \mathcal{B}_{\bm M} \left(\bm w^{\star}; \lambda_{\max} \left(\L \bm w^{\star} \right) \right)$. Applying Lemma \ref{lem2} with $\bm w_1 = \bm w_t$, $\bm w_2 = \bm w^\star$ and $r=  \lambda_{\max}(\L \bm w^{\star})$ yields
\begin{equation}\label{q1}
 t \left\langle -\L^{\ast}\left( \L \bm w_t + \bm J\right)^{-1} + \L^{\ast}\left( \L \bm w^\star + \bm J\right)^{-1}, \ \hat{\bm w} - \bm w^\star \right\rangle  \geq  \left( 2\lambda_{\max} \left(\L \bm w^{\star} \right) \right)^{-2} \norm{\mathcal{L} \bm w_t - \mathcal{L}\bm w^\star}_{\mathrm{F}}^2. 
\end{equation}
Let $q(a) = -\log \det \left(\L \big(\bm w^\star + a(\hat{\bm w} - \bm w^\star) \big) + \bm J \right) + a\langle \L^{\ast}(\L \bm w^\star + \bm J)^{-1}, 
\bm \hat{\bm w} - \bm w^\star \rangle$ and $a \in [0, 1]$. One has
\begin{equation}
q'(a)=\langle - \L^\ast\left( \L\bm w_a + \bm J\right)^{-1} + \L^\ast \left( \L \bm w^\star + \bm J \right)^{-1}, \ \hat{\bm w} - \bm w^\star \rangle, \label{q2}
\end{equation}
and
\begin{equation}
q''(a) = \left\langle \L^\ast \left( \left( \L \bm w_a + \bm J\right)^{-1} \left( \L \hat{\bm w} - \L \bm w^\star \right)\left( \L \bm w_a + \bm J\right)^{-1}\right), \ \hat{\bm w} - \bm w^\star \right\rangle = \tr{\bm A \bm B \bm A \bm B},   \nonumber
\end{equation}
where $\bm w_a=\bm w^\star + a \left(\hat{\bm w} - \bm w^\star \right)$, $\bm A = (\L \bm w_a + \bm J)^{-1}$ and $\bm B = \left( \L \hat{\bm w} - \L \bm w^\star \right)$. Note that $\bm A$ is symmetric and positive definite because $\bm w_t \in \mathcal{S}_{\bm w}$ and $\bm B$ is symmetric. Let $\bm C = \bm A \bm B$. According to Theorem 1 in \citet{drazin1962criteria}, all the eigenvalues of a matrix $\bm X \in \mathbb{R}^{p \times p}$ are real if there exists a symmetric and positive definite matrix $\bm Y \in \mathbb{R}^{p \times p}$ such that $\bm X \bm Y$ are symmetric. It is easy to check that the matrix $\bm C \bm A$ is symmetric with $\bm A$ symmetric and positive definite, and thus all the eigenvalues of $\bm C$ are real. Suppose $\lambda_1, \ldots, \lambda_p$ are the eigenvalues of $\bm C$. Then the eigenvalues of $\bm C \bm C$ are $\lambda_1^2, \ldots, \lambda_p^2$. Therefore, $q''(a) = \sum_{i=1}^p \lambda_i^2 \geq 0$, implying that $q'(a)$ is non-decreasing with the increase of $a$. Then one obtains
\begin{equation}\label{d4}
\begin{split}
&t \left\langle  \L^\ast \left( \L \bm w^\star + \bm J \right)^{-1} - \L^\ast\left( \L\hat{\bm w} + \bm J\right)^{-1}, \hat{\bm w} - \bm w^\star \right\rangle = tq'(1)\geq tq'(t) \\
& \qquad \qquad \qquad \geq \left( 2\lambda_{\max} \left(\L \bm w^{\star} \right) \right)^{-2} \norm{\mathcal{L} \bm w_t - \mathcal{L}\bm w^\star}_{\mathrm{F}}^2,
\end{split}
\end{equation}
where the first inequality holds because $q'(a)$ is non-decreasing and $t \leq 1$, and the second inequality follows from \eqref{q1}.

The Lagrangian of the optimization \eqref{w_hat} is
\begin{equation}
L(\bm w, \bm \upsilon) = - \log \det(\L \bm w + \bm J) + \tr{\L \bm w \bm S} + \bm z^{\top} \bm w - \bm \upsilon^{\top} \bm w, \nonumber
\end{equation}
where $\bm \upsilon$ is a KKT multiplier. Let $(\hat{\bm w}, \hat{\bm \upsilon})$ be the primal and dual optimal point. Then $(\hat{\bm w}, \hat{\bm \upsilon})$ must satisfy the KKT conditions as below
\begin{align}
-\L^{\ast} \left(  \left(\L \hat{\bm w} + \bm J \right)^{-1} \right) + \L^{\ast} \bm S + \bm z - \hat{\bm \upsilon} = \bm 0&; \label{kkt1}\\
\hat{w}_i \hat{\upsilon}_i =0, \ \text{for} \ i=1, \ldots, p(p-1)/2 &; \label{kkt2}\\
\hat{\bm w} \geq \bm 0, \ \hat{\bm \upsilon} \geq \bm 0&. \label{kkt3}
\end{align}
According to \eqref{kkt1}, one has
\begin{equation} \label{kkt1-new}
 \left\langle -\L^{\ast} \left( \L \hat{\bm w} + \bm J \right)^{-1} + \L^{\ast}\bm S, \, \hat{\bm w} - \bm w^{\star} \right\rangle = \left\langle \hat{\bm \upsilon} - \bm z, \hat{\bm w} - \bm w^{\star} \right\rangle.    
\end{equation}
Substituting \eqref{kkt1-new} into \eqref{d4} yields
\begin{align}
\norm{\L \bm w_t - \L \bm w^{\star}}_{\mathrm{F}}^2 & \leq  4t\lambda_{\max}^{2} \left(\L \bm w^{\star} \right) \left(  \left\langle \hat{\bm \upsilon} - \bm z, \hat{\bm w} - \bm w^{\star} \right\rangle +  \left\langle  \L^\ast \left( \left( \L \bm w^\star + \bm J \right)^{-1} - \bm S\right), \hat{\bm w} - \bm w^{\star} \right\rangle \right) \nonumber \\
& = 4t\lambda_{\max}^{2} \left(\L \bm w^{\star} \right) \bigg( \underbrace{ \left\langle \hat{\bm \upsilon}, \hat{\bm w} - \bm w^{\star} \right\rangle}_{\mathrm{term \; I}} -  \underbrace{ \left\langle \bm z, \hat{\bm w} - \bm w^{\star} \right\rangle}_{\mathrm{term \; II}} \nonumber \\
& \qquad \qquad  + \underbrace{\left\langle  \L^\ast \left( \left( \L \bm w^\star + \bm J \right)^{-1} - \bm S\right), \hat{\bm w} - \bm w^{\star} \right\rangle}_{\mathrm{term \; III}} \bigg). \label{temrs}
\end{align}

Next we will bound term I, II and III \eqref{temrs}, respectively. The term I can be directly bounded by
\begin{equation}\label{t1}
\left \langle \hat{\bm \upsilon}, \hat{\bm w} - \bm w^{\star} \right\rangle = - \left\langle \hat{\bm \upsilon}, \bm w^{\star} \right\rangle \leq 0, 
\end{equation}
where the equality follows from \eqref{kkt2} and the inequality follows from $\hat{\bm \upsilon} \geq \bm 0$ in \eqref{kkt3} and $\bm w^\star \geq \bm 0$.

For term II, we separate the support of $\bm z$ into two parts, $\mathcal{S}^\star$ and its complementary set $\left\{\mathcal{S}^\star \right\}^c$, where $\mathcal{S}^\star$ is the support of $\bm w^\star$ with $ \left|\mathcal{S}^\star \right| \leq s$. Take a set $\E$ satisfying $\mathcal{S}^\star \subseteq \E$ and $|\E| \leq 2s$. A simple algebra yields
\begin{equation}\label{t2}
\begin{split}
\left\langle \bm z, \hat{\bm w} - \bm w^{\star} \right\rangle &= \left\langle \bm z_{\mathcal{S}^\star}, \left(\hat{\bm w} - \bm w^{\star} \right)_{\mathcal{S}^\star} \right\rangle  + \Big\langle \bm z_{ \left\{\mathcal{S}^\star \right\}^c}, \left(\hat{\bm w} - \bm w^{\star} \right)_{ \left\{\mathcal{S}^\star \right\}^c} \Big\rangle   \\
& = \left\langle \bm z_{\mathcal{S}^\star}, \left(\hat{\bm w} - \bm w^{\star} \right)_{\mathcal{S}^\star} \right\rangle  + \Big\langle \bm z_{ \left\{\mathcal{S}^\star \right\}^c}, \hat{\bm w}_{ \left\{\mathcal{S}^\star \right\}^c} \Big\rangle        \\
& \geq - \norm{\bm z_{\mathcal{S}^\star}} \norm{ \left(\hat{\bm w} - \bm w^{\star} \right)_{\mathcal{S}^\star}} + \Big\langle \bm z_{ \left\{\mathcal{S}^\star \right\}^c}, \hat{\bm w}_{ \left\{\mathcal{S}^\star \right\}^c} \Big\rangle         \\
& \geq - \norm{\bm z_{\mathcal{S}^\star}} \norm{ \left(\hat{\bm w} - \bm w^{\star} \right)_{\mathcal{S}^\star}} + \left\langle \bm z_{\E^c}, \hat{\bm w}_{\E^c} \right\rangle, 
\end{split}
\end{equation}
where the first inequality follows from Cauchy–Schwarz inequality and the second inequality follows from $\bm z \geq \bm 0$, $\hat{\bm w} \geq \bm 0$ and $\E^c \subseteq \{\mathcal{S}^\star\}^c$. 

For term III, we separate the support of $\L^\ast \left( ( \L \bm w^\star + \bm J )^{-1} - \bm S \right)$ into parts, $\E$ and $\E^c$. Then one has
\begin{equation}\label{t3}
\begin{split}
&\left\langle \L^\ast \left( \left( \L \bm w^\star + \bm J \right)^{-1} - \bm S \right), \hat{\bm w} - \bm w^{\star} \right\rangle \\
= &\left\langle \left(\L^\ast \left(\left( \L \bm w^\star + \bm J \right)^{-1} - \bm S\right) \right)_{\E}, \left(\hat{\bm w} - \bm w^{\star} \right)_{\E} \right\rangle  \\
&\qquad \qquad \qquad \qquad + \left\langle \left( \L^\ast \left( \left( \L \bm w^\star + \bm J \right)^{-1} - \bm S\right) \right)_{\E^c}, \left(\hat{\bm w} - \bm w^{\star} \right)_{\E^c} \right\rangle    \\
\leq & \norm{ \left(\L^\ast \left(  \left(\L \bm w^\star + \bm J \right)^{-1} - \bm S \right) \right)_{\E}} \norm{ \left(\hat{\bm w} - \bm w^{\star} \right)_{\E}}  \\
&\qquad \qquad \qquad  \qquad + \left\langle \left( \L^\ast \left( \left( \L \bm w^\star + \bm J \right)^{-1} - \bm S\right) \right)_{\E^c}, \, \hat{\bm w} _{\E^c} \right\rangle . 
\end{split}
\end{equation}
Substituting \eqref{t1}, \eqref{t2} and \eqref{t3} into \eqref{temrs} yields
\begin{equation}\label{q4} 
\begin{split}
& \norm{ \L \bm{w}_t - \L \bm w^{\star}}_{\mathrm{F}}^2 \leq  ~ 4t\lambda_{\max}^{2} \left(\L \bm w^{\star} \right) \Big( \norm{ \left(  \L^{\ast} \left( \left( \L \bm w^{\star} + \bm J  \right)^{-1} - \bm S\right) \right)_{\E} } \norm{ \left( \hat{\bm w} - \bm w^{\star}  \right)_{\E} }  \\
 &\qquad \qquad + \left\langle  \left( \L^{\ast} \left( \left(\L \bm w^{\star} + \bm J \right)^{-1} - \bm S \right) \right)_{\E^c} - \bm z_{\E^c}, \hat{\bm w}_{\E^c} \right\rangle  + \norm{{\bm z}_{\mathcal{S}^\star}} \norm{ \left(\hat{\bm w} - {\bm w}^{\star} \right)_{\mathcal{S}^\star}}\Big).
 \end{split}
\end{equation}
Notice that the inequality
\begin{equation}\label{inq-lem}
\left\langle  \left( \L^{\ast} \left( \left(\L \bm w^{\star} + \bm J \right)^{-1} - \bm S \right) \right)_{\E^c} - \bm z_{\E^c}, \hat{\bm w}_{\E^c} \right\rangle \leq 0 
\end{equation}
holds because $\hat{\bm w} \geq \bm 0$, $\norm{\bm z_{\E^c}}_{\min} \geq \lambda/2$ and
\begin{equation}
\norm{\left( \L^{\ast} \left( \left(\L \bm w^{\star} + \bm J \right)^{-1} - \bm S \right)\right)_{\E^c} }_{\max} \leq \norm{ \L^{\ast} \left( \left(\L \bm w^{\star} + \bm J \right)^{-1} - \bm S \right) }_{\max}  \leq \frac{\lambda}{2},       \nonumber
\end{equation}
where the last inequality follows from the conditions in Lemma \ref{lem3}. Combining \eqref{q4} and \eqref{inq-lem} together yields
\begin{equation*}
\begin{split}
\norm{ \L \bm{w}_t - \L \bm w^{\star}}_{\mathrm{F}}^2 & \leq  4t\lambda_{\max}^{2} \left(\L \bm w^{\star} \right) \Big( \norm{{\bm z}_{\mathcal{S}^\star}} \norm{ \left(\hat{\bm w} - {\bm w}^{\star} \right)_{\mathcal{S}^\star}} \\
  & \qquad \qquad+ \norm{ \left(  \L^{\ast} \left( \left( \L \bm w^{\star} + \bm J  \right)^{-1} - \bm S  \right) \right)_{\E} }  \norm{ \left( \hat{\bm w} - \bm w^{\star}  \right)_{\E} }  \Big)  \\
 & \leq 4t\lambda_{\max}^{2} \left(\L \bm w^{\star} \right) \left( \norm{{\bm z}_{\mathcal{S}^\star}} + \norm{ \left(  \L^{\ast} \left( \left( \L \bm w^{\star} + \bm J  \right)^{-1} - \bm S  \right) \right)_{\E} }  \right) \norm{  \hat{\bm w} - \bm w^{\star}  }, 
 \end{split}
\end{equation*}
where the last inequality follows from $\norm{\hat{\bm w} - \bm w^\star} \geq \norm{(\hat{\bm w} - \bm w^\star)_{\mathcal{E}}} \geq \norm{(\hat{\bm w} - \bm w^\star)_{\mathcal{S}^\star}}$.
On the other hand, one has
\begin{equation*}
\norm{ \L \bm{w}_t - \L \bm w^{\star}}_{\mathrm{F}} = t \norm{\L \hat{\bm{w}} - \L \bm w^{\star}}_{\mathrm{F}}     
 \geq t \left(\sum_{i \neq j}\left( \left[\L \hat{\bm{w}} - \L \bm w^{\star} \right]_{ij}\right)^2 \right)^{\frac{1}{2}}   
 = \sqrt{2} t \norm{  \hat{\bm w} - \bm w^{\star} }.
\end{equation*}
Therefore, one has
\begin{equation}\label{eq1}
 \norm{ \L \bm{w}_t - \L \bm w^{\star}}_{\mathrm{F}}  \leq  2\sqrt{2}\lambda_{\max}^{2} \left(\L \bm w^{\star} \right) \left( \norm{{\bm z}_{\mathcal{S}^\star}} + \norm{ \left(  \L^{\ast} \left( \left( \L \bm w^{\star} + \bm J  \right)^{-1} - \bm S  \right) \right)_{\E} }  \right). 
\end{equation}

Recall that $\norm{\bm z}_{\max}\leq \lambda$ and $ \left|\mathcal{S}^\star \right| \leq s$. Thus one has
\begin{equation}
\norm{\bm z_{\mathcal{S}^\star}} \leq \sqrt{s} \lambda. \label{eq-z}
\end{equation}
One also has
\begin{equation}\label{q3}
\norm{ \left(  \L^{\ast} \left( \left( \L \bm w^{\star} + \bm J  \right)^{-1} - \bm S  \right) \right)_{\E} } \leq \left(  |\E| \  \norm{ \L^{\ast} \left( \left( \L \bm w^{\star} + \bm J  \right)^{-1} - \bm S \right) }_{\max}^2  \right)^{\frac{1}{2}}  \leq \frac{\sqrt{2}}{2} \sqrt{s} \lambda. 
\end{equation}

Substituting \eqref{eq-z} and \eqref{q3} into \eqref{eq1} yields
\begin{equation}\label{q30}
\norm{ \L \bm{w}_t - \L \bm w^{\star}}_{\mathrm{F}}  \leq 2(\sqrt{2} + 1) \lambda_{\max}^{2} \left(\L \bm w^{\star} \right) \sqrt{s} \lambda < \lambda_{\max}(\L \bm w^{\star}), 
\end{equation}
which implies that $t=1$ in \eqref{wt}, i.e., $\bm w_t = \hat{\bm w}$. The last inequality is established by plugging $\lambda = \sqrt{4 \alpha c_0^{-1} \log p /n}$ with $n \geq 94 \alpha c_0^{-1} \lambda_{\max}^2 \left(\L \bm w^{\star} \right) s \log p $. Therefore, we conclude that
\begin{equation*}
\begin{split}
\norm{\L \hat{\bm w} - \L \bm w^{\star}}_{\mathrm{F}} & \leq 2\sqrt{2} \lambda_{\max}^2 \left(\L \bm w^{\star} \right) \left( \norm{{\bm z}_{\mathcal{S}^\star}} + \norm{ \left(  \L^{\ast} \left( \left( \L \bm w^{\star} + \bm J  \right)^{-1} - \bm S  \right) \right)_{\E} }  \right)   \\
& \leq 2 (1+\sqrt{2})\lambda_{\max}^2 \left(\L \bm w^{\star} \right)\sqrt{s} \lambda,      
\end{split}
\end{equation*}
where the first inequality is established by \eqref{eq1} with $t=1$, and the second inequality is established by plugging \eqref{eq-z} and \eqref{q3}.
\end{proof}

\subsection{Proof of Lemma \ref{lem4}}
\begin{proof}
Recall that $\E^{(k)} = \left\{ \mathcal{S}^\star \cup \S^{(k)} \right\}$ and $\S^{(k)} = \left\{ i \in [p(p-1)/2] \, | \hat{w}_i^{(k-1)} \geq  b \right\}$ with $b = (2+\sqrt{2}) \lambda^2_{\max} \left(\L \bm w^{\star} \right)\lambda$. 

We prove $| \E^{(k)} | \leq 2s $ holds by induction. For $k=1$, $\forall i \notin \text{supp}^+(\hat{\bm w}^{(0)})$, i.e., $w_i^{(0)} \leq 0$, one has $w_i^{(0)} < b$, implying that $i \notin \S^{(1)}$. In other words, $\S^{(1)} \subseteq \text{supp}^+(\hat{\bm w}^{(0)})$. Then one has
\begin{equation}
|\E^{(1)}| = |\mathcal{S}^\star \cup \S^{(1)}| \leq |\mathcal{S}^\star \cup \text{supp}^+(\hat{\bm w}^{(0)}) | \leq s + s = 2s.    \nonumber
\end{equation} 
Therefore, $| \E^{(k)} | \leq 2s $ holds for $k=1$.

Assume $| \E^{(k-1)} | \leq 2s $ holds for some $k \geq 2$. We separate the set $\E^{(k)}$ into two parts, $\mathcal{S}^\star$ and $\S^{(k)} \backslash \mathcal{S}^\star$. For any $i \in \S^{(k)} \backslash \mathcal{S}^\star$, one has $\hat{w}_i^{(k-1)} \geq b$, and further obtains
\begin{equation}\label{q5}
\begin{split}
\sqrt{ \left |\S^{(k)} \backslash \mathcal{S}^\star \right|} & \leq \sqrt{\sum_{i \in \S^{(k)} \backslash \mathcal{S}^\star} \left( \frac{\hat{w}_i^{(k-1)}}{b} \right)^2 } = \frac{\norm{ \hat{\bm w}_{ \S^{(k)} \backslash \mathcal{S}^\star}^{(k-1)} }}{b}  \\
& = \frac{ \norm{ \left( \hat{\bm w}^{(k-1)} - \bm w^{\star} \right)_{ \S^{(k)} \backslash \mathcal{S}^\star} }}{b}  \leq \frac{ \norm{ \hat{\bm w}^{(k-1)} - \bm w^{\star} }}{b}.
\end{split}
\end{equation} 
Let $\bm z^{(k-2)}$ satisfy $z_i^{(k-2)} = h'_{\lambda} (\hat{w_i}^{(k-2)})$, $i \in [p(p-1)/2]$. By Assumption \ref{assumption 1}, one has $z_i^{(k-2)}  \in [0, \lambda]$ for $i \in [p(p-1)/2]$. For any $i \in \left\{ \S^{(k-1)}\right\}^c$, one further has
\begin{equation}\label{qq5}
z_i^{(k-2)} = h'_{\lambda} (\hat{w}_i^{(k-2)}) \geq h'_{\lambda} (b) \geq \frac{\lambda}{2}, 
\end{equation} 
where the first inequality holds because $\hat{w}_i^{(k-2)} < b$ for any $i \in \left\{ \S^{(k-1)}\right\}^c$ by the definition of $\S^{(k-1)}$, and $h'_{\lambda}$ is non-increasing by Assumption \ref{assumption 1}; the second inequality follows from Assumption \ref{assumption 1}. Therefore, one obtains
\begin{equation} 
\norm{ \bm z_{\left\{ \E^{(k-1)}\right\}^c}^{(k-2)} }_{\min} \geq \norm{ \bm z_{\left\{ \S^{(k-1)}\right\}^c}^{(k-2)} }_{\min} \geq \lambda/2,    \nonumber
\end{equation} 
where the first inequality follows from $\left\{ \E^{(k-1)}\right\}^c \subseteq \left\{ \S^{(k-1)}\right\}^c$ and the second inequality follows from \eqref{qq5}.
One also has $| \E^{(k-1)} | \leq 2s $ and $\mathcal{S}^\star \subseteq \E^{(k-1)}$. Hence we can apply Lemma \ref{lem3} with $\E = \E^{(k-1)}$ and $\bm z =\bm z^{(k-2)}$ and obtain
\begin{equation} \label{q6}
\norm{ \hat{\bm w}^{(k-1)} - \bm w^{\star} } \leq \frac{\sqrt{2}}{2} \norm{ \L \hat{\bm w}^{(k-1)} - \L \bm w^{\star} }_{\mathrm{F}} \leq  (2+\sqrt{2}) \lambda^2_{\max} \left(\L \bm w^{\star} \right) \sqrt{s} \lambda,  
\end{equation} 
where the first inequality holds with the proof similar to \eqref{lw-w}. Combining \eqref{q5} and \eqref{q6} together yields
\begin{equation}
\sqrt{ \left|\S^{(k)} \backslash \mathcal{S}^\star \right|} \leq \frac{(2+\sqrt{2}) \lambda^2_{\max} \left(\L \bm w^{\star} \right) \sqrt{s} \lambda}{b} = \sqrt{s},   \nonumber
\end{equation} 
where the last equality follows from $b = (2+\sqrt{2}) \lambda^2_{\max}(\L \bm w^{\star} )\lambda$. Therefore, one gets
\begin{equation}
\left|\E^{(k)}  \right| = \left|\mathcal{S}^\star \cup \S^{(k)} \backslash \mathcal{S}^\star \right| = \left|\mathcal{S}^\star \right| + \left|\S^{(k)} \backslash \mathcal{S}^\star \right| \leq s + s = 2s,   \nonumber
\end{equation}
completing the induction.
\end{proof}

\subsection{Proof of Lemma \ref{lem5}}

\begin{proof}
For any $k \geq 1$, one has $ \left| \E^{(k)} \right| \leq 2s$ by Lemma \ref{lem4}. According to the definition of $\E^{(k)}$ in \eqref{e-set}, one has $\mathcal{S}^\star \subseteq \E^{(k)}$. Let $z_i^{(k-1)} = h'_{\lambda} \left(\hat{w_i}^{(k-1)} \right)$, $i \in [p(p-1)/2]$. By Assumption \ref{assumption 1}, one has $z_i^{(k-1)}  \in [0, \lambda]$ for $i \in [p(p-1)/2]$. For any $i \in \left\{ \S^{(k)}\right\}^c$, one has
\begin{equation}
z_i^{(k-1)} = h'_{\lambda} \left(\hat{w}_i^{(k-1)} \right) \geq h'_{\lambda} (b) \geq \frac{\lambda}{2},    \nonumber
\end{equation} 
where the first inequality holds because $\hat{w}_i^{(k-1)} < b$ for any $i \in \left\{ \S^{(k)}\right\}^c$ by the definition of $\S^{(k)}$ in \eqref{e-set}, and $h'_{\lambda}$ is non-increasing by Assumption \ref{assumption 1}; the second inequality follows from Assumption \ref{assumption 1}. Therefore, $\Big \| \bm z_{\left\{ \E^{(k)}\right\}^c}^{(k-1)} \Big\|_{\min} \geq \Big \|\bm z_{\left\{ \S^{(k)}\right\}^c}^{(k-1)} \Big \|_{\min} \geq \lambda/2$. Applying Lemma \ref{lem3} with $\E = \E^{(k)}$ and $\bm z =\bm z^{(k-1)}$ yields
\begin{equation}\label{q7}
\norm{ \L \hat{\bm w}^{(k)} - \L \bm w^{\star} }_{\mathrm{F}} \leq 2\sqrt{2} \lambda_{\max}^2 \left(\L \bm w^{\star} \right) \left( \norm{ \bm z_{\mathcal{S}^\star}^{(k-1)} } + \norm{ \left(\L^{\ast} \left( \left( \L \bm w^{\star} + \bm J  \right)^{-1} - \bm S  \right) \right)_{\E^{(k)}} } \right).
\end{equation} 

We will show that the term $\Big \| \bm z_{\mathcal{S}^\star}^{(k-1)} \Big \|$ in \eqref{q7} can be bounded in terms of $ \norm{ \hat{\bm w}^{(k-1)} - \bm w^{\star} }$. For any given  $\bm w \in \mathbb{R}^{p(p-1)/2}$, if $ \left|w_i^{\star} - w_i \right| \geq b$, then one has
\begin{equation}
0 \leq h'_{\lambda}(w_i) \leq \lambda  \leq \lambda b^{-1} \left|w_i^{\star} - w_i \right|, \nonumber
\end{equation}
where $b= (2+\sqrt{2})\lambda^2_{\max} \left(\L \bm w^{\star} \right)\lambda$, and the first two inequalities follows from Assumption \ref{assumption 1}. Otherwise, one has $w_i^{\star} - w_i \leq \left|w_i^{\star} - w_i \right| \leq b$, then $0 \leq h'_{\lambda }(w_i) \leq h'_{\lambda } \left(w_i^{\star} - b \right)$ because $h'_{\lambda }$ is non-increasing. Totally, one has
\begin{equation}
h'_{\lambda }(w_i) \leq \lambda b^{-1} \left|w_i^{\star} - w_i \right| + h'_{\lambda } \left(w_i^{\star} - b \right), \quad \forall i \in [p(p-1)/2]. \label{eqlem7}
\end{equation}
Collecting the indices $i \in \mathcal{S}^\star$ together and applying \eqref{eqlem7} with $\bm w = \hat{\bm w}^{(k-1)}$ yields
\begin{equation}\label{q8}
\begin{split}
 \norm{ \bm z_{\mathcal{S}^\star}^{(k-1)}  } &= \norm{h'_{\lambda} \left(\hat{\bm w}^{(k-1)}_{\mathcal{S}^\star} \right)} \leq \frac{\lambda}{b} \norm{ \hat{\bm w}_{\mathcal{S}^\star}^{(k-1)} - \bm w_{\mathcal{S}^\star}^{\star} } + \norm{ h'_{\lambda}\left(\bm w_{\mathcal{S}^\star}^{\star} - \bm b \right) }   \\
&\leq \frac{\lambda}{b} \norm{ \hat{\bm w}^{(k-1)} - \bm w^{\star} } + \norm{ h'_{\lambda } \left( \bm w_{\mathcal{S}^\star}^{\star} - \bm b \right) }, 
\end{split}
\end{equation}
where $h'_{\lambda} \left(\hat{\bm w}^{(k-1)}_{\mathcal{S}^\star} \right) = \left(h'_{\lambda} \left(\hat{w}^{(k-1)}_i \right) \right)_{i \in \mathcal{S}^\star}$, and $\bm b =[b, \ldots, b]^{\top}$ is a constant vector. Combining \eqref{q7} and \eqref{q8} together yields
\begin{equation}\label{q10}
\begin{split}
& \norm{ \L \hat{\bm w}^{(k)} - \L \bm w^{\star} }_{\mathrm{F}} \leq  2\sqrt{2} \frac{\lambda}{b} \lambda_{\max}^2 \left(\L \bm w^{\star} \right) \; \norm{ \hat{\bm w}^{(k-1)} - \bm w^{\star} }\\
&  \qquad \qquad + 2\sqrt{2} \lambda_{\max}^2 \left(\L \bm w^{\star} \right) \left( \norm{ \left( \L^{\ast} \left( \left( \L \bm w^{\star} + \bm J  \right)^{-1} - \bm S  \right) \right)_{\E^{(k)}} } +  \norm{ h'_{\lambda} \left(\bm w_{\mathcal{S}^\star}^{\star} - \bm b \right) } \right).
\end{split}
\end{equation}
By separating the set $\E^{(k)}$ into two parts, ${\mathcal{S}^\star}$ and $\S^{(k)} \backslash {\mathcal{S}^\star}$, one has
\begin{equation*}
\begin{split}
\norm{ \left( \L^{\ast} \left( \left( \L \bm w^{\star} + \bm J  \right)^{-1} - \bm S  \right) \right)_{\E^{(k)}} } \leq & ~ \norm{ \left( \L^{\ast} \left( \left( \L \bm w^{\star} + \bm J  \right)^{-1} - \bm S  \right) \right)_{\mathcal{S}^\star} }   \\
&  + \norm{ \left( \L^{\ast} \left( \left( \L \bm w^{\star} + \bm J  \right)^{-1} - \bm S  \right) \right)_{\S^{(k)} \backslash {\mathcal{S}^\star}} }.
\end{split}
\end{equation*}
We show that the term $\norm{ \left(\L^{\ast} \left( \left( \L \bm w^{\star} + \bm J  \right)^{-1} - \bm S  \right) \right)_{\S^{(k)} \backslash {\mathcal{S}^\star}} }$ can be bounded in terms of $\norm{ \hat{\bm w}^{(k-1)} - \bm w^{\star} }$ as below.
\begin{equation*}
\begin{split}
\norm{ \left( \L^{\ast} \left( \left( \L \bm w^{\star} + \bm J  \right)^{-1} - \bm S  \right) \right)_{\S^{(k)} \backslash {\mathcal{S}^\star}}} & \leq \sqrt{ \left|\E^{(k)} \backslash {\mathcal{S}^\star} \right|} \, \norm{ \left(\L^{\ast} \left( \left( \L \bm w^{\star} + \bm J  \right)^{-1} - \bm S  \right) \right)_{\S^{(k)} \backslash {\mathcal{S}^\star}} }_{\max}         \\
& \leq \sqrt{ \left|\E^{(k)} \backslash {\mathcal{S}^\star} \right|}  \, \norm{\L^{\ast} \left( \left( \L \bm w^{\star} + \bm J  \right)^{-1} - \bm S  \right) }_{\max}        \\
& \leq \frac{1}{b}  \norm{ \hat{\bm w}^{(k-1)} - \bm w^{\star} } \, \norm{ \L^{\ast} \left( \left( \L \bm w^{\star} + \bm J  \right)^{-1} - \bm S  \right)  }_{\max}        \\
& \leq \frac{\lambda}{2b}  \norm{ \hat{\bm w}^{(k-1)} - \bm w^{\star} },     
\end{split}
\end{equation*}
where the last second equality follows from \eqref{q5}. Thus one has
\begin{equation} \label{q9}
\norm{ \left( \L^{\ast} \left( \left( \L \bm w^{\star} + \bm J  \right)^{-1} - \bm S  \right) \right)_{\E^{(k)}} } \leq \norm{ \left( \L^{\ast} \left( \left( \L \bm w^{\star} + \bm J  \right)^{-1} - \bm S  \right)  \right)_{\mathcal{S}^\star} } +  \frac{\lambda}{2b}\norm{ \hat{\bm w}^{(k-1)} - \bm w^{\star} }.
\end{equation}
Substituting \eqref{q9} into \eqref{q10} yields
\begin{equation*}
\begin{split}
\norm{ \L \hat{\bm w}^{(k)} - \L \bm w^{\star} }_{\mathrm{F}} & \leq 2\sqrt{2} \lambda_{\max}^2 \left(\L \bm w^{\star} \right)\left( \norm{ \left(\L^{\ast} \left( \left( \L \bm w^{\star} + \bm J  \right)^{-1} - \bm S  \right)  \right)_{\mathcal{S}^\star} } +   \norm{ h'_{\lambda} \left(\bm w_{\mathcal{S}^\star}^{\star} - \bm b \right) } \right) \\
& \qquad \qquad  + 3\sqrt{2}\frac{\lambda}{b} \lambda_{\max}^2 \left(\L \bm w^{\star} \right) \; \norm{ \hat{\bm w}^{(k-1)} - \bm w^{\star} }      \\
& = 2\sqrt{2} \lambda_{\max}^2 \left(\L \bm w^{\star} \right)  \norm{ \left(\L^{\ast} \left( \left( \L \bm w^{\star} + \bm J  \right)^{-1} - \bm S  \right) \right)_{\mathcal{S}^\star} } \\
& \qquad \qquad + \frac{3\sqrt{2}}{2+\sqrt{2}}  \norm{ \hat{\bm w}^{(k-1)} - \bm w^{\star} }   \\
& \leq 2\sqrt{2} \lambda_{\max}^2 \left(\L \bm w^{\star} \right)  \norm{ \left(\L^{\ast} \left( \left( \L \bm w^{\star} + \bm J  \right)^{-1} - \bm S  \right)  \right)_{\mathcal{S}^\star} } \\
& \qquad \qquad  + \frac{3}{2+\sqrt{2}}  \norm{ \L \hat{\bm w}^{(k-1)} - \L \bm w^{\star} }_{\mathrm{F}},
\end{split}
\end{equation*}
where the equality is established by plugging $b = (2+\sqrt{2}) \lambda^2_{\max} \left(\L \bm w^{\star} \right)\lambda$ and following from $\norm{ h'_{\lambda} \left(\bm w_{\mathcal{S}^\star}^{\star} - \bm b \right) } =0$ because $\norm{\bm w_{\mathcal{S}^\star}^{\star}}_{\min} - b \geq \gamma \lambda$ and $h'_{\lambda}(x)=0$ for any $x \geq \gamma \lambda$ following from Assumption \ref{assumption 2}; the last inequality follows from
\begin{equation}\label{lw-w}
\begin{split}
\norm{\L \hat{\bm{w}}^{(k)} - \L \bm w^{\star}}_{\mathrm{F}} &= \left( 2\sum_{i=1}^{p(p-1)/2} \left(\hat{w}_i^{(k)} - w^{\star}_i \right)^2 + \sum_{j=1}^{p} \left([\L \hat{\bm{w}}^{(k)} - \L \bm w^{\star}]_{jj}  \right)^2 \right)^{\frac{1}{2}} \\
&\geq \sqrt{2} \norm{ \hat{\bm{w}}^{(k)} - \bm w^{\star}}. 
\end{split}
\end{equation}
Similarly, one also obtains
\begin{equation*}
\begin{split}
 \norm{ \hat{\bm{w}}^{(k)} - \bm w^{\star}} & \leq \frac{\sqrt{2}}{2}  \norm{\L \hat{\bm w}^{(k)} - \L \bm w^{\star} }_{\mathrm{F}}    \\
& \leq 2 \lambda_{\max}^2 \left(\L \bm w^{\star} \right) \norm{ \left(\L^{\ast} \left( \left( \L \bm w^{\star} + \bm J  \right)^{-1} - \bm S  \right)  \right)_{\mathcal{S}^\star} } + \frac{3}{2+\sqrt{2}}  \norm{ \hat{\bm w}^{(k-1)} - \bm w^{\star} }. 
\end{split}
\end{equation*}
\end{proof}

\subsection{Proof of Lemma \ref{lem9}}
\begin{proof}
We apply Lemma \ref{lem8} with $t= \lambda/2$ and union sum bound, then get
\begin{equation}
\mathbb{P} \left[ \norm{ \L^{\ast} \left( \left( \L \bm w^{\star} + \bm J  \right)^{-1} - \bm S  \right)  }_{\max} \geq \lambda/2 \right] \leq p(p-1) \exp \left(- \frac{1}{4} c_0 n \lambda^2 \right) \leq p^2 \exp \left(- \frac{1}{4} c_0 n \lambda^2 \right),  \nonumber
\end{equation} 
for any $\lambda \leq   2t_0 $, where $t_0 = \norm{\L^{\ast} \left(\L \bm w^{\star} + \bm J \right)^{-1}}_{\max}$ and $c_0 = 1/\left( 8 \norm{\L^{\ast} \left(\L \bm w^{\star} + \bm J \right)^{-1}}_{\max}^2\right)$. Take $\lambda = \sqrt{4\alpha c_0^{-1} \log p /n} $ for some $\alpha >2$. To guarantee $\lambda \leq 2 t_0$, one takes $n \geq 8 \alpha \log p$. By calculation, we establish
\begin{equation}
\mathbb{P} \left[ \norm{\L^{\ast} \left( \left( \L \bm w^{\star} + \bm J  \right)^{-1} - \bm S  \right)  }_{\max} \leq \lambda/2 \right ] \geq 1- p^2 \exp \left(- \frac{1}{4} c_0 n \lambda^2 \right) \geq 1-1/p^{\alpha-2},    \nonumber
\end{equation}
completing the proof.
\end{proof}


\subsection{Proof of Lemma \ref{lem8}}
\begin{proof}
The LGMRF is a constrained GMRF model with $\bm 1^{\top} \bm x = 0$ and we can follow the method called \textit{conditioning by Kriging} \citep{rue2005gaussian} to sample LGMRF. More specifically, to sample $\bm x$ for LGMRF with precision matrix $\L \bm w^{\star}$,we could first sample from a unconstrained GMRF $\tilde{\bm x} \sim N \left(\bm 0, \left(\L \bm w^{\star} + \bm J \right)^{-1} \right)$ and then correct for the constraint $\bm 1^{\top} \bm x = 0$ by 
\begin{equation}
\bm x^{(k)} = \tilde{\bm x}^{(k)} - \frac{1}{p} \bm 1 \bm 1^{\top} \tilde{\bm x}^{(k)}, \quad \mathrm{for} \ k = 1, \ldots, n.  \nonumber
\end{equation} 
For any $i \in [p(p-1)/2]$, one has
\begin{equation}\label{q19}
\begin{split}
\left[ \L^{\ast} \bm S \right]_i & =  \left[ \L^{\ast} \left( \frac{1}{n} \sum_{k=1}^n \bm x^{(k)} \left( \bm x^{(k)}\right)^{\top} \right)\right]_i  = \frac{1}{n} \sum_{k=1}^n \left[ \L^{\ast} \left( \bm x^{(k)} \left(\bm x^{(k)} \right)^\top \right) \right]_i   =  \frac{1}{n} \sum_{k=1}^n \left( x_a^{(k)}- x_b^{(k)} \right)^2           \\
& =  \frac{1}{n} \sum_{k=1}^n \left( \left(\tilde{x}_a^{(k)} -  \left(\frac{1}{p} \bm 1 \bm 1^{\top} \tilde{\bm x}^{(k)} \right)_a \right)- \left(\tilde{x}_b^{(k)} -  \left(\frac{1}{p} \bm 1 \bm 1^{\top} \tilde{\bm x}^{(k)} \right)_b \right) \right)^2       \\
& = \frac{1}{n} \sum_{k=1}^n \left( \tilde{x}_a^{(k)}- \tilde{x}_b^{(k)} \right)^2,
\end{split}
\end{equation} 
where the indices $a$ and $b$ obey $i = a -b + (b-1)(2p-b)/2$ and $a>b$. 

Let $\tilde{\bm \Sigma } = \left(\L \bm w^{\star} + \bm J \right)^{-1}$ and thus $\tilde{\bm x} \sim N \left(\bm 0, \tilde{\bm \Sigma } \right)$. We first introduce two auxiliary random variables $Y_{k,i} := \tilde{x}_a^{(k)}- \tilde{x}_b^{(k)}$ and $Z_{k,i} := Y_{k,i}^2$. Together with \eqref{q19}, one has
\begin{equation} \label{new-q19}
\frac{1}{n}\sum_{k=1}^n Z_{k,i} = \left[\L^{\ast} \bm S \right]_i.
\end{equation} 

We can see $Y_{k, i} \sim N \left(0, \sigma^2_i \right)$ because of the fact that any linear combination of $p$ components in $\tilde{\bm x}$ has a univariate normal distribution. The variance of $Y_{k, i}$ is
\begin{equation}\label{q12}
\begin{split}
\sigma^2_i &=\mathbb{E} \left[ \left(Y_{k,i} - \mathbb{E} \left(Y_{k,i} \right) \right)^2 \right] = \mathbb{E} \left[ \left(\tilde{x}_a^{(k)}- \tilde{x}_b^{(k)} \right)^2 \right] \\
& = \tilde{\Sigma}_{aa} +  \tilde{\Sigma}_{bb} -  \tilde{\Sigma}_{ab} -  \tilde{\Sigma}_{ba} = \left( \L^{\ast} \tilde{\bm \Sigma}\right)_i. 
\end{split}
\end{equation} 
Therefore $Z_{k,i}/\sigma^2_i \sim \chi^2(1)$ and $\mathbb{E} \left[Z_{k,i}/\sigma^2_i \right] = 1$. We say a random variable $X$ is \textit{sub-exponential} if there are non-negative parameters $(\upsilon, \alpha)$ such that
\begin{equation}
 \mathbb{E} \left[ \exp \left (\lambda \left(X- \mathbb{E}[X] \right) \right) \right] \leq \exp \left(\frac{\upsilon^2 \lambda^2}{2} \right), \  \mathrm{for \ all} \  \left|\lambda \right| < \frac{1}{\alpha}. \label{q18}
\end{equation} 
By checking the condition in \eqref{q18}, one can conclude that $Z_{k,i}/\sigma^2_i$ is is sub-exponential with parameters $ (2, 4)$. Furthermore, if random variables $\{Y_k\}_{k=1}^n$ are independent and sub-exponential with parameters $\left(\upsilon_k, \alpha_k \right)$, then $\sum_{k=1}^n Y_k$ is still sub-exponential with parameters $\left(\upsilon_{\ast}, \alpha_{\ast} \right)$ where 
\begin{equation}
\upsilon_{\ast}: = \sqrt{\sum_{k=1}^n \upsilon_k^2} \quad \quad \mathrm{and} \quad \quad \alpha_{\ast} := \max_{k=1, \ldots, n} \alpha_k.  \nonumber
\end{equation} 
Thus, $\sum_{k=1}^n Z_{k,i}/\sigma^2_i$ is sub-exponential with parameters $(2\sqrt{n}, 4)$. The application of the sub-exponential tail bound in Lemma \ref{lem16} yields 
\begin{equation}
\mathbb{P} \left[ \left | \sum_{k=1}^n Z_{k,i}/\sigma^2_i - n \right | \geq t_0 \right] \leq 2\exp \left ( - \frac{t_0^2}{8n}\right ),  \quad  \mathrm{for} \ \  t_0 \in [0, n].   \nonumber
\end{equation} 
By taking $t_0 = nt / \max_i \sigma_i^2$, one has
\begin{equation}\label{pre-q20}
\begin{split}
\mathbb{P} \left[ \left | \frac{1}{n}\sum_{k=1}^n Z_{k,i} - \sigma_i^2 \right | \geq t \right] & = \mathbb{P} \left[ \left | \sum_{k=1}^n Z_{k,i}/\sigma^2_i - n \right | \geq \frac{nt}{\sigma_i^2} \right]  \\
& \leq \mathbb{P} \left[ \left | \sum_{k=1}^n Z_{k,i}/\sigma^2_i - n \right | \geq \frac{nt}{\max_i \sigma_i^2} \right]  \\
& \leq 2\exp \left ( - \frac{nt^2}{8 \left( \max_i \sigma_i^2\right)^2 }\right ) 
\end{split}
\end{equation}
holds for $t \in \left[0, \max_i \sigma_i^2 \right]$. Notice that $\sigma_i^2 =  \left( \L^{\ast} \left(\L \bm w^{\star} + \bm J  \right)^{-1} \right)_i $ according to \eqref{q12}. Substituting \eqref{new-q19} into \eqref{pre-q20} yields
\begin{equation}\label{q20}
\mathbb{P}\left [ \, \left | [\L^{\ast} \bm S]_i -  \left( \L^{\ast} \left(\L \bm w^{\star} + \bm J \right)^{-1} \right)_i \right | \geq t \right ]  \leq 2\exp  \left( - c_0 nt^2 \right),  \quad  \mathrm{for} \ \  t \in \left[0, t_0 \right],
\end{equation}
where $t_0 = \norm{\L^{\ast} \left(\L \bm w^{\star} + \bm J \right)^{-1}}_{\max}$ and $c_0 = 1/\left( 8 \norm{\L^{\ast} \left(\L \bm w^{\star} + \bm J \right)^{-1}}_{\max}^2 \right)$, completing the proof.
\end{proof}

\subsection{Proof of Lemma \ref{lem12}}
\begin{proof}
Recall that 
\begin{equation*}
\mathbb{B} \left(\bm w^{\star};r \right) = \left\{ \bm w \in \mathbb{R}^{p(p-1)/2} \, | \norm{ \bm w - \bm w^{\star}} \leq r \right\} \ \mathrm{and} \ \mathcal{S}_{\bm w} = \left\{ \bm w \, | \bm w \geq \bm 0, \left(\L\bm w + \bm J \right) \in \S_{++}^p \right\}.
\end{equation*}
It is easy to verify that the set $\mathbb{B} \left(\bm w^{\star}; r \right)$ is convex. For any $\bm x_1, \bm x_2 \in \mathcal{S}_{\bm w}$, define $\bm x_t = t\bm x_1 + (1-t)\bm x_2$, $t \in [0, 1]$. It is clear that $\bm x_t \geq \bm 0$. Since $\S_{++}^p$ is a convex cone, one has
\begin{equation} 
\L \bm x_t + \bm J = t \left(\L \bm x_1 + \bm J \right) + (1-t) \left(\L \bm x_2 + \bm J \right) \in \S_{++}^p,  \label{Lxt}
\end{equation} 
indicating that $\bm x_t \in \mathcal{S}_{\bm w}$ and thus the set $\mathcal{S}_{\bm w}$ is convex. Hence $\mathcal{B}\left(\bm w^{\star};r \right)$ is a convex set. For any $\bm w_1$, $\bm w_2 \in \mathcal{B} \left(\bm w^{\star}; \frac{1}{\sqrt{2p}\delta \tau} \right)$ with $\delta > 1$, by Mean Value Theorem, we have
\begin{equation}
f \left(\bm w_2 \right) = f \left(\bm w_1 \right) + \left\langle \nabla f \left(\bm w_1 \right), \bm w_2 - \bm w_1 \right\rangle + \frac{1}{2} \left\langle \bm w_2 - \bm w_1, \nabla^2 f \left(\bm w_t \right)  \left(\bm w_2 -\bm w_1 \right)  \right\rangle, \label{q23}
\end{equation}
where $\bm w_t = t\bm w_2 + (1-t)\bm w_1$ with $t \in [0, 1]$. Then one obtains
\begin{equation}\label{q21}
\begin{split}
\lambda_{\min}\left( \nabla^2 f \left(\bm w_t \right) \right) =& \inf_{\norm{\bm x}=1} \bm x^\top \nabla^2 f \left(\bm w_t \right) \bm x  \\
 =& \inf_{\norm{\bm x}=1} \left(\mathrm{vec} (\L \bm x) \right)^{\top} \left( \left( \L \bm w_t + \bm J \right)^{-1} \otimes \left( \L \bm w_t + \bm J \right)^{-1} \right)  \mathrm{vec} (\L \bm x)  \\
 \geq  & \inf_{\norm{\bm x} =1} \frac{ \left(\mathrm{vec} (\L \bm x) \right)^{\top}  \left(  \left(\L \bm w_t + \bm J \right)^{-1} \otimes \left(\L \bm w_t + \bm J \right)^{-1} \right) \mathrm{vec} (\L \bm x)}{\left( \mathrm{vec} (\L \bm x) \right)^{\top} \mathrm{vec} (\L \bm x)} \\
 & \qquad \times \inf_{\norm{\bm x} =1} \norm{\L \bm x}_{\mathrm{F}}^2       \\
= & \inf_{\norm{\bm x} =1} \frac{ \left(\mathrm{vec} (\L \bm x) \right)^{\top}  \left( \left( \left(\L \bm w_t \right)^{\dagger} + \bm J \right) \otimes \left( \left(\L \bm w_t \right)^{\dagger} + \bm J \right) \right) \mathrm{vec} (\L \bm x)}{\left( \mathrm{vec} (\L \bm x) \right)^{\top} \mathrm{vec} (\L \bm x)} \\
 & \qquad \times \inf_{\norm{\bm x} =1} \bm x^\top \bm M \bm x     \\
= & 2\inf_{\norm{\bm x} =1} \frac{ \left(\mathrm{vec} (\L \bm x) \right)^{\top}  \left( \left(\L \bm w_t \right)^{\dagger}  \otimes \left(\L \bm w_t \right)^{\dagger}  \right) \mathrm{vec} (\L \bm x)}{\left( \mathrm{vec} (\L \bm x) \right)^{\top} \mathrm{vec} (\L \bm x)}  \\
 \geq & 2 \lambda_{2}^2 \left( \left(\L \bm w_t \right)^{\dagger} \right),  
\end{split}
\end{equation}
where $\lambda_{2}  \left( \left(\L \bm w_t \right)^{\dagger} \right)$ denotes the second smallest eigenvalue of $ \left(\L \bm w_t \right)^{\dagger}$ and $ \left(\L \bm w_t \right)^{\dagger}$ is the pseudo inverse of $\L \bm w_t$. The second equality is according to Lemma \ref{lem6}; the third equality is established by $\left(\L \bm w_t + \bm J \right)^{-1} = \left(\L \bm w_t \right)^{\dagger} + \bm J$ because the row spaces as well as column spaces of $\L \bm w_t$ and $\bm J$ are orthogonal with each other, and $\bm M$ is defined in Lemma \ref{lem11} with $\lambda_{\min}(\bm M)=2$; the last equality follows from $\left( \mathrm{vec} (\L \bm x) \right)^{\top}  \left( \left(\L \bm w_t \right)^{\dagger} \otimes \bm J \right) \mathrm{vec} (\L \bm x) = 0$, $\left( \mathrm{vec} (\L \bm x) \right)^{\top}  \left(  \bm J \otimes \left(\L \bm w_t \right)^{\dagger} \right) \mathrm{vec} (\L \bm x)=0$, and $\left( \mathrm{vec} (\L \bm x)\right)^{\top}  (  \bm J \otimes \bm J )  \mathrm{vec} (\L \bm x) =0$ which are easy to verify; the last inequality holds because of the property of Kronecker product that the eigenvalues of $\bm A \otimes \bm B$ are $\lambda_i \mu_j$ with the corresponding eigenvector $\bm a_i \otimes \bm b_j$, where $\lambda_1, \ldots, \lambda_p$ are the eigenvalues of $\bm A \in \mathbb{R}^{p \times p}$ with the corresponding eigenvectors $\bm a_1, \ldots, \bm a_p$, and $\mu_1, \ldots, \mu_p$ are the eigenvalues of $\bm B \in \mathbb{R}^{p \times p}$ with the corresponding eigenvectors $\bm b_1, \ldots, \bm b_p$. Notice that there is one and only one zero eigenvalue for $\L \bm w_t$ because $\bm w_t \in \mathcal{S}_{\bm w}$. Assume $\lambda_1, \ldots, \lambda_p$ are the eigenvalues of $\left(\L \bm w_t \right)^{\dagger}$ with the corresponding eigenvectors $\bm a_1, \ldots, \bm a_p$. Without loss of generality, let $\lambda_1=0$ and then $\bm a_1 = \frac{1}{\sqrt{p}}\bm 1$. By calculation, one obtains
\begin{equation}
 \left(\mathrm{vec} (\L \bm x) \right)^{\top} \mathrm{vec} \left(\bm a_i \otimes \bm a_1 \right)=0, \quad \mathrm{and} \quad \left(\mathrm{vec} (\L \bm x) \right)^{\top} \mathrm{vec} \left(\bm a_1 \otimes \bm a_i \right)=0,  \nonumber
\end{equation}
for any $i=1, \ldots, p$ and any $\bm x \in \mathbb{R}^{p(p-1)/2}$, indicating that $\mathrm{vec} (\L \bm x)$ is orthogonal to all the eigenvectors of $ \left(\L \bm w_t \right)^{\dagger} \otimes \left(\L \bm w_t \right)^{\dagger}$ corresponding to zero eigenvalues. The smallest nonzero eigenvalue of $ \left(\L \bm w_t \right)^{\dagger} \otimes \left(\L \bm w_t \right)^{\dagger}$ is $\lambda_{2}^2 \left( \left(\L \bm w_t \right)^{\dagger} \right)$, establishing \eqref{q21}. One further has
\begin{equation}\label{q22}
\begin{split}
\lambda_{2}^2 \left( \left(\L \bm w_t \right)^{\dagger} \right) & = \norm{  \L \bm w_t }_2^{-2}  =\norm{\L \bm w_1 +t \L \left(\bm w_2 -\bm w_1 \right)}_2^{-2}            \\
& \geq \left( \norm{\L \bm w^{\star} }_2  +  (1-t) \norm{\L \bm w_1 - \L \bm w^{\star}}_2 + t\norm{\L \bm w_2 - \L \bm w^{\star}}_2  \right)^{-2}    \\
& \geq \left( \norm{\L \bm w^{\star} }_2 + (1-t)\sqrt{2p} \norm{\bm w_1 - \bm w^{\star}}+  t\sqrt{2p}\norm{\bm w_2 - \bm w^{\star}} \right)^{-2}          \\
& \geq \left( \tau + \frac{1}{\delta \tau} \right)^{-2}  \geq \frac{1}{\left( 1 + \delta^{-1}\right)^2 \tau^2}, 
\end{split}
\end{equation}
where the second inequality follows from $\norm{\L \bm x}_2 \leq \norm{\L \bm x}_{\mathrm{F}} \leq \left(\bm x^{\top} \bm M \bm x \right)^{\frac{1}{2}} \leq \sqrt{2p} \norm{\bm x}$ according to Lemma \ref{lem11}; the third inequality holds because both $\bm w_1$, $\bm w_2 \in \mathcal{B} \left(\bm w^{\star}; \frac{1}{\sqrt{2p}\delta \tau} \right)$, and $\norm{\L \bm w^{\star} }_2 \leq \tau$ following from Assumption \ref{assumption 2}; the last inequality follows from $\frac{1}{\delta \tau} \leq \delta^{-1} \tau$, where $\tau >1$. Substituting \eqref{q22} into \eqref{q21} yields
\begin{equation}\label{qq22}
\lambda_{\min} \left( \nabla^2 g \left(\bm w_t \right)\right) \geq \frac{2}{\left( 1 + \delta^{-1}\right)^2 \tau^2}. 
\end{equation}
Combining \eqref{q23} and \eqref{qq22} together yields
\begin{equation}
f \left(\bm w_2 \right) \geq f \left(\bm w_1 \right) + \left\langle \nabla f \left(\bm w_1 \right), \bm w_2 - \bm w_1 \right\rangle +\frac{1}{\left( 1 + \delta^{-1}\right)^2 \tau^2} \norm{\bm w_2 -\bm w_1}^2.   \nonumber
\end{equation}

Next, we will bound $\lambda_{\max} \left(\nabla^2 f \left(\bm w_t \right) \right)$. Similarly, one has
\begin{equation}\label{q24}
\begin{split}
\lambda_{\max} \left( \nabla^2 f \left(\bm w_t \right)\right) = & \sup_{\norm{\bm x} = 1} \bm x^{\top} \nabla^2 f \left(\bm w_t \right) \bm x     \\
=& \sup_{\norm{\bm x} =1} \left(\mathrm{vec} (\L \bm x) \right)^{\top} \left( \left( \L \bm w_t + \bm J \right)^{-1} \otimes \left( \L \bm w_t + \bm J \right)^{-1} \right)  \mathrm{vec} (\L \bm x)    \\
\leq & \sup_{\norm{\bm x} =1} \frac{ \left(\mathrm{vec} (\L \bm x) \right)^{\top}  \left( \left( \left(\L \bm w_t \right)^{\dagger} + \bm J \right) \otimes \left( \left(\L \bm w_t \right)^{\dagger} + \bm J \right) \right) \mathrm{vec} (\L \bm x)}{\left( \mathrm{vec} (\L \bm x) \right)^{\top} \mathrm{vec} (\L \bm x)}  \\
& \qquad \times \sup_{\norm{\bm x} =1} \norm{\L \bm x}_{\mathrm{F}}^2  \\
= & \sup_{\norm{\bm x} =1} \frac{ \left(\mathrm{vec} (\L \bm x) \right)^{\top}  \left( \left(\L \bm w_t \right)^{\dagger}  \otimes \left(\L \bm w_t \right)^{\dagger}  \right) \mathrm{vec} (\L \bm x)}{\left( \mathrm{vec} (\L \bm x) \right)^{\top} \mathrm{vec} (\L \bm x)}\cdot \lambda_{\max} (\bm M)  \\
\leq & \lambda_{\max}  \left( \left(\L \bm w_t \right)^{\dagger} \otimes \left(\L \bm w_t \right)^{\dagger} \right) \cdot \lambda_{\max} (\bm M)    \\
= & 2 p \lambda_{\max} \left( \left(\L \bm w_t \right)^{\dagger} \otimes \left(\L \bm w_t \right)^{\dagger} \right), 
\end{split} 
\end{equation}
where the last equality follows from Lemma \ref{lem11}. One further obtains,
\begin{equation}\label{qn24}
\lambda_{\max} \left( \left(\L \bm w_t \right)^{\dagger} \otimes \left(\L \bm w_t \right)^{\dagger} \right) = \lambda_{\max}^2 \left( \left(\L \bm w_t \right)^{\dagger}  \right) = \lambda_{2}^{-2} \left(\L \bm w_t \right),
\end{equation}
where $\lambda_{2} \left(\L \bm w_t \right)$ denotes the second smallest eigenvalue of $\L \bm w_t$, i.e., the minimum nonzero eigenvalue, which could be bounded by
\begin{equation}\label{q25}
\begin{split}
\lambda_2 \left(\L \bm w_t \right) &= \inf_{\norm{\bm x} =1, \bm x \bot \bm 1} \bm x^{\top} \left(\L \bm w_t \right) \bm x     \\
& \geq \inf_{\norm{\bm x} =1, \bm x \bot \bm 1} \bm x^{\top} \left(\L \bm w^{\star} \right) \bm x + (1-t)\inf_{\norm{\bm x} =1, \bm x \bot \bm 1} \bm x^{\top} \left(\L \bm w_1 - \L \bm w^{\star} \right) \bm x  \\
& \qquad \qquad \qquad + t\inf_{\norm{\bm x} =1, \bm x \bot \bm 1} \bm x^{\top} \left(\L \bm w_2 - \L \bm x^{\star} \right) \bm x          \\
& \geq \lambda_2 \left(\L \bm w^{\star} \right) - (1-t)\norm{\L \bm w_1 - \L \bm w^{\star}}_2 -t \norm{\L \bm w_2 - \L \bm w^{\star}}_2    \\
& \geq \lambda_2 \left(\L \bm w^{\star} \right) -  \frac{1}{\delta \tau}   \geq \frac{ \left(1 - \delta^{-1}  \right) }{\tau}, 
\end{split}
\end{equation}
where the first equality holds because $\L \bm w_t$ has only one zero eigenvalue and its eigenvector is $\frac{1}{\sqrt{p}}\bm 1$, and the eigvectors of $\L \bm w_t$ associated with different eigenvalues are orthogonal with each other; the last inequality follows from Assumption \ref{assumption 2}. Substituting \eqref{qn24} and \eqref{q25} into \eqref{q24} yields
\begin{equation}\label{qq26}
\lambda_{\max} \left( \nabla^2 f \left(\bm w_t \right)\right) \leq \frac{ 2p \tau^2}{ \left(1-\delta^{-1} \right)^2}. 
\end{equation}
Combining \eqref{q23} and \eqref{qq26} together yields
\begin{equation}
f \left(\bm w_2 \right) \leq f \left(\bm w_1 \right) + \left\langle \nabla f \left(\bm w_1 \right), \bm w_2 - \bm w_1 \right\rangle + \frac{ p \tau^2}{ \left(1-\delta^{-1} \right)^2} \norm{\bm w_2 -\bm w_1}^2,   \nonumber
\end{equation}
completing the proof.
\end{proof}

\subsection{Proof of Lemma \ref{lem2}}
\begin{proof}
Recall that $\mathbb{B}_{\bm M} \left(\bm w^{\star};r \right) = \left\{ \bm w \in \mathbb{R}^{p(p-1)/2} \, | \norm{ \bm w - \bm w^{\star}}_{\bm M} \leq r \right\}$, where $\norm{\bm x}_{\bm M}^2 = \left \langle \bm x, \bm M \bm x \right\rangle = \norm{\L \bm x}_{\mathrm{F}}^2$ with $\bm M \succ \bm 0$ defined in Lemma \ref{lem11}, and $\mathcal{S}_{\bm w} = \left\{ \bm w \, | \bm w \geq \bm 0, (\L\bm w + \bm J) \in \S_{++}^p \right\}$ 
 We can see $\mathcal{B}_{\bm M} \left(\bm w^{\star};r \right)$ is a convex set because both $\mathbb{B}_{\bm M} \left(\bm w^{\star}; r \right)$ and $\mathcal{S}_{\bm w}$ are convex. It is easy to check that $\mathcal{S}_{\bm w}$ is convex (See \eqref{Lxt} for more details). For any $\bm w_1$, $\bm w_2 \in \mathcal{B}_{\bm M} \left(\bm w^{\star}; r \right)$, by Mean Value Theorem, one obtains
\begin{equation}\label{nq23}
f \left(\bm w_2 \right) = f \left(\bm w_1 \right) + \left\langle \nabla f \left(\bm w_1 \right), \bm w_2 - \bm w_1 \right\rangle + \frac{1}{2} \left\langle \bm w_2 - \bm w_1, \nabla^2 f \left(\bm w_t \right)  \left(\bm w_2 -\bm w_1 \right) \right\rangle, 
\end{equation}
where $\bm w_t = t\bm w_2 + (1-t)\bm w_1$ with $t \in [0, 1]$. For any nonzero $\bm x \in \mathbb{R}^{p(p-1)/2}$, one has
\begin{equation} \label{nq21}
\begin{split}
\bm x^\top \nabla^2 f \left(\bm w_t \right) \bm x & = \left(\mathrm{vec} (\L \bm x) \right)^{\top} \left( \left( \L \bm w_t + \bm J \right)^{-1} \otimes \left( \L \bm w_t + \bm J \right)^{-1} \right)  \mathrm{vec} (\L \bm x)  \\
& = \frac{ \left(\mathrm{vec} (\L \bm x) \right)^{\top}  \left( \left(\L \bm w_t + \bm J \right)^{-1} \otimes \left(\L \bm w_t + \bm J \right)^{-1} \right) \mathrm{vec} (\L \bm x)}{\left( \mathrm{vec} (\L \bm x) \right)^{\top} \mathrm{vec} (\L \bm x)} \cdot \norm{\L \bm x}_{\mathrm{F}}^2  \\
&\geq \inf_{\bm y} \frac{ \left(\mathrm{vec} (\L \bm y) \right)^{\top}  \left( \left( \left(\L \bm w_t \right)^{\dagger} + \bm J \right) \otimes \left( \left(\L \bm w_t \right)^{\dagger} + \bm J \right) \right) \mathrm{vec} (\L \bm y)}{\left( \mathrm{vec} (\L \bm y) \right)^{\top} \mathrm{vec} (\L \bm y)} \cdot \norm{\L \bm x}_{\mathrm{F}}^2  \\
& \geq \lambda_{2}^2 \left( \left(\L \bm w_t \right)^{\dagger} \right)  \cdot \norm{\L \bm x}_{\mathrm{F}}^2, 
\end{split}
\end{equation}
where $\lambda_{2}  \left( \left(\L \bm w_t \right)^{\dagger} \right)$ denotes the second smallest eigenvalue of $ \left(\L \bm w_t \right)^{\dagger}$, and $\left(\L \bm w_t \right)^{\dagger}$ is the pseudo inverse of $\L \bm w_t$. The first equality follows from Lemma \ref{lem6} and the last inequality is according to \eqref{q21}. Notice that \eqref{nq21} also holds with $\bm x = \bm 0$. One further obtains
\begin{equation}\label{nq22}
\lambda_{2}^2 \left( \left(\L \bm w_t \right)^{\dagger} \right) \geq \left( \norm{\L \bm w^{\star} }_2  +  (1-t) \norm{\L \bm w_1 - \L \bm w^{\star}}_2 + t\norm{\L \bm w_2 - \L \bm w^{\star}}_2  \right)^{-2}  \geq \left( \norm{\L \bm w^{\star} }_2 + r \right)^{-2}, 
\end{equation}
where the second inequality is established by $\norm{\L \bm x}_2 \leq \norm{\L \bm x}_{\mathrm{F}}$ and the fact that both $\bm w_1$, $\bm w_2 \in \mathcal{B}_{\bm M} \left(\bm w^{\star}; r \right)$. Substituting \eqref{nq22} into \eqref{nq21} yields
\begin{equation}\label{nq27}
\bm x^\top \nabla^2 f \left(\bm w_t \right) \bm x \geq \left( \norm{\L \bm w^{\star} }_2 + r \right)^{-2} \cdot \norm{\L \bm x}_{\mathrm{F}}^2.  
\end{equation}
Combining \eqref{nq23} and \eqref{nq27} yields
\begin{equation}\label{nq24}
f \left(\bm w_2 \right) \geq f \left(\bm w_1 \right) + \left \langle \nabla f \left(\bm w_1 \right), \bm w_2 - \bm w_1 \right\rangle + \frac{1}{2} \left( \norm{\L \bm w^{\star} }_2 + r \right)^{-2} \norm{\mathcal{L} \bm w_1 -\mathcal{L} \bm w_2}_{\mathrm{F}}^2, 
\end{equation}
and
\begin{equation}\label{nq25}
f \left(\bm w_1 \right) \geq f \left(\bm w_2 \right) + \left\langle \nabla f \left(\bm w_2 \right), \bm w_1 - \bm w_2 \right\rangle + \frac{1}{2}  \left( \norm{\L \bm w^{\star} }_2 + r \right)^{-2} \norm{\mathcal{L} \bm w_1 -\mathcal{L} \bm w_2}_{\mathrm{F}}^2, 
\end{equation}
Combining \eqref{nq24} and \eqref{nq25}, we establish
\begin{equation}\label{nq26}
\left\langle \nabla f \left(\bm w_1 \right) - \nabla f \left(\bm w_2 \right), \bm w_1 - \bm w_2 \right\rangle \geq  \left( \norm{\L \bm w^{\star} }_2 + r \right)^{-2} \norm{\mathcal{L} \bm w_1 - \mathcal{L}\bm w_2}_{\mathrm{F}}^2, 
\end{equation}
completing the proof.
\end{proof}

\bibliographystyle{plainnat}

\end{document}